%% file: main.tex
\pdfoutput=1
\documentclass[10pt,twocolumn,letterpaper]{article}

\usepackage[pagenumbers]{cvpr} %

\input{preamble}
\usepackage[titletoc]{appendix}

\definecolor{cvprblue}{rgb}{0.21,0.49,0.74}
\usepackage[pagebackref,breaklinks,colorlinks,citecolor=cvprblue]{hyperref}

\def\paperID{*****} %
\def\confName{CVPR}
\def\confYear{2024}

\title{Be Yourself: 
Bounded Attention for Multi-Subject Text-to-Image Generation}

\author{Omer Dahary$^1$ \hspace{6mm}
        Or Patashnik$^{1,2}$ \hspace{6mm}
        Kfir Aberman$^2$ \hspace{6mm}
        Daniel Cohen-Or$^{1,2}$ \\[4pt]
        $^1$Tel Aviv University \hspace{10mm}
        $^2$Snap Research
        \\[-14pt]
}

\begin{document}
\input{figures/teaser}

\begin{abstract}
    \input{0-abstract}
\end{abstract}

\input{1-intro_new}

\input{2-related_work}
\input{3-prelim}

\input{4-analysis}

\input{5-method}

\input{6-experiments}

\input{7-conclusions}

{
    \small
    \bibliographystyle{ieeenat_fullname}
    \bibliography{main}
}
\clearpage
\begin{appendices}
\section*{\LARGE Appendix}
\input{8-supp}
\end{appendices}

\end{document}

%% file: preamble.tex
\usepackage[dvipsnames]{xcolor}
\usepackage{multirow}
\usepackage{mathtools}
\usepackage{bbm}
\usepackage{tikz}
\usepackage{comment}
\usepackage{scalerel}

%% file: figures/teaser.tex
\twocolumn[{%
    \renewcommand\twocolumn[1][]{#1}%
    \maketitle
    \begin{center}
        \includegraphics[width=1.0\textwidth]{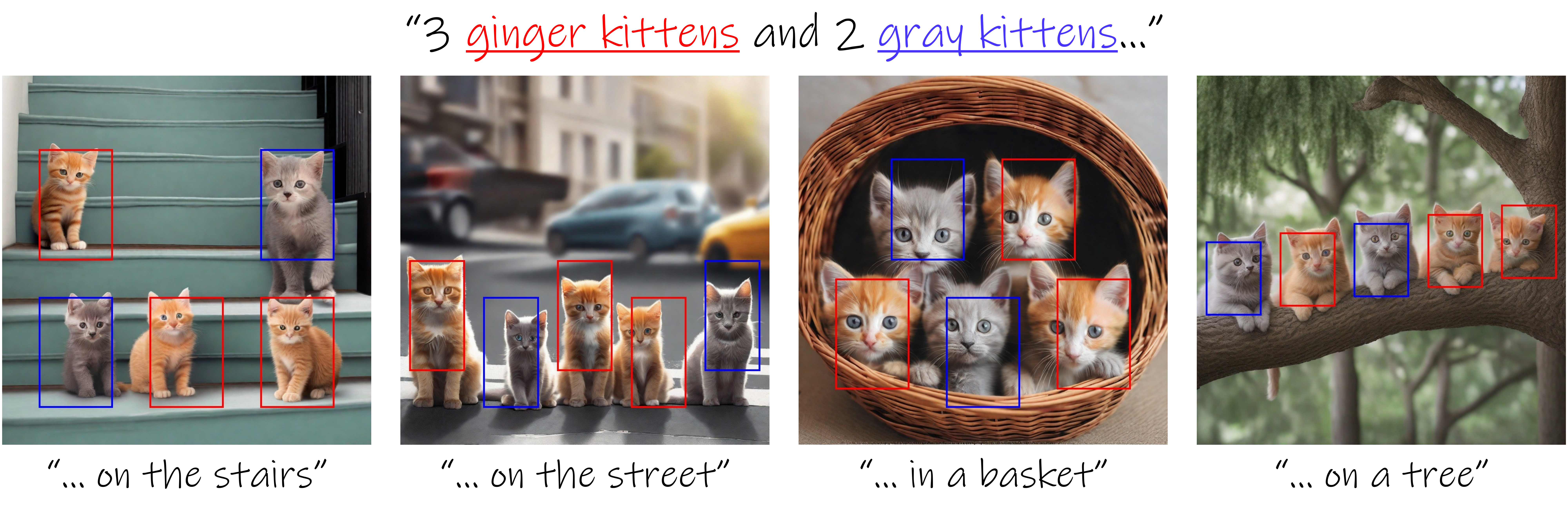}
        \vspace{-20pt}
        \captionof{figure}{Our method bounds the attention to enable layout control over a pre-trained text-to-image diffusion model. \textit{Bounded Attention} effectively reduces the impact of the innate semantic leakage during denoising, encouraging each subject to be itself. Our method can faithfully generate challenging layouts featuring multiple similar subjects with different modifiers (e.g., \textcolor{red}{\textit{\underline{ginger}}} and \textcolor{blue}{\textit{\underline{gray}}} kittens).
        }
    \label{fig:teaser}
    \end{center}
}]

%% file: 0-abstract.tex
\vspace{-9pt}

Text-to-image diffusion models have an unprecedented ability to generate diverse and high-quality images. However, they often struggle to faithfully capture the intended semantics of complex input prompts that include multiple subjects. Recently, numerous layout-to-image extensions have been introduced to improve user control, aiming to localize subjects represented by specific tokens. Yet, these methods often produce semantically inaccurate images, especially when dealing with multiple semantically or visually similar subjects. In this work, we study and analyze the causes of these limitations. Our exploration reveals that the primary issue stems from inadvertent semantic leakage between subjects in the denoising process. This leakage is attributed to the diffusion model’s attention layers, which tend to blend the visual features of different subjects. To address these issues, we introduce Bounded Attention, a training-free method for bounding the information flow in the sampling process. Bounded Attention prevents detrimental leakage among subjects and enables guiding the generation to promote each subject's individuality, even with complex multi-subject conditioning. Through extensive experimentation, we demonstrate that our method empowers the generation of multiple subjects that better align with given prompts and layouts.

\vspace{-9pt}

%% file: 1-intro_new.tex
\section{Introduction}

In recent years, text-to-image generation has undergone a significant shift with the integration of conditional diffusion models~\cite{saharia2022photorealistic,ramesh2022hierarchical,rombach2022high,podell2023sdxl}, allowing for the facile generation of high-quality and diverse images. 
The use of attention layers in architectures of such generative models has been identified as a major factor contributing to improved quality of generated images~\cite{rombach2022high, kang2023gigagan}.
However, these models struggle to accurately generate scenes containing multiple subjects, especially when they are semantically or visually similar.

In this work, we study the problem of multi-subject image generation in attention-based diffusion models.
Our contribution is twofold. First, we recognize the underlying reasons for the difficulty in generating images containing multiple subjects, especially those sharing semantic similarities.
Second, building on our insights, we present a method aimed at mitigating semantic leakage in the generated images, allowing control over the generation of multiple subjects (see Figure~\ref{fig:teaser}).

We demonstrate an innate bias within the common attention-based architectures utilized in diffusion models, which predisposes them to leak visual features between subjects. 
In particular, the functionality of attention layers is designed to blend features across the image. Therefore, they inherently lead to information leakage between subjects. This phenomenon is particularly noticeable when subjects are semantically similar and, therefore, attend to each other (Figure~\ref{fig:misalignment_examples}).

A plethora of works tries to mitigate the cross-subject leakage issue, either by modifying the sampling process to better follow different subjects in the prompt~\cite{feng2022training,chefer2023attend}, or by coupling the global prompt with layout information via segmentation maps or bounding boxes labeled with subject classes or local prompts~\cite{li2023gligen,kim2023dense,chen2023training,phung2023grounded}. 
However, the majority of these methods still encounter difficulties in accurately aligning to input layouts, particularly in scenarios involving two or more semantically similar subjects.

In our approach, we guide the image generation with a spatial layout~\cite{li2023gligen, phung2023grounded}.
To address cross-subject leakage, we introduce the \textit{Bounded Attention} mechanism, utilized during the denoising process to generate an image.
This method bounds the influence of irrelevant visual and textual tokens on each pixel, which otherwise promotes leakage.
By applying this mechanism, we encourage each subject \emph{to be itself}, in the sense that it hinders the borrowing of features from other subjects in the scene. We show that bounding the attention is needed both in the cross- and self-attention layers. Moreover, we find additional architectural components that amplify leakage, modify their operation, and present remedies to them.

We show that our method succeeds in facilitating control over multiple subjects within complex, coarse-grained layouts comprising numerous bounding boxes with similar semantics. Particularly challenging examples are demonstrated in Figure \ref{fig:teaser}, where we successfully generate five kittens with a mix of adjectives. We conduct experiments on both Stable Diffusion~\cite{rombach2022high} and SDXL~\cite{podell2023sdxl} architectures and demonstrate the advantage of our method compared to previous ones, both supervised and unsupervised.

\input{figures/misalignment_examples/full_figure}

%% file: figures/misalignment_examples/full_figure.tex
\begin{figure}
    \input{figures/misalignment_examples/figure}
    \captionof{figure}{
    \protect\input{figures/misalignment_examples/caption}
    }
    \label{fig:misalignment_examples}
\end{figure}

%% file: figures/misalignment_examples/figure.tex
\centering
\setlength{\tabcolsep}{0.001\textwidth}
{\scriptsize
\begin{tabular}{c c c c c}
    {\small Catastrophic} &
    \hspace{0.01\textwidth} &
    {\small Incorrect} &
    \hspace{0.01\textwidth} &
    {\small Subject}
    \\
    {\small neglect} &
    \hspace{0.01\textwidth} &
    {\small attribute binding} &
    \hspace{0.01\textwidth} &
    {\small fusion}
    \\
    \includegraphics[width=0.15\textwidth]{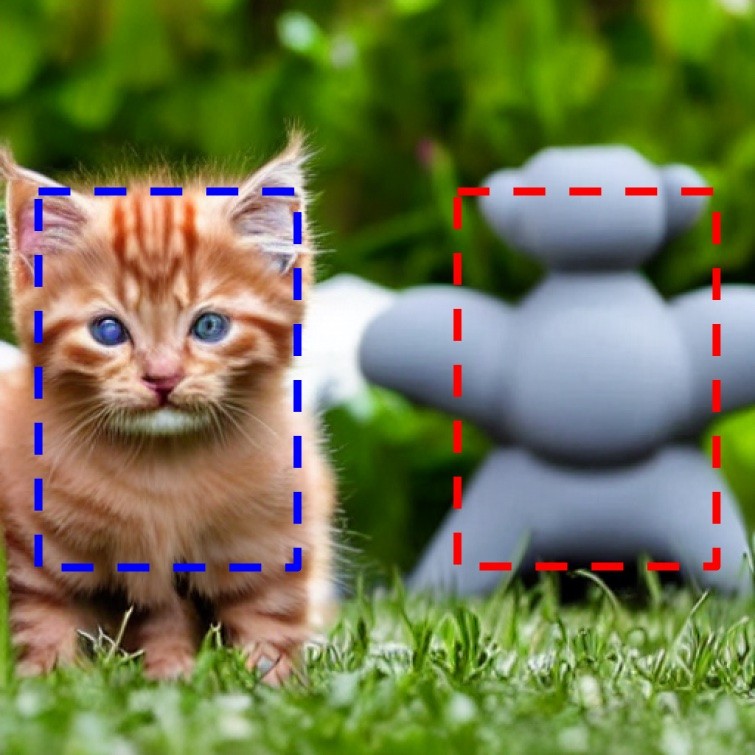}
    &
    \hspace{0.01\textwidth} &
    \includegraphics[width=0.15\textwidth]{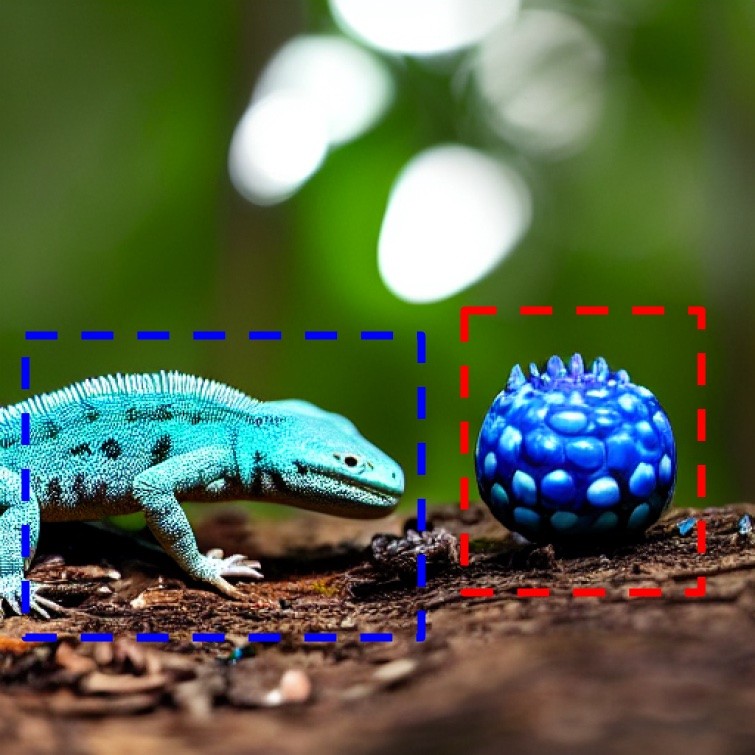}
    &
    \hspace{0.01\textwidth} &
    \includegraphics[width=0.15\textwidth]{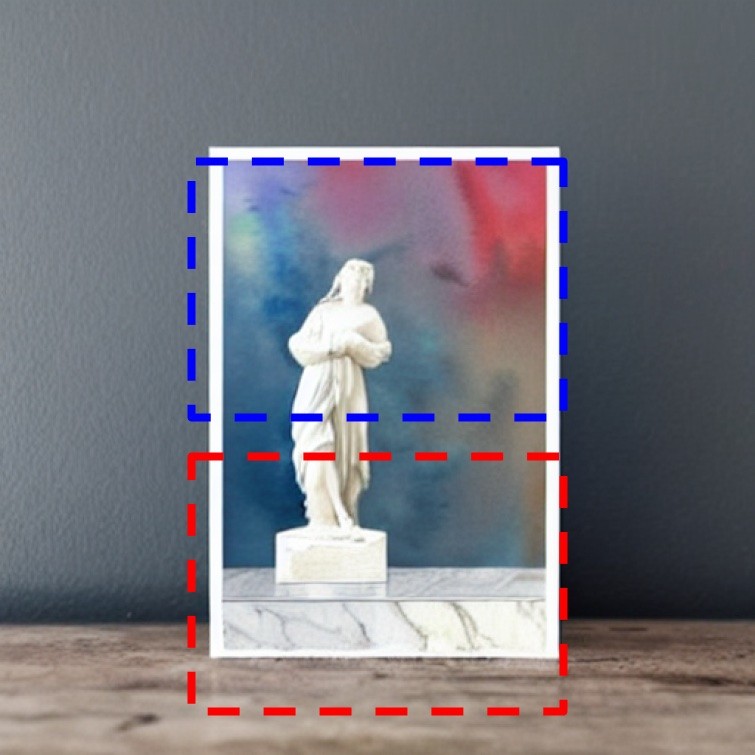}
    \\
    ``A \textcolor{blue}{\textit{\underline{ginger kitten}}} &
    \hspace{0.01\textwidth} &
    ``A \textcolor{blue}{\textit{\underline{spotted lizard}}} &
    \hspace{0.01\textwidth} &
    ``A \textcolor{blue}{\textit{\underline{watercolor painting}}} \\
    and a \textcolor{red}{\textit{\underline{gray puppy}}}'' &
    \hspace{0.01\textwidth} &
    and a \textcolor{red}{\textit{\underline{blue fruit}}}'' &
    \hspace{0.01\textwidth} &
    and a \textcolor{red}{\textit{\underline{marble statue}}}''
\end{tabular}
}

%% file: figures/misalignment_examples/caption.tex
\textbf{Misalignment in layout-to-image generation} include (i) \textit{catastrophic neglect}~\cite{chefer2023attend} where the model fails to include one or more subjects mentioned in the prompt within the generated image, (ii) \textit{incorrect attribute binding}~\cite{chefer2023attend,rassin2023linguistic} where attributes are not correctly matched to their corresponding subjects, and (iii) \textit{subject fusion}~\cite{zhao2023loco} where the model merges multiple subjects into a single, larger subject.

%% file: 2-related_work.tex
\section{Related work}

\paragraph{\textbf{Text-to-image diffusion models.}}

Diffusion models, trained on huge datasets~\cite{schuhmann2022laion5b}, have demonstrated their power in learning the complex distributions of diverse natural images~\cite{saharia2022photorealistic,ramesh2022hierarchical,balaji2022ediffi,rombach2022high,podell2023sdxl}.  Augmenting attention layers into diffusion models and conditioning them on textual prompts by coupling them with visually-aligned text encoders~\cite{radford2021learning} leads to powerful text-to-image models~\cite{ho2022classifier}. In this work, we specifically examine two such open-source text-to-image diffusion models: Stable Diffusion~\cite{rombach2022high}, and the more recent SDXL~\cite{podell2023sdxl}.

\paragraph{\textbf{Semantic alignment in text-to-image synthesis.}}
A critical drawback of current text-to-image models pertains to their limited ability to faithfully represent the precise semantics of input prompts. Various studies have identified common semantic failures and proposed mitigating strategies, such as adjusting text embeddings~\cite{tunanyan2023multi,feng2022training,white2022schr}, or optimizing noisy signals to strengthen or align cross-attention maps~\cite{chefer2023attend,rassin2023linguistic}. Nevertheless, these methods often fall short in generating multiple subjects, and do not adhere to positional semantics, such as subjects' number or location.

\paragraph{\textbf{Layout-guided image synthesis.}}
Addressing the semantic alignment concern, alternative approaches advocate for conditioning the diffusion process on layout information, either by training new models from scratch~\cite{zheng2023layoutdiffusion,qu2023layoutllm} or fine-tuning an existing one~\cite{zhang2023adding,li2023gligen,avrahami2023spatext,yang2023reco}. Despite their promise, these methods demand extensive computational resources and prolonged training times. Moreover, they are constrained by the layout distribution of the training data and the models' architectural bias to blend subject features, a limitation that our Bounded Attention aims to overcome.

To circumvent these challenges, numerous researchers explore training-free techniques, where the generation process itself is modified to enforce layout constraints. Several optimization-based works employ techniques similar to classifier-free guidance to localize the cross-attention~\cite{endo2023masked,chen2023training,xie2023boxdiff,zhao2023loco} and/or self-attention maps~\cite{phung2023grounded}. While effective in aligning random noise with the intended layout, guiding attention maps to distinct regions may inadvertently lead to undesired behavior, particularly when different subjects share similar semantics and visual features. Furthermore, these methods often exhibit a deteriorating effect on visual quality, thereby limiting their applicability to only the initial denoising steps and neglecting finer control over shape and visual details, which are determined only in later stages~\cite{patashnik2023localizing}. Bounded Attention addresses these shortcomings by regulating attention computation throughout the entire denoising process. 

Another approach involves generating each subject separately in its own denoising process~\cite{bar2023multidiffusion,lian2023llm}. While these methods inherently address catastrophic neglect, they tend to generate disharmonious images, and remain susceptible to leakage when merging subjects in subsequent stages. In contrast, masking attention maps to input bounding boxes~\cite{he2023localized} or attenuating attention in specific segments~\cite{kim2023dense} represents a milder variant of this strategy, aimed at avoiding visible stitching. However, these methods often fall short of fully mitigating subject leakage, compromising semantic alignment. In comparison, Bounded Attention is able to carefully govern information propagation among subjects in a single denoising process.

While both the trained models and training-free techniques aim to generate numerous objects, they do not mitigate the inherent leakage caused by attention mechanisms. Unlike our Bounded Attention technique, these methods encounter challenges in effectively generating and controlling a multitude of subjects, especially when they share semantic similarity. Notably, existing techniques struggle to accurately generate even two semantically similar subjects, whereas our method, as demonstrated succeeds in generating five and even more subjects.

%% file: 3-prelim.tex
\section{Preliminaries}

\paragraph{\textbf{Latent diffusion models.}} In this work, we examine Stable Diffusion~\cite{rombach2022high} and SDXL~\cite{podell2023sdxl}, which are both publicly available latent diffusion models. These models operate in the latent space of a pretrained image autoencoder, and are thus tasked with denoising a latent representation of the image, where each latent pixel corresponds to a patch in the generated image. Starting from pure random noise $\mathbf{z}_T$, at each timestep $t$, the current noisy latent $\mathbf{z}_t$ is passed to a denoising UNet $\epsilon_\theta$, trained to predict the current noise estimate $\epsilon_\theta\left(\mathbf{z}_t,y,t\right)$ using the guiding prompt encoding $y$. 

\vspace{-5pt}
\paragraph{\textbf{Attention layers.}} At each block, the UNet utilizes residual convolution layers, producing intermediate features $\phi^{(l)}(\mathbf{z}_t)$, where $l$ denotes the layer's index. These, in turn, are passed to attention layers, which essentially average different values $\mathbf{V^{\left(l\right)}_t}$ according to pixel-specific weights:
\begin{gather}
\phi^{(l+1)}(\mathbf{z}_t) = \mathbf{A}^{\left(l\right)}_t \mathbf{V}^{\left(l\right)}_t, \text{ where}
\\
\mathbf{A}^{\left(l\right)}_t = \operatorname{softmax}\left(\mathbf{Q}^{\left(l\right)}_t{\mathbf{K}^{\left(l\right)}_t}^\intercal\right).
\label{eq:attention}
\end{gather}

Here, the keys $\mathbf{K}^{\left(l\right)}_t = f^{\left(l\right)}_K\left(\mathbf{C}^{\left(l\right)}_t\right)$ and values $\mathbf{V}^{\left(l\right)}_t = f^{\left(l\right)}_V\left(\mathbf{\mathbf{C}^{\left(l\right)}_t}\right)$ are linear projections of context vectors $\mathbf{C}^{\left(l\right)}_t$. In the cross-attention layers we inject semantic context from the prompt encoding $\mathbf{C}^{\left(l\right)}_t\equiv y$, while self-attention layers utilize global information from the latent itself $\mathbf{C}^{\left(l\right)}_t\equiv \phi^{(l)}(\mathbf{z}_t)$. %

The weighting scheme is determined by the attention map $\mathbf{A}^{\left(l\right)}_t$ which represents the probability of each pixel to semantically associate with a key-value pair. This is done by linearly projecting the latent noisy pixels to queries $\mathbf{Q}^{\left(l\right)}_t=f^{\left(l\right)}_Q\left( \phi^{(l)}(\mathbf{z}_t) \right)$ and computing their inner product with the keys. It has been widely demonstrated that the cross-attention maps are highly indicative of the semantic association between the image layout and the prompt tokens~\cite{hertz2022prompt}. Meanwhile, the self-attention maps govern the correspondence between pixels, and thus form the image's structure~\cite{tumanyan2023plug}.

%% file: 4-analysis.tex
\section{Semantic Leakage}

\label{sec:analysis}

We begin by studying the causes of semantic leakage in Stable Diffusion~\cite{rombach2022high}, and examine the limitations of existing layout-to-image approaches.

\subsection{On Subject Similarity}

Figure~\ref{fig:misalignment_examples} illustrates various misalignment failures observed in state-of-the-art layout-to-image training-free methods. 
As we shall show, these failures are more prevalent for subjects that share semantic or visual similarity.

Let $x_{s_1},x_{s_2}\in\mathbb{R}^2$ be 2D latent coordinates corresponding to two semantically similar subjects $s_1, s_2$ in the generated image. Intuitively, we expect that along the denoising process, the queries corresponding to these pixels, $\mathbf{Q}^{\left(l\right)}_t\left[x_{s_1}\right], \mathbf{Q}^{\left(l\right)}_t\left[x_{s_2}\right]$, will be similar and hence also their attention responses. This, in turn, also implies that they will share semantic information from the token embeddings through the cross-attention layers or visual information via the self-attention layers.

To explore this hypothesis, we investigate the model's behavior when tasked with generating two subjects and analyze their attention features in both cross- and self-attention layers. Subsequently, we meticulously examine these features and demonstrate how their behavior sheds light on the leakage observed in generated images.

\subsection{Cross-Attention Leakage}

To analyze the leakage caused by cross-attention layers, we examine the cross-attention queries.
We depict these queries in the plots in Figure~\ref{fig:analysis_cross}, where each point corresponds to a single query projected to 2D with PCA.
To label each point with its corresponding subject,
we compute the subject masks by averaging cross-attention maps~\cite{hertz2022prompt} and color each projected query according to the subject's text color.
The leftmost plot, in which the two subjects were generated separately, serves as a reference point to the relation between the queries when there is no leakage.
For comparative analysis, we also present results for Layout-guidance~\cite{chen2023training}, as a simple representative of current training-free layout-guided approaches, and Bounded Attention, which we shall cover in the sequel.

\input{figures/analysis_cross/full_figure}

We consider the cross-attention queries in two examples: ``a kitten'' and ``a puppy'', and ``a hamster'' and a ``squirrel''.
As can be seen in the reference plots, the kitten and puppy queries share some small overlap, and the hamster and squirrel queries are mixed together.
The level of separation between the red and blue dots in the plots reveals the semantic similarity of the two forming subjects.

Clearly, vanilla Stable Diffusion struggles to adequately generate the two subjects within the same image. This is apparent by the visual leakage between the two subjects, as the model cannot avoid averaging their distinct visual properties. For example, the puppy has the visual features of a kitten, like raised ears and a triangular nose, while the squirrel loses its distinct ear shape and its feet are now pink like the hamster. 
Respectively, it can be seen that the queries of the two subjects are mixed, even for the kitten and the puppy which are more separated in the reference plot.

Meanwhile, Layout Guidance (LG), which optimizes $\mathbf{z}_t$ to have attention peaks for each noun token at their corresponding region, exhibits interesting results. Its optimization objective implicitly encourages the separation between subjects' cross-attention queries. This can have positive effects, like the hamster and squirrel having unique colors, but at the same time yields unwanted artifacts, like the puppy losing its face altogether. Moreover, it can inadvertently push the latent signal out of distribution, causing quality degradation, as evident by the hamster's and squirrel's cartoonish texture. When it overly pushes the latent out of distribution, it leads to the catastrophic neglect phenomenon (Figure \ref{fig:misalignment_examples}).

In comparison, when examining the plots of our method alongside the reference plots, our approach preserves the feature distribution of the subjects' queries, and successfully generates the two subjects, even when the queries are as mixed as in the hamster and the squirrel.

The above analysis yields two immediate conclusions: 
(i) Semantic similarity between subjects is reflected by their queries proximity, and leads to mixed queries when the subjects are generated together. This in turn leads to leakage between the subjects in the generated images,
and (ii) enforcing semantic separation by modifying the semantic meaning of the cross-attention queries is harmful. The former observation represents a crucial architectural limitation in current diffusion models, and the latter pinpoints to a previously unexplored weakness in the widely used latent optimization methodology. Bounded Attention is designed to overcome these limitations.

\subsection{Self-Attention Leakage}

We now turn to analyze the leakage caused by self-attention layers.
It has been shown that self-attention features exhibit dense correspondences within the same subject~\cite{cao2023masactrl} and across semantic similar ones~\cite{alaluf2023cross}. 
Hence, they are suspected as another source of leakage, that we shall study next. 

Here, we choose to examine the self-attention maps as a means to understand the leakage. In Figure~\ref{fig:analysis_self}
we focus on representative pixels (marked in yellow) associated with the subjects' eyes and legs, where visual leakage is most pronounced. As expected, features from one subject's eye or leg attend to the semantic similar body parts of the other. 
As a result, the features of each of the yellow points are directly affected by the features of the counterpart subject, causing leakage. In both images, the crab and the frog have similar appearances. In vanilla SD, both have a crab-like color and limbs with frog's toe pads. In LG, both have frog-like eyes and crab legs.

\input{figures/analysis_self/full_figure}

Notably, this tendency to rely on similar patches aids the model in denoising the latent signal and is essential for achieving coherent images with properly blended subjects and backgrounds. Nevertheless, it has the drawback of leaking features between disjointed subjects. Therefore, completely disjointing the visual features during denoising~\cite{bar2023multidiffusion}, or naively pushing self-attention queries apart through optimization~\cite{phung2023grounded}, can lead to subpar results. Consequently, we introduce the Bounded Attention mechanism to mitigate leakage and guide the latent signal towards subject separability, while avoiding detrimental artifacts. 

It's important to highlight that the leakage resulting from both cross- and self-attention layers is intertwined and mutually reinforced. Therefore, addressing the leakage caused by only one of these layers is insufficient to prevent the leakage in the generated images.

\subsection{Levels of Similarity}
\input{figures/analysis_layers/full_figure}
In the two previous sections, our focus was primarily on subjects that share semantic similarity. Building upon the observation that the UNet's inner layers dictate the subject's semantics and shape, while the outer layers control its style and appearance~\cite{voynov2023p+}, we analyzed the inner UNet layers. However, leakage can also occur prominently when generating subjects that share visual similarity rather than semantic. We now turn to explore this scenario and demonstrate that, in such cases, the leakage originates from the UNet's outer layers.

In Figure \ref{fig:analysis_layers}, we visualize cross-attention queries at different decoder's layers, when generating the kitten, puppy, lizard, and fruit in isolation. As can be seen, the queries of the kitten and the puppy are mixed across all UNet's layers, aligning with the visual and semantic similarity between these animals.
On the other hand, the queries of the lizard and the fruit are overlapped only in the highest-resolution layer, aligning with the lack of semantic similarity between them. A closer look reveals that the lizard and the fruit share surprisingly similar textures, which explains the overlap of the queries in the highest-resolution layer. As explained in the previous sections, this overlap causes leakage between the two subjects, in this case, a visual rather than a semantic leakage (see Figure~\ref{fig:misalignment_examples}).

%% file: figures/analysis_cross/full_figure.tex
\begin{figure}
        \input{figures/analysis_cross/figure}
    \captionof{figure}{
    \protect\input{figures/analysis_cross/caption}
    }
    \label{fig:analysis_cross}
\end{figure}

%% file: figures/analysis_cross/figure.tex
\setlength{\tabcolsep}{0.001\textwidth}
\begin{tabular}{c c c c c}
    A \textcolor{red}{\textit{\underline{kitten}}} &
    A \textcolor{blue}{\textit{\underline{puppy}}} &
    \multicolumn{3}{c}{A \textcolor{red}{\textit{\underline{kitten}}} and a \textcolor{blue}{\textit{\underline{puppy}}}} \\[2pt]
    \includegraphics[width=0.09\textwidth]{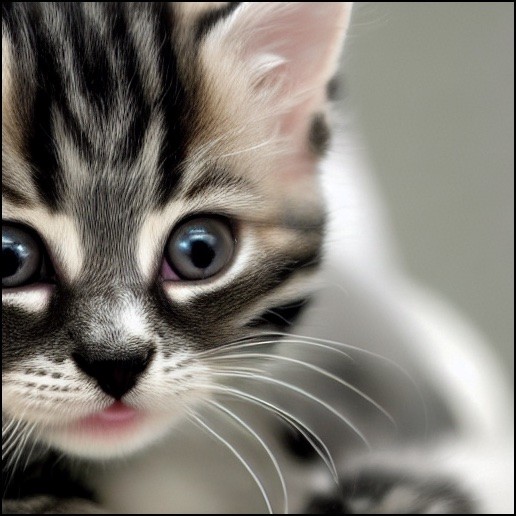} &
    \includegraphics[width=0.09\textwidth]{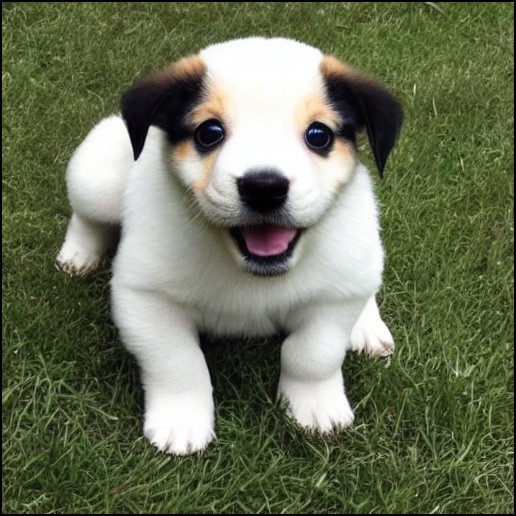} &
    \includegraphics[width=0.09\textwidth]{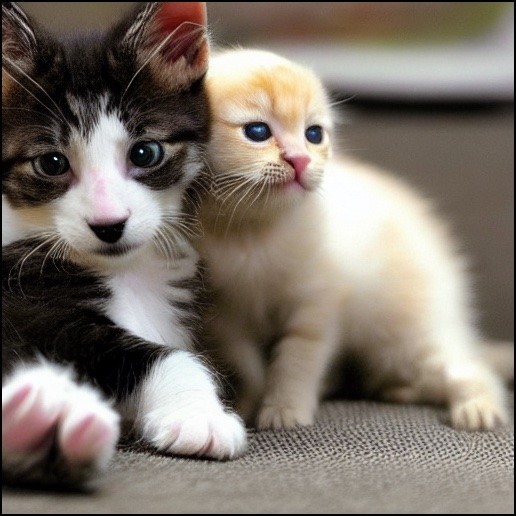} &
    \includegraphics[width=0.09\textwidth]{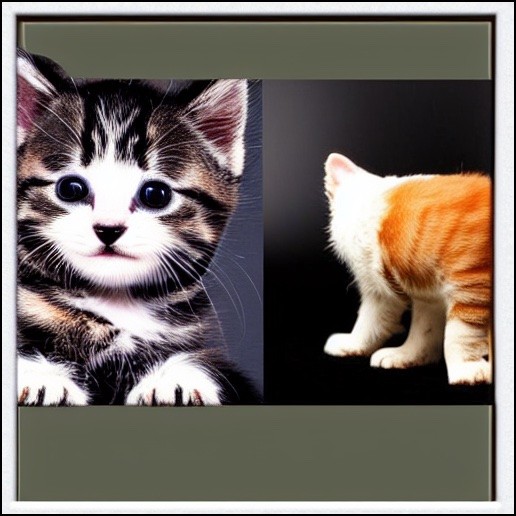} &
    \includegraphics[width=0.09\textwidth]{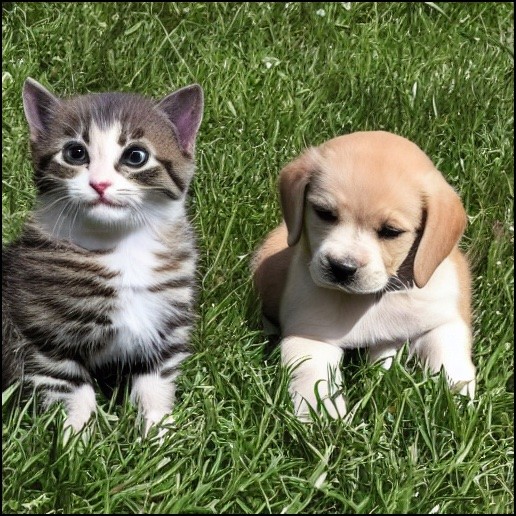} \\[-2pt]
    \multicolumn{2}{c}{\includegraphics[width=0.09\textwidth]{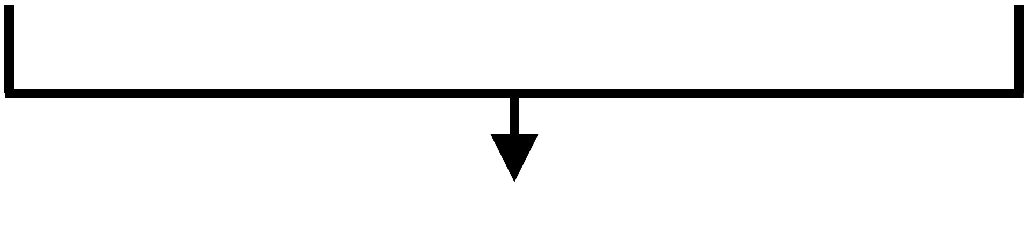}} &
    \includegraphics[width=0.09\textwidth]{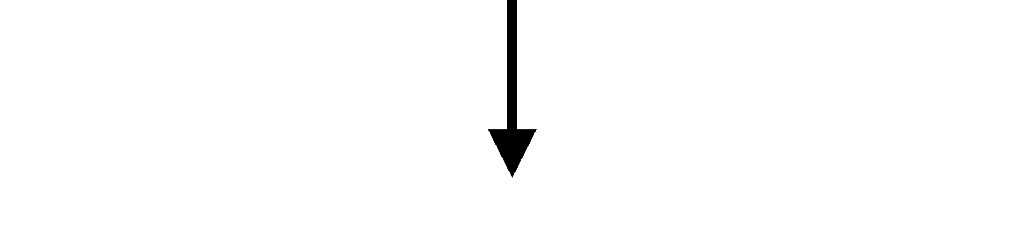} &
    \includegraphics[width=0.09\textwidth]{images/analysis/single_arrow.png} &
    \includegraphics[width=0.09\textwidth]{images/analysis/single_arrow.png} \\[-4pt]
    \multicolumn{2}{c}{\includegraphics[width=0.09\textwidth]{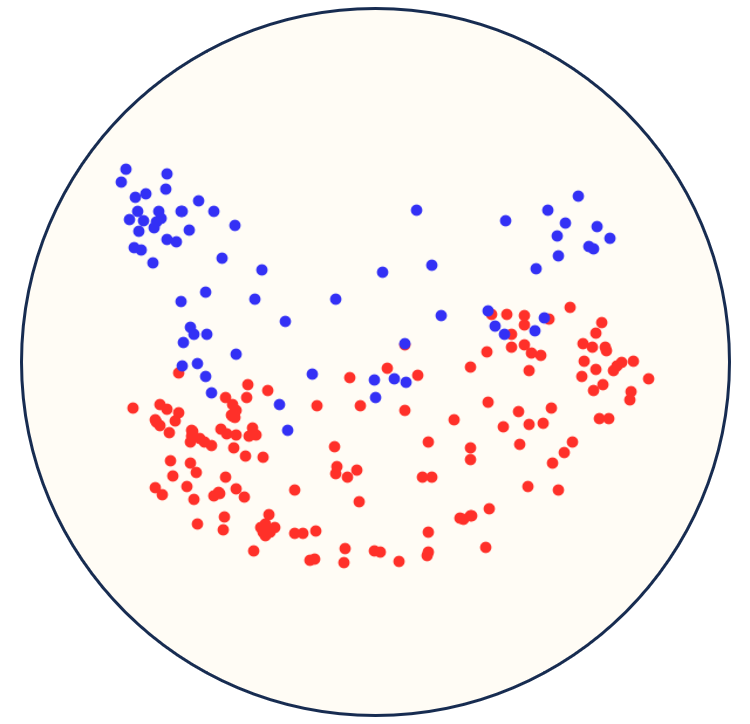}} &
    \includegraphics[width=0.09\textwidth]{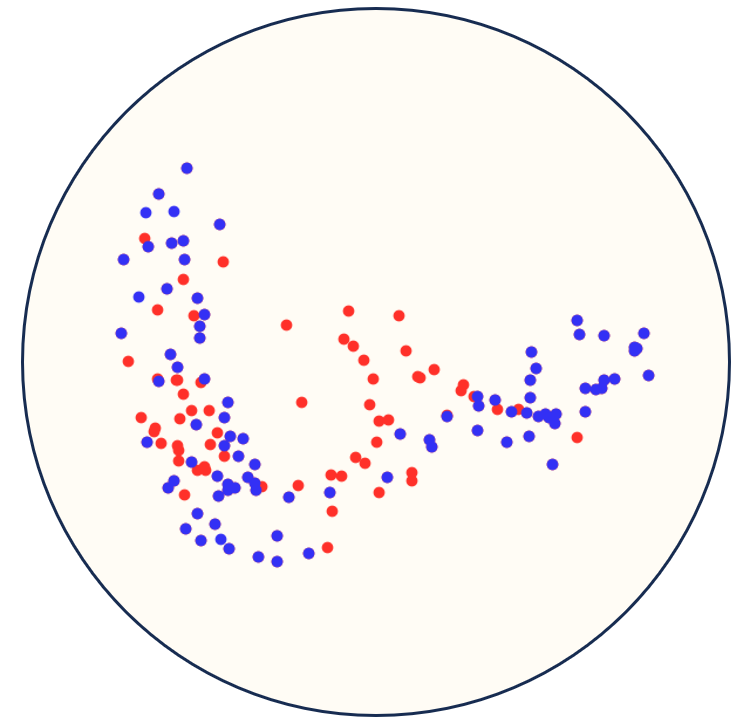} &
    \includegraphics[width=0.09\textwidth]{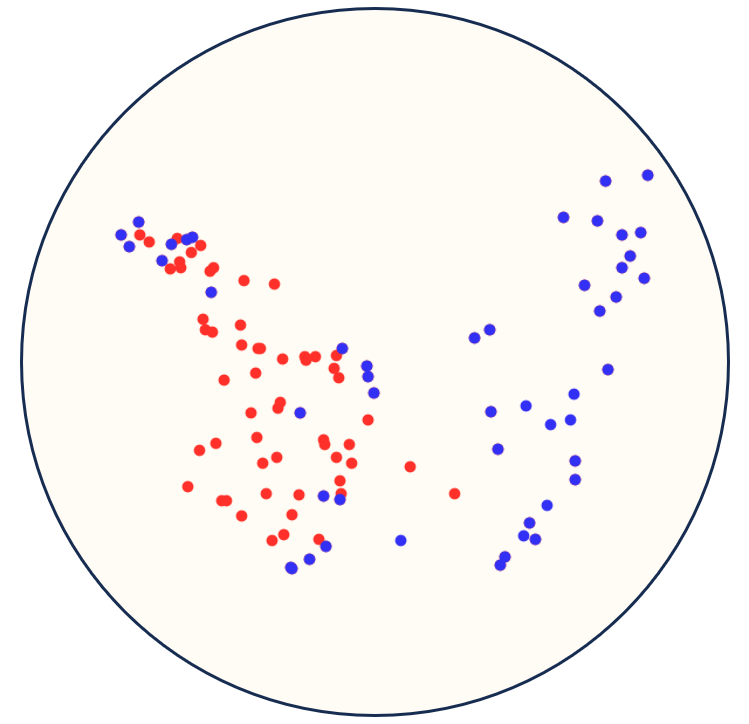} &
    \includegraphics[width=0.09\textwidth]{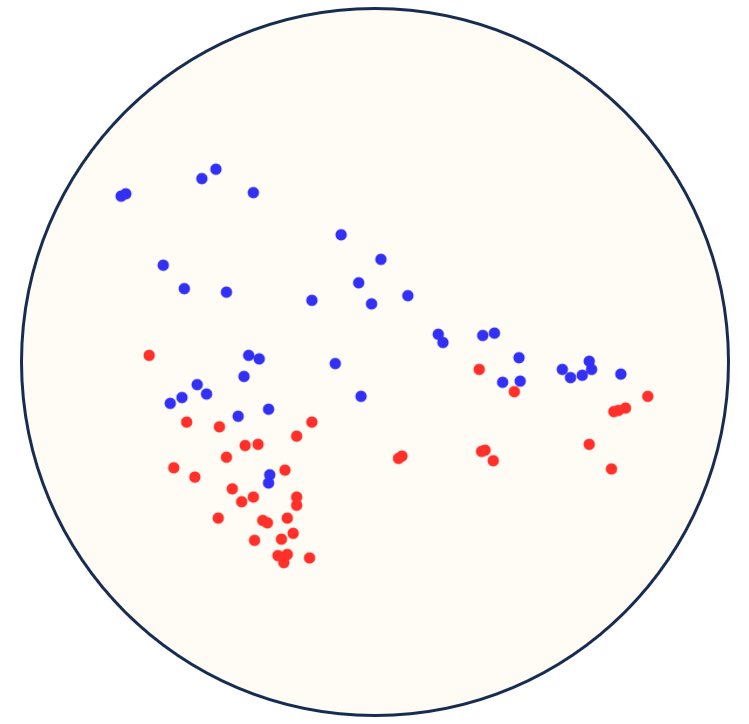}
    \\
    A \textcolor{red}{\textit{\underline{hamster}}} &
    A \textcolor{blue}{\textit{\underline{squirrel}}} &
    \multicolumn{3}{c}{A \textcolor{red}{\textit{\underline{hamster}}} and a \textcolor{blue}{\textit{\underline{squirrel}}}} \\
    \includegraphics[width=0.09\textwidth]{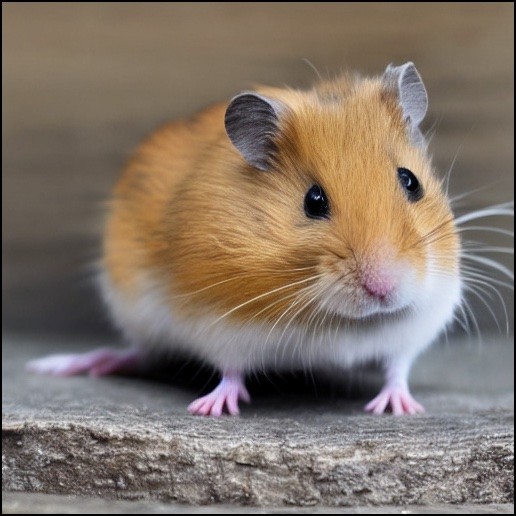} &
    \includegraphics[width=0.09\textwidth]{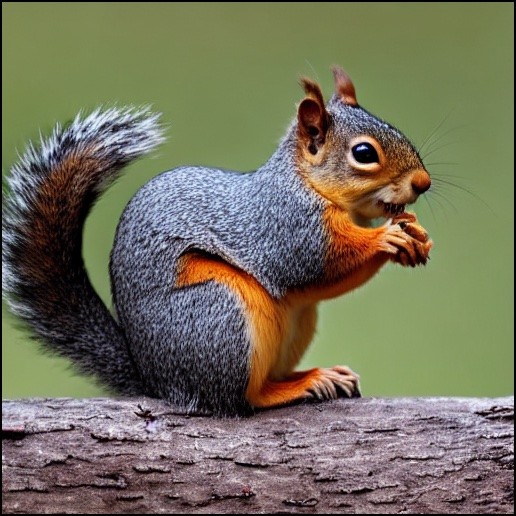} &
    \includegraphics[width=0.09\textwidth]{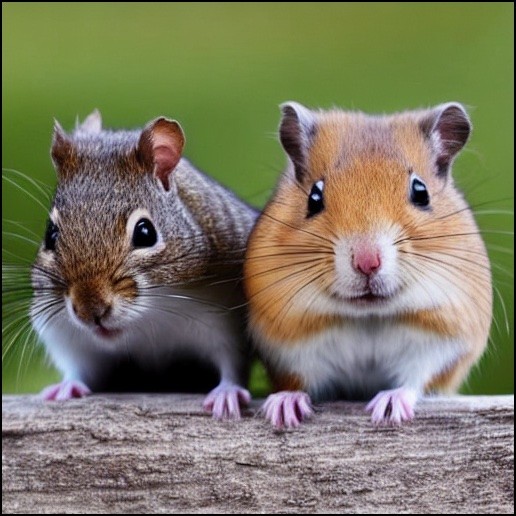} &
    \includegraphics[width=0.09\textwidth]{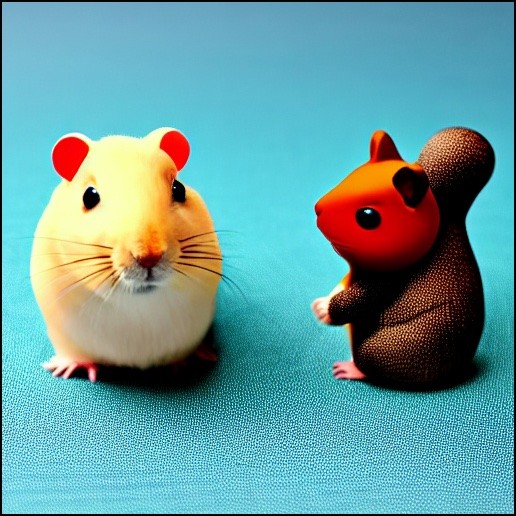} &
    \includegraphics[width=0.09\textwidth]{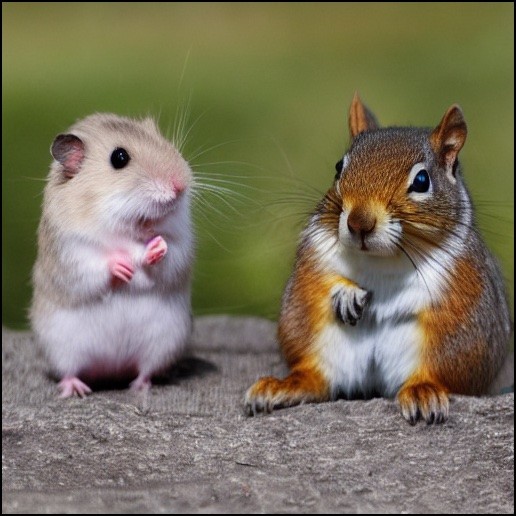} \\[-2pt]
    \multicolumn{2}{c}{\includegraphics[width=0.09\textwidth]{images/analysis/double_arrow.png}} &
    \includegraphics[width=0.09\textwidth]{images/analysis/single_arrow.png} &
    \includegraphics[width=0.09\textwidth]{images/analysis/single_arrow.png} &
    \includegraphics[width=0.09\textwidth]{images/analysis/single_arrow.png} \\[-4pt]
    \multicolumn{2}{c}{\includegraphics[width=0.09\textwidth]{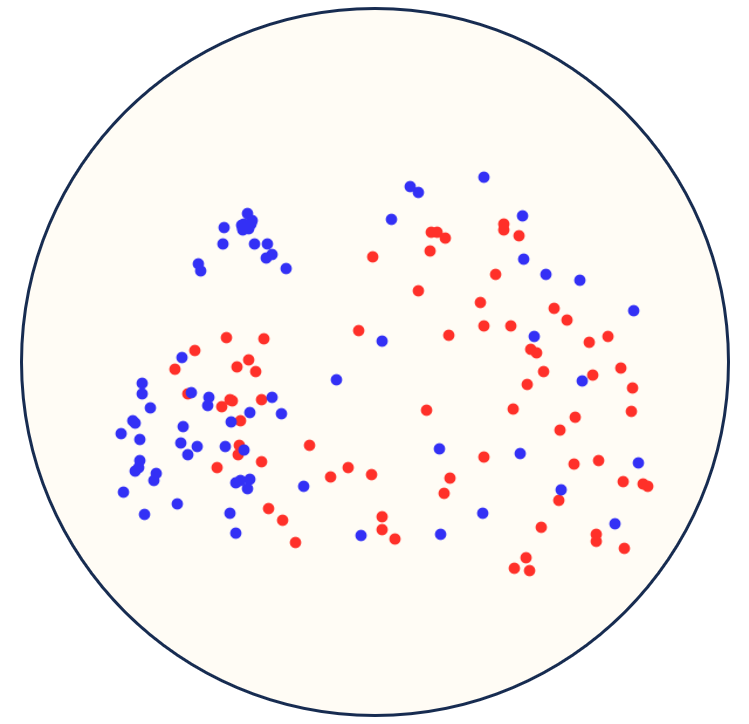}} &
    \includegraphics[width=0.09\textwidth]{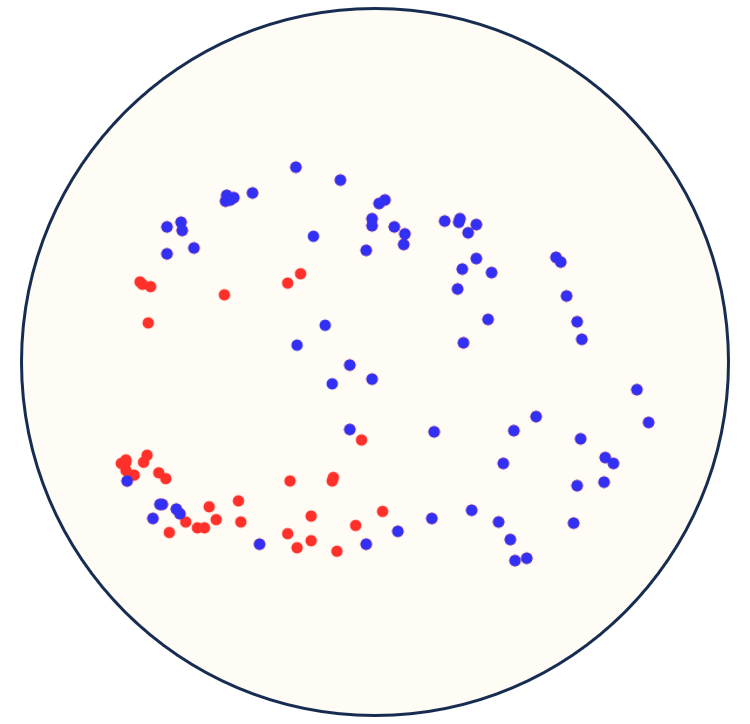} &
    \includegraphics[width=0.09\textwidth]{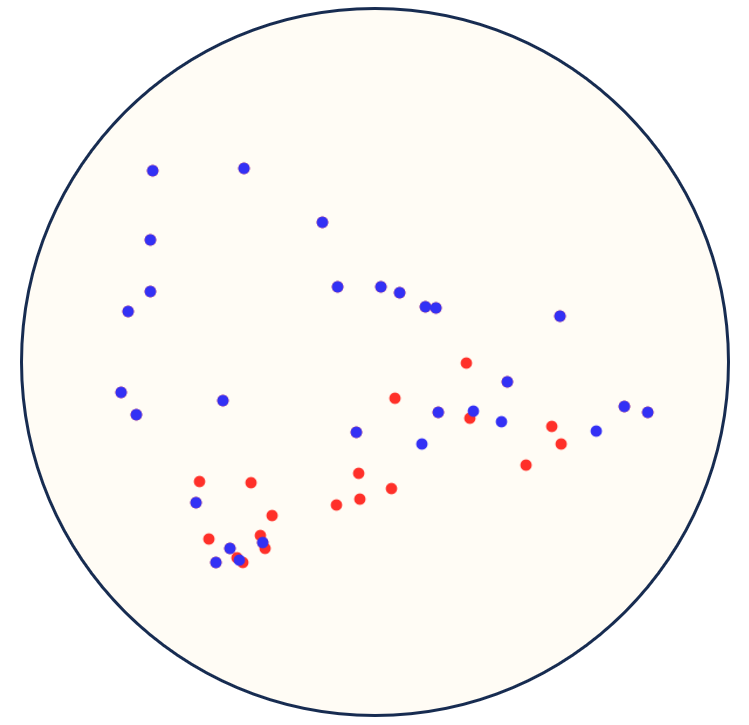} &
    \includegraphics[width=0.09\textwidth]{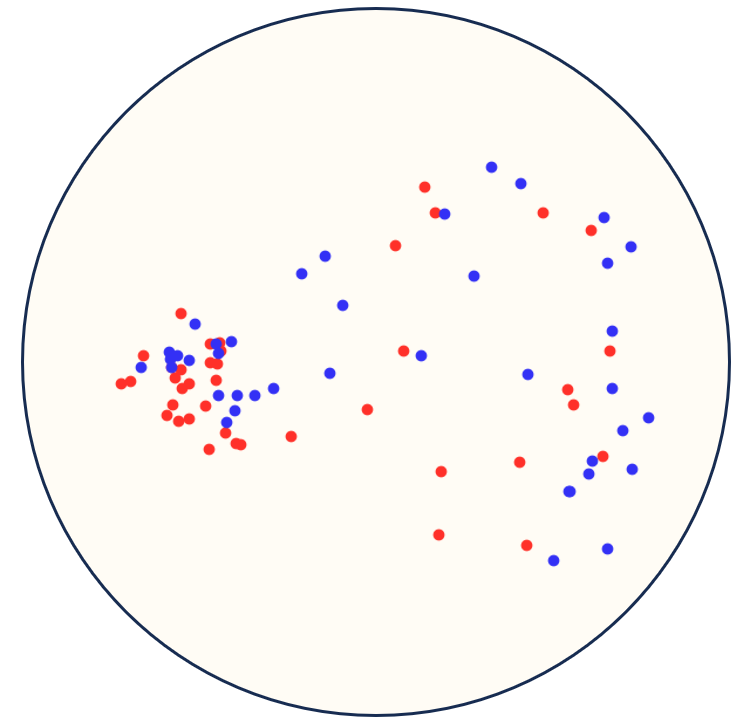} \\
    \multicolumn{2}{c}{SD} & SD & LG & BA \\[-6pt]
\end{tabular}

%% file: figures/analysis_cross/caption.tex
\textbf{Cross-Attention Leakage.} We demonstrate the emergence of semantic leakage at the cross-attention layers. We show two examples: a puppy a kitten, and a hamster and a squirrel.
In the two leftmost columns, the subjects were generated separately using Stable Diffusion (SD). In the right three columns, we generate a single image with the two subjects using three different methods: Stable Diffusion (SD), Layout Guidance (LG), and Bounded Attention (BA, ours). 
Under each row, we plot the two first principal components of the cross-attention queries. As can be seen, the separation of the queries (blue and red) reflects the leakage between the subjects in the generated images.

\vspace{-16pt}

%% file: figures/analysis_self/full_figure.tex
\begin{figure}
    \input{figures/analysis_self/figure}
    \captionof{figure}{
    \protect\input{figures/analysis_self/caption}
    }
    \label{fig:analysis_self}
\end{figure}

%% file: figures/analysis_self/figure.tex
\setlength{\tabcolsep}{0.001\textwidth}
{\scriptsize\centering
\begin{tabular}{c c c c}
    \multicolumn{2}{c}{SD} & \multicolumn{2}{c}{LG} \\
    \begin{tikzpicture}
        \node[anchor=north west,inner sep=0] at (0,0) {\includegraphics[width=0.11\textwidth]{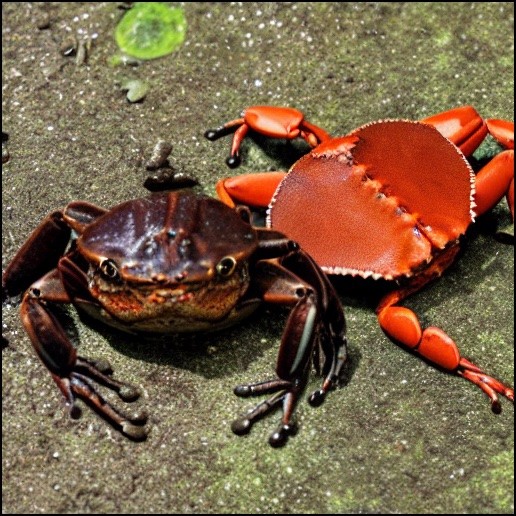}};
        \fill[fill=yellow, draw=black] (0.875,-1.04) circle (0.08);
    \end{tikzpicture} &
    \begin{tikzpicture}
        \node[anchor=north west,inner sep=0] at (0,0) {\includegraphics[width=0.11\textwidth]{images/analysis/crab_frog/sd.jpg}};
        \fill[fill=yellow, draw=black] (1,-0.44) circle (0.08);
    \end{tikzpicture} &
            \begin{tikzpicture}
        \node[anchor=north west,inner sep=0] at (0,0) {\includegraphics[width=0.11\textwidth]{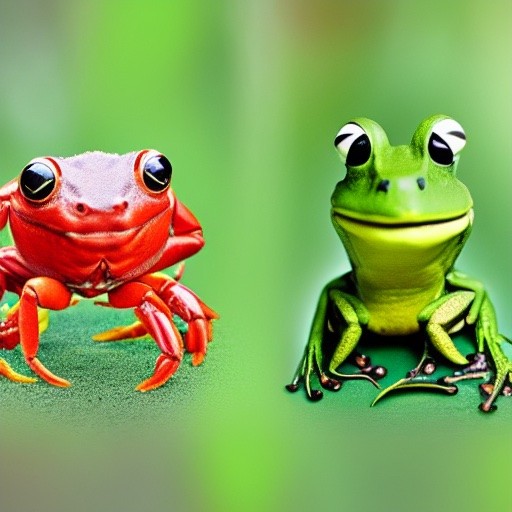}};
        \fill[fill=yellow, draw=black] (0.145,-0.723) circle (0.08);
    \end{tikzpicture} &
    \begin{tikzpicture}
        \node[anchor=north west,inner sep=0] at (0,0) {\includegraphics[width=0.11\textwidth]{images/analysis/crab_frog/guidance.jpg}};
        \fill[fill=yellow, draw=black] (1.67,-1.43) circle (0.08);
    \end{tikzpicture} \\
    \includegraphics[width=0.11\textwidth]{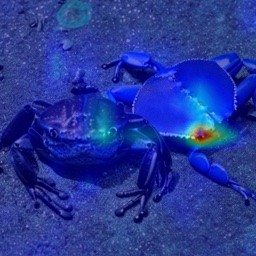} &
    \includegraphics[width=0.11\textwidth]{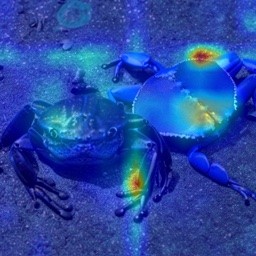} &
    \includegraphics[width=0.11\textwidth]{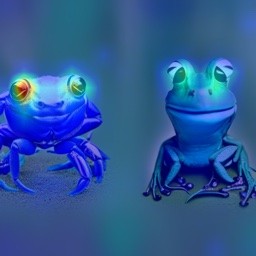} &
    \includegraphics[width=0.11\textwidth]{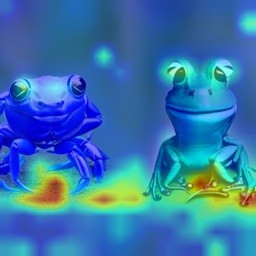} \\[-6pt]
\end{tabular}
}

%% file: figures/analysis_self/caption.tex
\textbf{Self-Attention Leakage.} We demonstrate the emergence of semantic leakage at the self-attention maps of two
subjects: a crab and a frog. The images are generated by Stable Diffusion (SD) and Layout-guidance (LG). The top row highlights specific pixels, such as those of a subject's eye or leg, while the bottom row present their respective self-attention maps. %

%% file: figures/analysis_layers/full_figure.tex
\begin{figure}
        \input{figures/analysis_layers/figure}
    \captionof{figure}{
    \protect\input{figures/analysis_layers/caption}
    }
    \label{fig:analysis_layers}
\end{figure}

%% file: figures/analysis_layers/figure.tex
\setlength{\tabcolsep}{0.001\textwidth}
\begin{tabular}{c c c c c}
    A \textcolor{red}{\textit{\underline{kitten}}} &
    A \textcolor{blue}{\textit{\underline{puppy}}} &
    \scriptsize{$64\times64$} &
    \scriptsize{$32\times32$} &
    \scriptsize{$16\times16$} \\
    \includegraphics[width=0.09\textwidth]{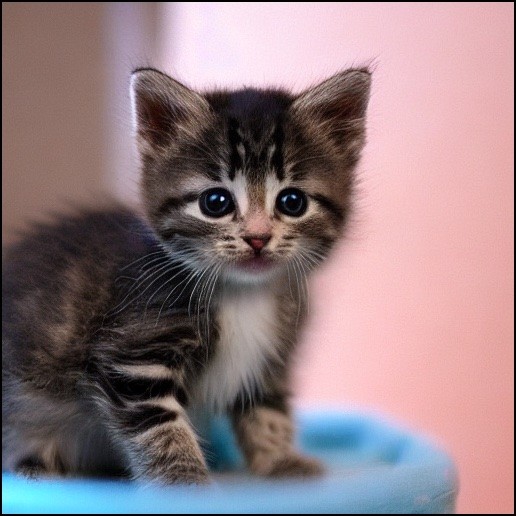} &
    \includegraphics[width=0.09\textwidth]{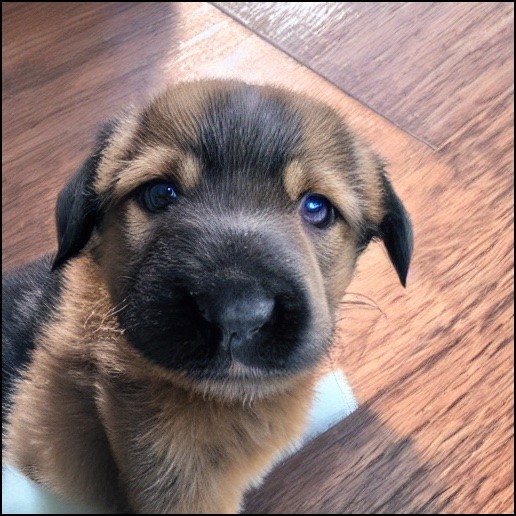} &
    \raisebox{0.5pt}{{\includegraphics[width=0.09\textwidth]{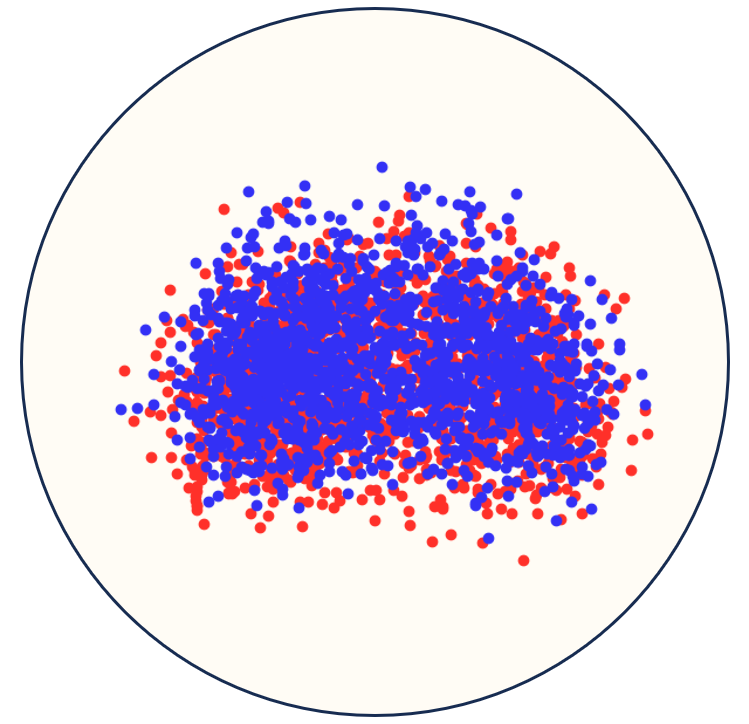}}} &
    \raisebox{0.5pt}{{\includegraphics[width=0.09\textwidth]{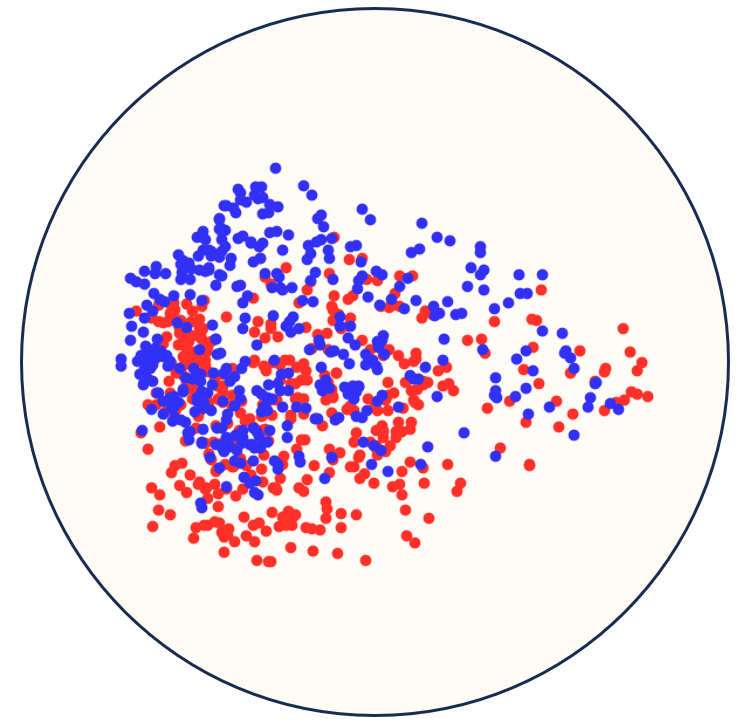}}} &
    \raisebox{0.5pt}{{\includegraphics[width=0.09\textwidth]{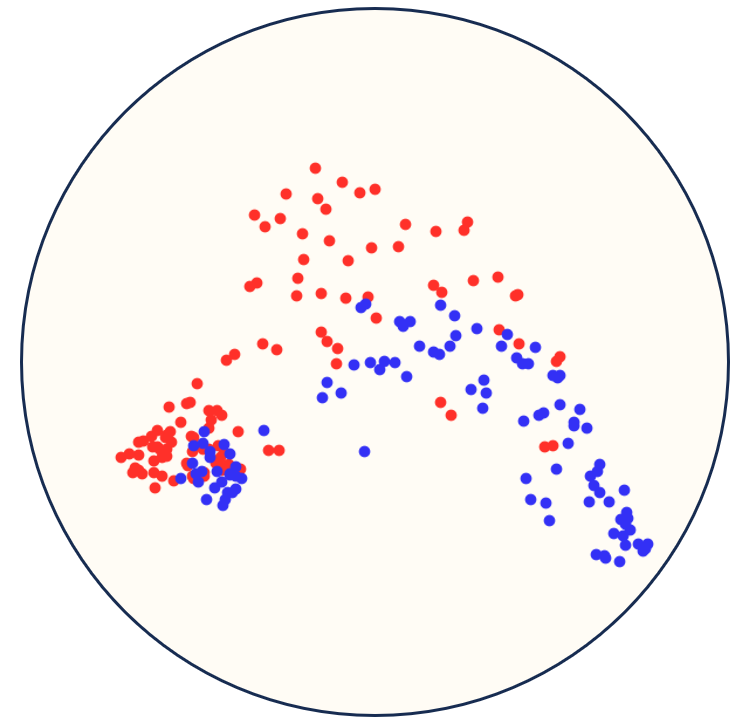}}} \\
    A \textcolor{red}{\textit{\underline{lizard}}} &
    A \textcolor{blue}{\textit{\underline{fruit}}} &
    \scriptsize{$64\times64$} &
    \scriptsize{$32\times32$} &
    \scriptsize{$16\times16$} \\
    \includegraphics[width=0.09\textwidth]{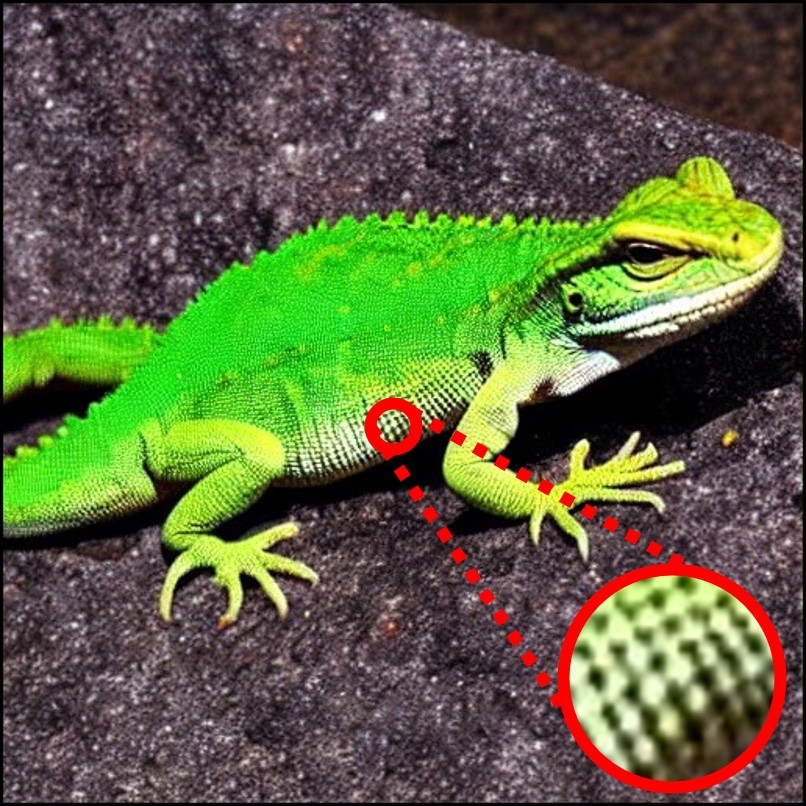} &
    \includegraphics[width=0.09\textwidth]{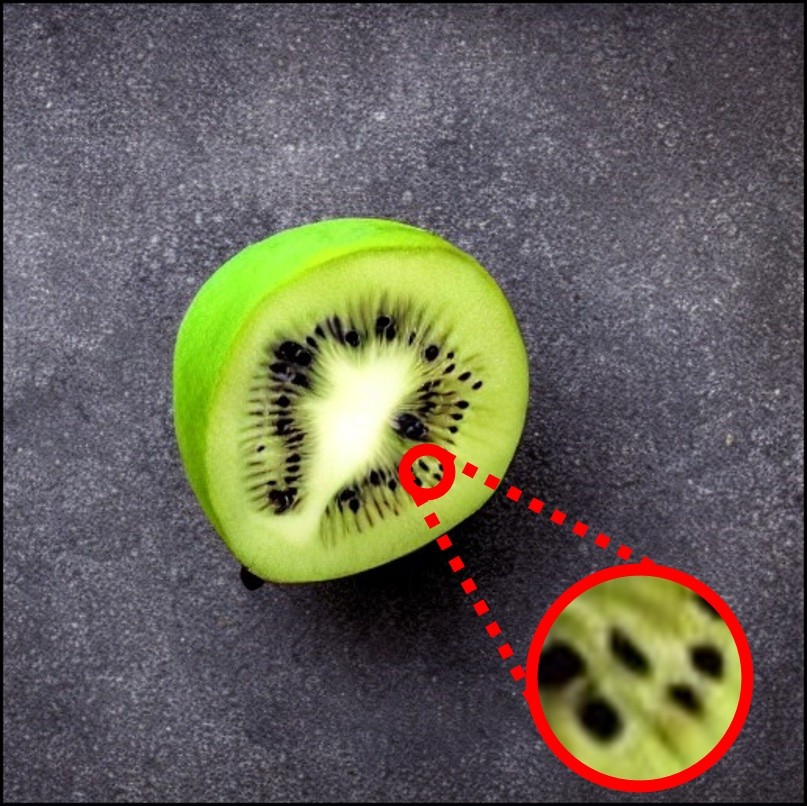} &
    \raisebox{0.5pt}{{\includegraphics[width=0.09\textwidth]{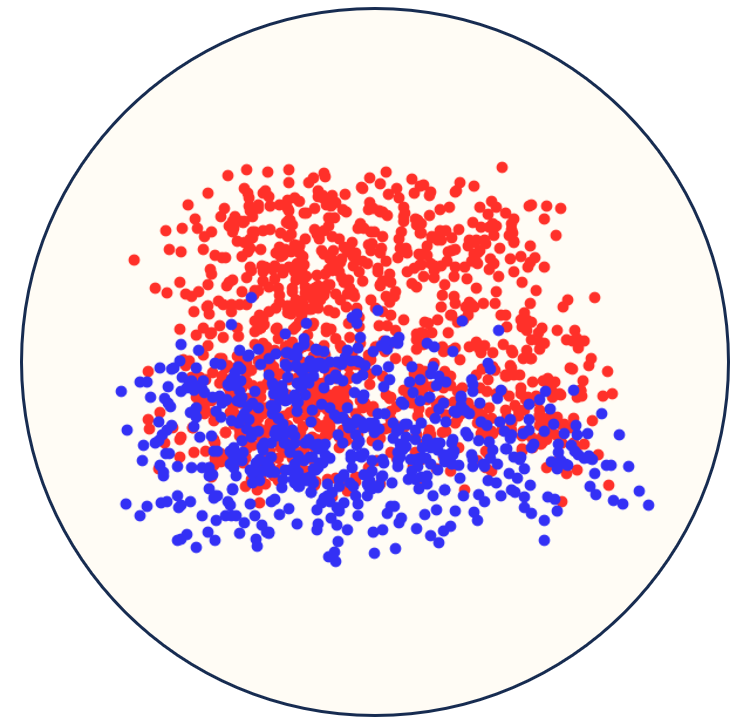}}} &
    \raisebox{0.5pt}{{\includegraphics[width=0.09\textwidth]{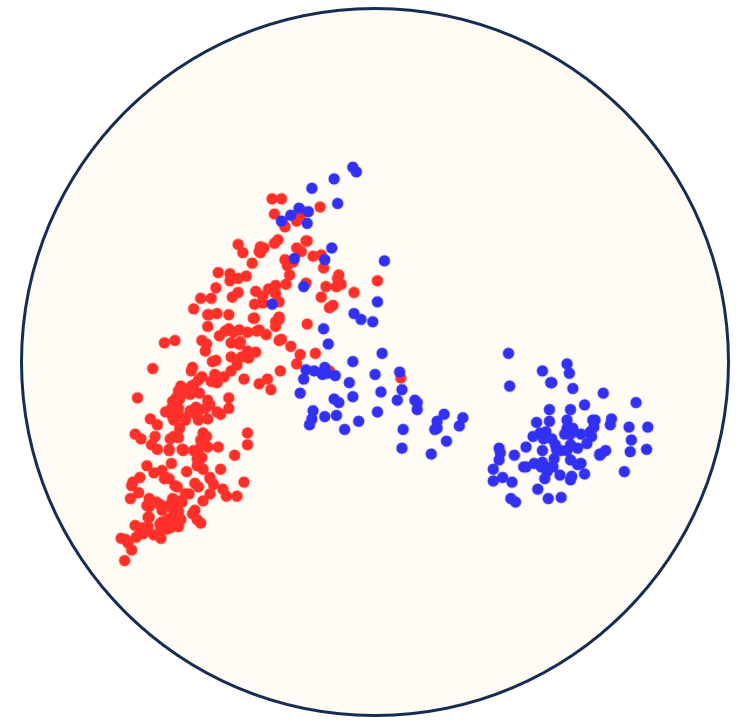}}} &
    \raisebox{0.5pt}{{\includegraphics[width=0.09\textwidth]{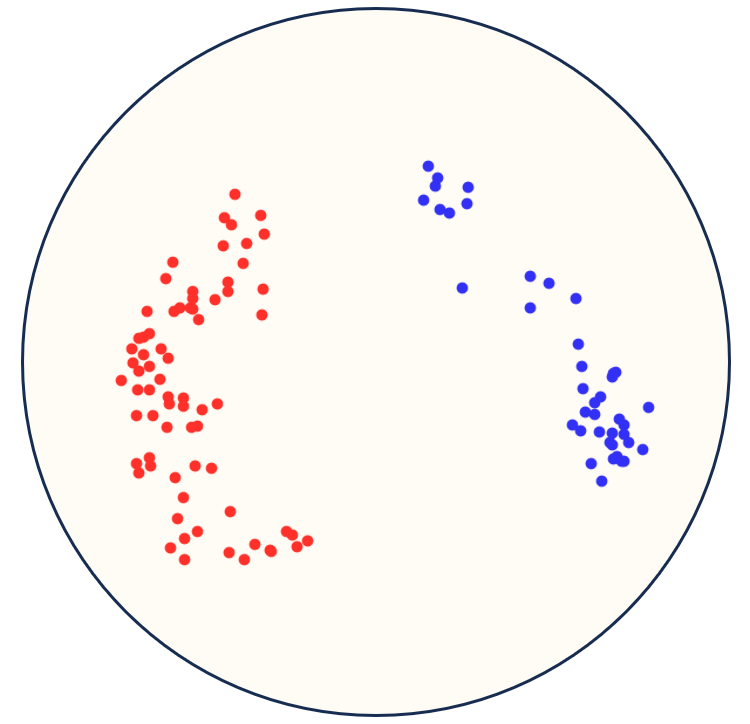}}} \\[-6pt]
\end{tabular}

%% file: figures/analysis_layers/caption.tex
We generate different subjects, and plot the first two principal components of the cross-attention queries at different layers of the UNet, where each layer is of different resolution. The high semantic similarity between the kitten and the puppy is expressed by the proximity of their queries through all layers. Meanwhile, the lizard and fruit share similar texture, and hence only their high-resolution queries are entangled.
\vspace{-12pt}

%% file: 5-method.tex
\section{Bounded Attention}

\input{figures/method_overview}

Our method takes as input $n$ distinct textual subjects $S=\left\{s_i\right\}_{i=1}^n$ contained within a global prompt $y$, along with their corresponding bounding boxes $B=\left\{b_i\right\}_{i=1}^n$.
Our objective is to condition the generation process on $y$, $S$, and $B$, while preserving the intended semantics of each subject, all without requiring any training or fine-tuning. 

Figure~\ref{fig:method-overview} illustrates an overview of our method. There, the input prompt $y$ is ``A kitten and a puppy'', $S = \{\text{``kitten''}, \text{``puppy''}\}$, and the two corresponding bounding boxes $\{b_1, b_2\}$ are illustrated in the top left corner.

Bounded Attention operates in two modes: Bounded Guidance and Bounded Denoising.
Specifically, at the beginning of the denoising process, for $t \in \left[T,T_{\textit{guidance}}\right]$, we perform a Bounded Guidance step followed by a Bounded Denoising step. In the guidance step, we utilize the Bounded Guidance loss. This interval of timesteps constitutes the optimization phase. 
Then, for $t \in \left[T_{\textit{guidance}}, 0\right]$, we apply only Bounded Denoising steps.

In both modes, we manipulate the model's forward pass by adopting an augmented weighting scheme in the attention layers, that safeguard the flow of information between the queries and the keys:

\begin{equation}
\mathbf{A}^{\left(l\right)}_t
=
\operatorname{softmax}\left(
\mathbf{Q}^{\left(l\right)}_t {\mathbf{K}^{\left(l\right)}_t}^\intercal + \mathbf{M}_t \right)
,
\label{eq:bounded_attention}
\end{equation}
where $l$ represents the layer index, $t$ represents the diffusion timestep, and $\mathbf{Q}^{\left(l\right)}_t, \mathbf{K}^{\left(l\right)}_t$ are the queries and keys of the $l$-th attention layer. 
$\mathbf{M}_t$ represents time-specific masks composed of zeros and $-\infty$ elements.
We refer to $\mathbf{A}^{\left(l\right)}_t$ as the \textit{Bounded Attention map}. 

When indexing $\mathbf{A}_t^{(l)}$, we use pixel coordinates $\mathbf{x}$ for rows, and attention-type-specific context vectors $\mathbf{c}$ for columns. In locations $\left[ \mathbf{x}, \mathbf{c} \right]$, where $\mathbf{M}_t \left[ \mathbf{x}, \mathbf{c} \right] = -\infty$, it holds that $\mathbf{A}^{\left(l\right)}_t \left[ \mathbf{x}, \mathbf{c} \right] = 0$.
Therefore, these masks prevent harmful information flow
between pixels in self-attention layers, and between pixels and token embeddings in cross-attention layers.

\subsection{Bounded Guidance}

In Bounded Guidance, we backpropagate through the diffusion model to steer the latent signal toward the desired layout, using Gradient Descent.
Our Bounded Guidance loss encourages the Bounded Attention map of each key corresponding to subject $s_i$, to be within the $b_i$ bounding box. To this end, for each subject key we consider the ratio between the attention within the corresponding bounding box, to the entire Bounded Attention map (see Figure \ref{fig:method-overview}).

Formally, we aggregate the following loss on the different subjects: 
\begin{equation} \label{eq:guidance-loss}
\mathcal{L}_i = 
1 - \frac{
\sum\limits_{\substack{\mathbf{x} \in b_i,\\ \mathbf{c} \in C_i}}
\hat{\mathbf{A}}\left[ \mathbf{x}, \mathbf{c} \right]
}{
\sum\limits_{\substack{\mathbf{x} \in b_i, \\ \mathbf{c} \in C_i}}
\hat{\mathbf{A}}\left[ \mathbf{x}, \mathbf{c} \right]
+
\alpha
\sum\limits_{\substack{\mathbf{x} \notin b_i,\\ \mathbf{c} \in C_i}}
\hat{\mathbf{A}}\left[ \mathbf{x}, \mathbf{c} \right]
}
,
\end{equation}
where $i$ denotes the index of subject $s_i$, $\hat{\mathbf{A}}$ is the mean Bounded Attention map, averaged across heads and layers, and $\alpha$ is a hyperparameter that magnifies the significance of disregarding attention towards the background, as we explain later. 
Similarly to the above, When indexing $\hat{\mathbf{A}}$, pixel coordinates $\mathbf{x}$ represent rows, and attention-type-specific context vectors $\mathbf{c}$ represent columns. We designate $C_i$ as the set of all $s_i$-related context vectors, i.e., pixel coordinates in $b_i$ for self-attention layers, and the token identifiers of $s_i$ in cross-attention layers. Additionally, for cross-attention layers, we include the first padding token [EoT] in $C_i$ to enhance layout alignment~\cite{zhao2023loco}.

For each subject $s_i$, the mask $\mathbf{M}_t$ should block the influence of opposing keys (tokens in $s_j$ and pixels in $b_j$ for $j \neq i$), to avoid the artifacts illustrated in Figure \ref{fig:analysis_cross}.
This fosters that the queries of different subjects, including similar ones, are not erroneously forced to be far apart. 
Utilizing this loss, our Bounded Guidance step is defined as
\begin{equation} \label{eq:guidance}
    z_t^\text{opt} = z_t - \beta \nabla_{z_t} \textstyle\sum_i {\mathcal{L}_i^2}.
\end{equation}

Integrating this loss within the cross-attention layers encourages the localization of each subject's semantics within its bounding boxes~\cite{chen2023training}. However, as cross-attention responses tend to peak around more typical patches associated with the subject's semantics (e.g., the face of a human, the legs of a crab, etc.), it may lack control over the subject's boundaries. By applying the loss within the self-attention layers, we encourage each subject to establish its own boundaries close to its bounding box, thereby discouraging subject fusion (see Figure \ref{fig:misalignment_examples}).

In the computation of the loss, we also introduce a hyperparameter $\alpha$ to reinforce attention to the background. This adjustment aids in preventing subject amalgamation, where a redundant subject is realized from different subject semantics in the background.

To preserve image quality, we limit the application of this mode to an initial time interval $\left[T,T_{\textit{guidance}}\right]$, following similar works \cite{chen2023training,chefer2023attend}.

\subsection{Bounded Denoising}

\input{figures/denoising_steps/full_figure}

In Bounded Denoising, we compute the model's output and use it as the next latent in the series. Here, the masks aim to reduce semantic leakage between subjects, as detailed in Section \ref{sec:analysis}, and to prevent unintended semantics from leaking to the background. Unlike Bounded Guidance and typical attention-based guidance approaches, Bounded Denoising can be applied throughout all time steps to mitigate leaks in fine details, which emerge only in later stages~\cite{meng2021sdedit}.

However, coarse masking in later stages may degrade image quality and result in noticeable stitching. To address this, after the optimization phase, for $t \in \left[T_{\textit{guidance}},0\right]$, we replace each bounding box with a fine segmentation mask obtained by clustering the self-attention
maps~\cite{patashnik2023localizing} (see Figure~\ref{fig:method-denoising-steps}).
Since the subject outlines are roughly determined in the initial time steps and evolve gradually thereafter \cite{patashnik2023localizing}, we refine the masks periodically. The resulting Bounded Attention maps of cross-attention layers are visualized in Figure~\ref{fig:cross_maps}.

Notably, this mechanism also addresses imperfect alignments between subjects and bounding boxes after the guidance phase, which are more common when generating numerous subjects. Thus, employing this technique enhances the robustness of our method to seed selection, ensuring proper semantics even when subjects extend beyond their initial confines (see Figures \ref{fig:teaser},\ref{fig:sdxl2}). Compared to methods that necessitate strict input masks yet remain susceptible to leakage \cite{bar2023multidiffusion,kim2023dense,couairon2023zero}, our method offers greater user control with simpler inputs and more satisfactory outcomes.
\input{figures/cross_attention_maps/full_figure}

\paragraph{\textbf{Method Details.}}
Further details on the adaptation of Bounded Attention to the cross- and self-attention layers, along with a description of the subject mask refinement process, are provided in 
Appendix~\ref{sec:app-method-details}.

%% file: figures/method_overview.tex
\begin{figure*}
    \centering
    \includegraphics[width=0.85\textwidth]{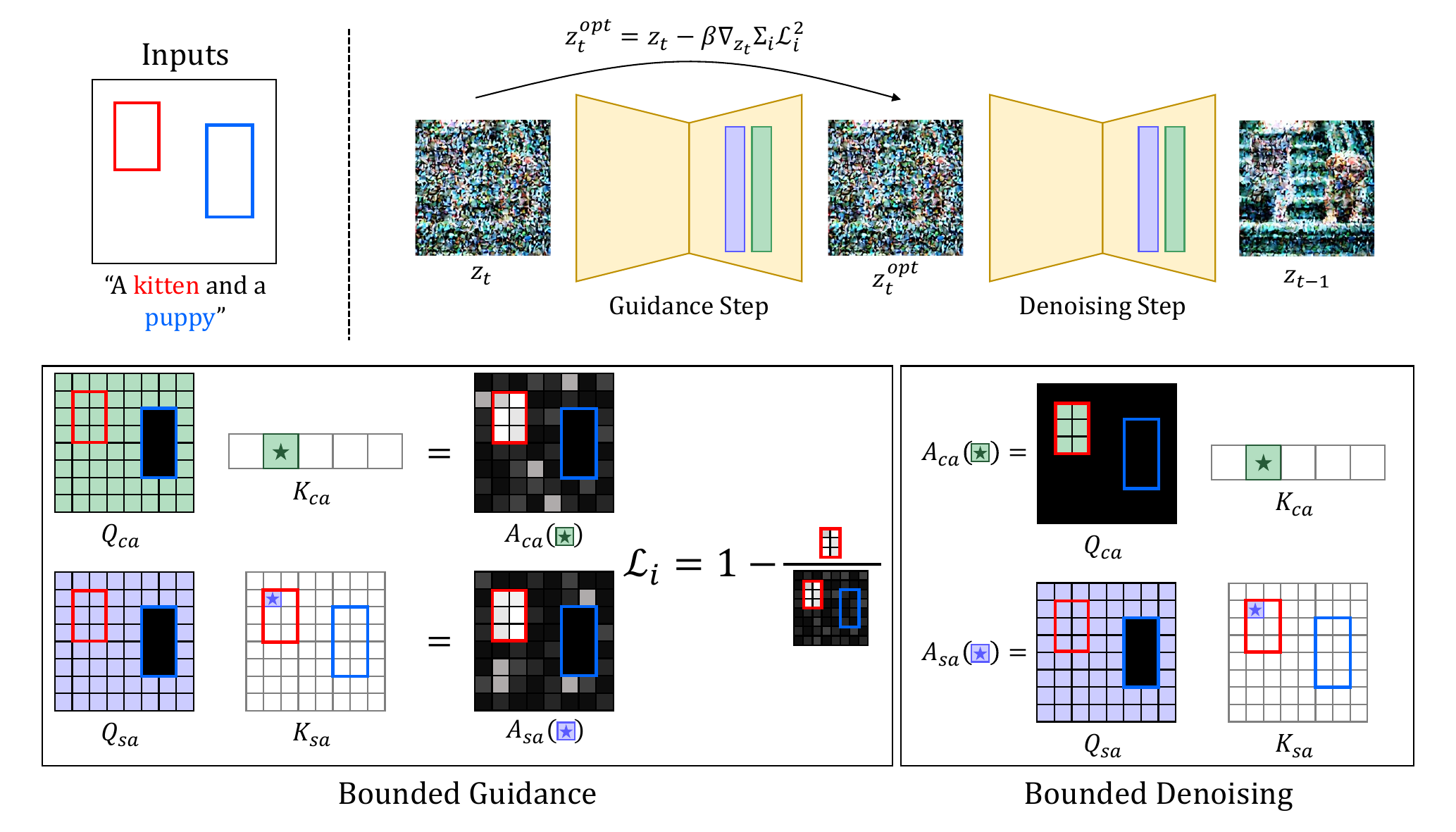}
    \vspace{-12pt}
    \caption{
    Bounded Attention operates in two modes: guidance and denoising. In each mode, strict constraints are imposed to bound the attention of each subject solely to itself and, possibly, to the background, thereby preventing any influence from other subjects' features.
    In guidance mode, we minimize a loss that encourages each subject's attention to concentrate within its corresponding bounding box. When calculating this loss, we mask the other bounding boxes, to prevent artifacts
    as shown in Figure~\ref{fig:analysis_cross}. 
    We simplify the visualization of the loss by setting $\alpha = 1$ in Equation~\ref{eq:guidance-loss}.
    During the denoising step, we confine the attention of each subject solely to its bounding box, along with the background in the self-attention. This strategy effectively prevents feature leakage while maintaining the natural immersion of the subject within the image.
    To demonstrate each of these modes, we show the attention map of a specific key, marked with $\star$. 
    For the cross-attention map, we show the key corresponding to the ``kitten'' token, and for the self-attention map, we show a key that lies in the kitten's target bounding box. 
    }
    \vspace{-10pt}
    \label{fig:method-overview}
\end{figure*}

%% file: figures/denoising_steps/full_figure.tex
\begin{figure}
    \input{figures/denoising_steps/figure}
    \captionof{figure}{
    \protect\input{figures/denoising_steps/caption}
    }
    \label{fig:method-denoising-steps}
\end{figure}

%% file: figures/denoising_steps/figure.tex
\centering
\includegraphics[width=1.0\linewidth]{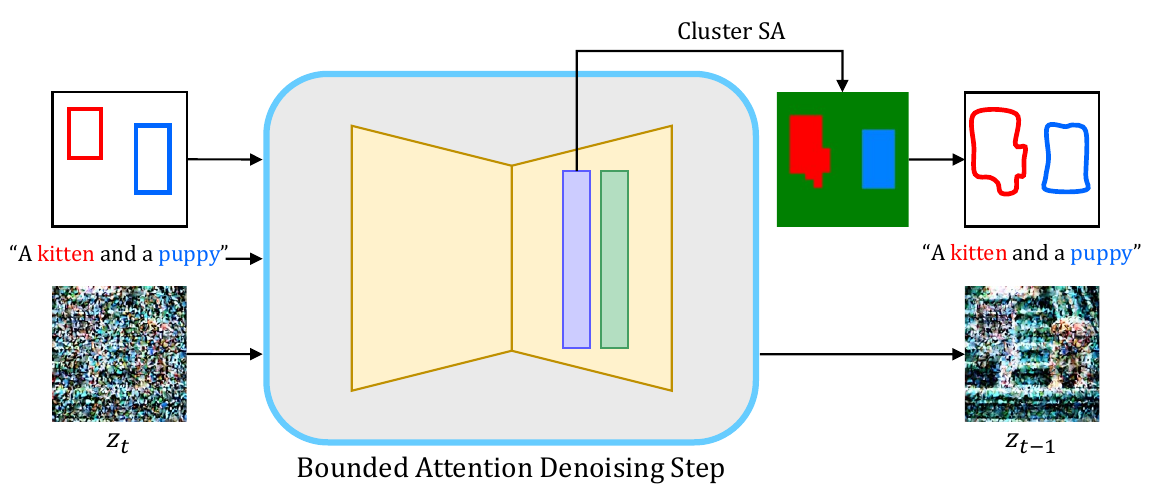}
\vspace{-8pt}

%% file: figures/denoising_steps/caption.tex
In the first phase, the coarse shapes of the subjects are formed. Then, we move to the next phase, illustrated here. During this phase, we apply Bounded Denoising steps using the fine-grained subject masks. At regular intervals, we refine these masks by clustering the self-attention (SA) maps.
\vspace{-8pt}

%% file: figures/cross_attention_maps/full_figure.tex
\begin{figure}
    \input{figures/cross_attention_maps/figure}
    \captionof{figure}{
    \protect\input{figures/cross_attention_maps/caption}
    }
    \label{fig:cross_maps}
\end{figure}

%% file: figures/cross_attention_maps/figure.tex
\begin{center}
\setlength{\tabcolsep}{0.003\textwidth}
{\scriptsize\centering
\begin{tabular}{c c c c}
    \multicolumn{2}{c}{Generated Image} &
    \multicolumn{2}{c}{Bounded Self-Attention Map} \\
    \multicolumn{2}{c}{\includegraphics[width=0.2\textwidth]{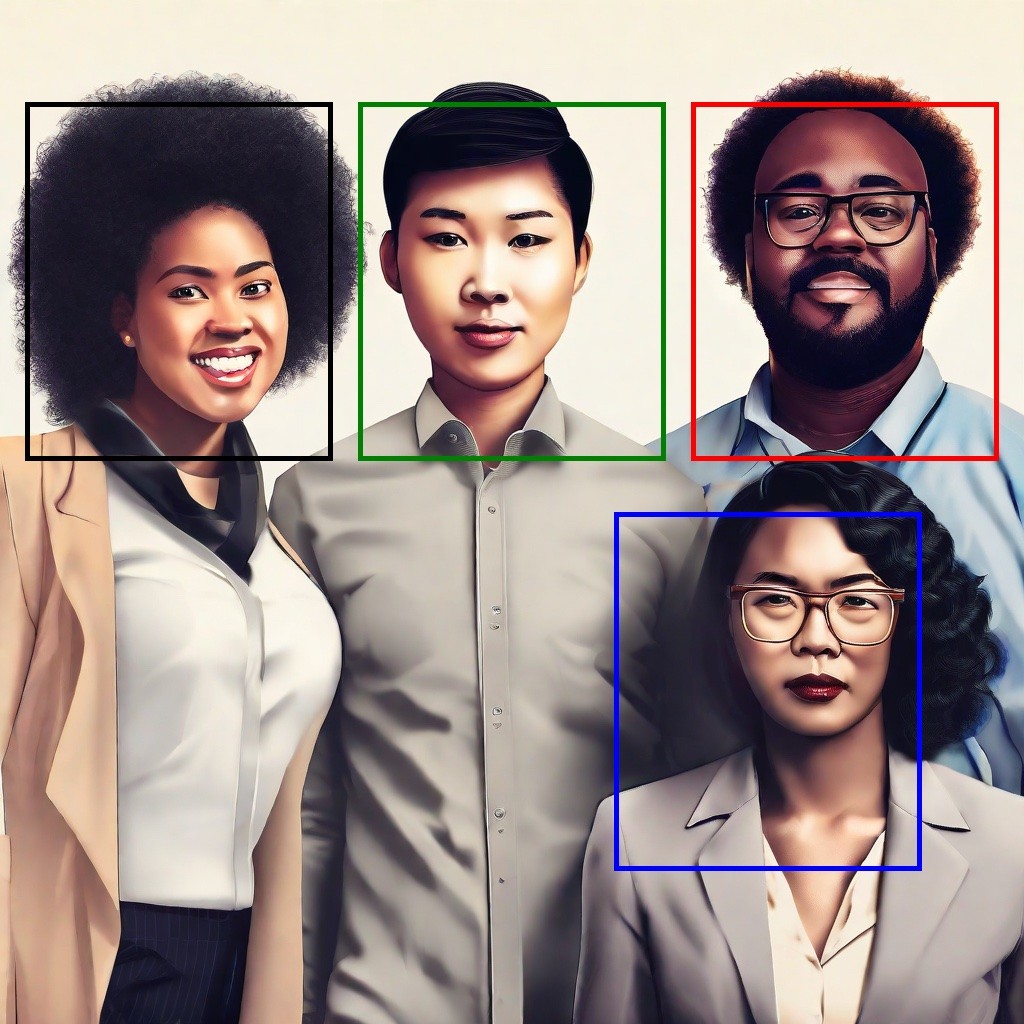}} &
    \multicolumn{2}{c}{\includegraphics[width=0.2\textwidth]
    {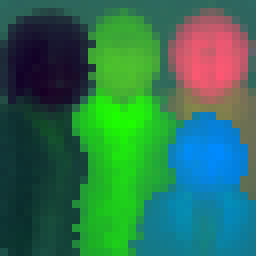}} \\
    \\
    \multicolumn{4}{c}{Bounded Cross-Attention Maps} \\
    \hspace*{2pt}
    \includegraphics[width=0.08\textwidth]{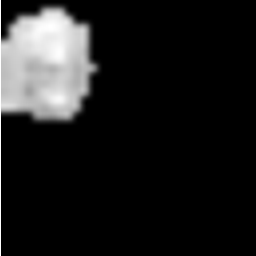} &
    \includegraphics[width=0.08\textwidth]
    {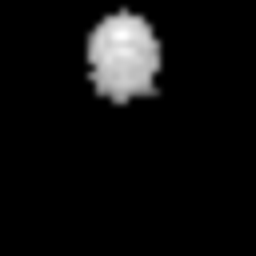} &
    \includegraphics[width=0.08\textwidth]
    {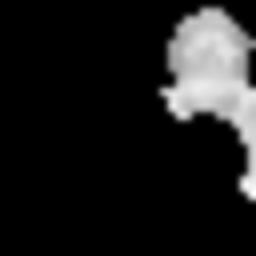} &
    \includegraphics[width=0.08\textwidth]
    {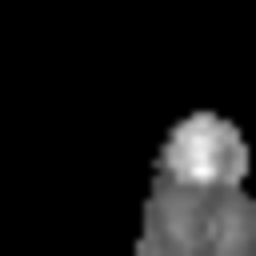} \\
    \color{black}{black} & \color{green}{asian} & \color{red}{black} & \color{blue}{asian} \\
    \\
    \hspace*{2pt}
    \includegraphics[width=0.08\textwidth]
    {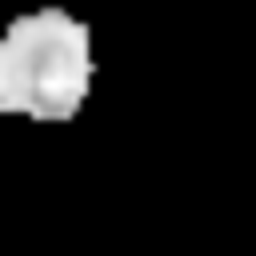} &
    \includegraphics[width=0.08\textwidth]{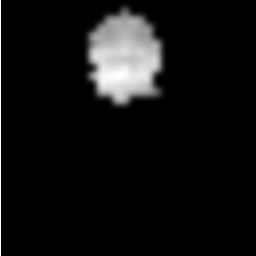} &
    \includegraphics[width=0.08\textwidth]{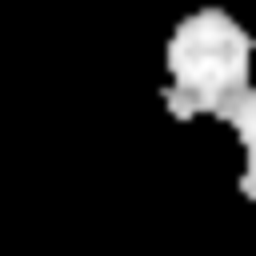} &
    \includegraphics[width=0.08\textwidth]{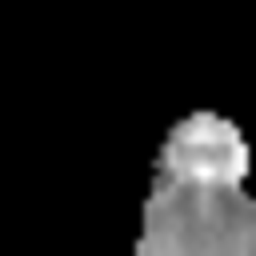} \\
    \color{black}{woman} & \color{green}{man} & \color{red}{man} & \color{blue}{woman} \\[-12pt]
\end{tabular}
}
\end{center}

%% file: figures/cross_attention_maps/caption.tex
By segmenting the average self-attention maps from the UNet's bottleneck (top-right corner), we can refine the input bounding boxes into dense segmentation masks, even amidst high-magnitude noise. Bounded Attention utilizes these masks to align the evolving image structure (represented by the self-attention maps) with its semantics (reflected by the cross-attention maps), ensuring each subject retains its distinct characteristics during denoising.
\vspace{-12pt}

%% file: 6-experiments.tex
\section{Experiments}
\input{figures/sdxl_results/full_figure}

In this section, we conduct both qualitative and quantitative experiments to assess the effectiveness of our Bounded Attention method. We compare our approach with three training-free baseline methods: Layout-guidance (LG)~\cite{chen2023training}, BoxDiff (BD)~\cite{xie2023boxdiff}, and MultiDiffusion (MD)~\cite{bar2023multidiffusion}. Additionally, we include comparisons with GLIGEN~\cite{li2023gligen} and ReCo~\cite{yang2023reco}, which necessitate training. Since Attention-refocusing (AR)~\cite{phung2023grounded} is based on GLIGEN, we categorize it as a trained method for the purpose of our evaluation.
For fairness, when comparing our method with other methods, we use Stable Diffusion.

\input{figures/sdxl_results2}

\subsection{Qualitative Results}

\paragraph{\textbf{SDXL results.}}
We begin our experiments by showcasing the effectiveness of our method in challenging scenarios, particularly when tasked with generating multiple semantically similar subjects using SDXL.

In Figure~\ref{fig:sdxl}, we demonstrate that Bounded Attention is capable of generating subjects with complex positional relations and occlusions, such as stacked cake layers, as well as visually similar subjects that naturally blend into the background, like dogs of various breeds partially submerged in a pool. Our approach produces all subjects with their distinctive characteristics, even when altering the prompt, seed, or bounding box assignment.

Moving to Figure~\ref{fig:sdxl2}, we generate multiple semantically similar subjects with different modifiers across various seeds. It is evident that Vanilla SDXL fails to follow the prompts due to semantic leakage. In the first row, it inaccurately generates the number of dogs and kittens and mixes between their colors. In the middle row, the clothing compositions combine fabrics, silhouettes, and colors mentioned in the prompt. Finally, in the last row, it merges the appearance of the subjects while leaking the pink attribute into the background.

We provide more results in Appendix \ref{sec:more-results}.

\input{figures/seed0/full_figure}

\paragraph{\textbf{Non-curated results.}}

Next, we conduct a non-curated comparison with the training-free baseline methods and present the results in Figure~\ref{fig:seed0}. We showcase the initial six images sampled from seed 0 for each method. 
We anticipate that in the absence of Bounded Attention,
semantic leakage may freely blend subject features, hindering the intended layout's formation.

It is evident from the results that none of the competing methods is able to consistently construct the input layout. Layout Guidance~\cite{chen2023training} frequently neglects one of the subjects, and even when it generates three subjects, it struggles to avoid leakage, resulting in puppies with kitten-like features or incorrect color assignments. BoxDiff~\cite{xie2023boxdiff} often generates the correct number of subjects but suffers from artifacts in the form of blobs. Similar to Layout Guidance, it encounters difficulties in properly constructing the puppy. Surprisingly, even MultiDiffusion~\cite{bar2023multidiffusion}, which generates the subjects separately, faces challenges in generating them all, with some disappearing or merging together in its bootstrapping phase.

In contrast, our method consistently outperforms these approaches, producing three subjects that align with the both prompt and layout in all six images.

\paragraph{\textbf{Comparisons with baselines.}}

We present a qualitative comparison in Figure~\ref{fig:comparisons}.
All competing methods, including those trained specifically for the layout-to-image task, exhibit significant visual and semantic leakage. The training-free methods perform the worst: MultiDiffusion produces disharmonious, low-quality images, while optimization-based methods often result in object fusion, combining different semantics without adhering to the layout.

The training-based approaches closely follow the layout but fail to convery the correct semantics. In the first row, they neglect the corduroy jacket, leaking the denim texture into the other jacket, or even fusing them together.
In the other rows, the elephant's or penguin's features leak into the rhino or bear, respectively. Moreover, due to the rigidity of these approaches, stemming from being trained on perfect bounding boxes, they are unable to depict the penguin riding the bear.

In comparison, our method generates images that align with the input layout and prompt, ensuring each subject retains its unique attributes, semantics, and appearance.

\input{figures/comparison}

\subsection{Quantitative Results}

\paragraph{\textbf{Dataset evaluation.}} 

We evaluate our method's effectiveness using the DrawBench dataset~\cite{saharia2022photorealistic}, known for its challenging prompts designed to test a model's ability to compose multiple subjects with specific quantities and relations. We use the evaluation procedure from previous work~\cite{wu2019detectron2,phung2023grounded}.

Our results, alongside those of other training-free methods, are summarized in Table \ref{table:drawbench}. Unlike other approaches that do not account for semantic leakage, our method demonstrates notable improvements in the counting category. While other methods struggle to surpass the recall rates of vanilla SD, Bounded Attention enhances recall by 0.1, representing a noteworthy advancement. Moreover, it improves counting precision and spatial accuracy, highlighting the effectiveness of Bounded Attention in addressing semantic misalignments.

\input{tables/drawbench}
\input{tables/user_study}

\input{sub_sections/user_study}

\subsection{Ablation Studies}

\input{figures/ablation}

To assess the significance of each component, we conduct an ablation study where we systematically vary our method's configurations by omitting one component in each setting.
    Two examples, generated with SDXL (top row) and SD (bottom row), are illustrated in Figure~\ref{fig:ablation}.

Guidance is crucial for aligning the latent signal with the intended layout. However, attempting to guide the latent signal without our Bounded Guidance mechanism leads to subpar results, as seen in the partial alignment of the lizard with its bounding
    box, and the puppy's distorted form. The issue arises from the inherent query entanglement between the two semantically similar subjects in each example. Without Bounded Guidance, the optimization in the top row reaches a plateau, where aligning the lizard with its bounding box reduces its loss but also increases the turtle's. In the bottom row, the optimization pushes the two subject queries away from each other, creating artifacts.

Meanwhile, forgoing Bounded Denoising results in noticeable semantic
leakage. In the top example, the lizard is replaced by a turtle, with the ``red'' attribute erroneously leaking to the wrong subject. Similarly, in the bottom example, the puppy is replaced with a kitten.

Lastly, incorporating mask refinement in the later stages
    preserves fine details and prevents them from leaking. Without mask refinement, the kitten's legs lose the details of their ginger fur texture, the turtle's facial features resemble those of a lizard, and the lizard exhibits shell-like contours on its back.

%% file: figures/sdxl_results/full_figure.tex
\begin{figure*}[t]
\centering

\input{figures/sdxl_results/figure}
    \captionof{figure}{
        \protect\input{figures/sdxl_results/caption}
    }
    \label{fig:sdxl}
\end{figure*}

%% file: figures/sdxl_results/figure.tex
\setlength{\tabcolsep}{0.001\textwidth}
{\centering\small
\begin{tabular}{c c c c c c}
    \multicolumn{6}{c}{``a \textcolor{red}{\textit{\underline{[base]}}} with \textcolor{blue}{\textit{\underline{[cake]}}} and \textcolor{green}{\textit{\underline{[icing]}}} and \textcolor{orange}{\textit{\underline{[toppings]}}} on a table''} \\
    \includegraphics[width=0.16\textwidth]{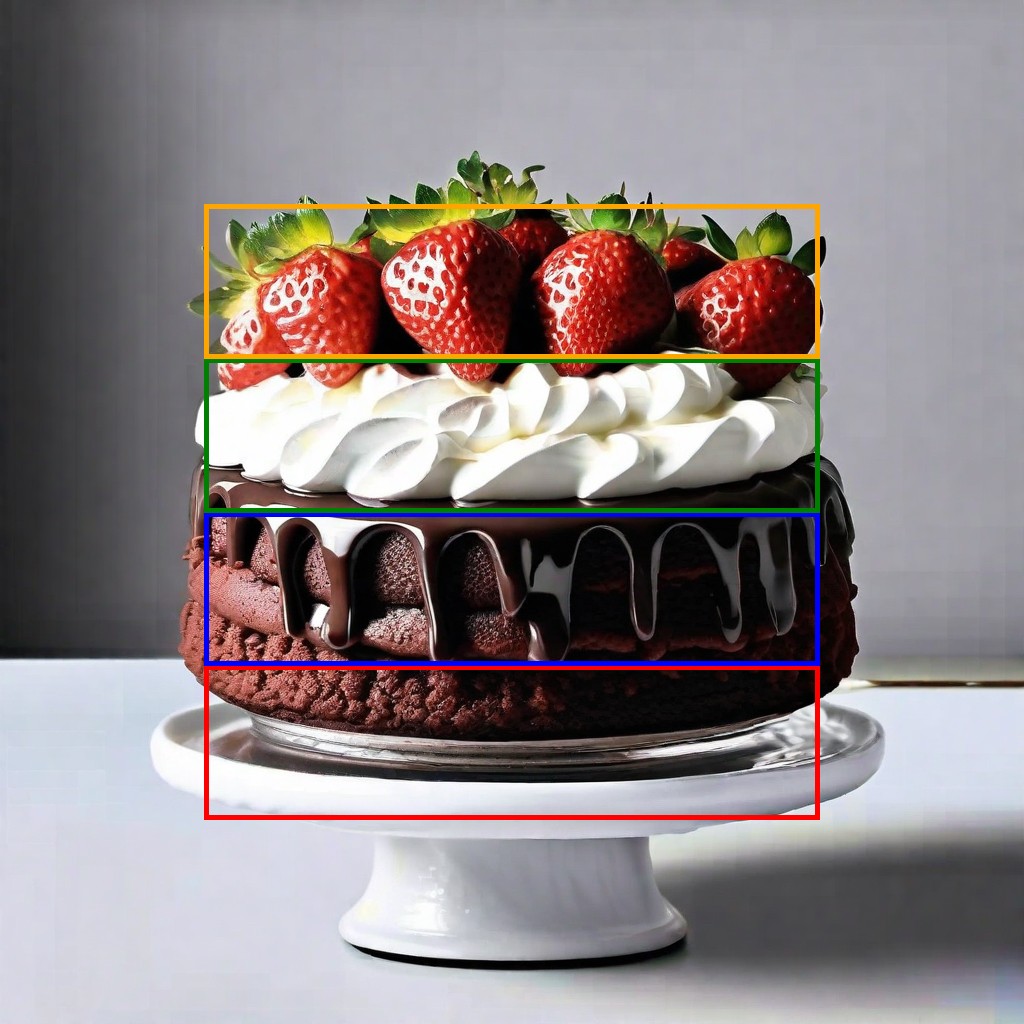} &
    \includegraphics[width=0.16\textwidth]
    {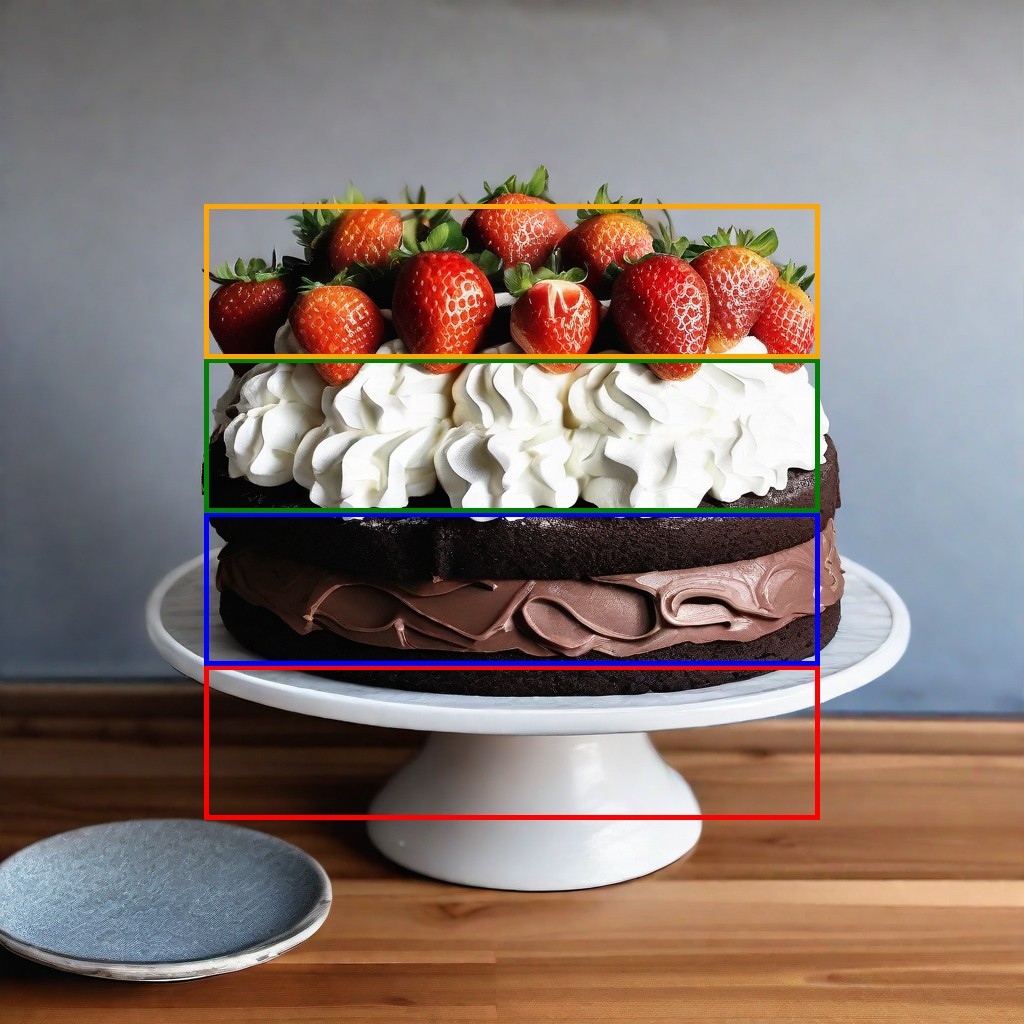} &
    \includegraphics[width=0.16\textwidth]{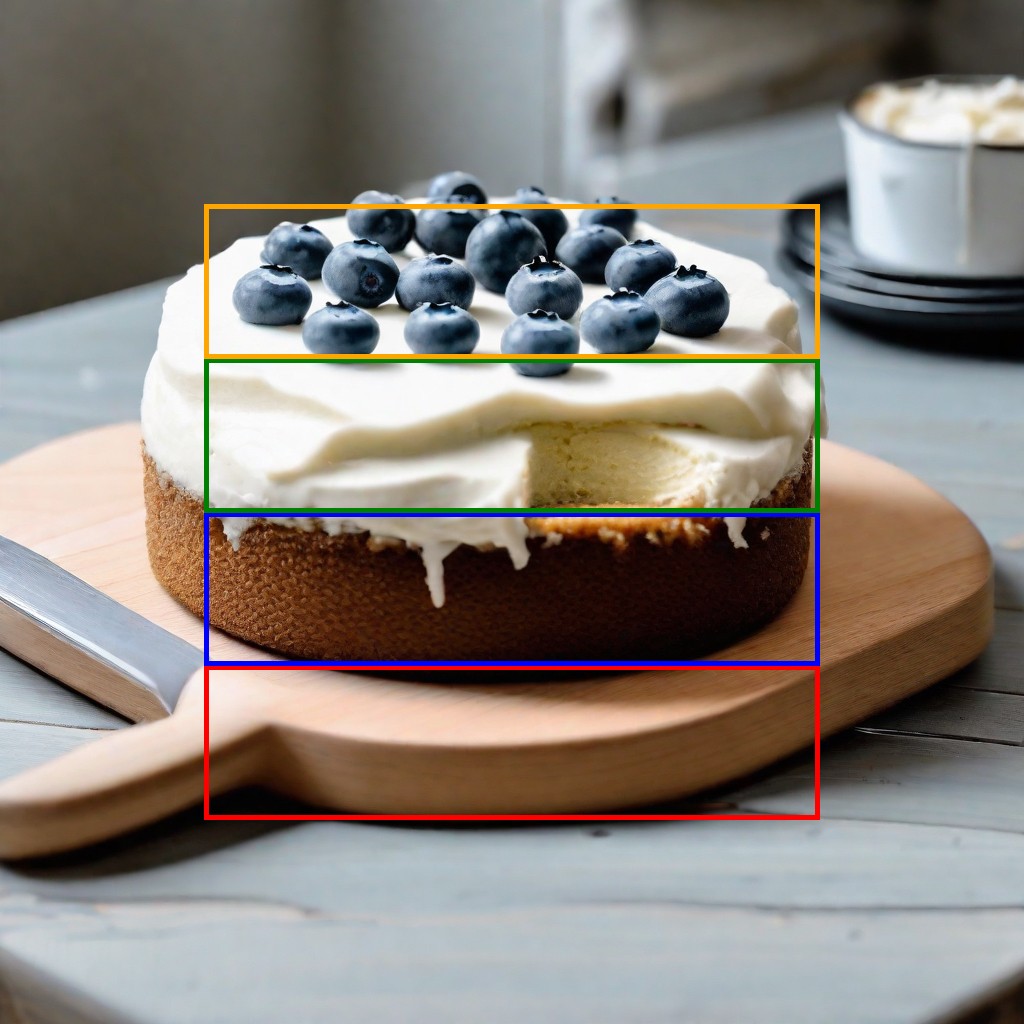} &
    \includegraphics[width=0.16\textwidth]
    {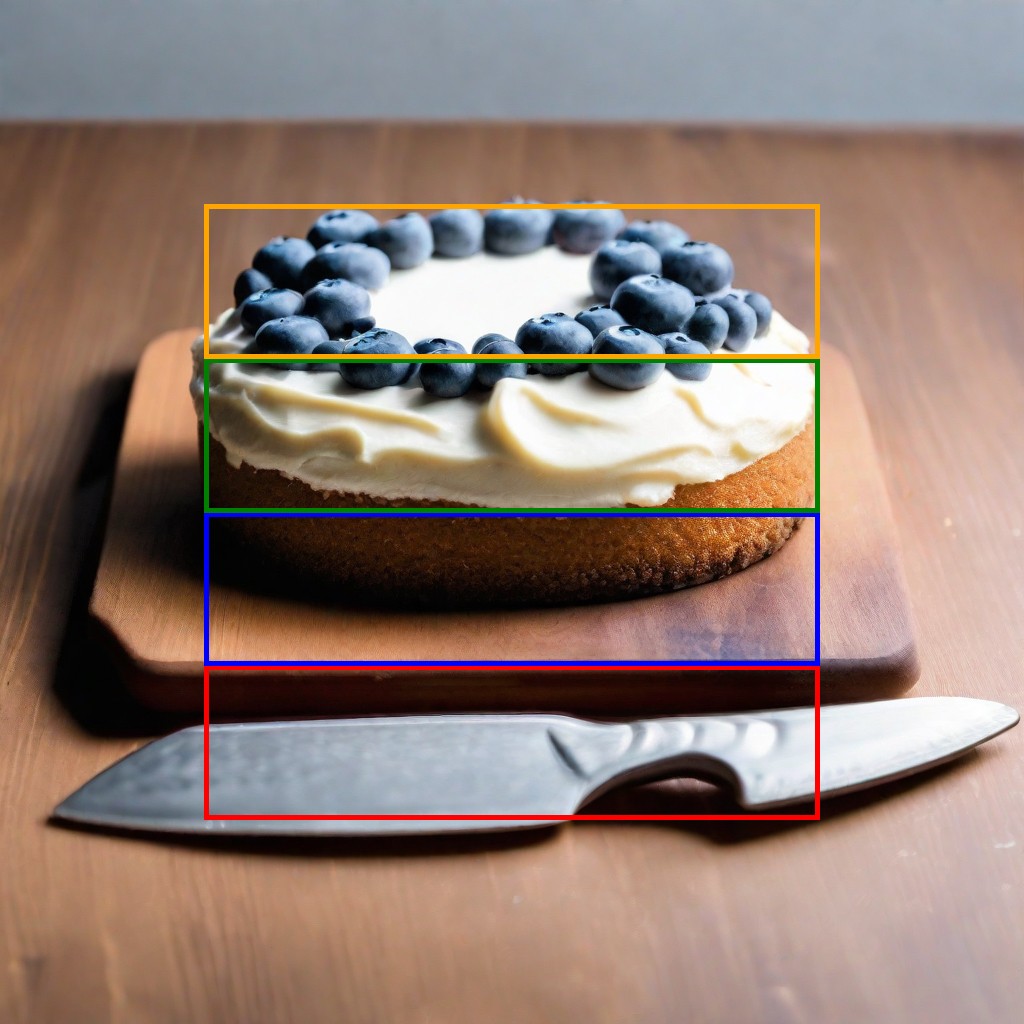} &
    \includegraphics[width=0.16\textwidth]{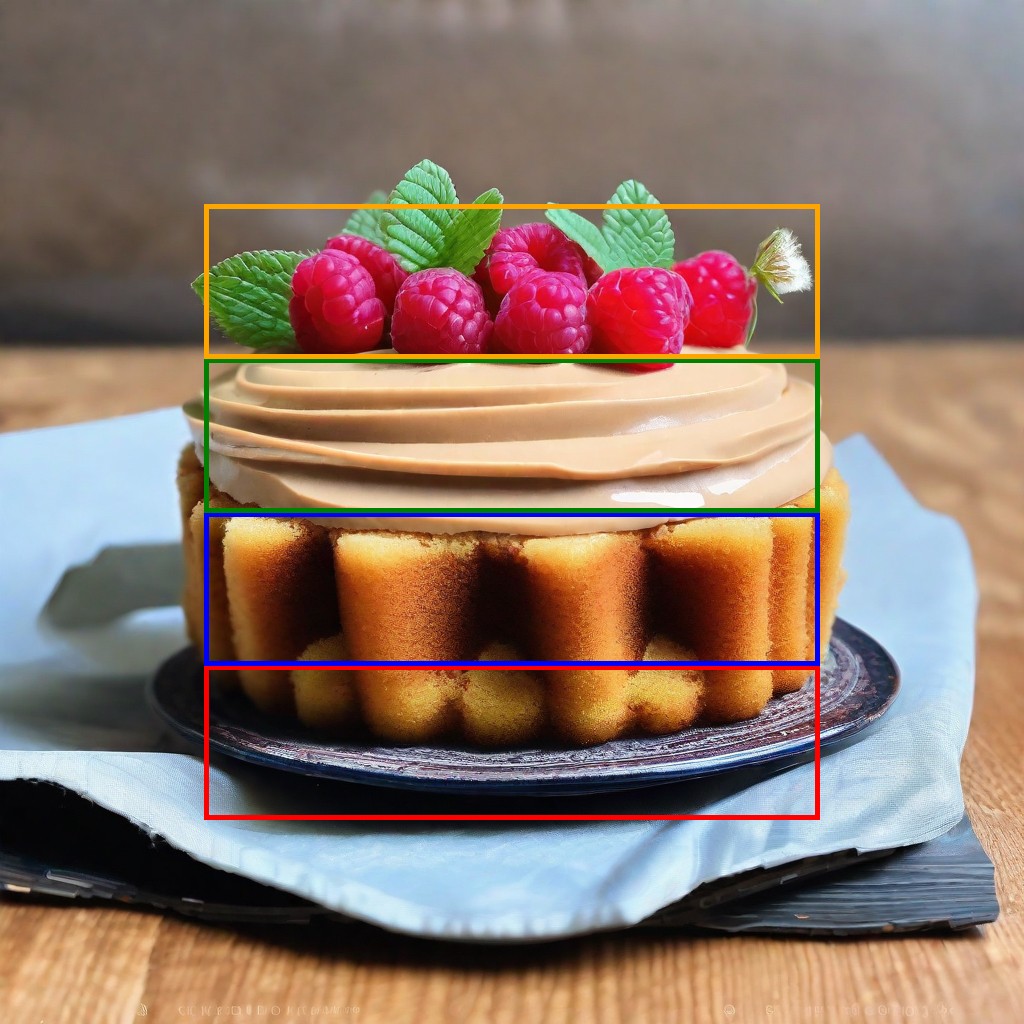} &
    \includegraphics[width=0.16\textwidth]
    {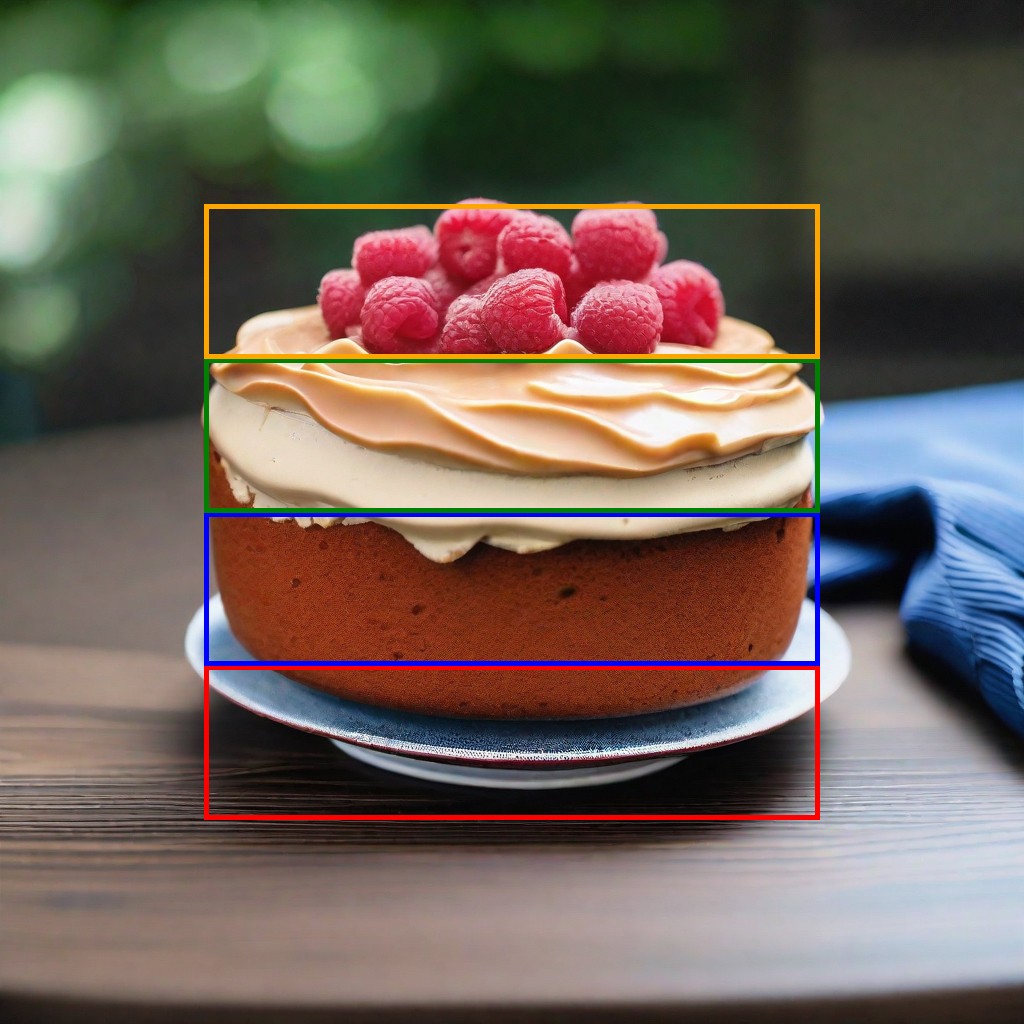} \\
    \multicolumn{2}{c}{\scriptsize``\textcolor{red}{\textit{\underline{cake stand}}}'', ``\textcolor{blue}{\textit{\underline{chocolate cake}}}'',} & \multicolumn{2}{c}{\scriptsize``\textcolor{red}{\textit{\underline{cutting board}}}'', ``\textcolor{blue}{\textit{\underline{cheese cake}}}'',} &
    \multicolumn{2}{c}{\scriptsize``\textcolor{red}{\textit{\underline{ceramic plate}}}'', ``\textcolor{blue}{\textit{\underline{sponge cake}}}'',} \\
    \multicolumn{2}{c}{\scriptsize``\textcolor{green}{\textit{\underline{whipped cream}}}'', ``\textcolor{orange}{\textit{\underline{strawberries}}}''} & \multicolumn{2}{c}{\scriptsize``\textcolor{green}{\textit{\underline{vanilla frosting}}}'', ``\textcolor{orange}{\textit{\underline{blueberries}}}''} & \multicolumn{2}{c}{\scriptsize``\textcolor{green}{\textit{\underline{caramel frosting}}}'', ``\textcolor{orange}{\textit{\underline{raspberries}}}''} \\
    \\
   \multicolumn{6}{c}{``a \textcolor{red}{\textit{\underline{golden retriever}}} and a \textcolor{blue}{\textit{\underline{german shepherd}}} and a \textcolor{green}{\textit{\underline{boston terrier }}} and an \textcolor{orange}{\textit{\underline{english bulldog}}} and a \textcolor{purple}{\textit{\underline{border collie}}} in a pool''} \\
    \includegraphics[width=0.16\textwidth]
    {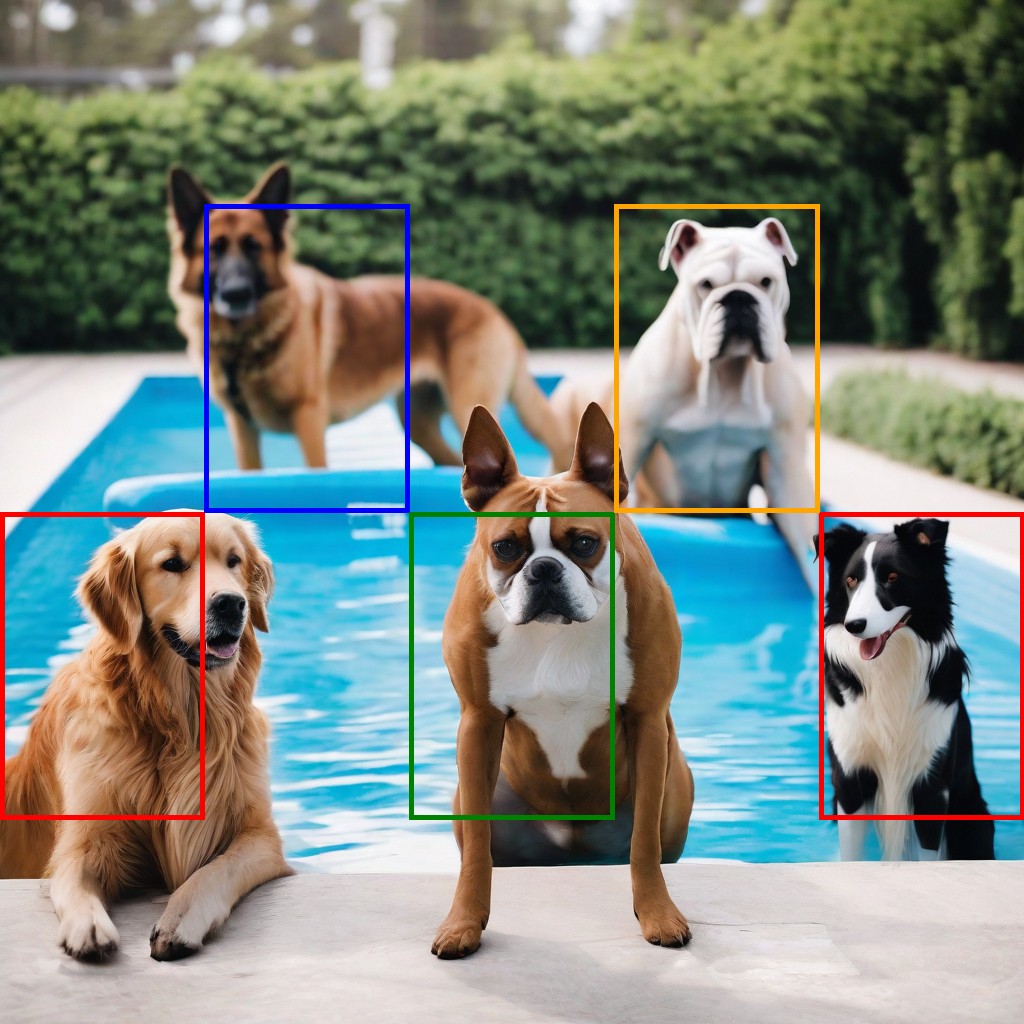} &
    \includegraphics[width=0.16\textwidth]
    {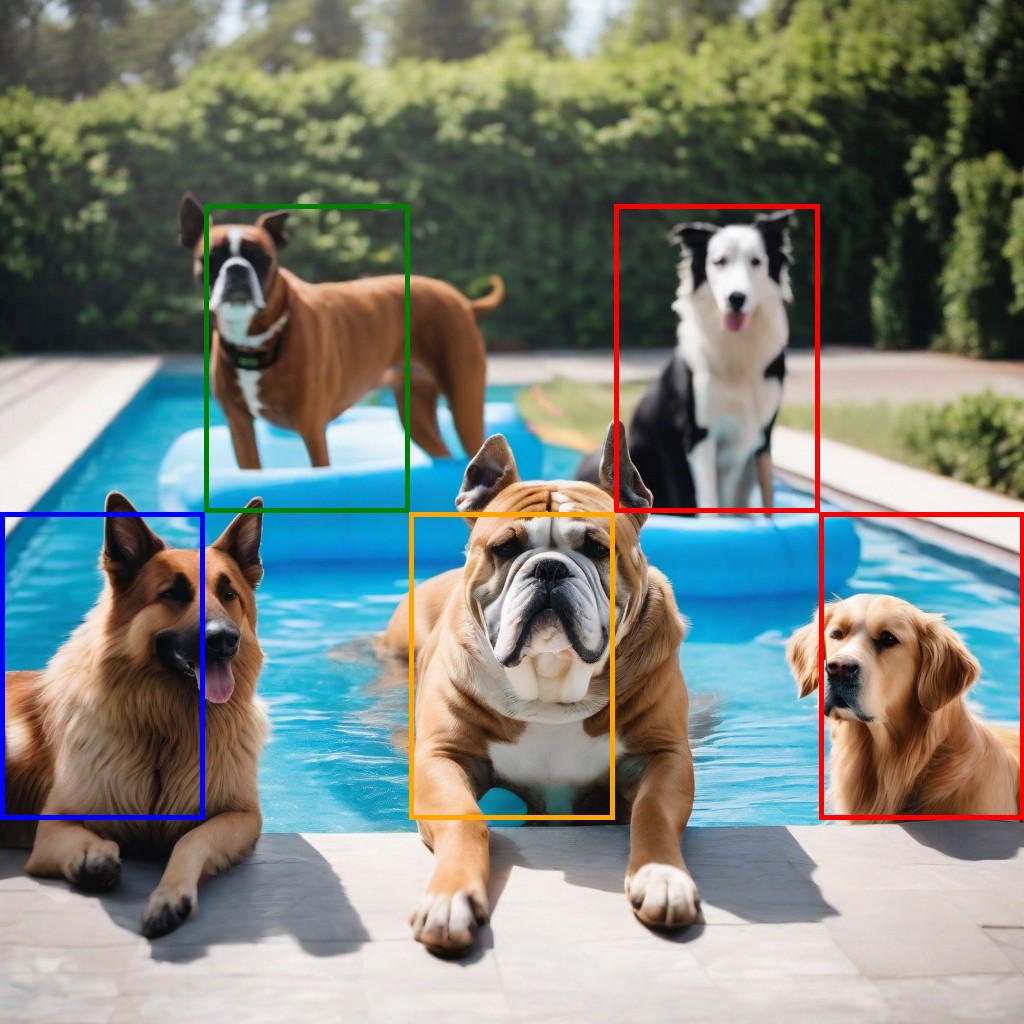} &
    \includegraphics[width=0.16\textwidth]
    {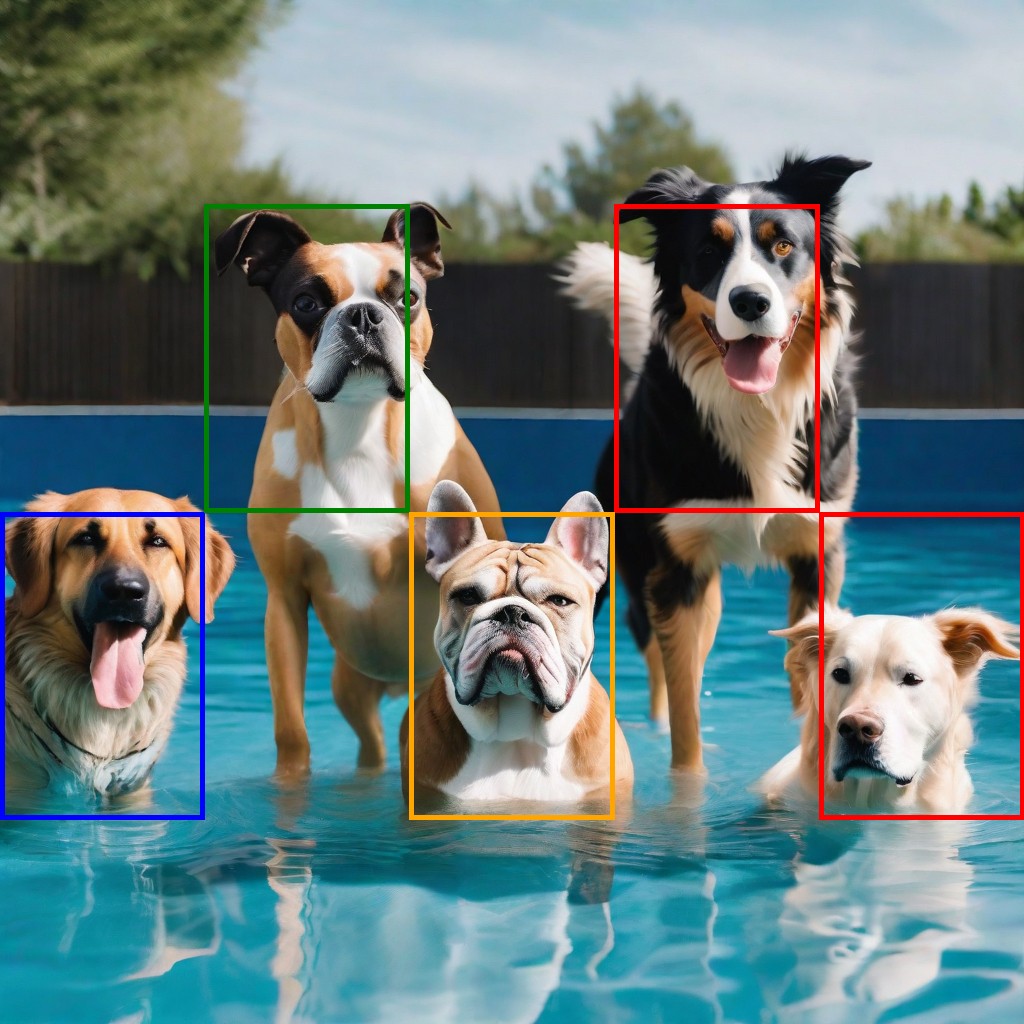} &
    \includegraphics[width=0.16\textwidth]
    {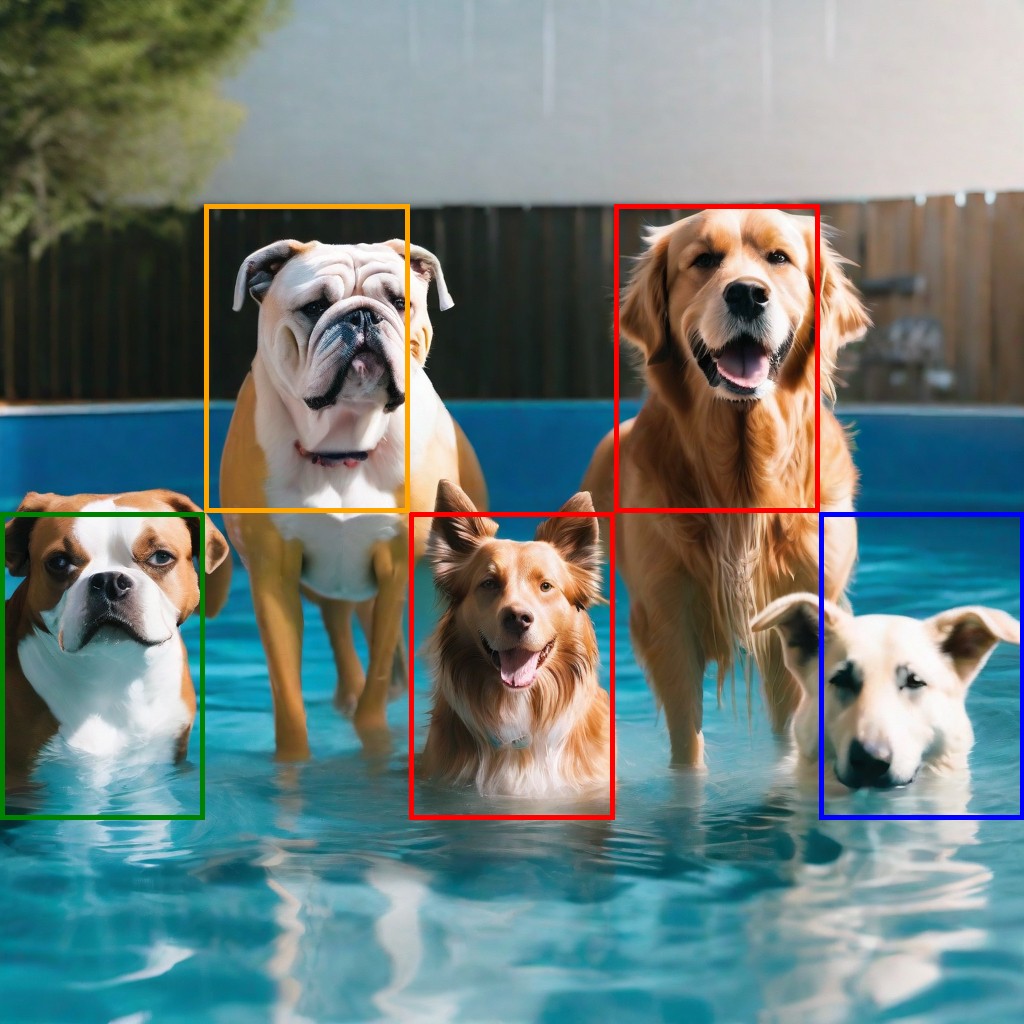} &
    \includegraphics[width=0.16\textwidth]
    {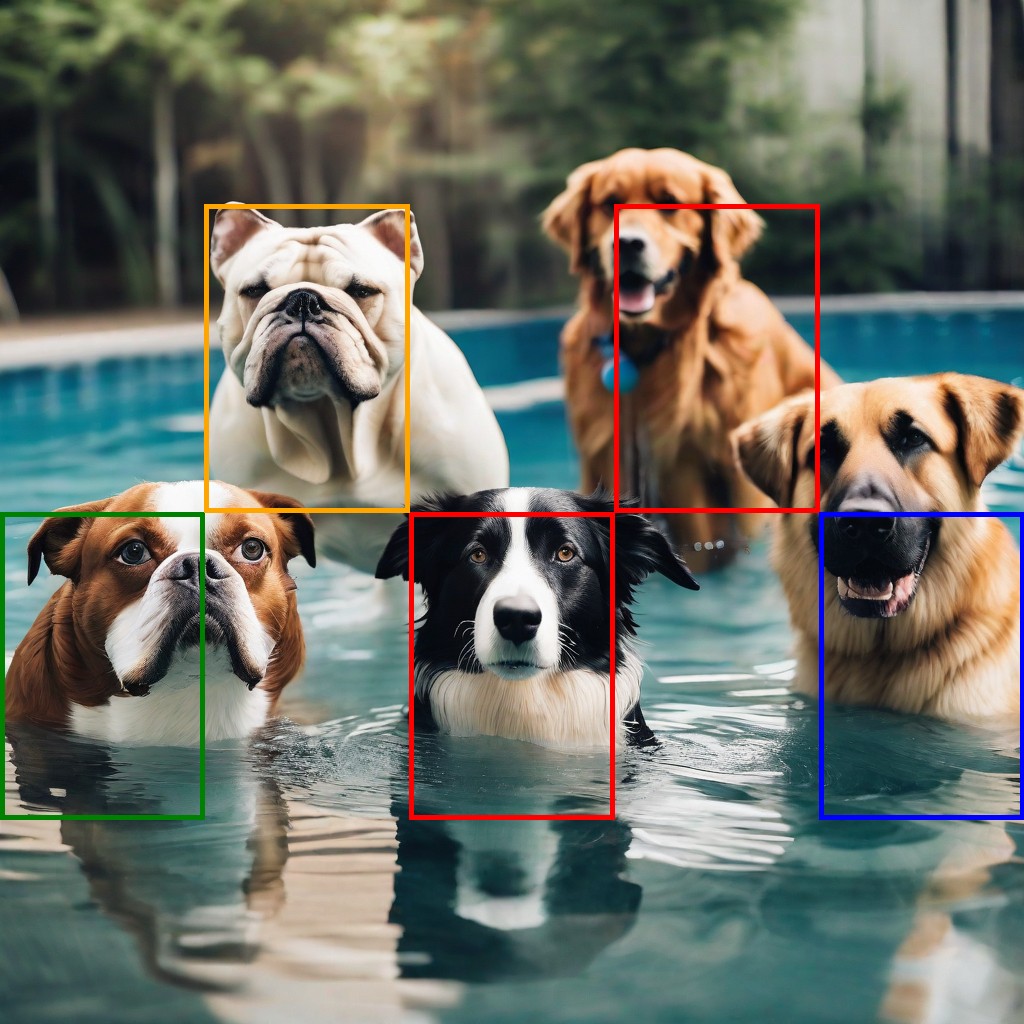} &
    \includegraphics[width=0.16\textwidth]
    {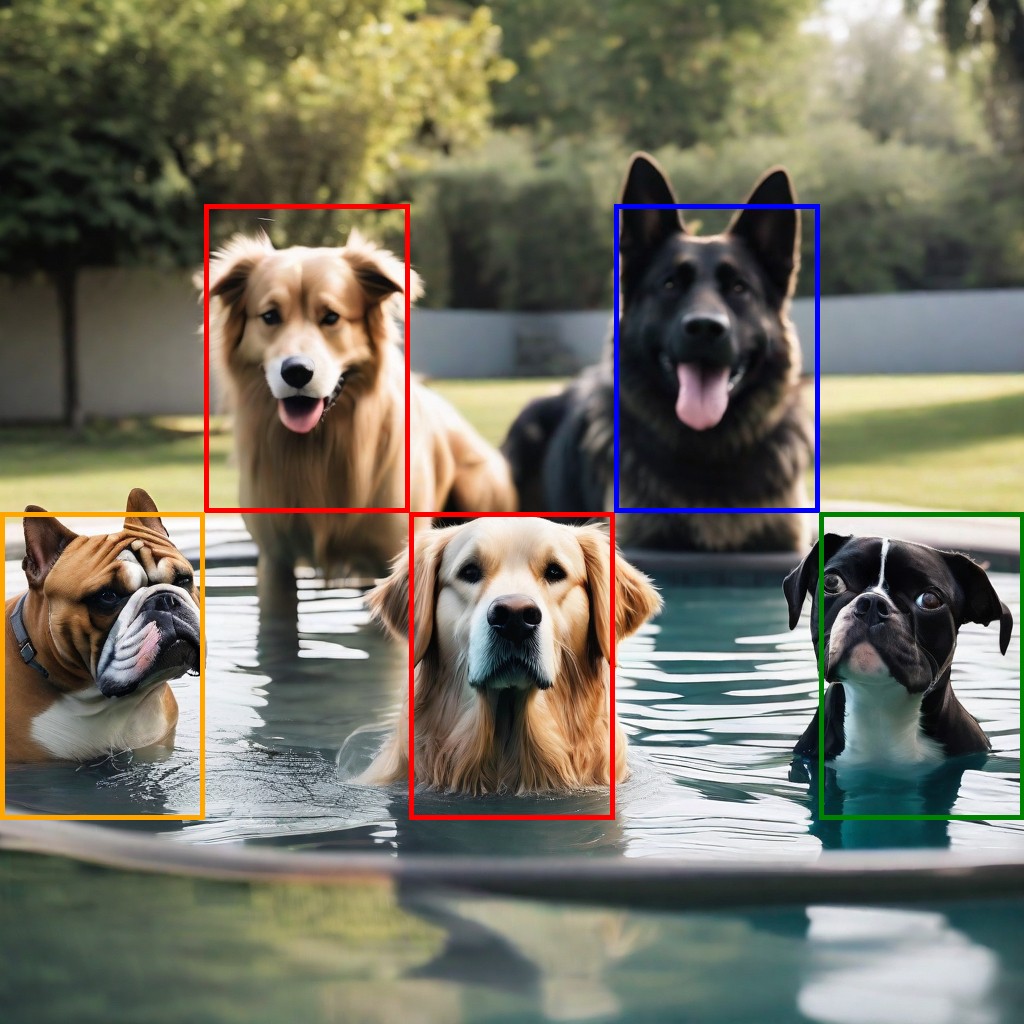} \\
\end{tabular}
}

%% file: figures/sdxl_results/caption.tex
Bounded Attention in SDXL enables precise control over multiple subjects, producing natural images where each subject seamlessly integrates into the scene while retaining its distinct features.
\vspace{-10pt}

%% file: figures/sdxl_results2.tex
\begin{figure}
    \setlength{\tabcolsep}{1pt}
    {\scriptsize\centering
    \begin{tabular}{c c c c c}
        &
        \multicolumn{4}{c}{"A \textcolor{red}{\textit{\underline{gray kitten}}} and a \textcolor{blue}{\textit{\underline{ginger kitten}}} and a \textcolor{green}{\textit{\underline{black kitten}}}} \\
        &
        \multicolumn{4}{c}{and a \textcolor{orange}{\textit{\underline{white dog}}} and a \textcolor{purple}{\textit{\underline{brown dog}}} on a bed"} \\
       \raisebox{1pt}{\rotatebox{90}{Bounded Attention}} &
        \includegraphics[width=0.11\textwidth]{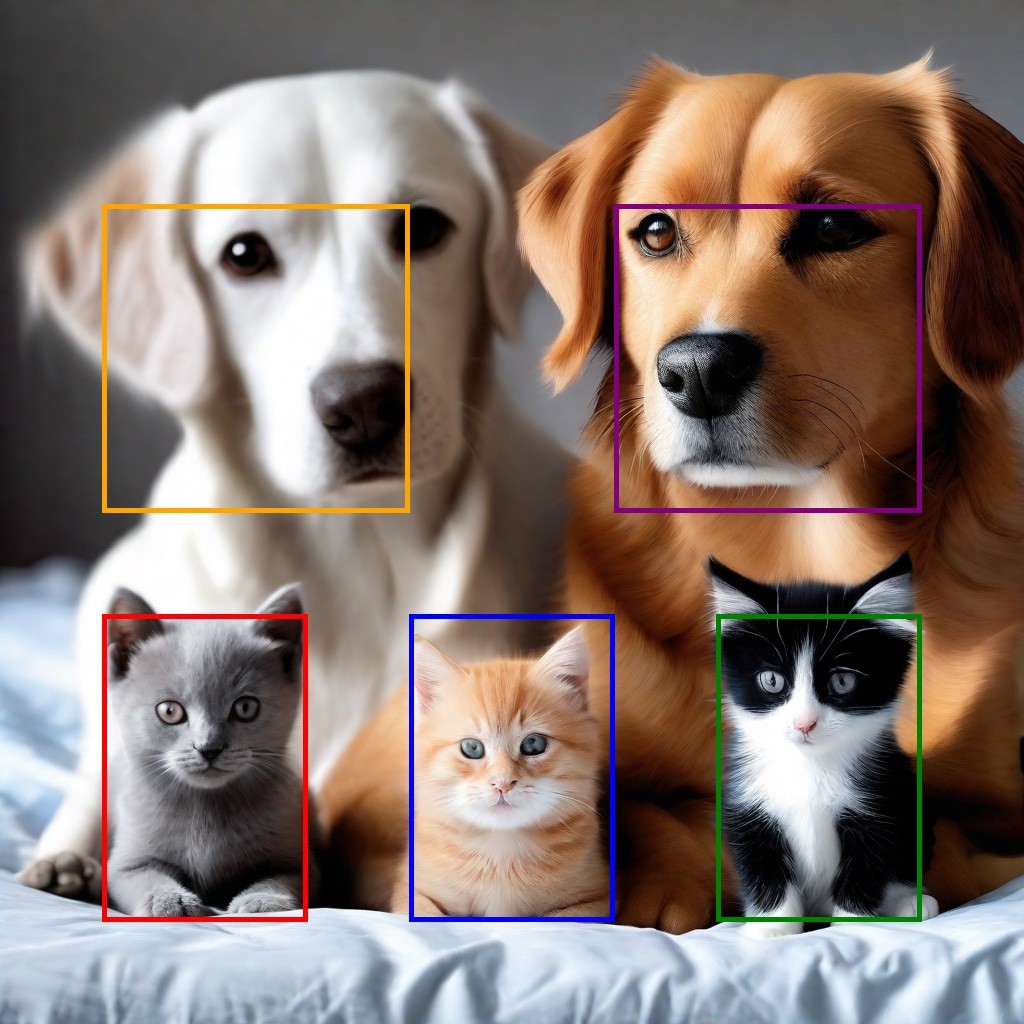} &
        \includegraphics[width=0.11\textwidth]{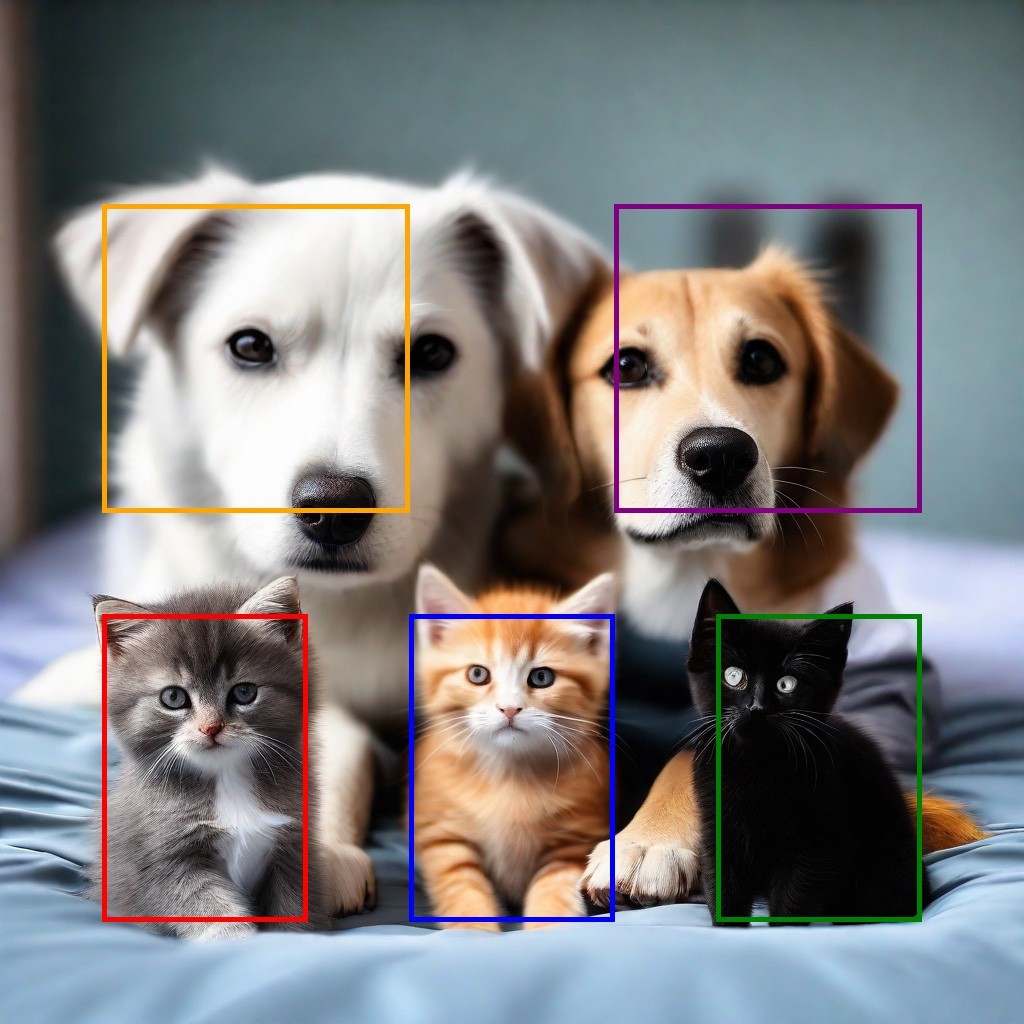} &
        \includegraphics[width=0.11\textwidth]{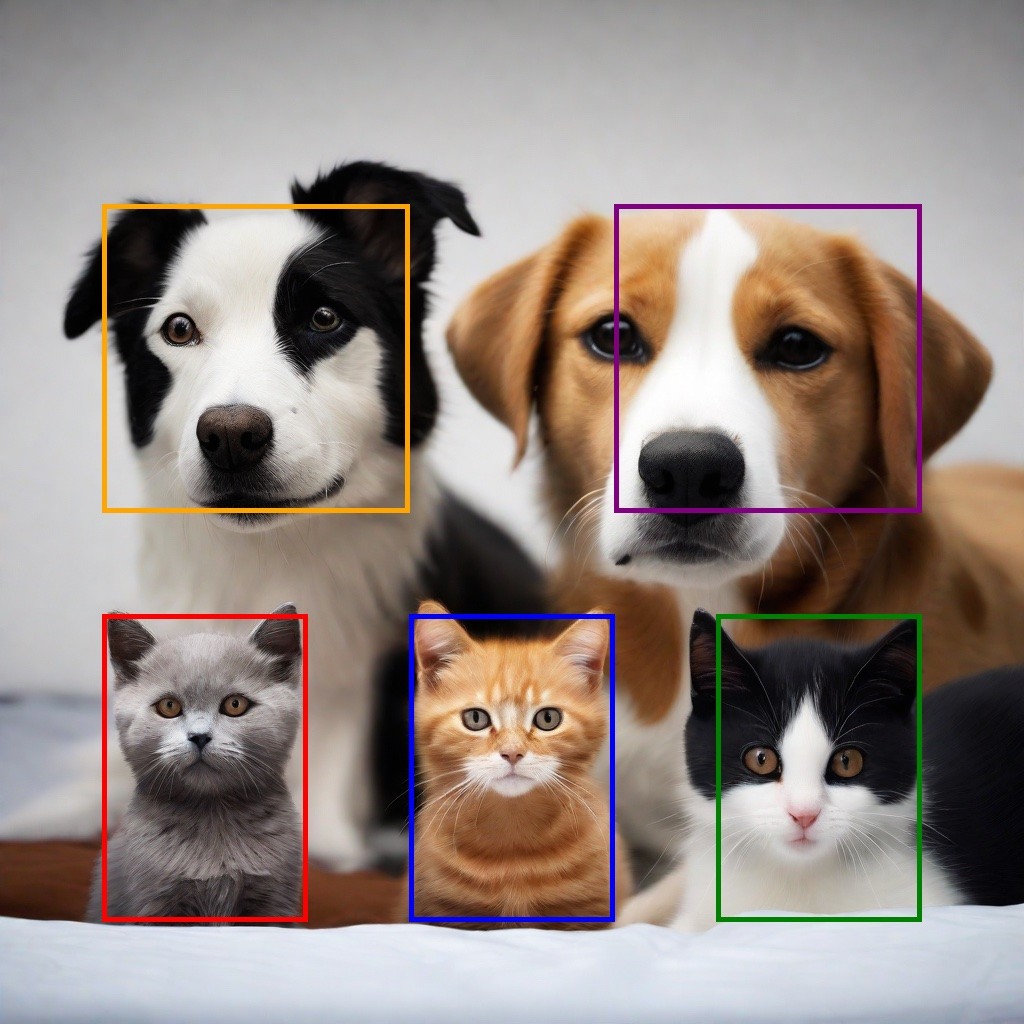} &
        \includegraphics[width=0.11\textwidth]{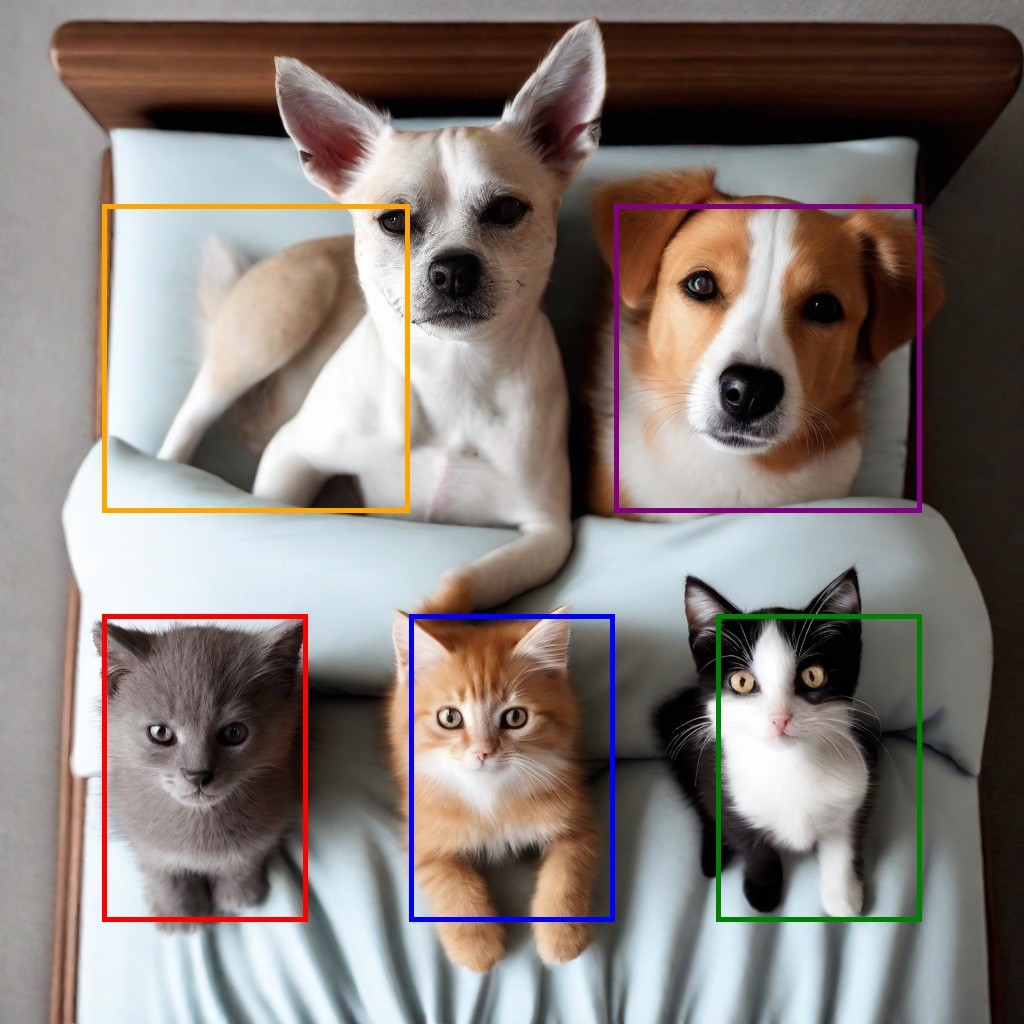} \\
        \raisebox{7pt}{\rotatebox{90}{Vanilla SDXL}} &
        \includegraphics[width=0.11\textwidth]{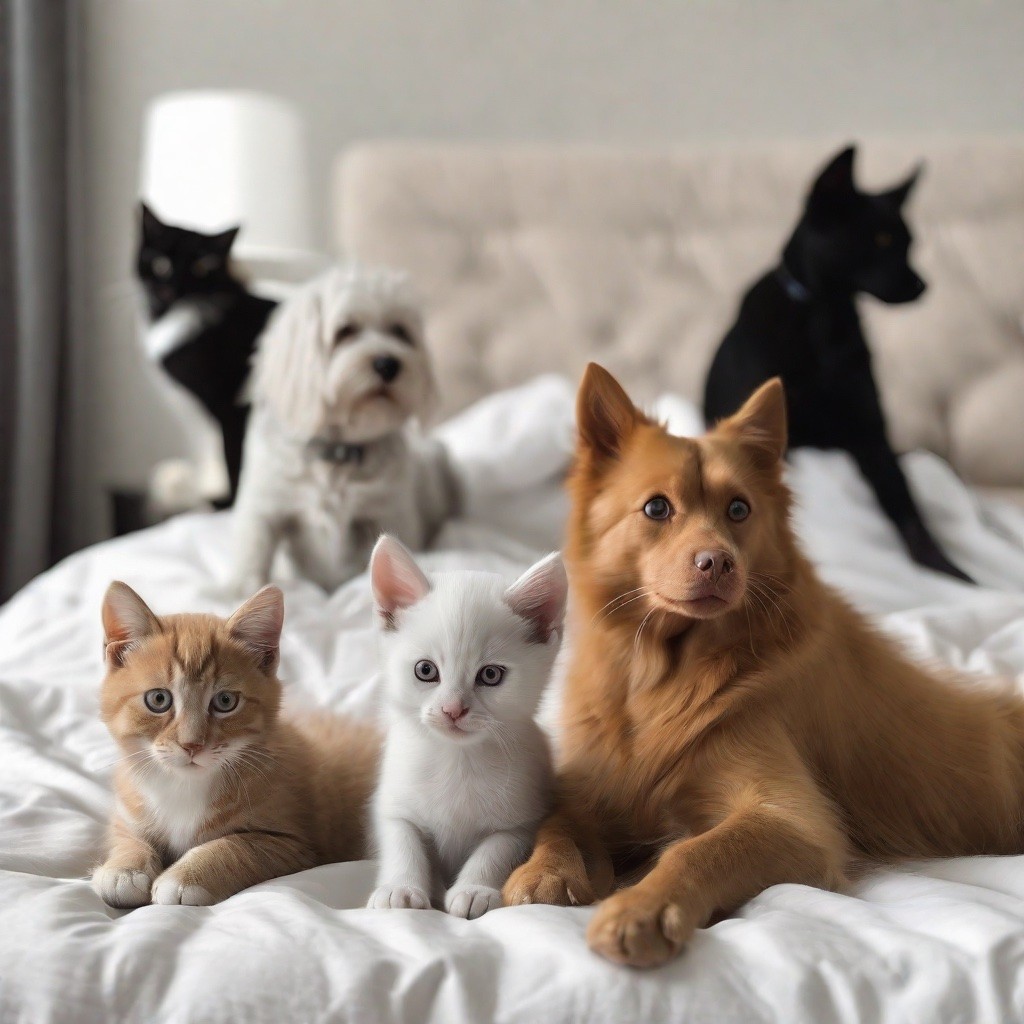} &
        \includegraphics[width=0.11\textwidth]{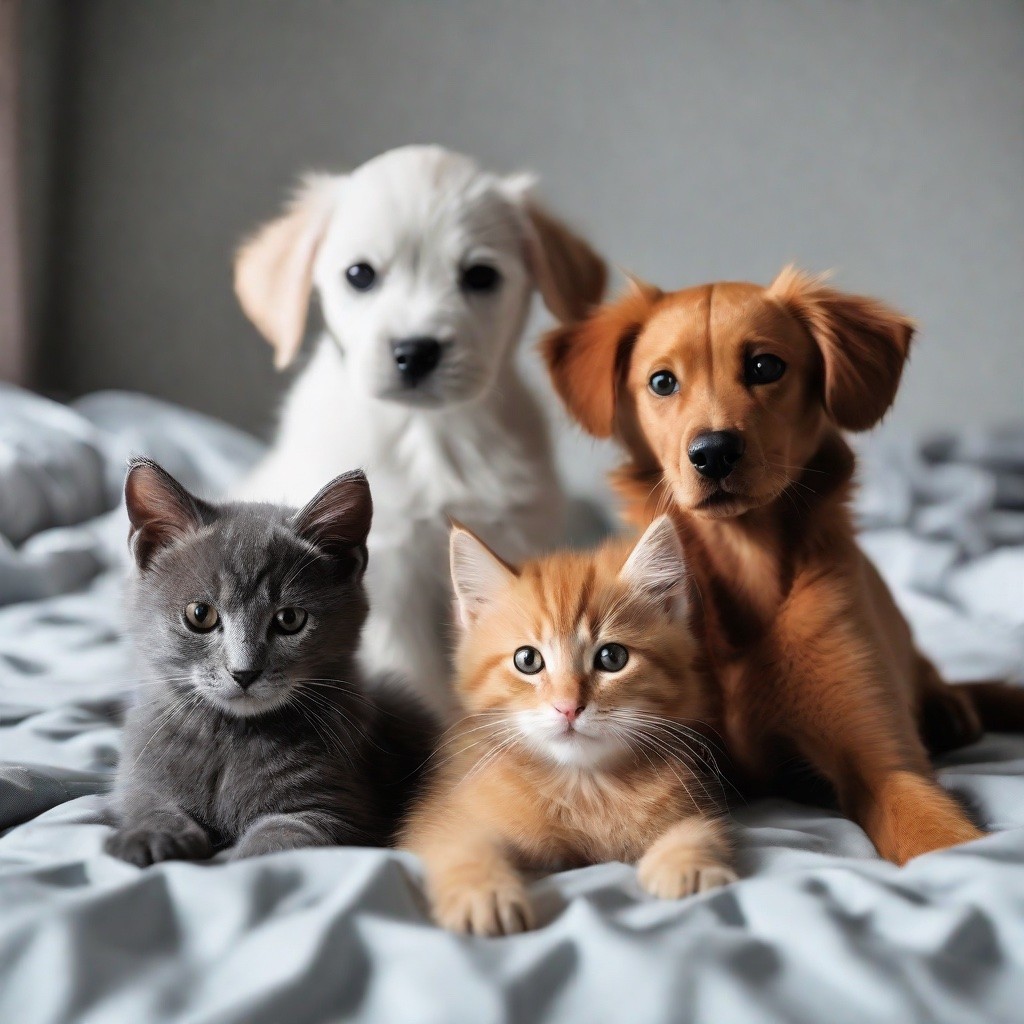} &
        \includegraphics[width=0.11\textwidth]{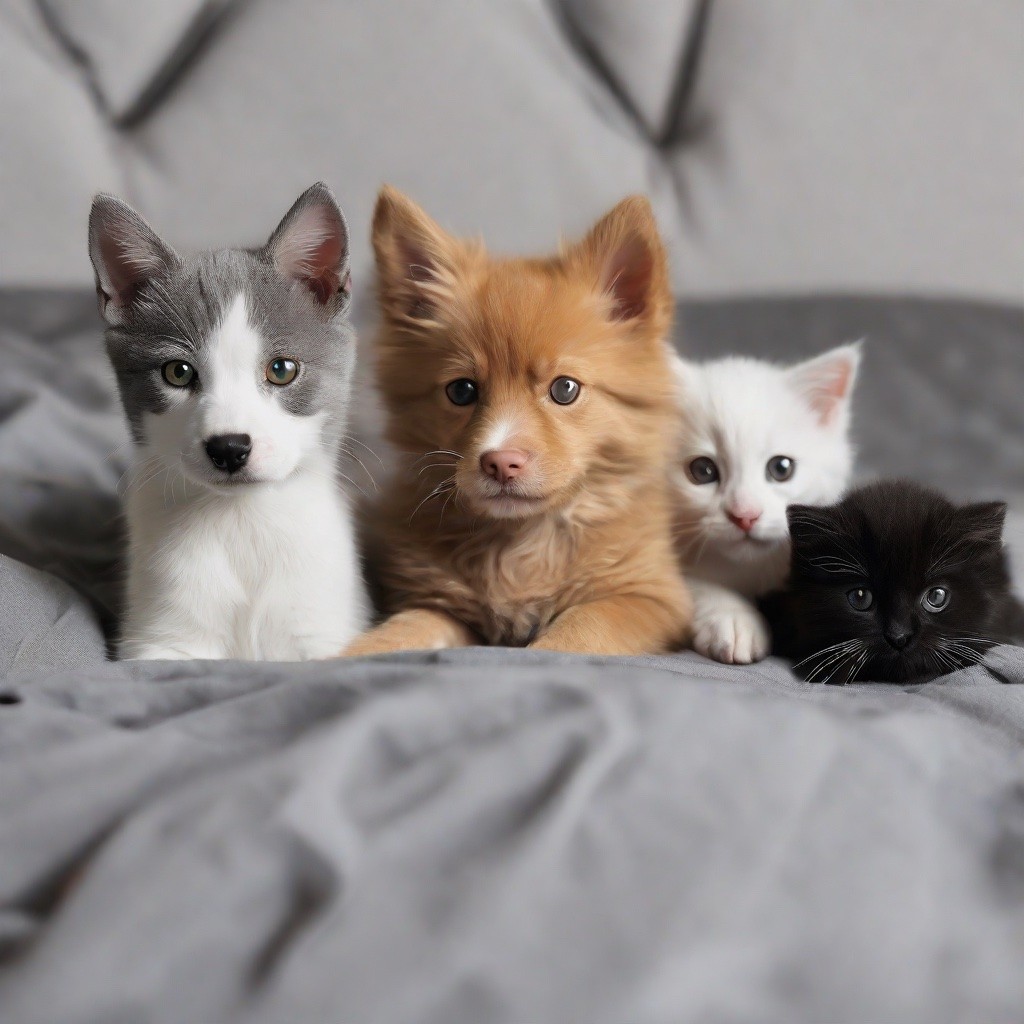} &
        \includegraphics[width=0.11\textwidth]{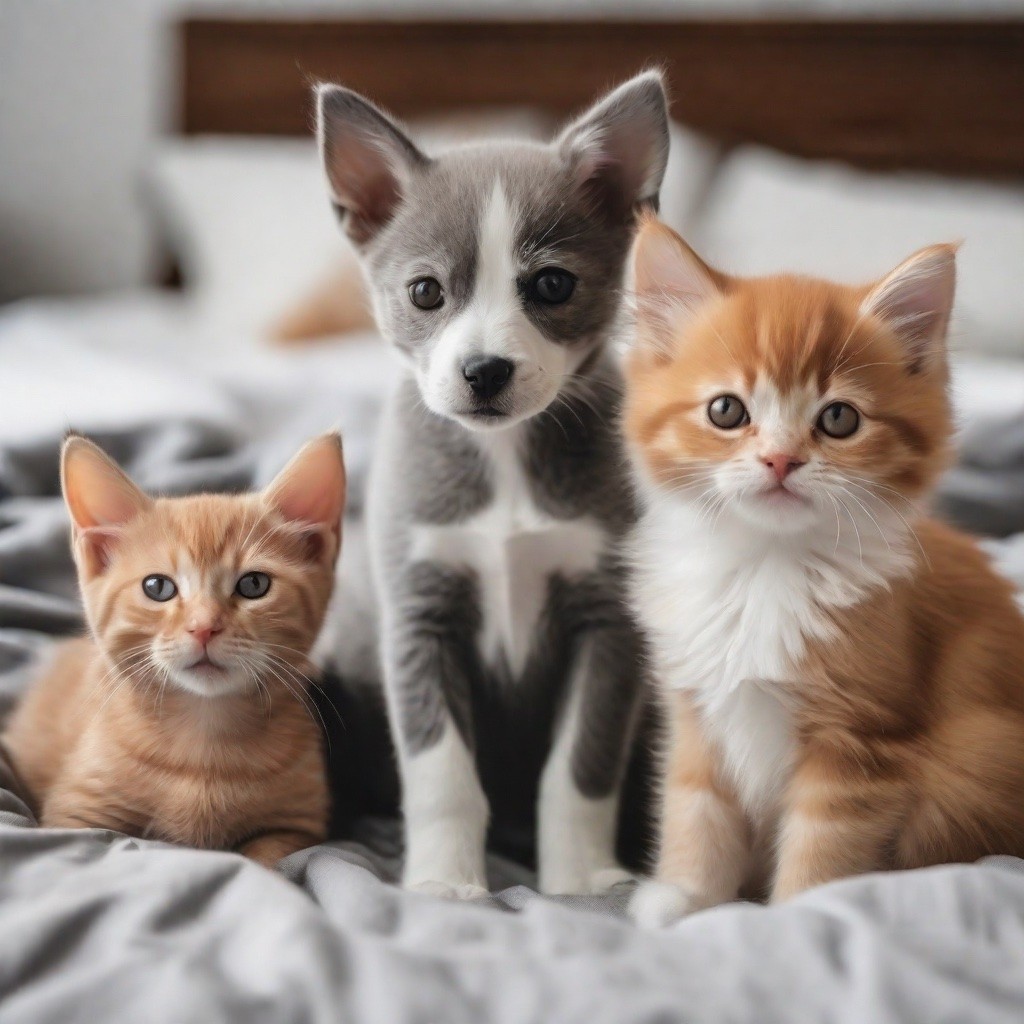} \\
        \\[-6pt]
        
        &
        \multicolumn{4}{c}{``A realistic photo of a window dressing with three mannequins wearing a}
        \\
        &
        \multicolumn{4}{c}{\textcolor{red}{\textit{\underline{blue velvet dress}}} and a \textcolor{blue}{\textit{\underline{pink tulle gown}}} and a \textcolor{green}{\textit{\underline{brown fur coat}}}.''} \\
       \raisebox{1pt}{\rotatebox{90}{Bounded Attention}} &
        \includegraphics[width=0.11\textwidth]{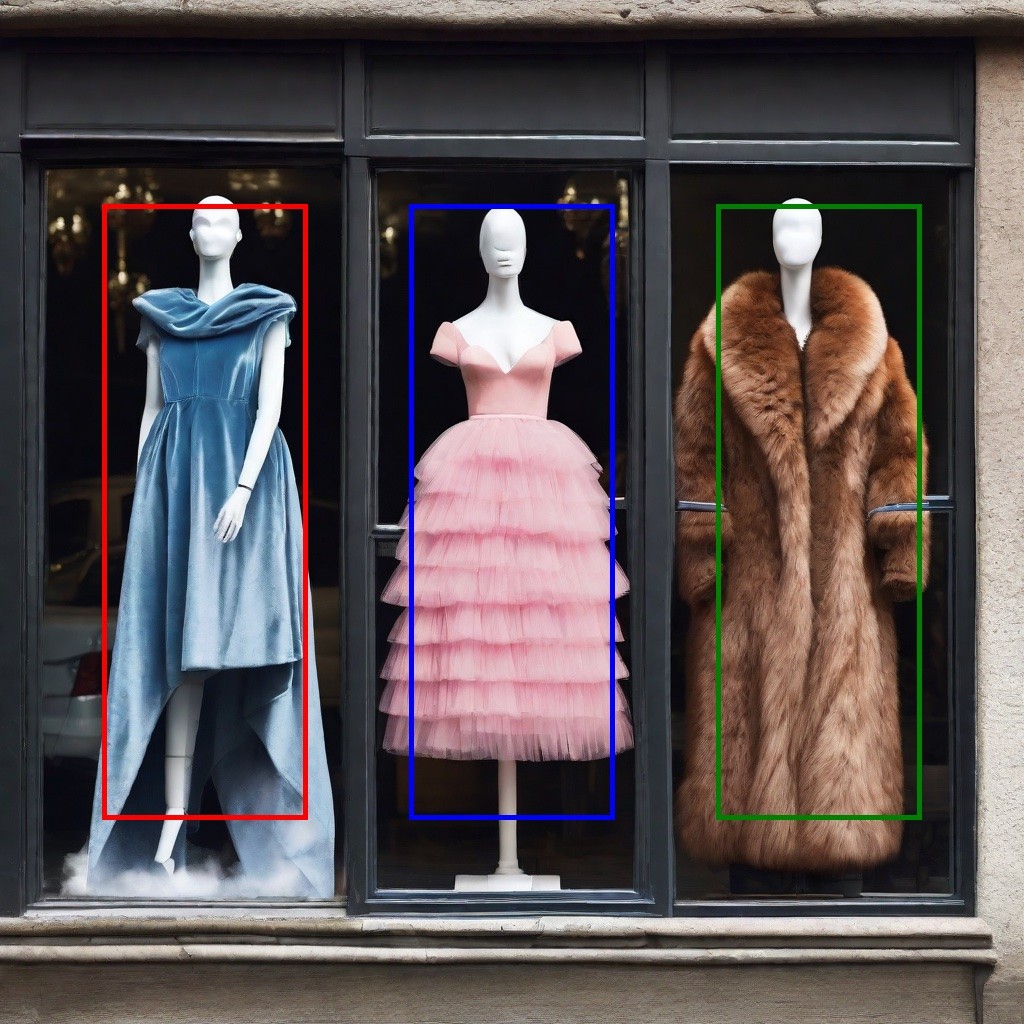} &
        \includegraphics[width=0.11\textwidth]{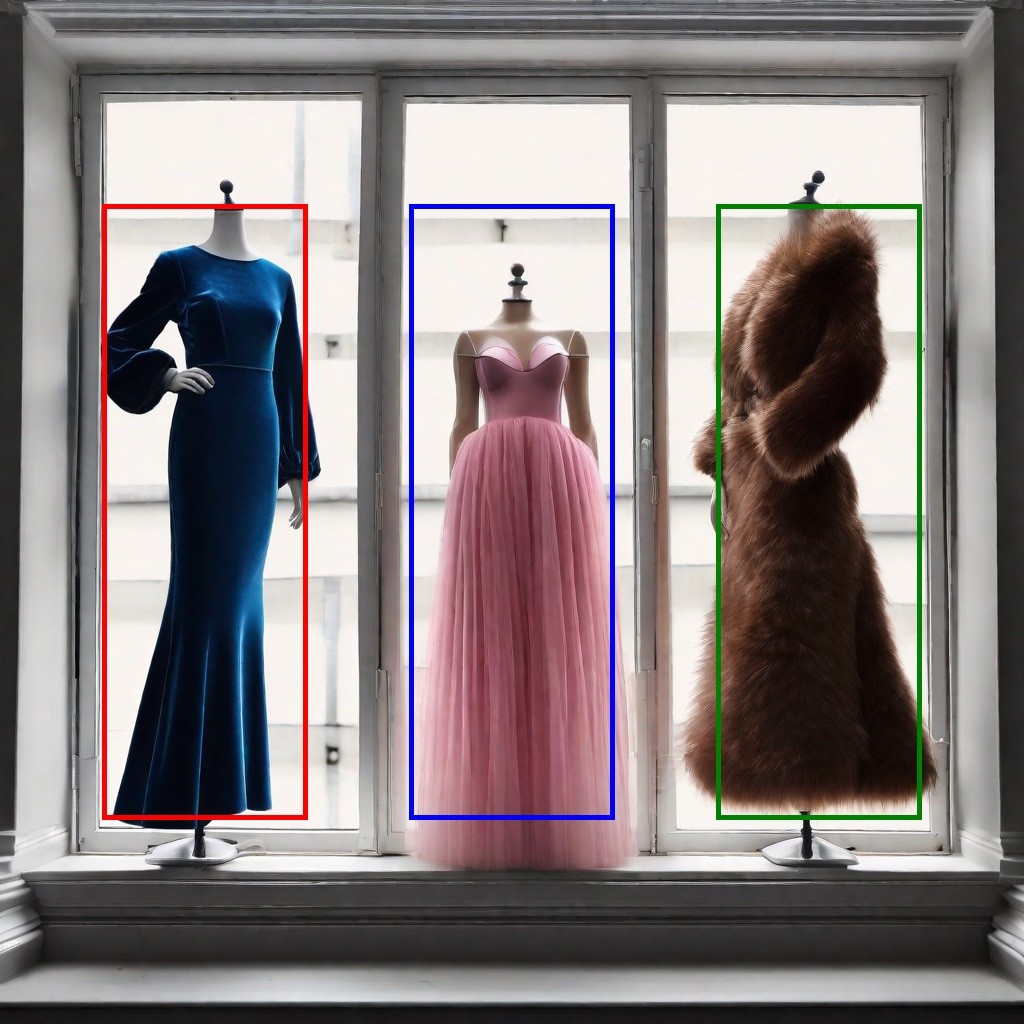} &
        \includegraphics[width=0.11\textwidth]{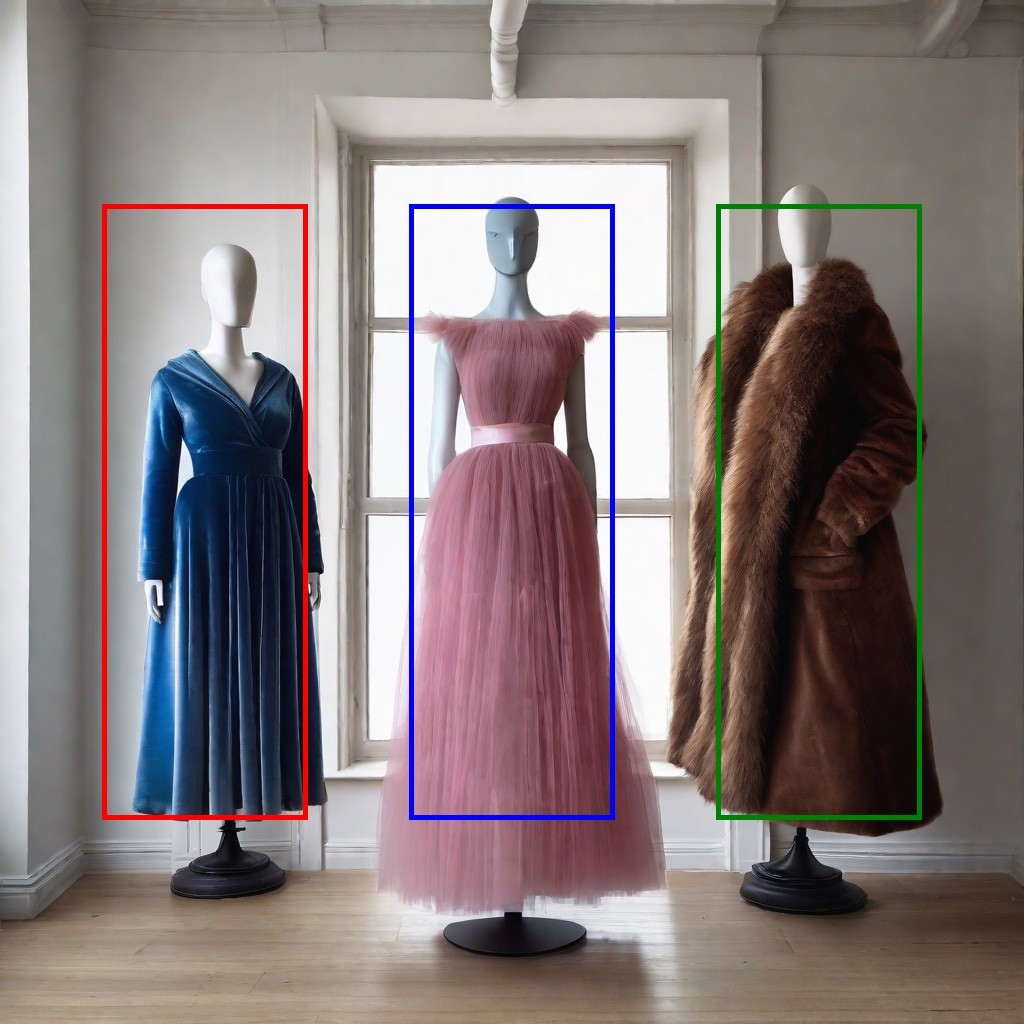} &
        \includegraphics[width=0.11\textwidth]{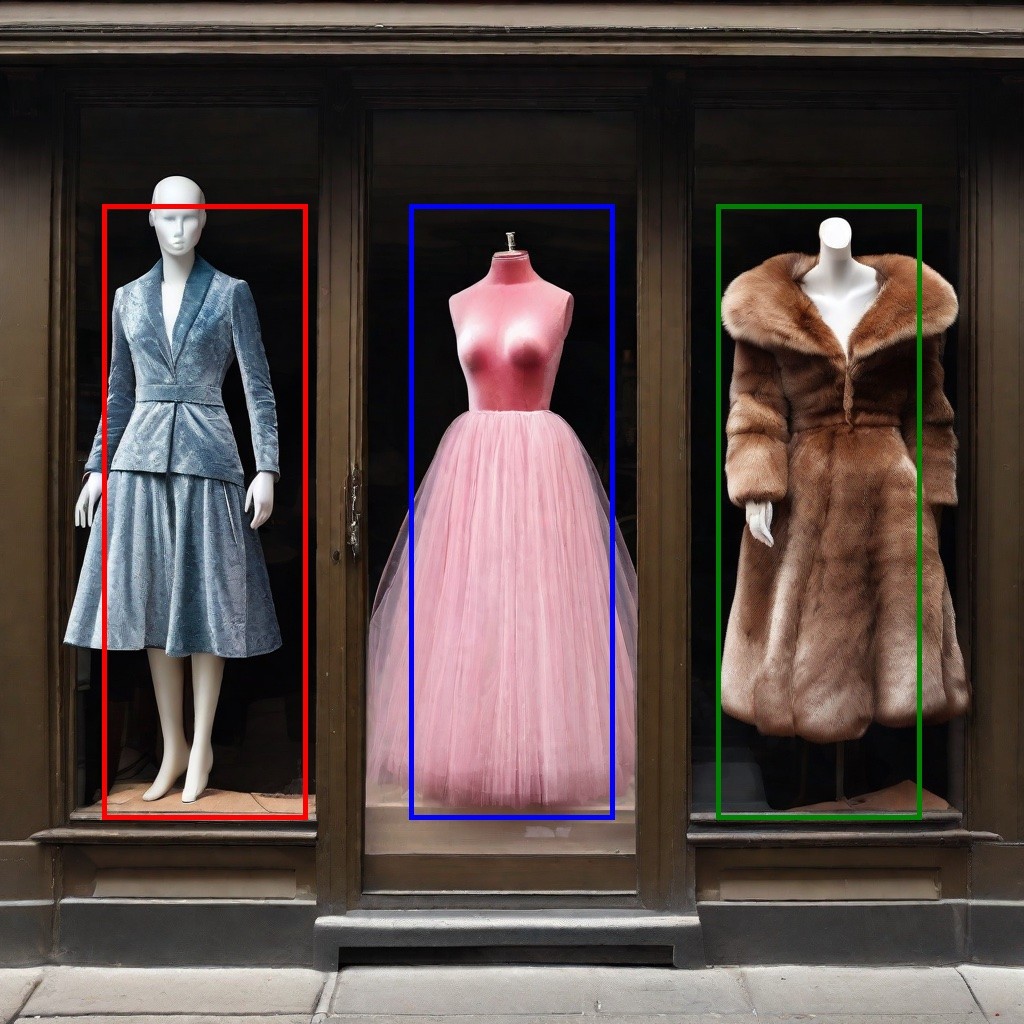} \\
        \raisebox{7pt}{\rotatebox{90}{Vanilla SDXL}} &
        \includegraphics[width=0.11\textwidth]{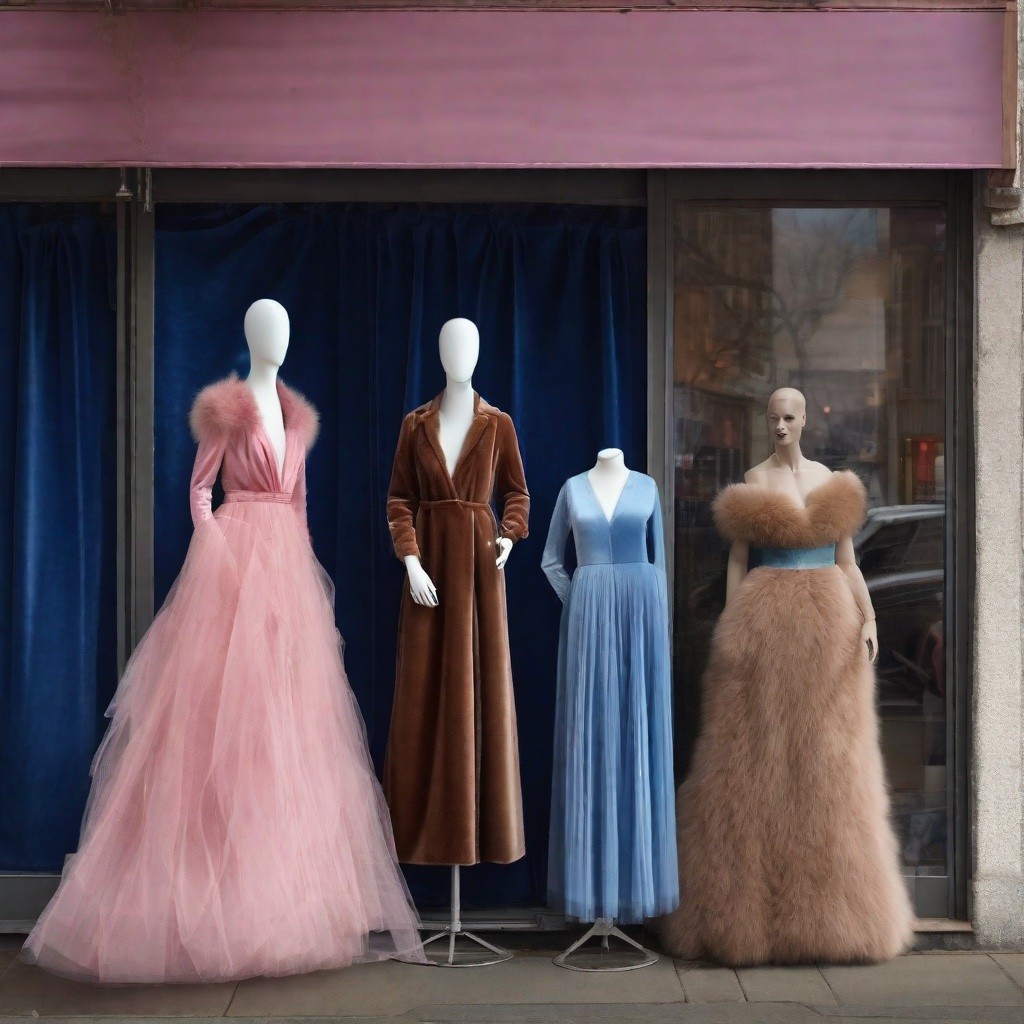} &
        \includegraphics[width=0.11\textwidth]{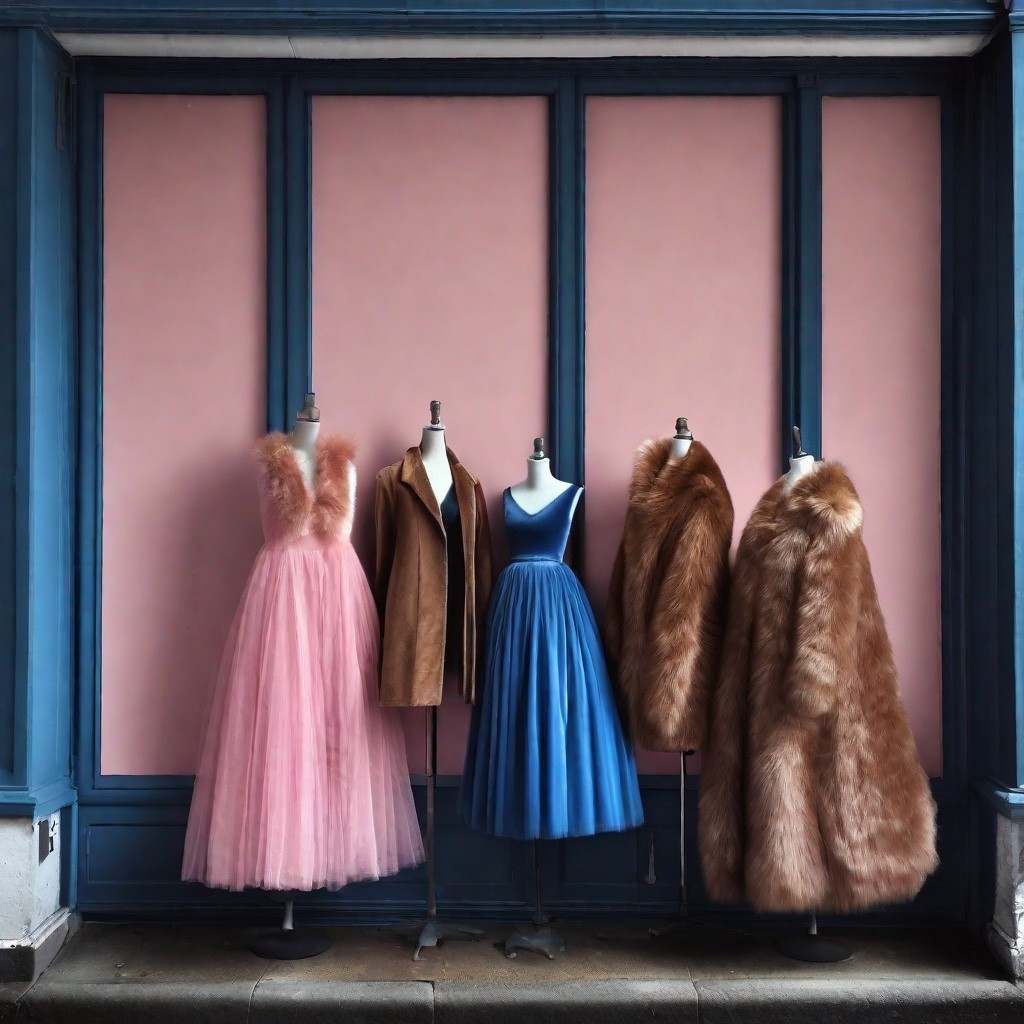} &
        \includegraphics[width=0.11\textwidth]{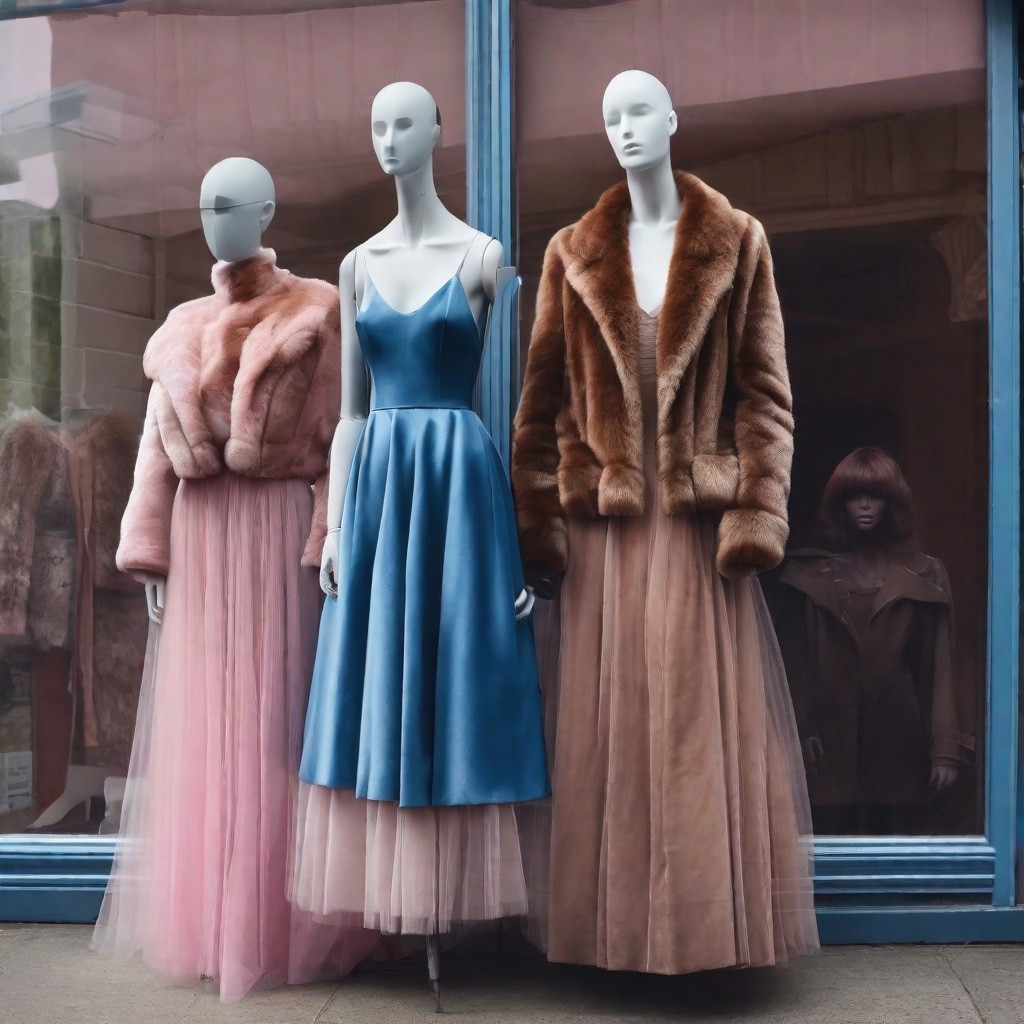} &
        \includegraphics[width=0.11\textwidth]{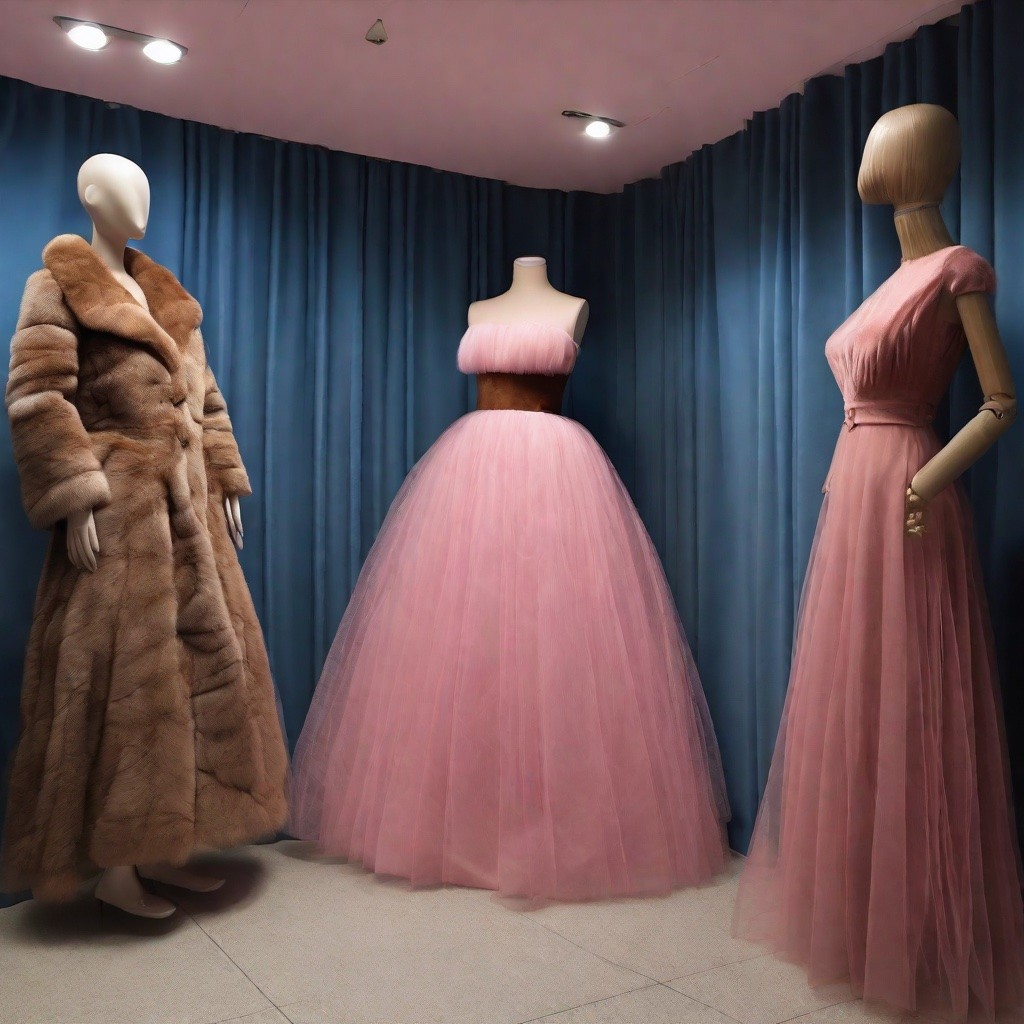} \\
        \\[-6pt]

        &
        \multicolumn{4}{c}{``3D Pixar animation of a \textcolor{red}{\textit{\underline{cute unicorn}}} and a \textcolor{blue}{\textit{\underline{pink hedgehog}}}}
        \\
        &
        \multicolumn{4}{c}{ and a \textcolor{green}{\textit{\underline{nerdy owl}}} traveling in a magical forest.''}
        \\
       \raisebox{1pt}{\rotatebox{90}{Bounded Attention}} &
        \includegraphics[width=0.11\textwidth]{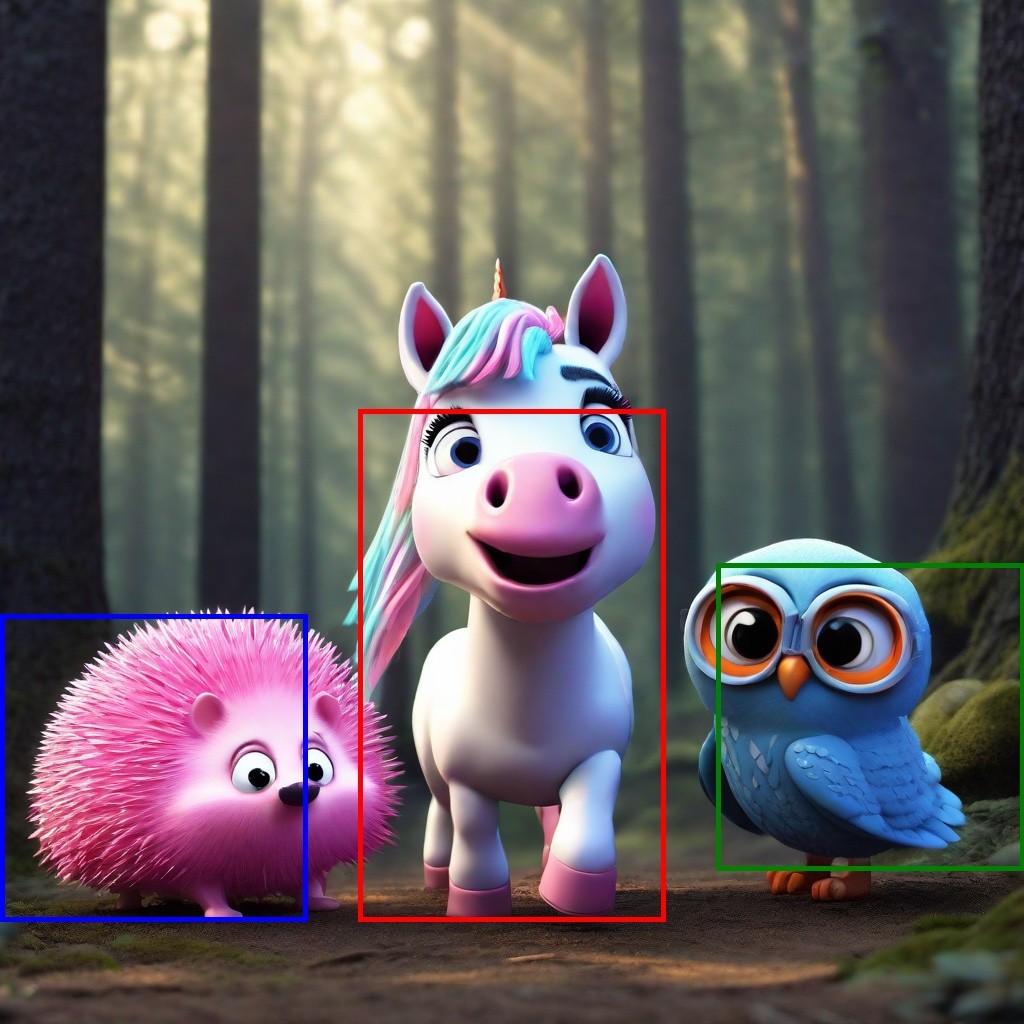} &
        \includegraphics[width=0.11\textwidth]{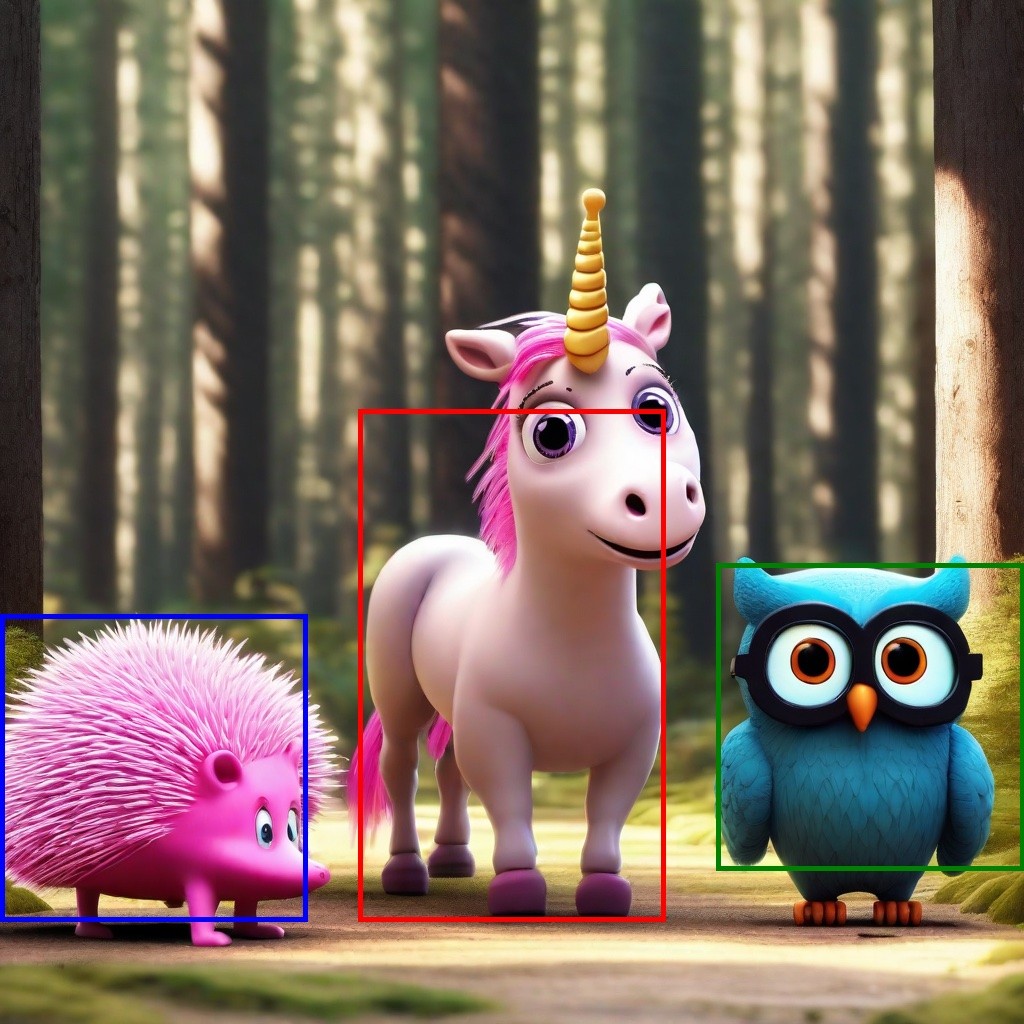} &
        \includegraphics[width=0.11\textwidth]{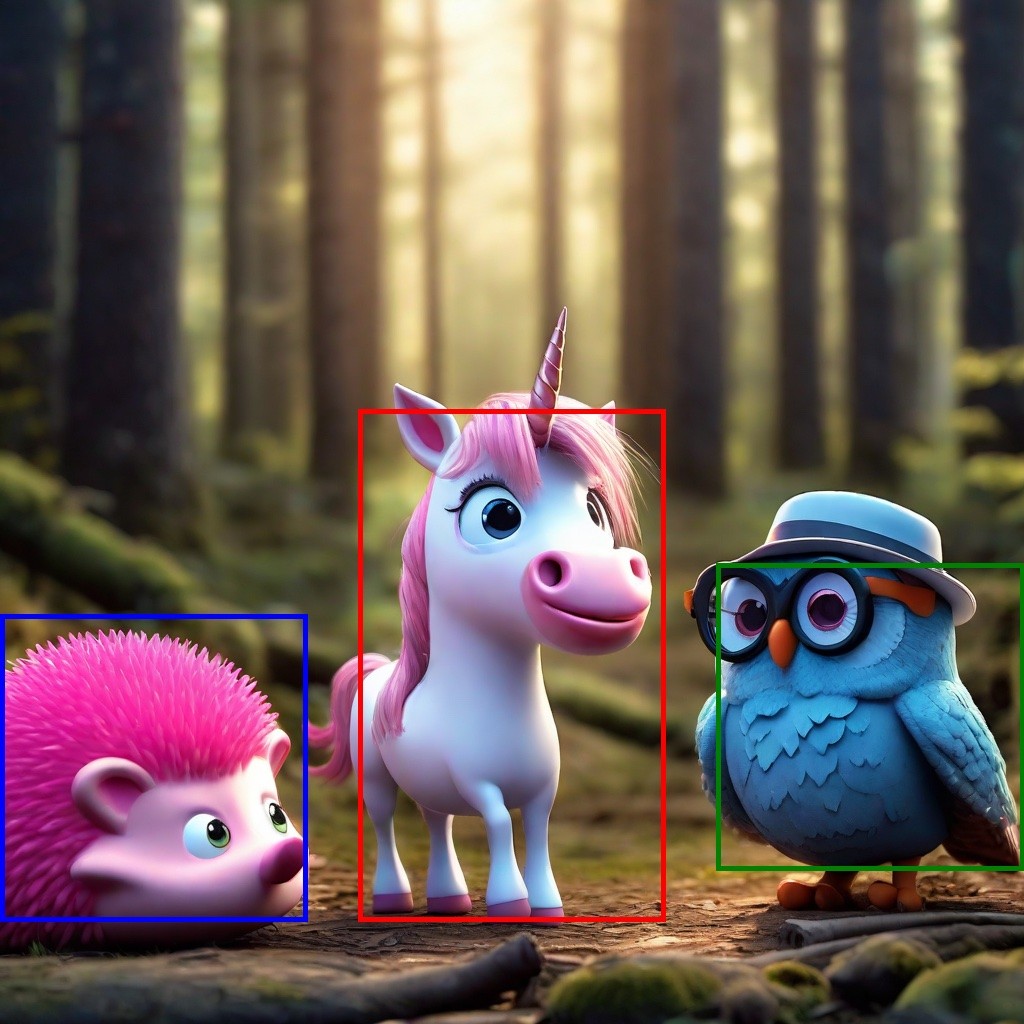} &
        \includegraphics[width=0.11\textwidth]{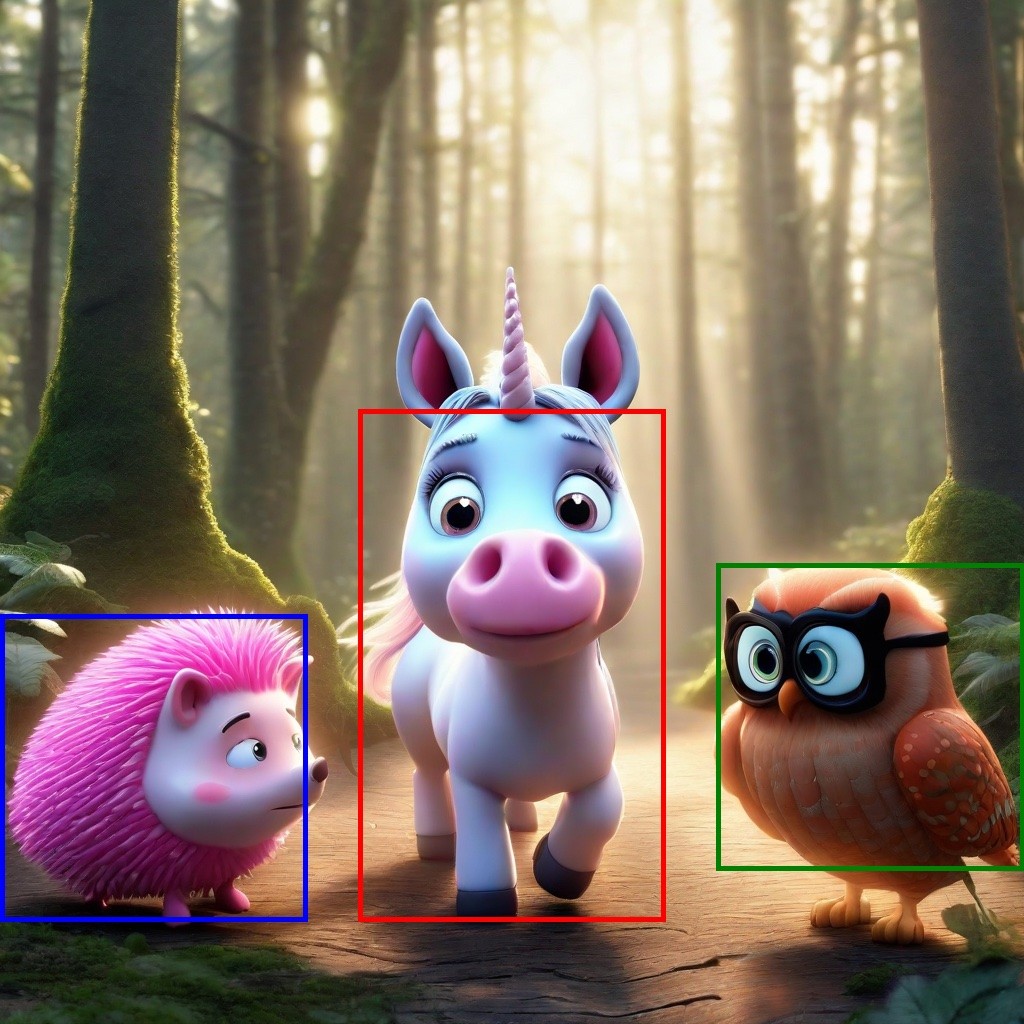} \\
        \raisebox{7pt}{\rotatebox{90}{Vanilla SDXL}} &
        \includegraphics[width=0.11\textwidth]{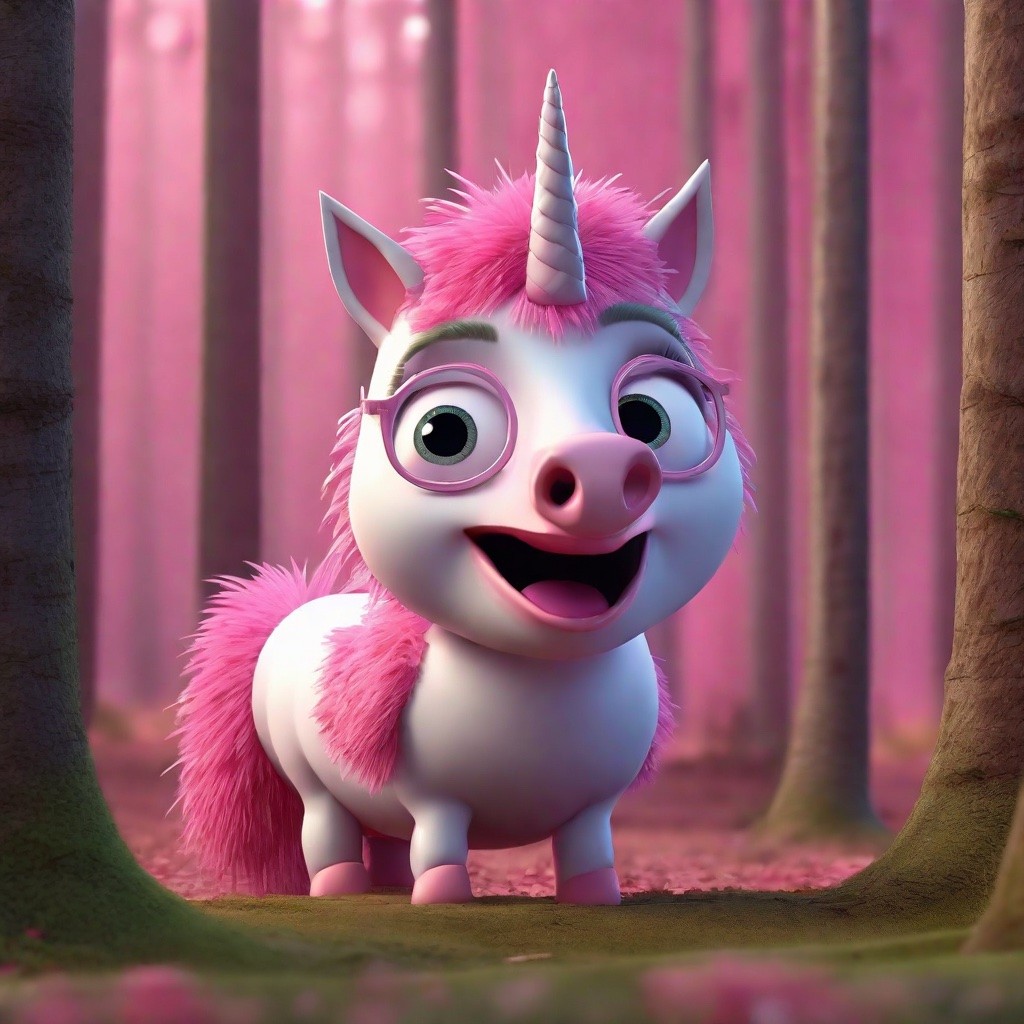} &
        \includegraphics[width=0.11\textwidth]{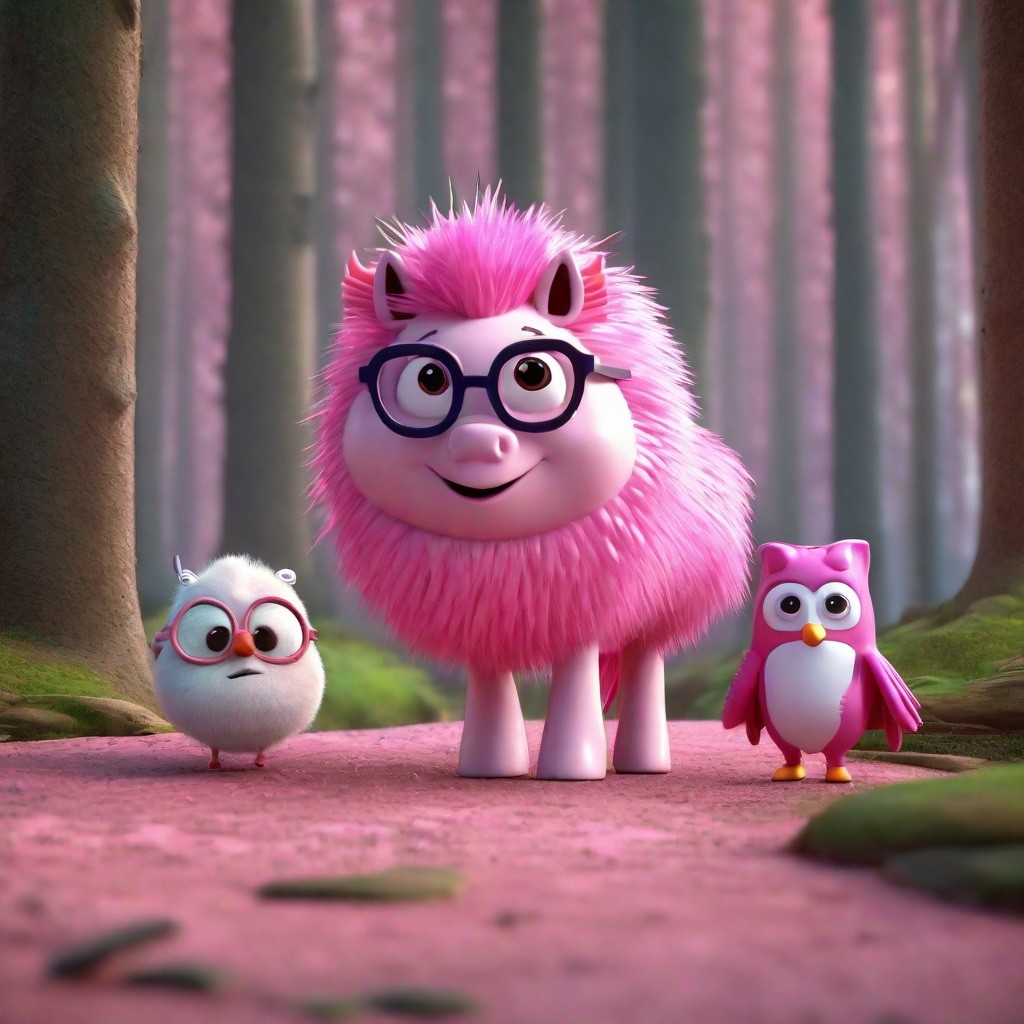} &
        \includegraphics[width=0.11\textwidth]{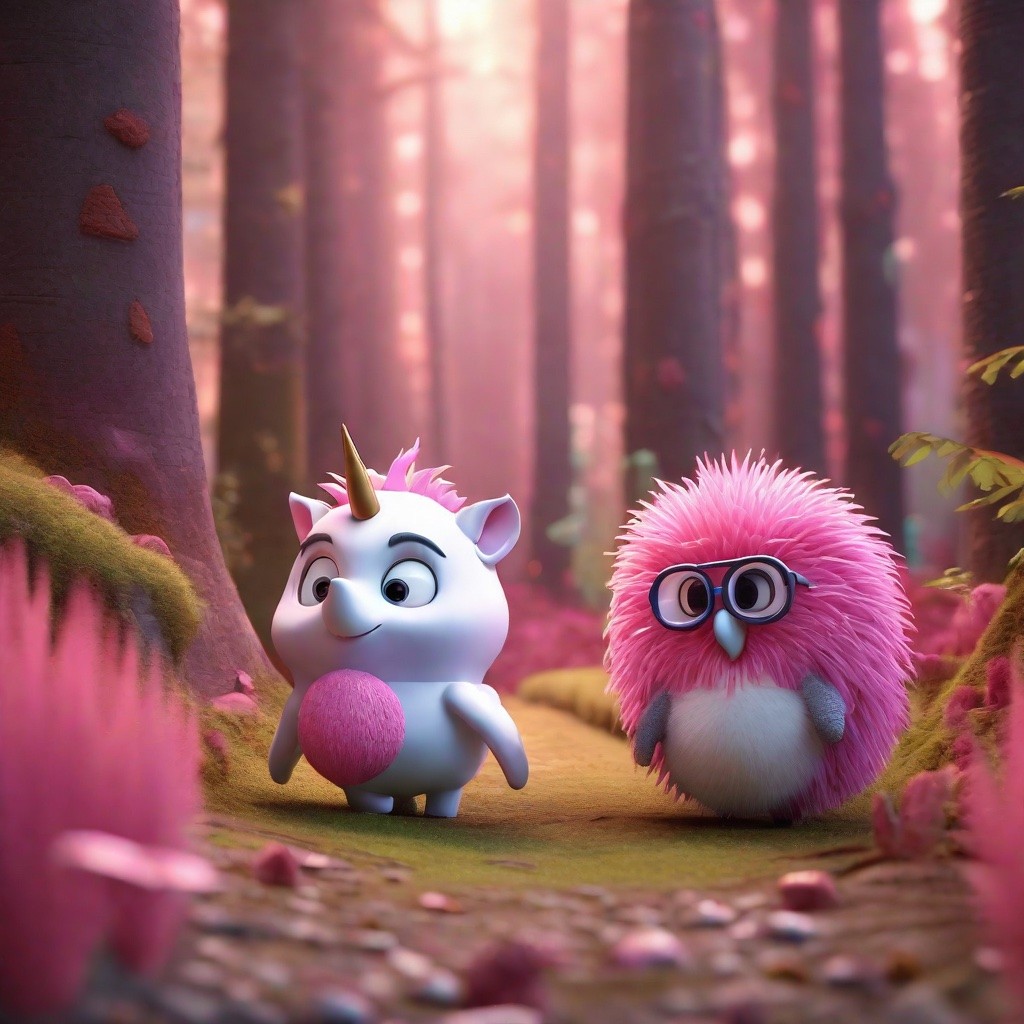} &
        \includegraphics[width=0.11\textwidth]{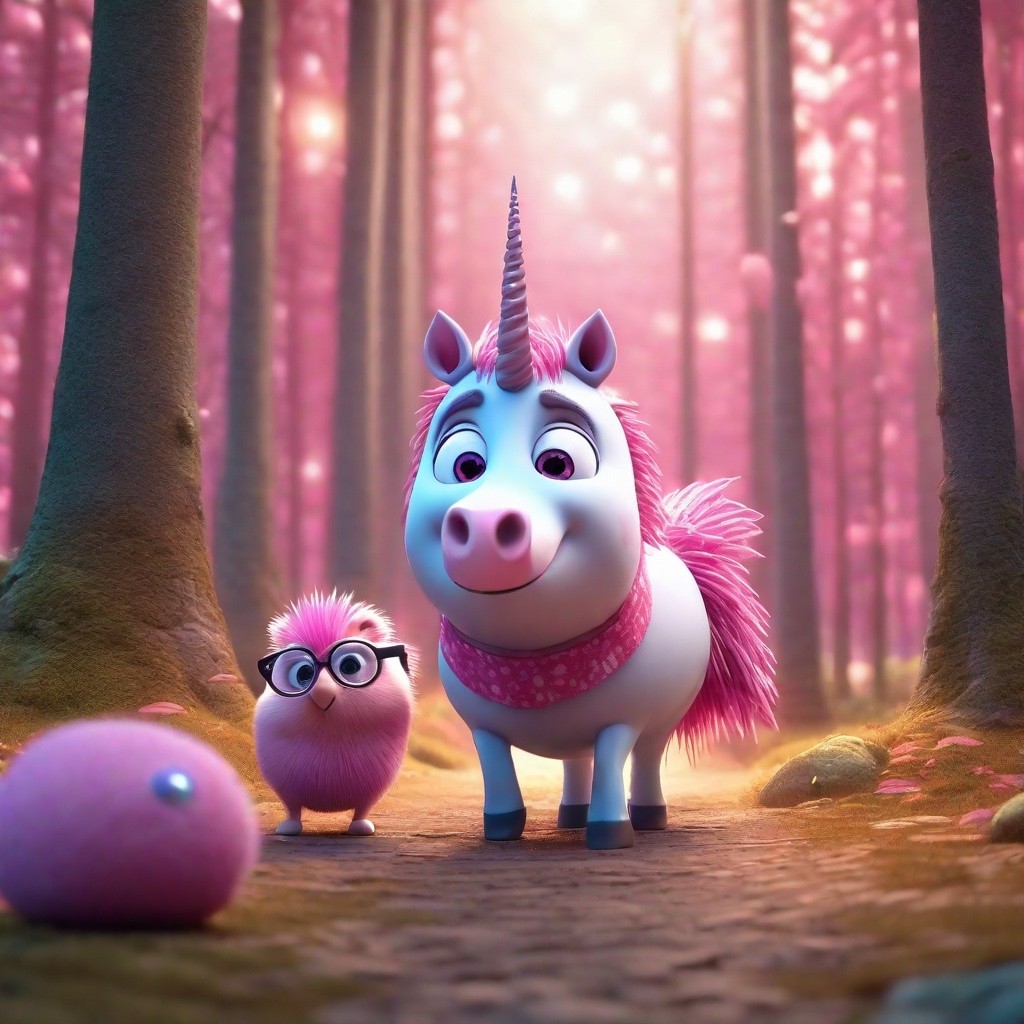} \\[-10pt]
    \end{tabular}
    }
    \captionof{figure}{
        Bounded Attention in SDXL promotes faithful generation of all subjects and their respective modifiers.
    }
    \vspace{-8pt}
    \label{fig:sdxl2}
\end{figure}

%% file: figures/seed0/full_figure.tex
\input{figures/seed0/figure_t}

%% file: figures/seed0/figure_t.tex
\begin{figure*}
    \centering
    {\small ``A \textcolor{red}{\textit{\underline{gray kitten}}} and a \textcolor{blue}{\textit{\underline{ginger kitten}}} and \textcolor{green}{\textit{\underline{black puppy}}} in a yard.''}
    \setlength{\tabcolsep}{1pt}
    {\small 
    \begin{tabular}{ccc ccc ccc cc}
        \includegraphics[width=0.115\linewidth]{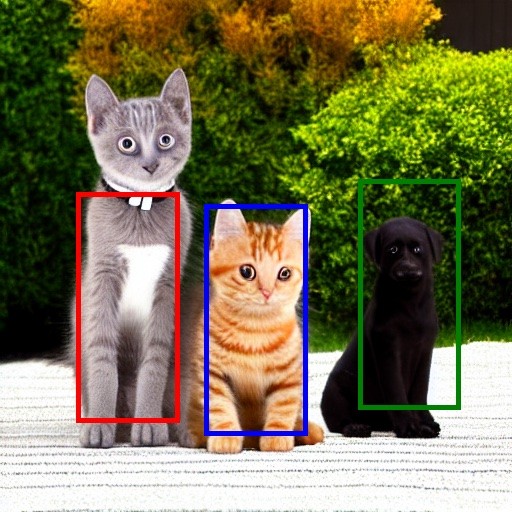} &
        \includegraphics[width=0.115\linewidth]{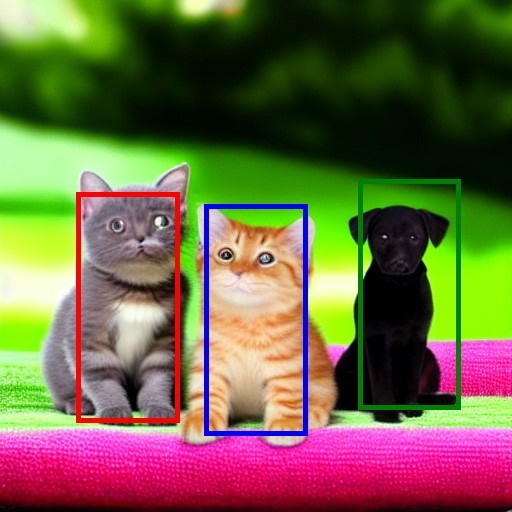} &
        { } &
        \includegraphics[width=0.115\linewidth]{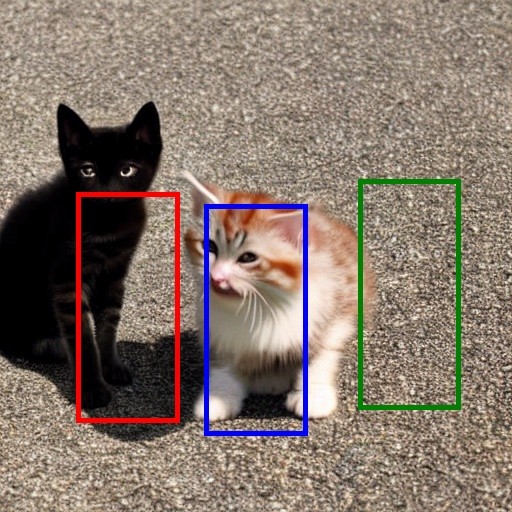} &
        \includegraphics[width=0.115\linewidth]{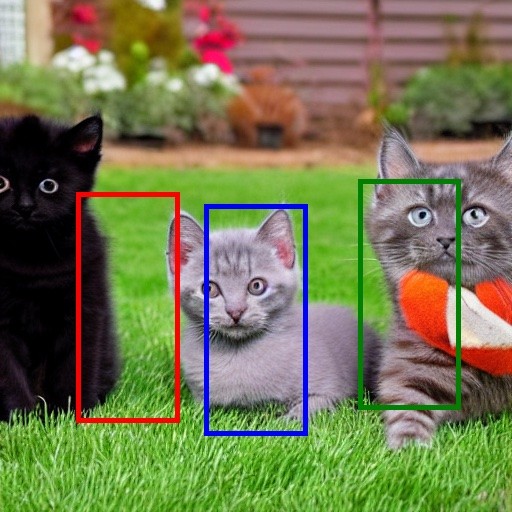} &
        { }&
        \includegraphics[width=0.115\linewidth]{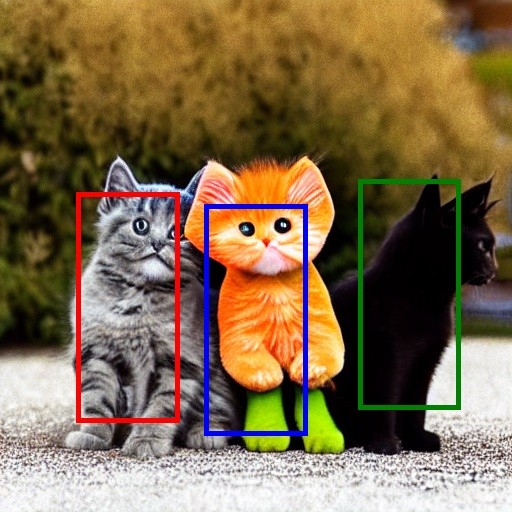} &
        \includegraphics[width=0.115\linewidth]{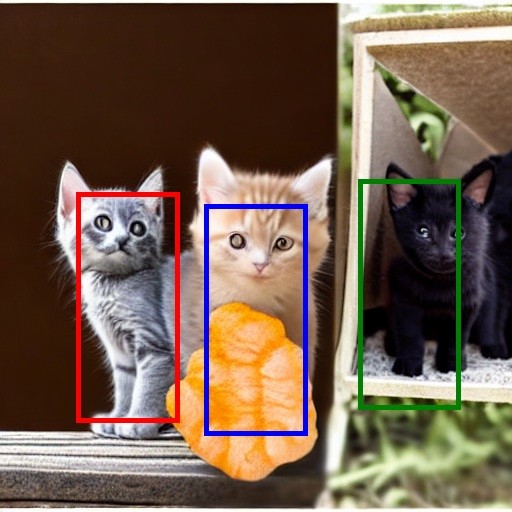} &
        { }&
        \includegraphics[width=0.115\linewidth]{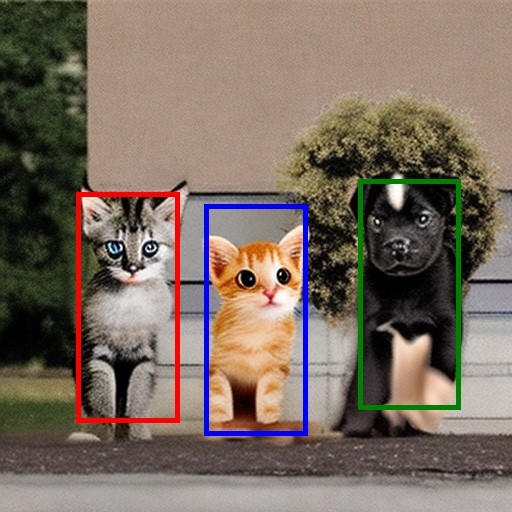} 
        \includegraphics[width=0.115\linewidth]{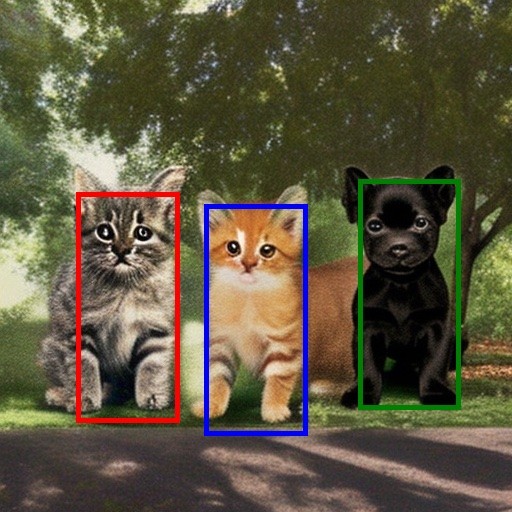} \\
        \includegraphics[width=0.115\linewidth]{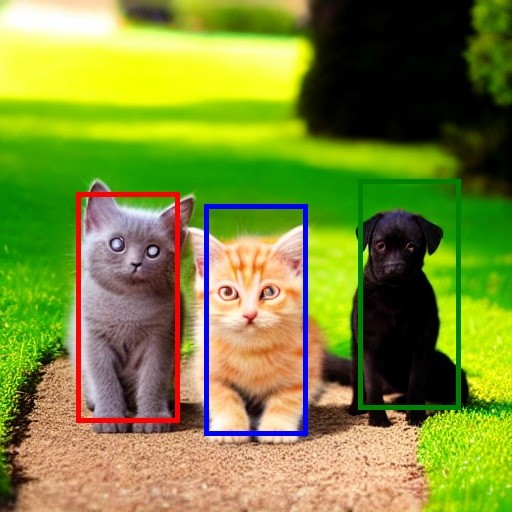} &
        \includegraphics[width=0.115\linewidth]{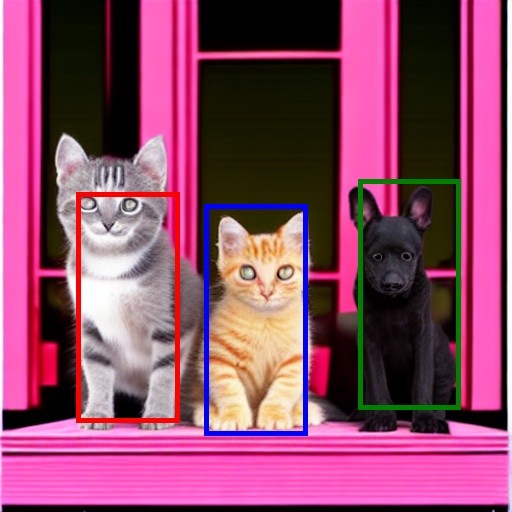} &
        { }&
        \includegraphics[width=0.115\linewidth]{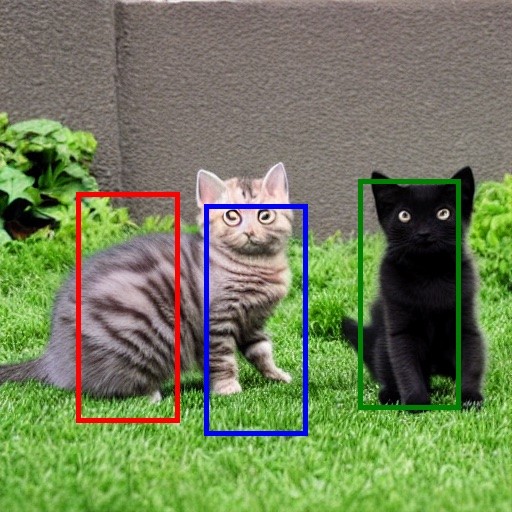} &
        \includegraphics[width=0.115\linewidth]{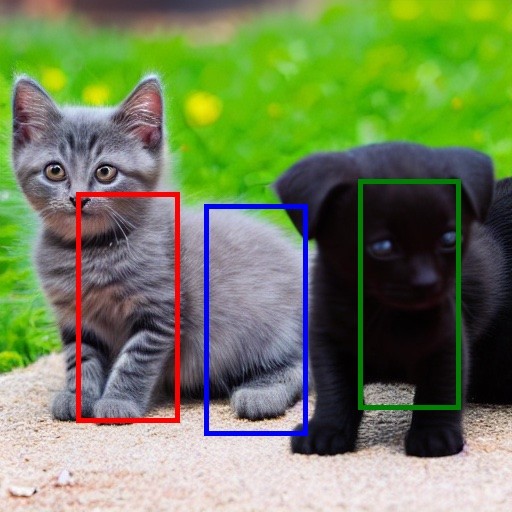} &
        { }&
        \includegraphics[width=0.115\linewidth]{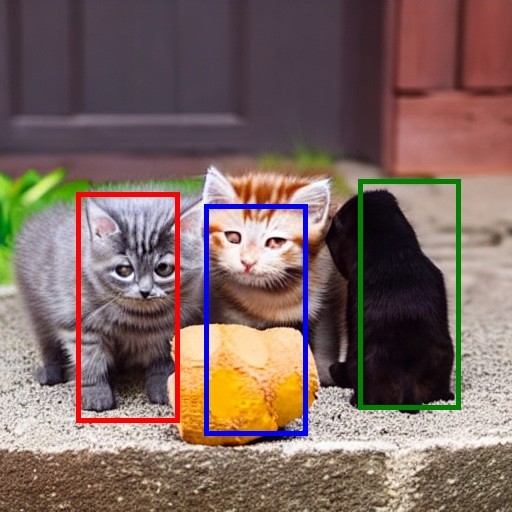} &
        \includegraphics[width=0.115\linewidth]{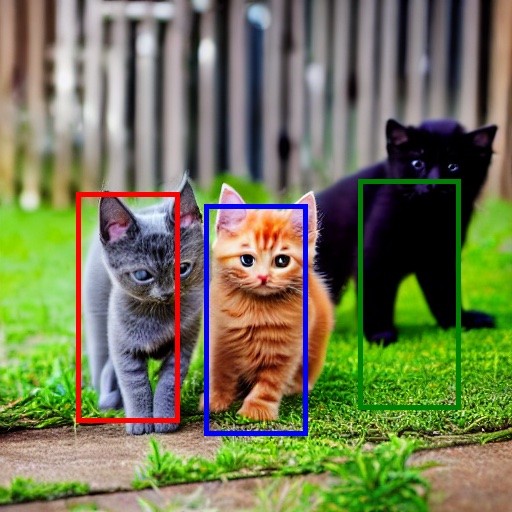} &
        { }&
        \includegraphics[width=0.115\linewidth]{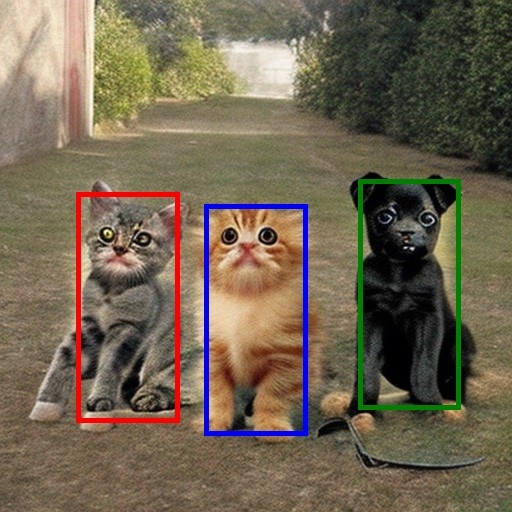} 
        \includegraphics[width=0.115\linewidth]{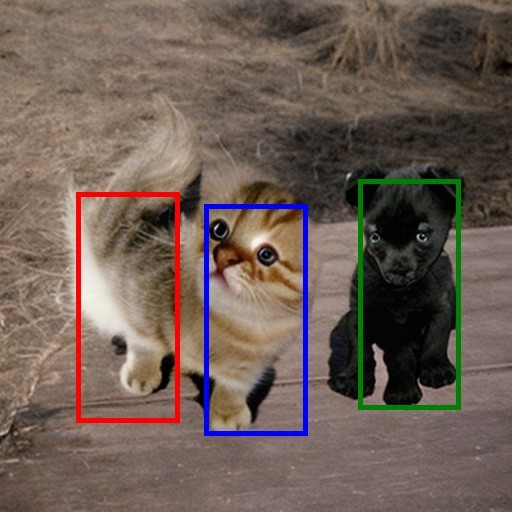} \\
        \includegraphics[width=0.115\linewidth]{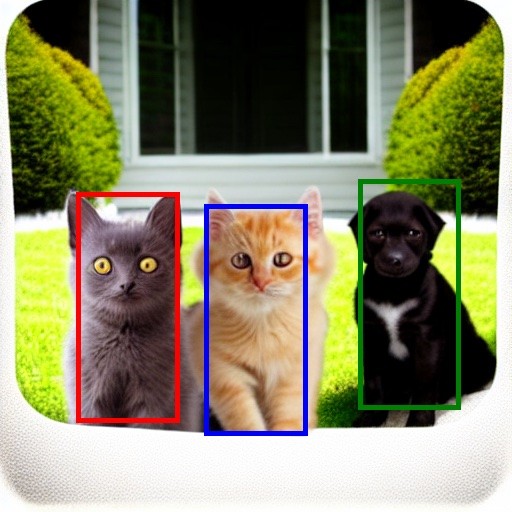} &
        \includegraphics[width=0.115\linewidth]{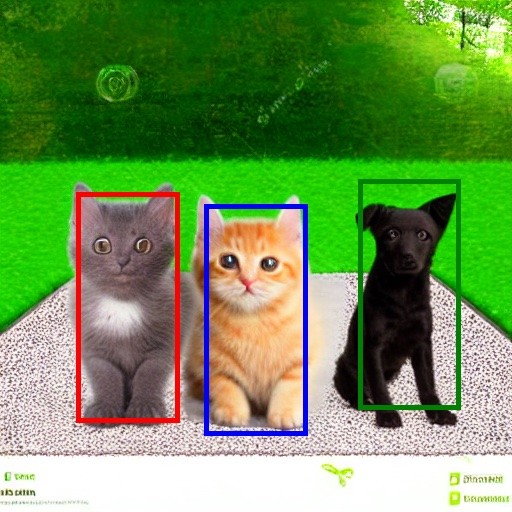} &
        { }&
        \includegraphics[width=0.115\linewidth]{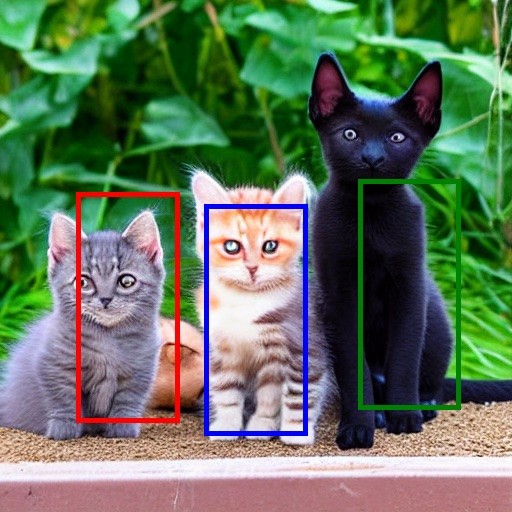} &
        \includegraphics[width=0.115\linewidth]{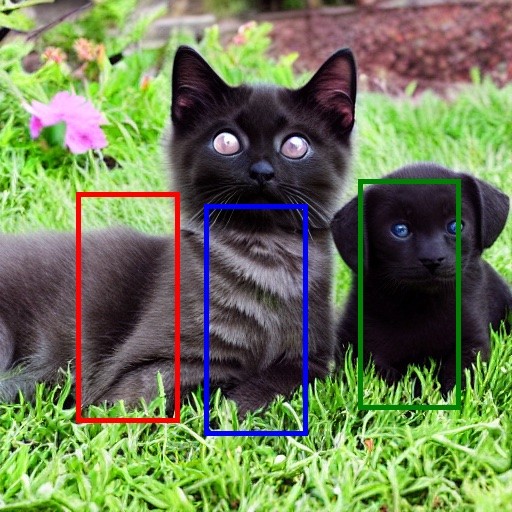} &
        { }&
        \includegraphics[width=0.115\linewidth]{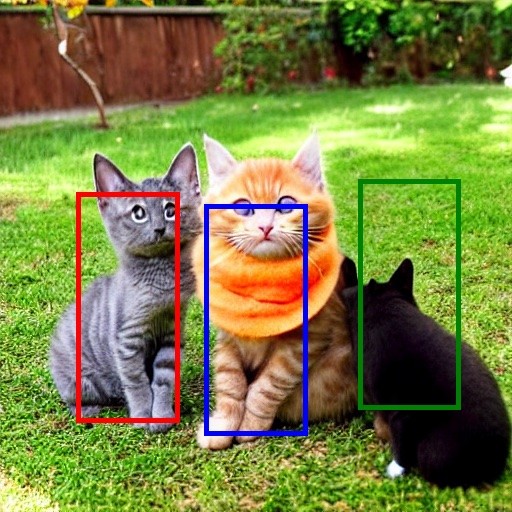} &
        \includegraphics[width=0.115\linewidth]{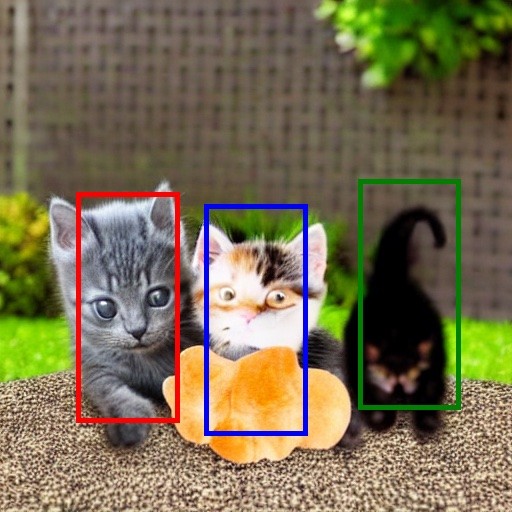} &
        { }&
        \includegraphics[width=0.115\linewidth]{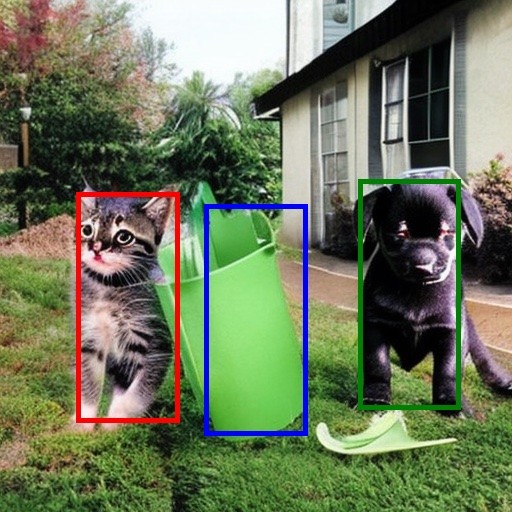} 
        \includegraphics[width=0.115\linewidth]{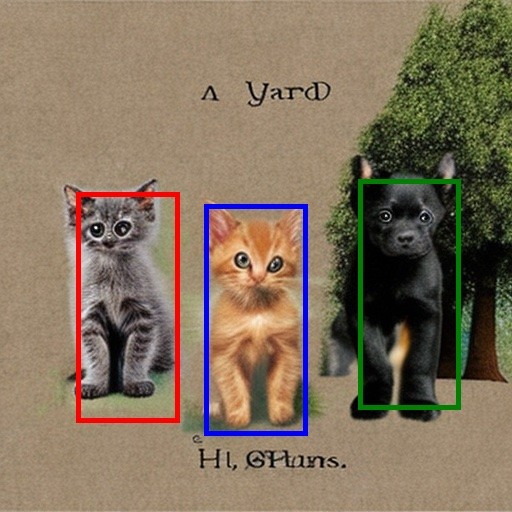} \\
        \multicolumn{2}{c}{Bounded Attention} & {}&
        \multicolumn{2}{c}{Layout Guidance} &{}&
        \multicolumn{2}{c}{BoxDiff} &{}&
        MultiDiffusion
        
    \end{tabular}
    }
    \vspace{-6pt}
    \captionof{figure}{
    Comparison of the first six images generated from the seed $0$. 
    }
    \label{fig:seed0}
\end{figure*}

%% file: figures/comparison.tex
\begin{figure*}[t]
    \centering
    \setlength{\tabcolsep}{0.001\textwidth}
    {\small
    \begin{tabular}{c c c c c c c}
        \multicolumn{7}{c}{``A \textcolor{red}{\textit{\underline{denim jacket}}} and a \textcolor{blue}{\textit{\underline{corduroy jacket}}} and a \textcolor{green}{\textit{\underline{leather handbag}}} in a closet.''} \\
        \includegraphics[width=0.14\textwidth]{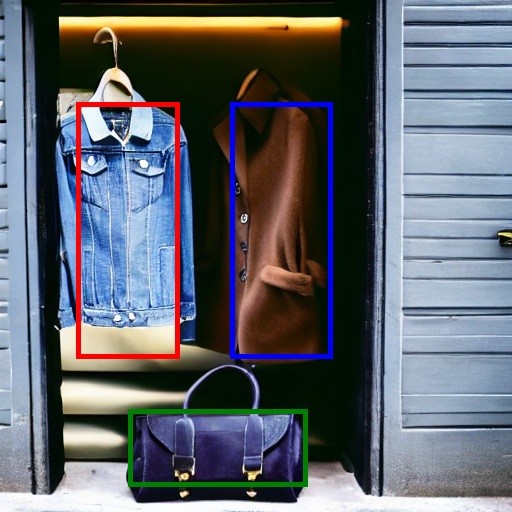} &
        \includegraphics[width=0.14\textwidth]
        {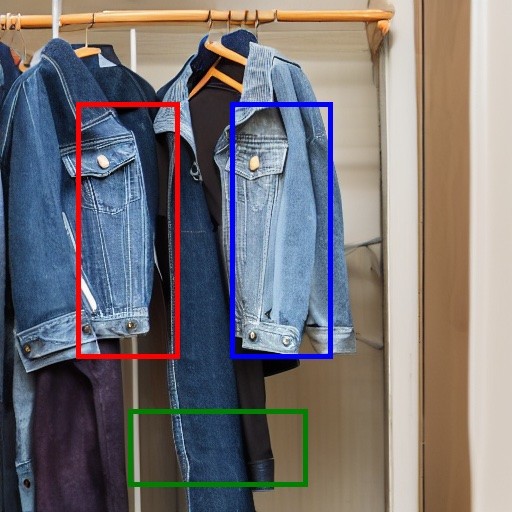} &
        \includegraphics[width=0.14\textwidth]
        {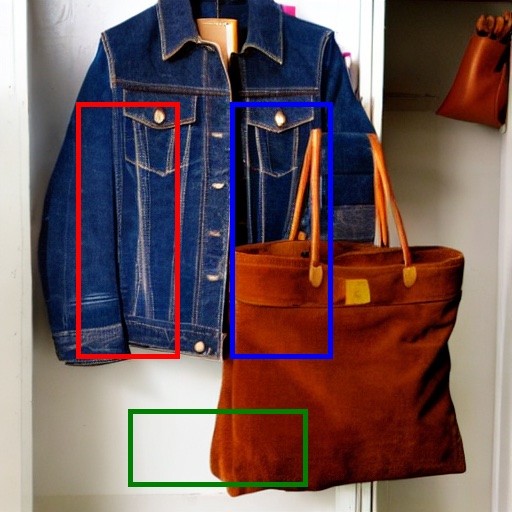} &
        \includegraphics[width=0.14\textwidth]
        {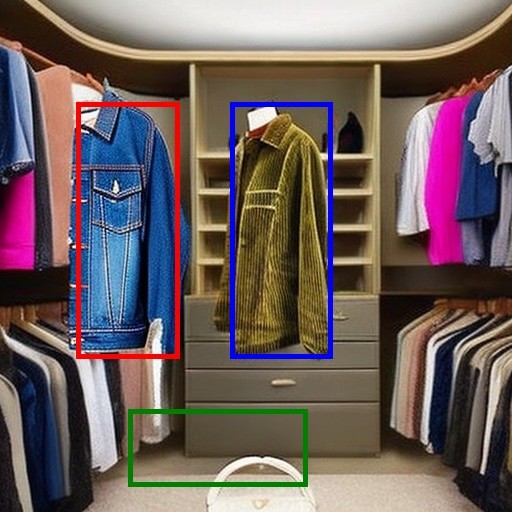} &
        \includegraphics[width=0.14\textwidth]
        {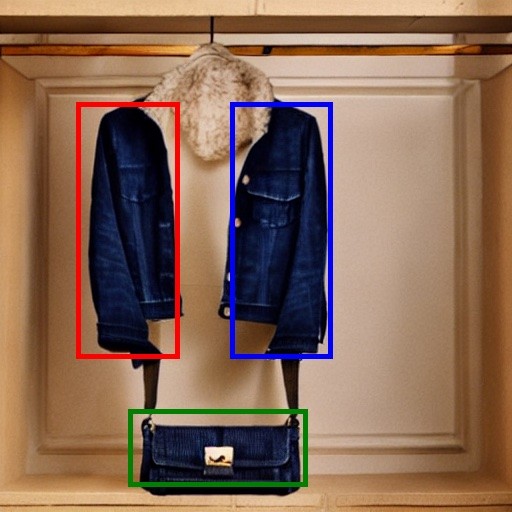} &
        \includegraphics[width=0.14\textwidth]
        {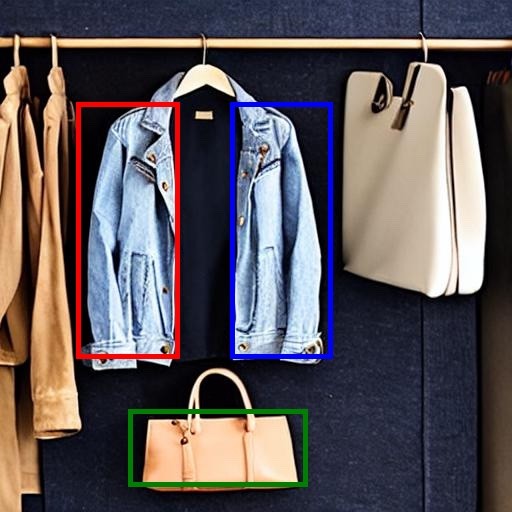} &
        \includegraphics[width=0.14\textwidth]
        {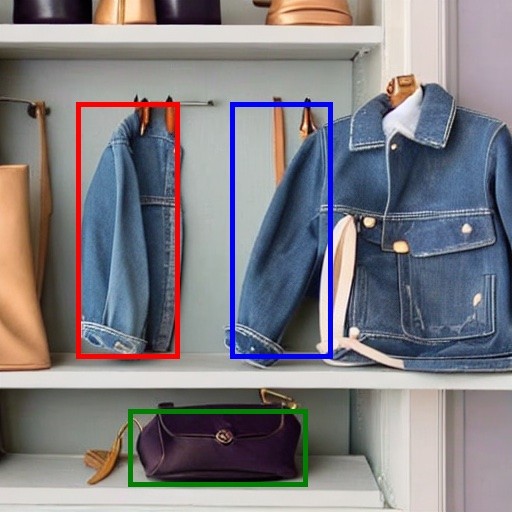} \\
        \multicolumn{7}{c}{``A \textcolor{red}{\textit{\underline{penguin riding}}} a \textcolor{blue}{\textit{\underline{white bear}}} in the north pole.''} \\
        \includegraphics[width=0.14\textwidth]{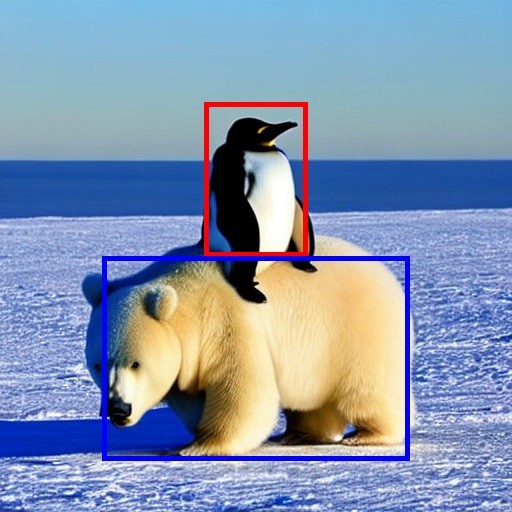} &
        \includegraphics[width=0.14\textwidth]
        {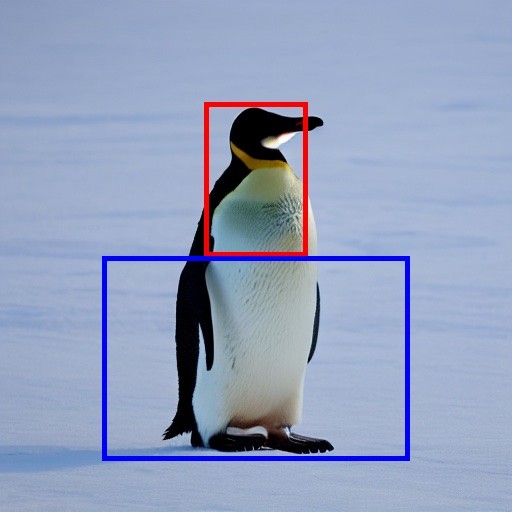} &
        \includegraphics[width=0.14\textwidth]
        {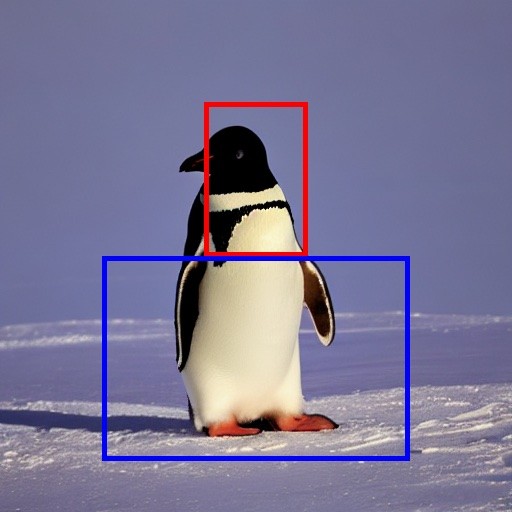} &
        \includegraphics[width=0.14\textwidth]
        {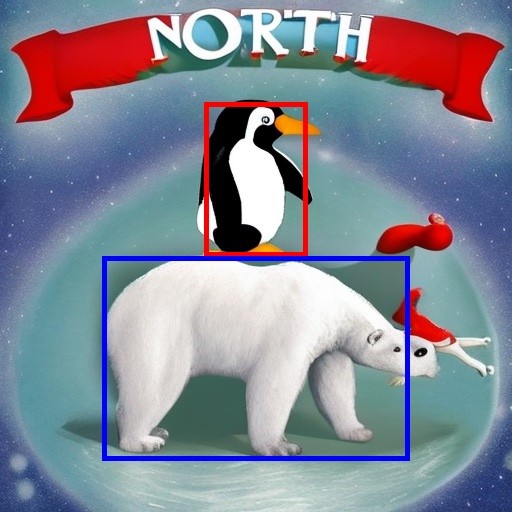} &
        \includegraphics[width=0.14\textwidth]
        {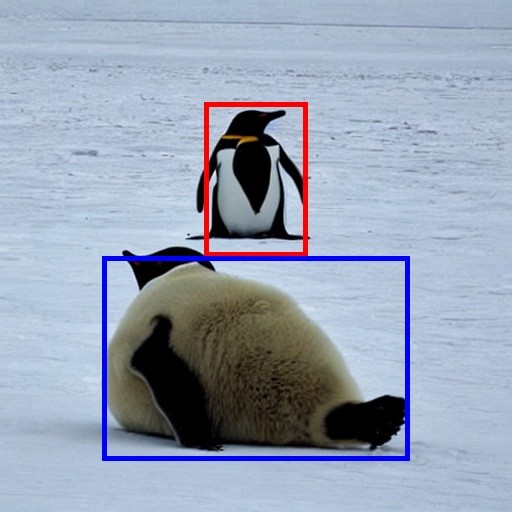} &
        \includegraphics[width=0.14\textwidth]
        {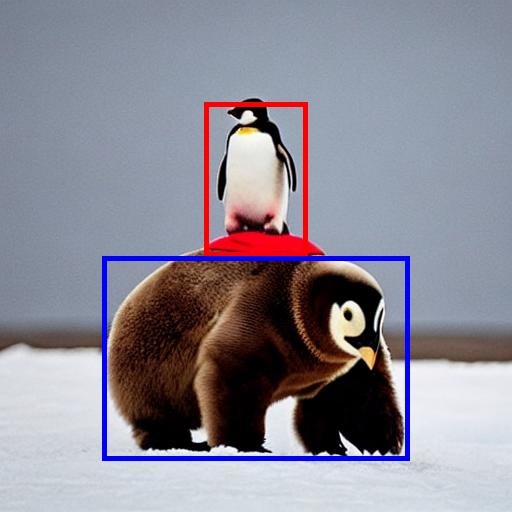} &
        \includegraphics[width=0.14\textwidth]
        {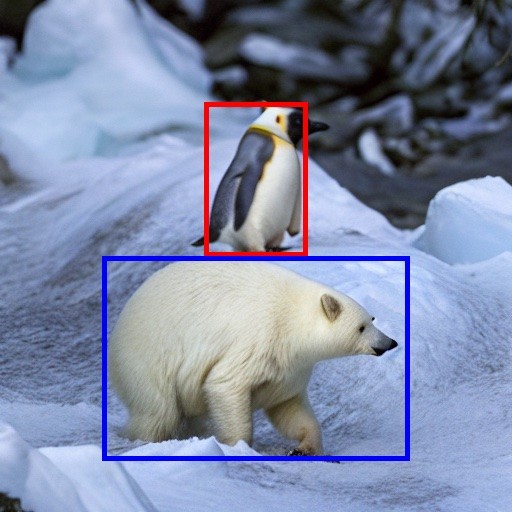} \\
        \multicolumn{7}{c}{``A \textcolor{red}{\textit{\underline{big red elephant}}} and a \textcolor{blue}{\textit{\underline{far away rhino}}} in a jungle.''} \\
        \includegraphics[width=0.14\textwidth]{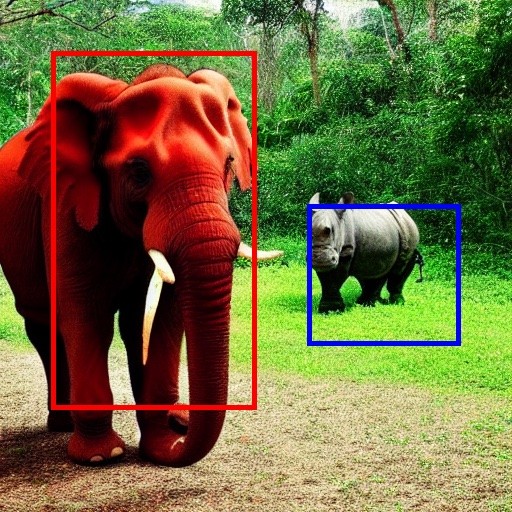} &
        \includegraphics[width=0.14\textwidth]
        {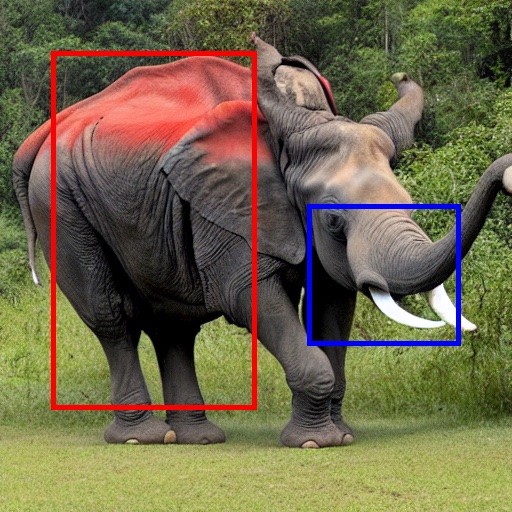} &
        \includegraphics[width=0.14\textwidth]
        {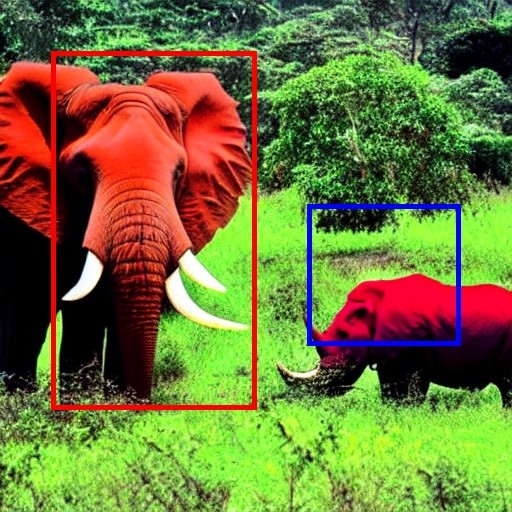} &
        \includegraphics[width=0.14\textwidth]
        {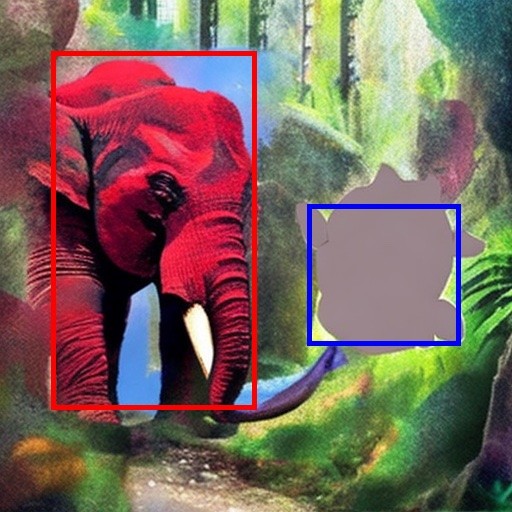} &
        \includegraphics[width=0.14\textwidth]
        {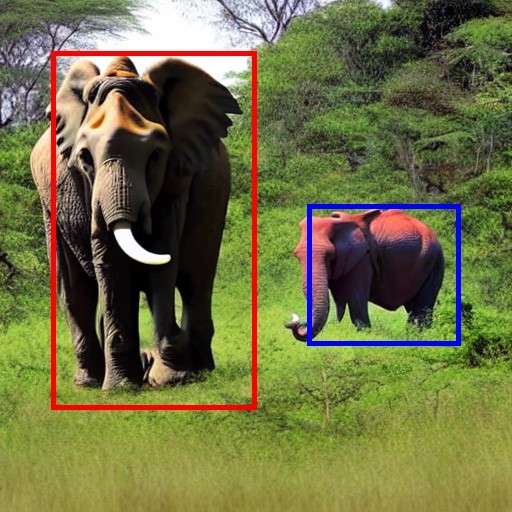} &
        \includegraphics[width=0.14\textwidth]
        {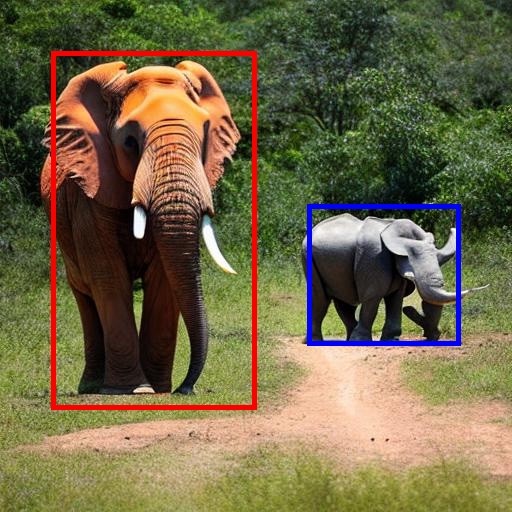} &
        \includegraphics[width=0.14\textwidth]
        {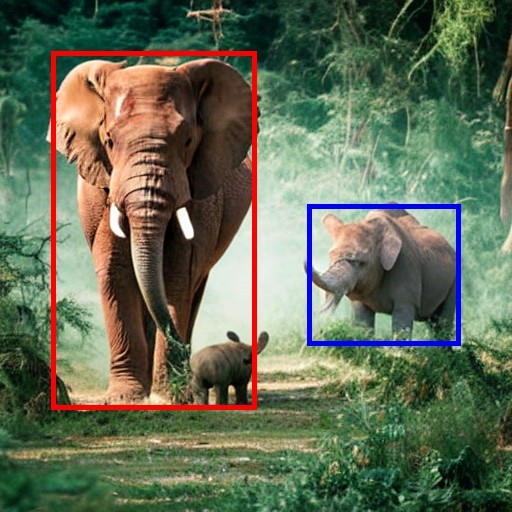} \\
        BA (ours) & LG~\cite{chen2023training} & BD~\cite{xie2023boxdiff} & MD~\cite{bar2023multidiffusion} & GLIGEN~\cite{li2023gligen} & AR~\cite{phung2023grounded} & ReCo~\cite{yang2023reco} \\
    \end{tabular}
    \vspace{-6pt}
    }
    \captionof{figure}{\textbf{Qualitative comparison} of our method with baseline methods: Above each row, we display the input prompt, where each subject's color matches the color of its corresponding bounding box. We compare with both training-free methods (2nd-4th columns) and trained models (5th-7th columns). See Appendix \ref{sec:more-results} for more results.}
    \vspace{-8pt}
    \label{fig:comparisons}
\end{figure*}

%% file: tables/drawbench.tex
\begin{table}
    \centering
    \setlength{\tabcolsep}{0.003\textwidth}
    \begin{tabular}{l c c c c}
        \toprule
        Method & \multicolumn{3}{c}{Counting} & Spatial \\
        \cmidrule{2-5}
        & Precision & Recall & F1 & Accuracy \\
        \midrule
        Stable Diffusion~\cite{rombach2022high} & 0.74 & 0.78 & 0.73 & 0.19 \\
        Layout-guidance~\cite{chen2023training} & 0.72 & 0.78 & 0.72 & 0.35 \\
        BoxDiff~\cite{xie2023boxdiff} & 0.81 & 0.78 & 0.76 & 0.28 \\
        MultiDiffusion~\cite{bar2023multidiffusion} & 0.70 & 0.55 & 0.57 & 0.15 \\
        Ours & \textbf{0.83} & \textbf{0.88} & \textbf{0.82} & \textbf{0.36} \\
        \bottomrule
    \end{tabular}
    \captionof{table}{
    Quantitative evaluation
    on the DrawBench dataset.
    }
    \label{table:drawbench}
\end{table}

%% file: tables/user_study.tex
\begin{table}
{\small
\centering
    \setlength{\tabcolsep}{0.003\textwidth}
    \begin{tabular}{l c c c c}
        \toprule
        & LG~\cite{chen2023training} & BD~\cite{xie2023boxdiff} & MD~\cite{bar2023multidiffusion} \\
        \midrule
        Our score vs. & 0.85 & 0.72 & 0.95\\
        \bottomrule
    \end{tabular}
    \captionof{table}{User study results.}
    \vspace{-16pt}
    \label{table:user-study}
    }
\end{table}

%% file: sub_sections/user_study.tex
\paragraph{\textbf{User study.}}

The automatic metrics utilized in
Table \ref{table:drawbench}
have limitations in capturing semantic leakage, as they rely on an object detector~\cite{wu2019detectron2} trained on real images, where such erroneous leakage between subjects is absent. To address this issue, we conducted a user study.

For a comprehensive evaluation, we enlisted ChatGPT to provide us with five pairs of visually similar yet distinctly recognizable animals, along with a suitable background prompt and four different layouts. Subsequently, for each subject pair and its corresponding background prompt, we generated six non-curated images for each layout using our training-free baseline. For each prompt-layout condition, users were presented with an image set comprising the six images generated by our method and six images produced by a competing method. Users were then tasked with selecting realistic images that accurately depicted the prompt and layout, devoid of any leakage between subjects.

In total, responses were collected from 32 users, resulting in 330 responses. The results of the user study are summarized in Table~\ref{table:user-study}, where we report the estimated conditional probability of a selected image being generated by our method compared to corresponding competing methods.

%% file: figures/ablation.tex
\begin{figure*}
    \centering
    \setlength{\tabcolsep}{0.002\textwidth}
    {\small
    \begin{tabular}{c c c c c c c c}
        &
        \multicolumn{7}{c}{``A \textcolor{red}{\textit{\underline{red lizard}}} and a \textcolor{blue}{\textit{\underline{turtle}}} on the grass.''} \\
        \raisebox{22pt}{\rotatebox{90}{SDXL}} &
        \includegraphics[width=0.135\textwidth]{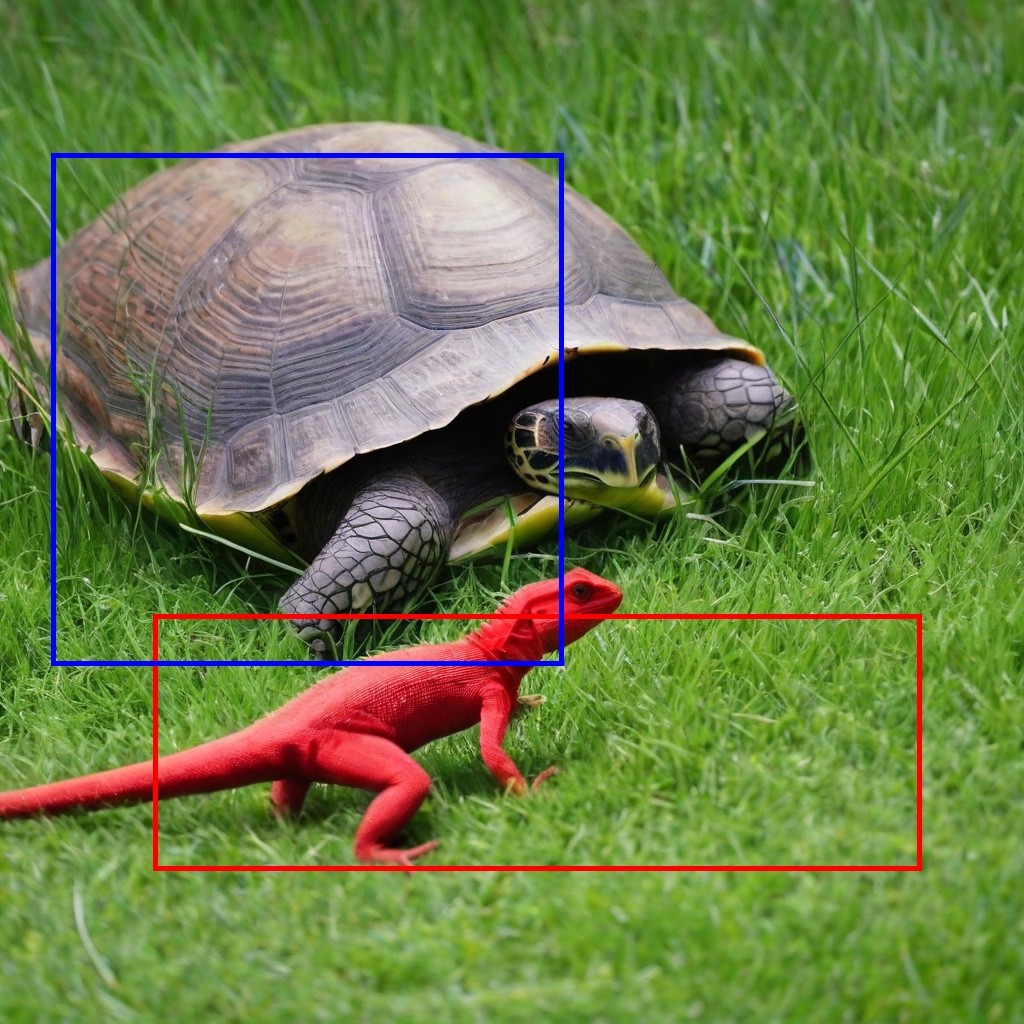}  &
        \includegraphics[width=0.135\textwidth]{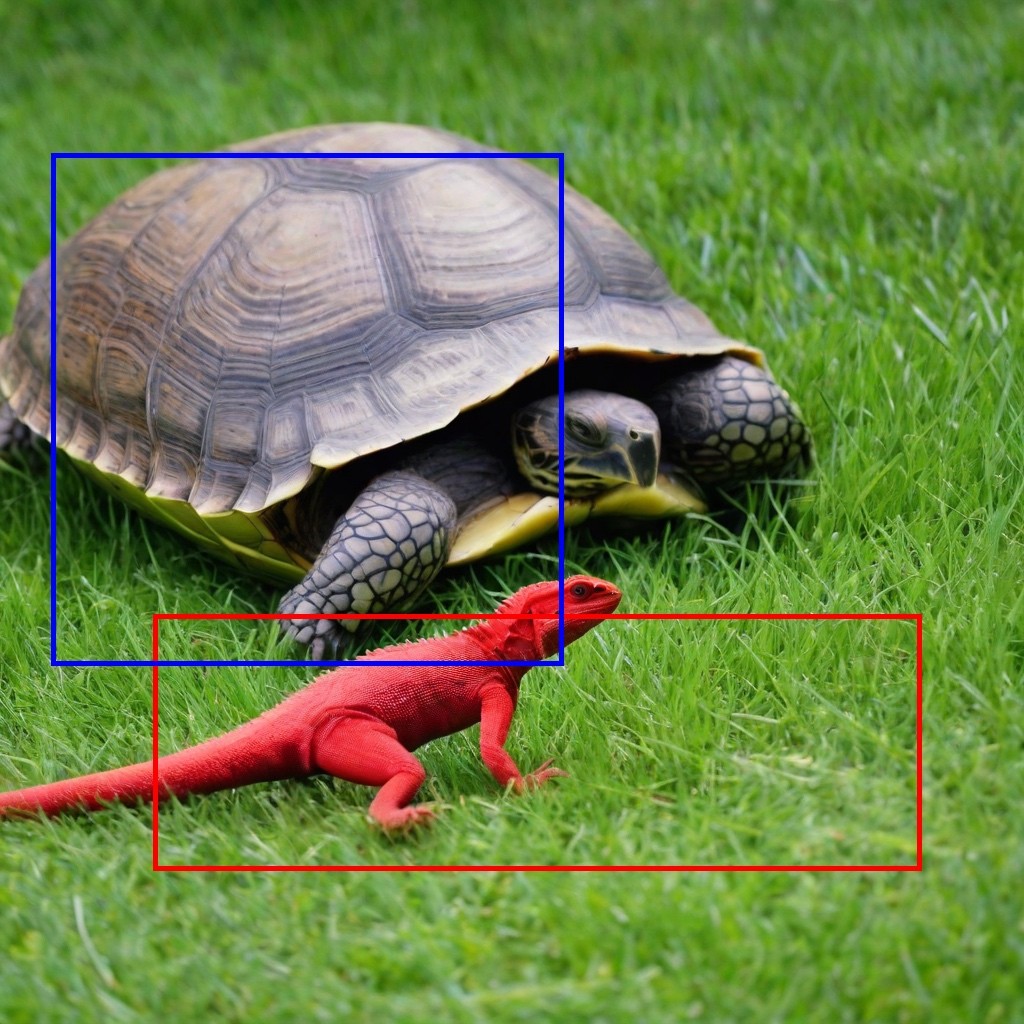} &
        \includegraphics[width=0.135\textwidth]{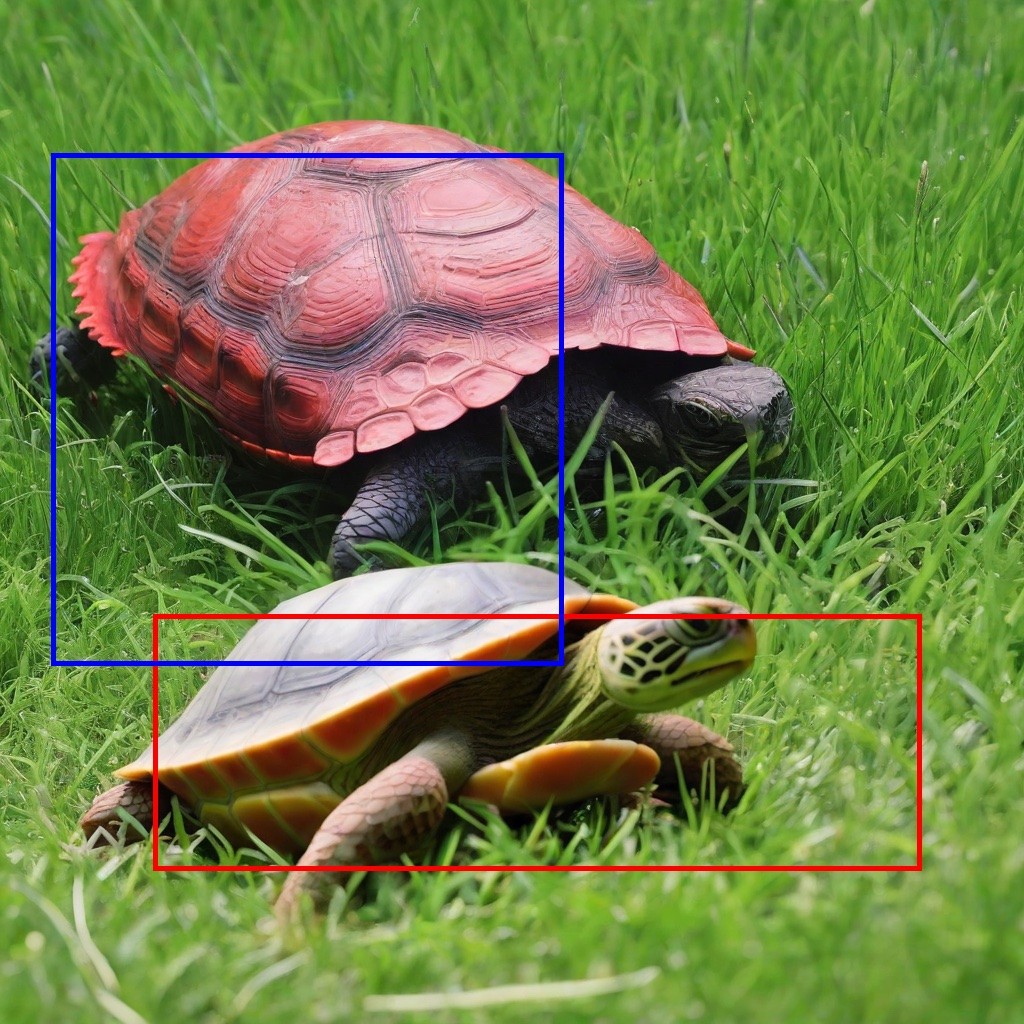} &
        \includegraphics[width=0.135\textwidth]{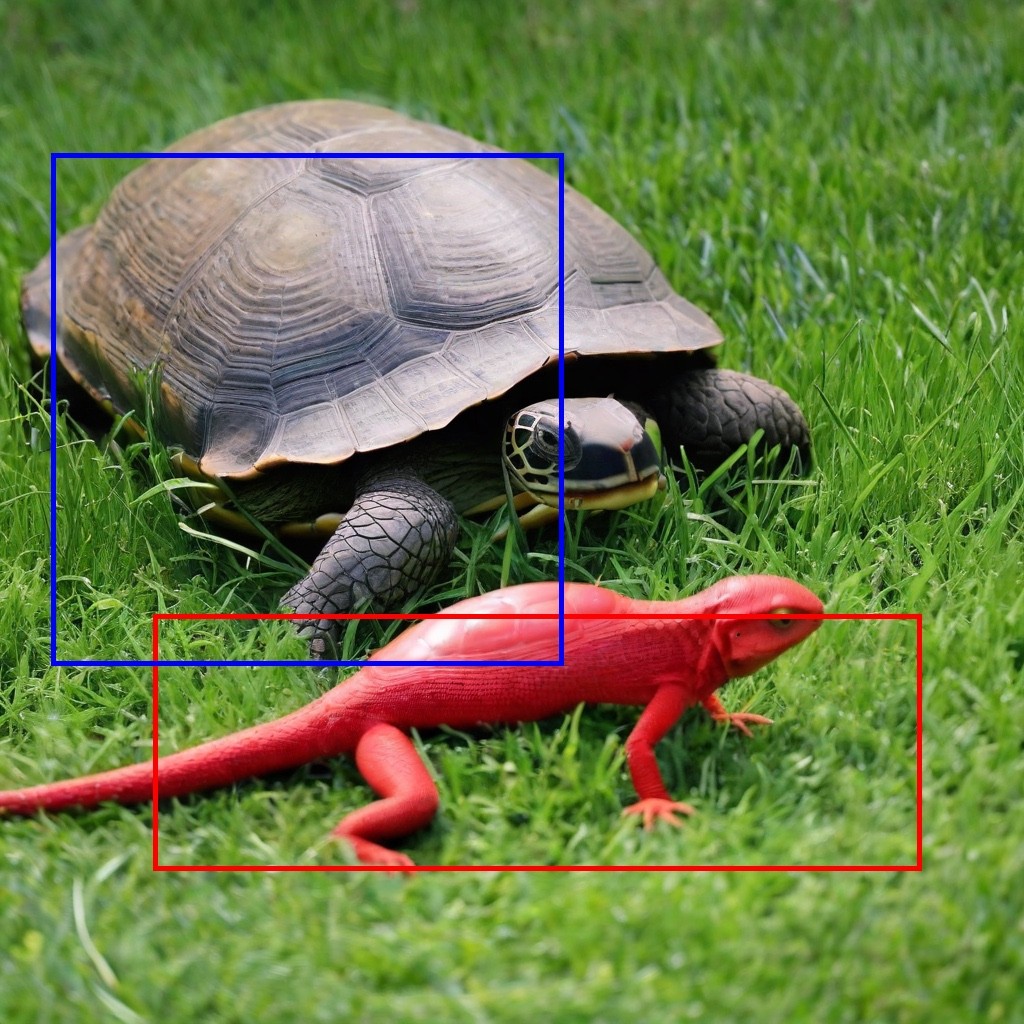} &
        \includegraphics[width=0.135\textwidth]{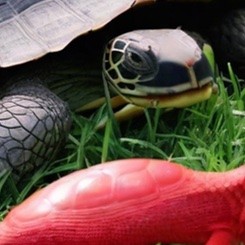} &
        \includegraphics[width=0.135\textwidth]{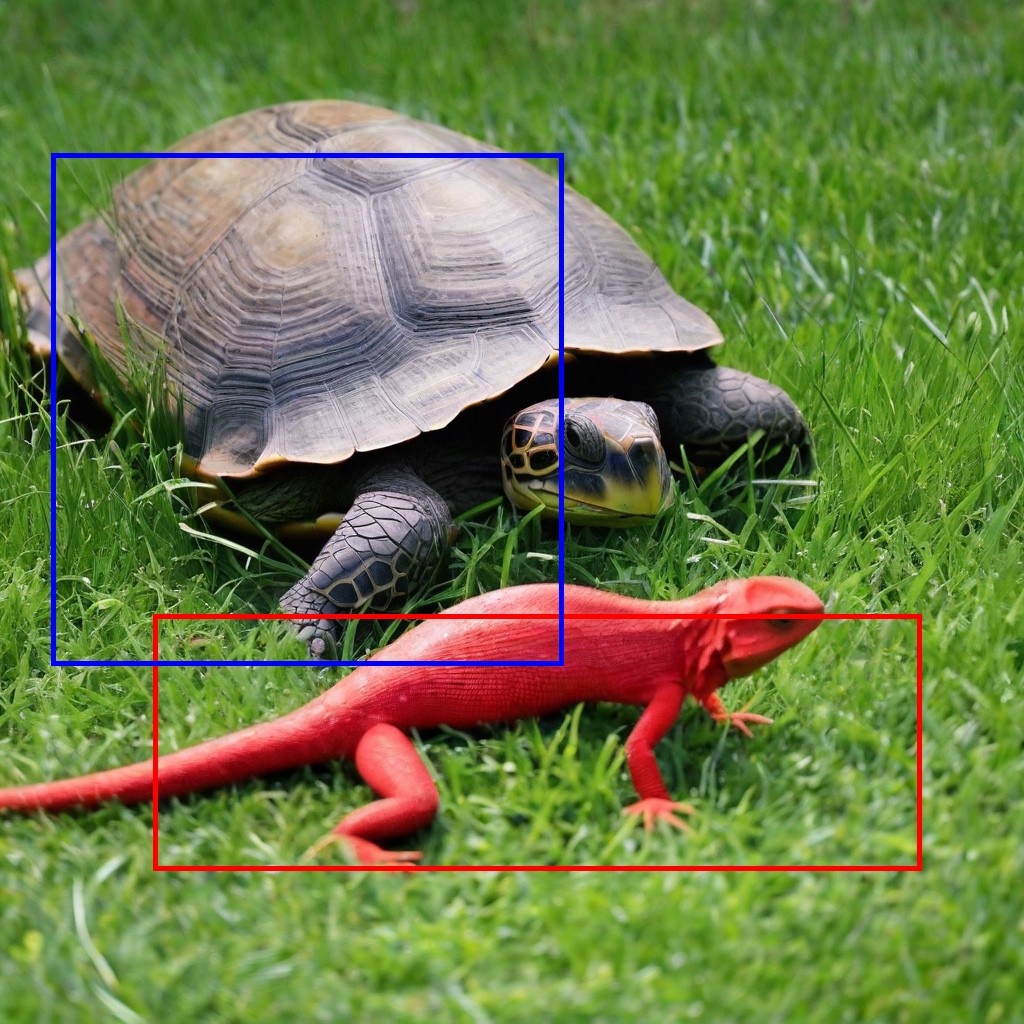} &
        \includegraphics[width=0.135\textwidth]{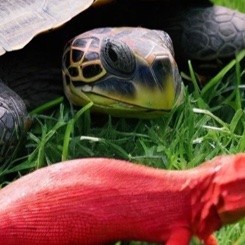} 
        \\
        \\
        &
        \multicolumn{7}{c}{``A \textcolor{red}{\textit{\underline{ginger kitten}}} and a \textcolor{blue}{\textit{\underline{gray puppy}}} on the front stairs.''} \\
        \raisebox{4pt}{\rotatebox{90}{Stable Diffusion}} &
        \includegraphics[width=0.135\textwidth]{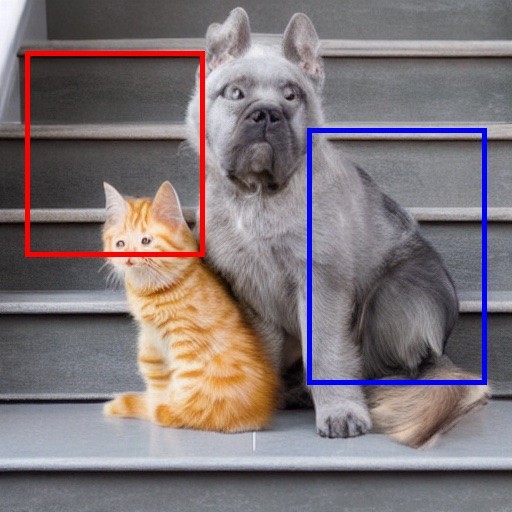}  &
        \includegraphics[width=0.135\textwidth]{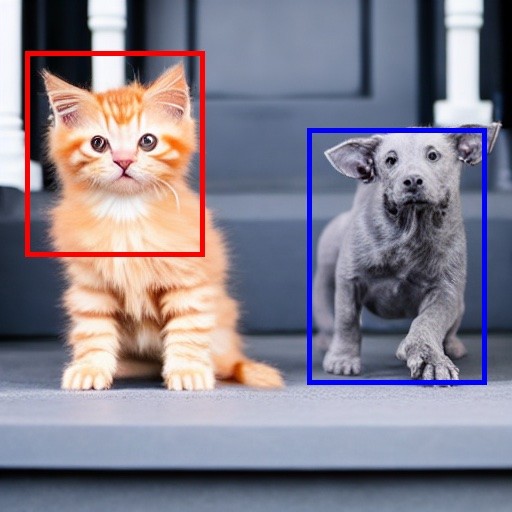} &
        \includegraphics[width=0.135\textwidth]{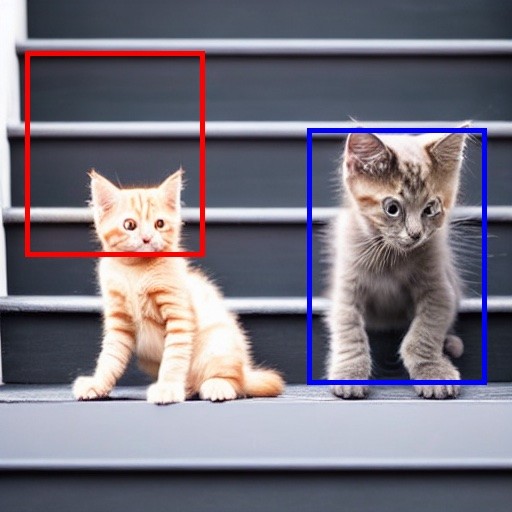} &
        \includegraphics[width=0.135\textwidth]{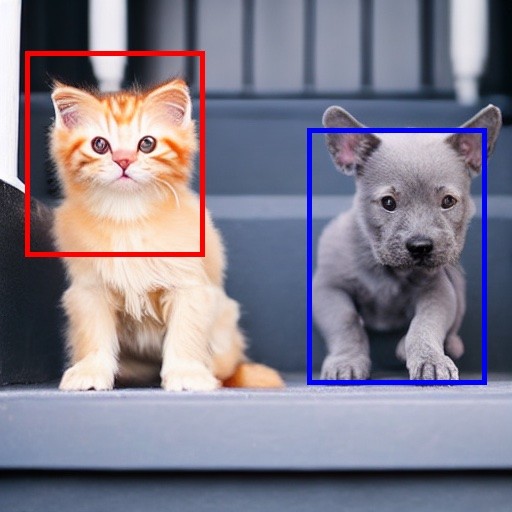} &
        \includegraphics[width=0.135\textwidth]{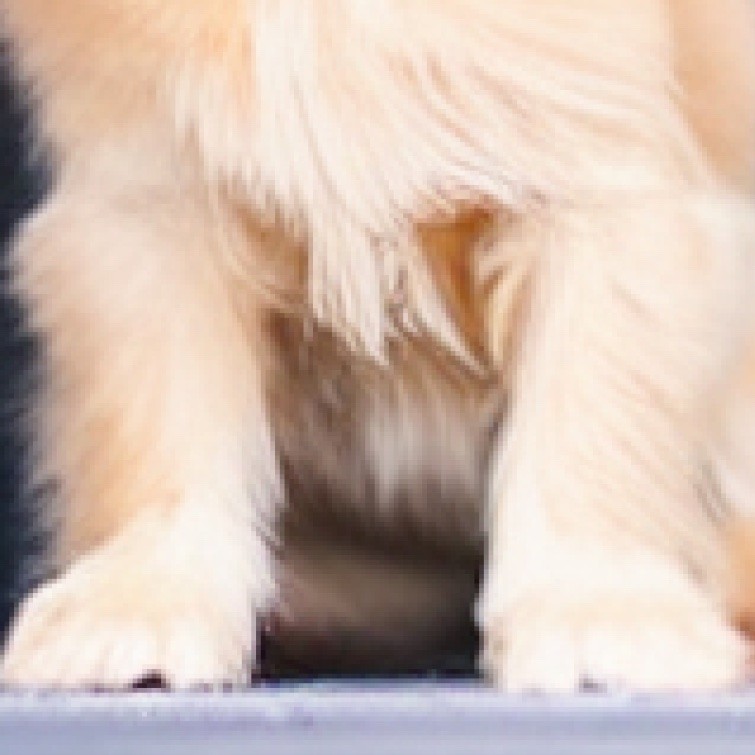} &
        \includegraphics[width=0.135\textwidth]{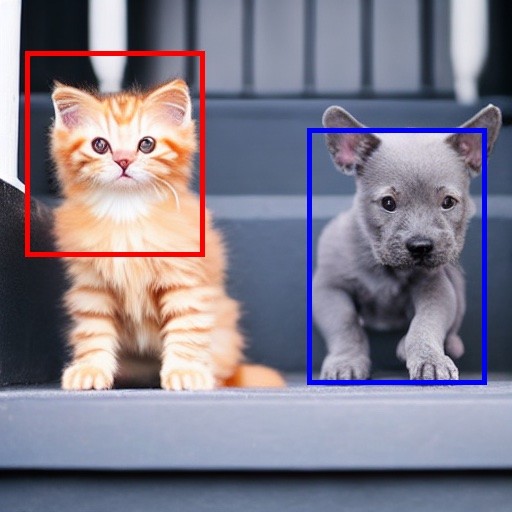} &
        \includegraphics[width=0.135\textwidth]{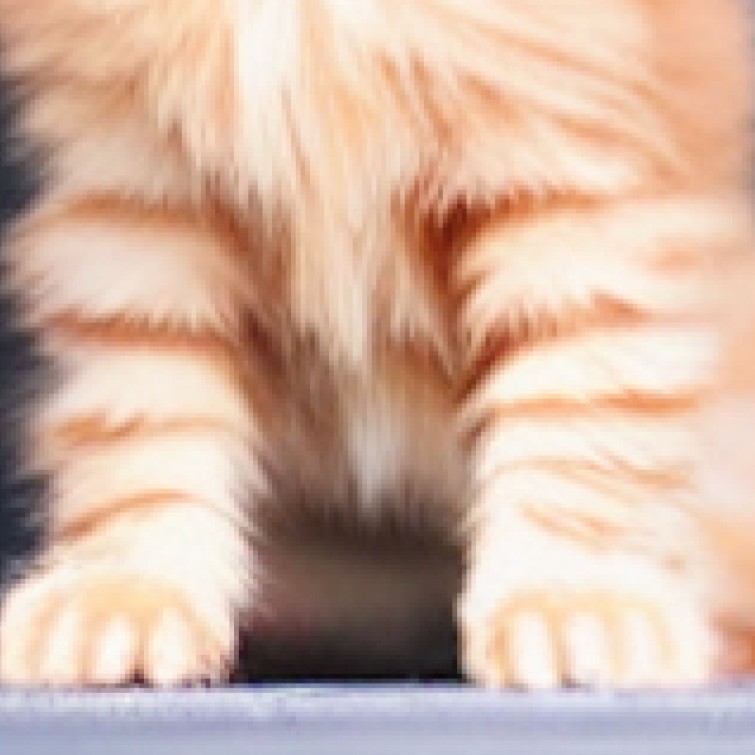} \\
        &
        {w/o G} &
        {w/o BG} &
        {w/o BD} &
        \multicolumn{2}{c}{w/o MR} &
        \multicolumn{2}{c}{Full Method} \\
    \end{tabular}
    }
    \captionof{figure}{\textbf{Qualitative ablation.} We ablate our method by skipping the guidance step (G),
    performing guidance without Bounded Guidance (BG), not applying Bounded Denoising (BD), and not performing Mask Refinement (MR). We show zoomed-in images of the two rightmost configurations.}
    \label{fig:ablation}
\end{figure*}

%% file: 7-conclusions.tex
\section{Conclusions}

We introduce Bounded Attention, a technique designed to regulate the accurate generation of multiple subjects within an image. This approach encourages each subject to ``be yourself'', emphasizing the importance of preserving individuality and uniqueness without being excessively influenced by other subjects present in the image.
Our development of the Bounded Attention technique stemmed from an in-depth analysis of the root causes behind the misalignment observed between the provided prompt and the resulting generated image. Our investigation revealed that this misalignment primarily arises due to semantic leakage among the generated subjects, a phenomenon observed in both the cross and self-attention layers.

While Bounded Attention effectively mitigates a significant portion of semantic leakage, it does not entirely eliminate it. Our findings demonstrate a marked improvement in performance compared to other methods that seek to achieve semantic alignment. However, residual leakage persists, which we attribute to imperfect optimization during the guidance mode and inaccurate segmentation of the subject prior to the second phase.

While Bounded Attention excels in generating multiple subjects with plausible semantic alignment, its performance may vary across different layouts. Achieving success with Bounded Attention hinges on a strong match between the seed and the layout. Moving forward, we aim to explore methods for generating well-suited seeds tailored to specific layouts. One potential avenue is to introduce noise to the layout image, thereby creating a seed that aligns more closely with the desired outcomes.

\section*{Acknowledgement}

We thank Guy Tevet for fruitful discussions and useful suggestions. 
This work is partly supported by a research gift from Snap Inc.

%% file: 8-supp.tex
\section{Method Details} \label{sec:app-method-details}
\input{sub_sections/bounded_attention}

\subsection{Subject Mask Refinement}

To derive the segmentation masks in $\left[T_{\textit{guidance}},0\right]$, we employ a technique akin to self-segmentation, previously introduced for localized image editing~\cite{patashnik2023localizing}.

In self-segmentation, first, all self-attention maps are averaged across heads, layers and timesteps. Then, each pixel is associated with its corresponding averaged self-attention map, which serves as input for a clustering method, such as KMeans. Each cluster is labeled as belonging to the background or to a specific subject's noun, by aggregating and comparing the corresponding cross-attention maps of these nouns within each cluster.

When adapting this technique for generation, we encounter the challenge of lacking the more reliable late-stage attention maps. To address this, we observe that self-attention maps formed at the UNet's middle block and first up-block are more robust to noise. Therefore, starting at $T_{\textit{guidance}}$, we cluster the time-specific self-attention maps, averaging only at the mentioned layers.

For labeling, we initially compute cross-attention masks $\mathbf{M}^{\textit{cross}}_i$ for each subject $s_i$, by applying a soft threshold to the subject's cross-attention maps~\cite{epstein2024diffusion},

\begin{equation}
\mathbf{M}^{\textit{cross}}_i =
\operatorname{norm}\left(
\operatorname{sigmoid}\left(
s \cdot \operatorname{norm}\left(
\hat{\mathbf{A}}^{\textit{cross}}_i
\right)
-
\sigma_{\textit{noun}}
\right)
\right).
\label{eq:cross_masks}
\end{equation}

Here, $\hat{\mathbf{A}}^{\textit{cross}}_i$ is the mean Boudned Cross-Attention map of the subject's last noun in the prompt, averaged across heads and layers, and $\operatorname{norm}$ denotes the $\mathcal{L}_1$ normalization operation. We use hyperparameters $s,\sigma_{\textit{noun}}$, where $\sigma_{\textit{noun}}$ defines the soft threshold, and $s$ controls the binarization sharpness.

Then, we calculate the Intersection over Minimum (IoM) between each self-attention cluster $\mathbf{C}^{\textit{self}}_j$ and cross-attention mask $\mathbf{M}^{\textit{cross}}_i$,

\begin{equation}
\operatorname{IoM}\left(i,j\right)=
\frac{
\sum_\mathbf{x}
\left(\mathbf{M}^{\textit{cross}}_i \left[ \mathbf{x} \right]
\cdot
\mathbf{C}^{\textit{self}}_j \left[ \mathbf{x} \right]
\right)
}{
\min \left\{
\sum_\mathbf{x} \mathbf{M}^{\textit{cross}}_i \left[ \mathbf{x} \right],
\sum_\mathbf{x} \mathbf{C}^{\textit{self}}_j \left[ \mathbf{x} \right]
\right\}
}.
\end{equation}

For each cluster $\mathbf{C}^{\textit{self}}_j$, we determine the subject index with the highest $\operatorname{IoM}$, $i_{\textit{max}}\left(j\right)=\arg\max_{i} \operatorname{IoM}\left(i,j\right)$. We assign cluster $\mathbf{C}^{\textit{self}}_j$ to subject $s_{i_{\textit{max}}\left(j\right)}$, if $\operatorname{IoM}\left( {i_{\textit{max}}\left(j\right)}, j \right) \geq \sigma_{\textit{cluster}}$, where $\sigma_{\textit{cluster}}$ is a hyperparameter.

We repeat this process at uniform time intervals to enable continuous evolution of the shapes. To maintain real-time performance, we execute KMeans directly on the GPU. For temporal consistency and faster convergence, we initialize the cluster centers with the centers computed at the previous interval.

\subsection{Implementation Details}

In our experiments, we utilized 50 denoising steps with $T_{\textit{guidance}}=0.7$, conducting 5 Gradient Descent iterations for each Bounded Guidance step.
In Eq. \ref{eq:guidance-loss},
we set $\alpha$ to the number of input subjects. The loss computation was performed using attention layers 12 to 19 in SD and 70 to 81 in SDXL.

For the step size
    $\beta$ in Eq. \ref{eq:guidance},
we initiated the optimization process with $\beta \in \left[8,15\right]$ for SD and $\beta \in \left[10,25\right]$ for SDXL, with higher values utilized for more challenging layouts. We employed a linear scheduler, concluding the optimization with $\beta \in \left[2,5\right]$ for SD and $\beta \in \left[5,10\right]$ for SDXL. Early stopping was applied if the average loss between subjects dropped to $0.2$ or lower.

Subject mask refinement was conducted every 5 steps, computing the cross and self-attention masks from attention layers 14 to 19 in SD and 70 to 81 in SDXL. We set $s=10$ and $\sigma_{\textit{noun}}=0.2$ when computing the cross-attention masks in Eq. \ref{eq:cross_masks}. Additionally, when running KMeans, the number of self-attention clusters was set to 2 to 3 times the number of input subjects, with a threshold of $\sigma_{\textit{cluster}}=0.2$.

\section{Additional Results} \label{sec:more-results}

\input{figures/failures/full_figure}

\subsection{Failure Cases}

In Figure \ref{fig:failures},
we present two failure cases of our method. In these instances, we attempted to generate five similar stringed instruments within a challenging layout. Although each bounding box contains the correct instrument, the generated images do not accurately adhere to the prompt.

In the left image, the two guitars appear as paintings on the wall, deviating from the intended representation. Similarly, in the right image, redundant parts of instruments are generated at the margins of the bounding boxes, leading to inconsistencies in the output.

\subsection{Qualitative Results}

\paragraph{\textbf{Comparisons with baselines.}}

\input{figures/comparison2}

In Figure \ref{fig:comparisons2}, we present additional qualitative comparison results. While all competing methods fail to accurately generate all subjects from the prompt, our method successfully preserves each subject's intended semantics. For example, in the
third
row, none of the methods accurately generate the correct type of pasta in each dinnerware, and in the fourth row, none achieve the correct number of watercolor paintings.

\paragraph{\textbf{SDXL results.}}

\input{figures/sup/sdxl_results3}
\input{figures/sup/sdxl_results4}

We present additional results in SDXL in Figures
\ref{fig:sdxl3} and \ref{fig:sdxl4}. In these figures, Vanilla SDXL exhibits evident semantic leakage. For example, in Figure
\ref{fig:sdxl3}, it doesn't accurately generate all subjects with their intended modifiers. In the top row, a turtle is depicted with octopus tentacles in the second-to-right column. In the middle row, the green modifier intended for the alien spills over into the background. Lastly, none of the images in the last row include orchids as intended.

%% file: sub_sections/bounded_attention.tex
\subsection{Bounded Cross-Attention}

In the following, we elaborate on our design choices regarding the implementation of Bounded Attention into cross-attention layers. 

First, we have observed that tokens in the prompt $y$ that do not correspond to attributes of the subjects or the subjects themselves can unexpectedly result in leakage between subjects within the image.

Specifically, we found that conjunctions, positional relations, and numbers, carry semantics related to other subjects around them. Consequently, they can freely introduce conflicting features into the background and subjects, disrupting the intended layout. Given that vision-language models like CLIP struggle with encapsulating compositional concepts~\cite{paiss2023teaching}, such tokens are generally irrelevant for multi-subject generation, prompting us to exclude them after computing the token embeddings. We achieve this automatically using an off-the-shelf computational POS tagger.
Moreover, padding tokens can significantly impact the resulting layout. Specifically, the first [EoT] token typically attends to the image's foreground subjects in SD and can even generate subjects on its own~\cite{chen2023training,zhao2023loco}. However, it remains crucial for subject generation, particularly those specified last in the prompt~\cite{tunanyan2023multi}. We also observe similar behavior in SDXL. To address this, we confine this token's attention within the union of bounding boxes.

Second, our masking strategy in Bounded Attention for cross-attention layers exhibits slight variations between Bounded Guidance and Bounded Denoising.
In Bounded Denoising, we confine each subject token in $s_i$ to interact solely with its corresponding box pixels $b_i$.
Conversely, in Bounded Guidance we employ a similar strategy but allow all tokens to attend to the background. Thus, the two modes complement each other: Bounded Guidance refines the random latent signal to achieve a reasonable initial alignment with the input layout, while Bounded Denoising enforces the layout constraints.

\subsection{Bounded Self-Attention}

In self-attention layers, we use the same masking scheme for both Bounded Guidance and Bounded Denoising modes. Specifically, for each subject $s_i$, we exclude all keys corresponding to $b_j$ where $j\neq i$. We found that enabling subjects to interact with the background is crucial for generating natural-looking images, where the subjects seamlessly blend into the scene. Subjects should attend to the background to maintain high visual quality, as overly restrictive masking can lead to degradation, especially for small bounding boxes. Moreover, the background should attend to the subjects to integrate them into the environment and facilitate lighting effects such as shadows and reflections.

In contrast to Bounded Cross-Attention, we apply Bounded Self-Attention both to the prompt-conditioned noise estimation $\epsilon_\theta\left(\mathbf{z}_t,y,t\right)$, and its unconditional counterpart $\epsilon_\theta\left(\mathbf{z}_t,\phi,t\right)$. We observed that excluding the unconditional noise estimation introduces significant disparities between the conditional and unconditional directions in classifier-free guidance, consequently leading to noticeable artifacts.

%% file: figures/failures/full_figure.tex
\input{figures/failures/figure}

%% file: figures/failures/figure.tex
\begin{figure}
    \centering
    \setlength{\tabcolsep}{0.002\textwidth}
    {\scriptsize
    \centering
    \begin{tabular}{c c c}
        \multicolumn{3}{c}{\small ``A \textcolor{red}{\textit{\underline{classic guitar}}} and an \textcolor{blue}{\textit{\underline{electric guitar}}} and a \textcolor{green}{\textit{\underline{cello}}}} \\
        \multicolumn{3}{c}{\small  and a \textcolor{orange}{\textit{\underline{violin}}} on the wall of a music shop.''} \\
        \includegraphics[width=0.2\textwidth]{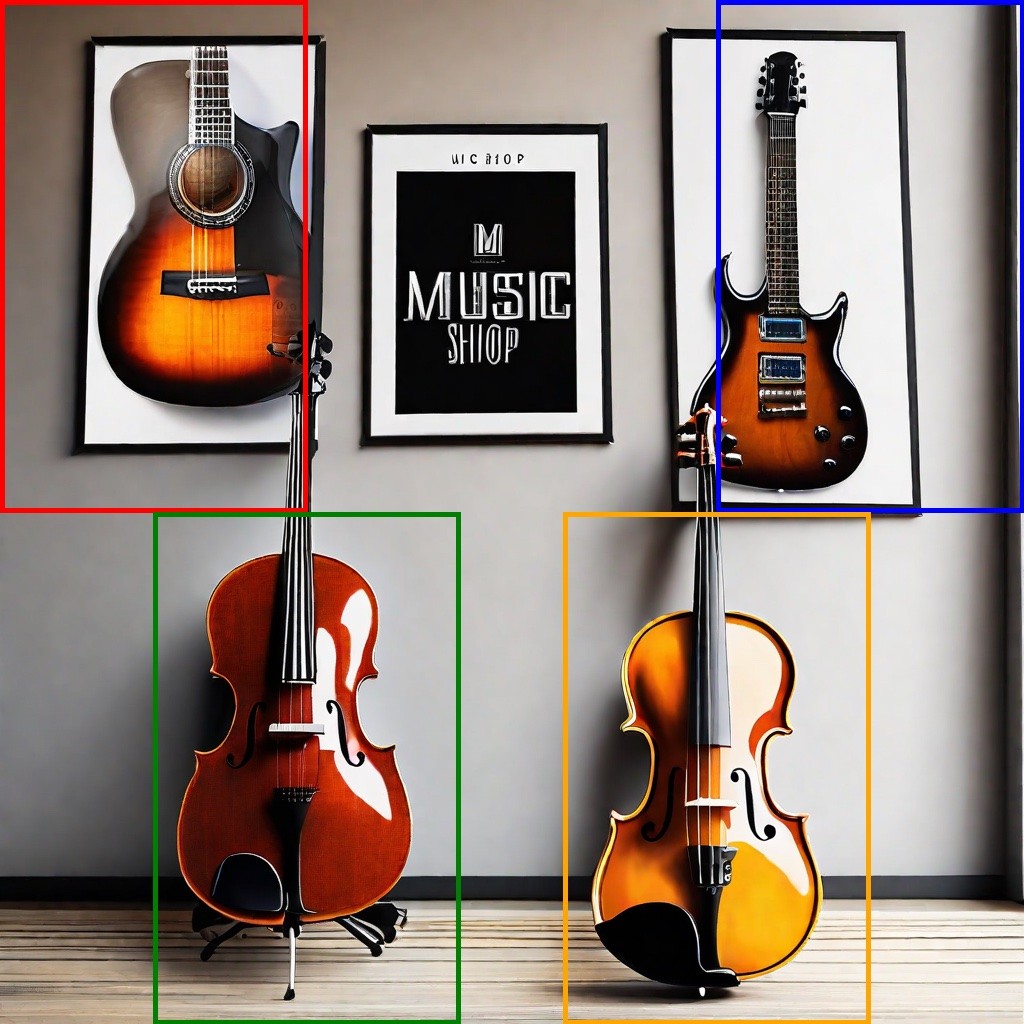}  &
        \hfill &
        \includegraphics[width=0.2\textwidth]{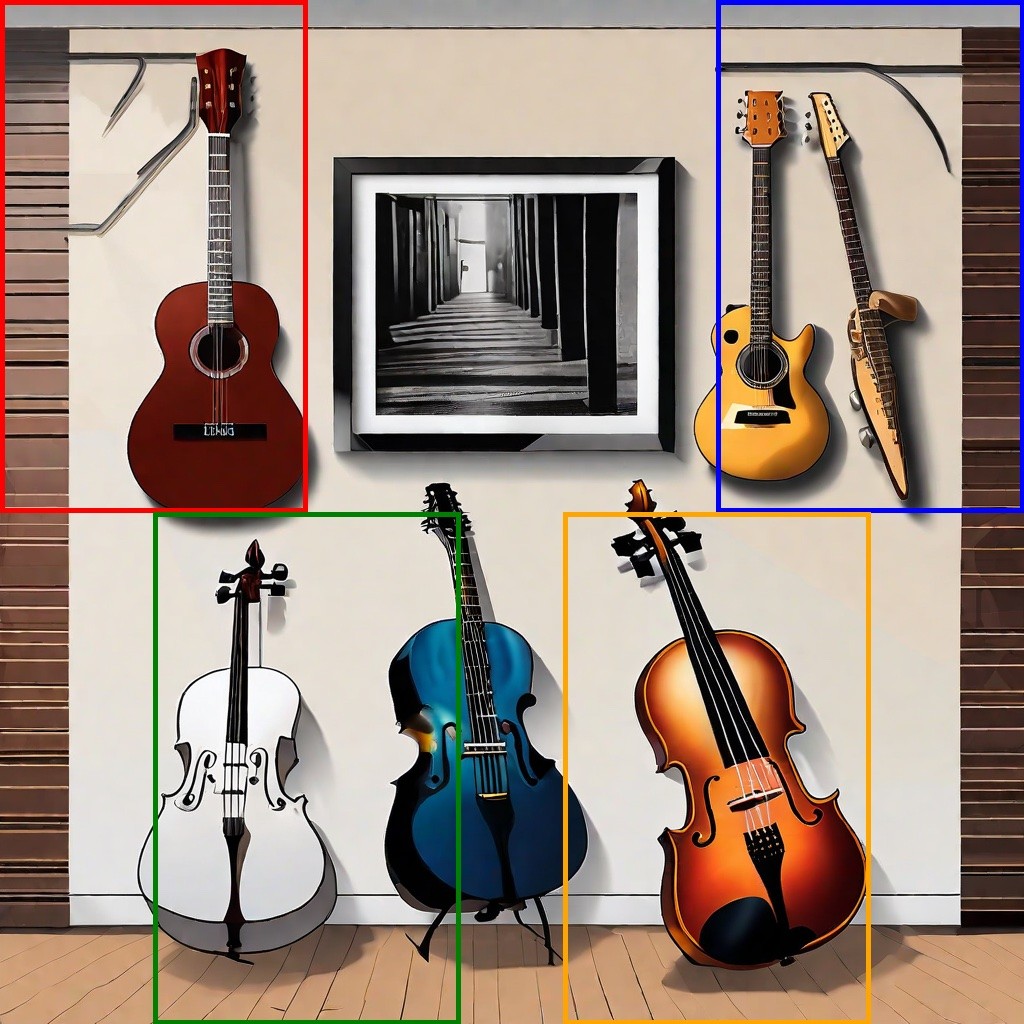}  
    \end{tabular}
    }
    \captionof{figure}{Failure cases.}
    \label{fig:failures}
\end{figure}

%% file: figures/comparison2.tex
\begin{figure*}[t]
    \centering
    \setlength{\tabcolsep}{0.001\textwidth}
    {\small
    \begin{tabular}{c c c c c c c}
        \multicolumn{7}{c}{``A \textcolor{red}{\textit{\underline{cactus}}} in a \textcolor{blue}{\textit{\underline{clay pot}}} and a \textcolor{green}{\textit{\underline{fern}}} in a \textcolor{orange}{\textit{\underline{porcelain pot}}}."} \\
        \includegraphics[width=0.135\textwidth]{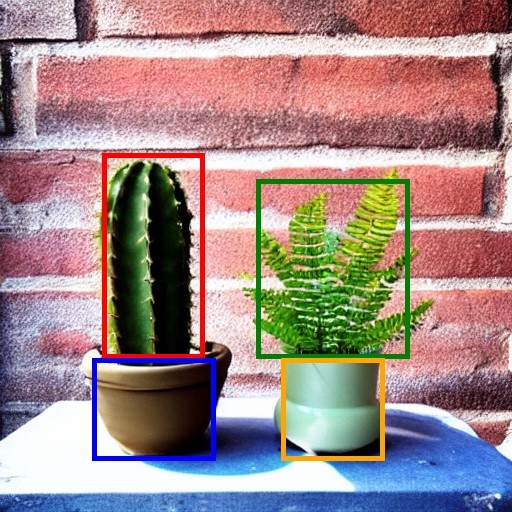} &
        \includegraphics[width=0.135\textwidth]
        {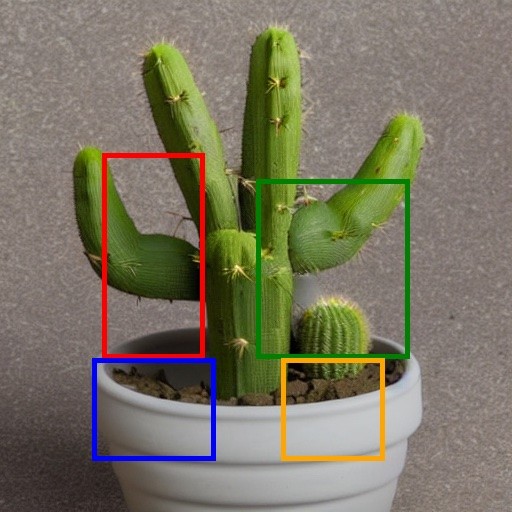} &
        \includegraphics[width=0.135\textwidth]
        {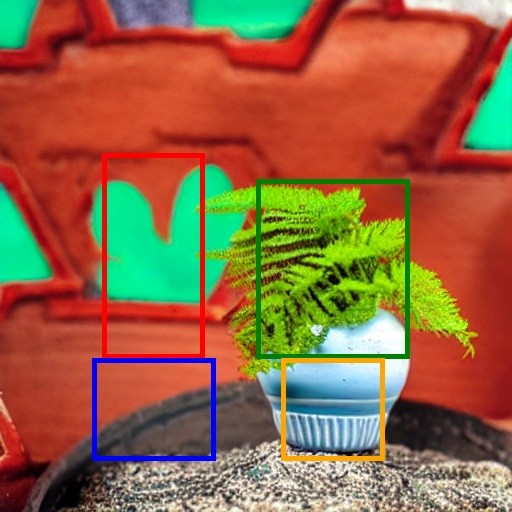} &
        \includegraphics[width=0.135\textwidth]
        {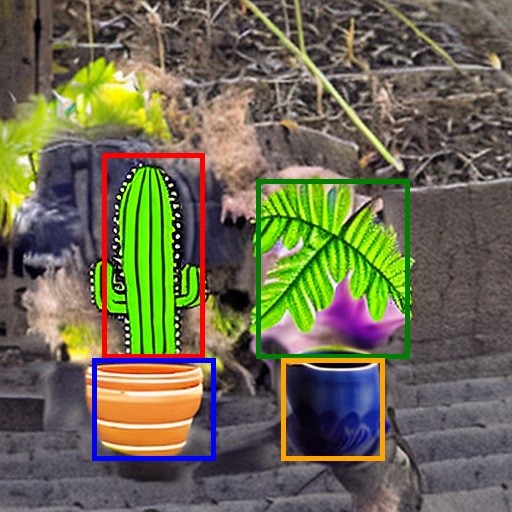} &
        \includegraphics[width=0.135\textwidth]
        {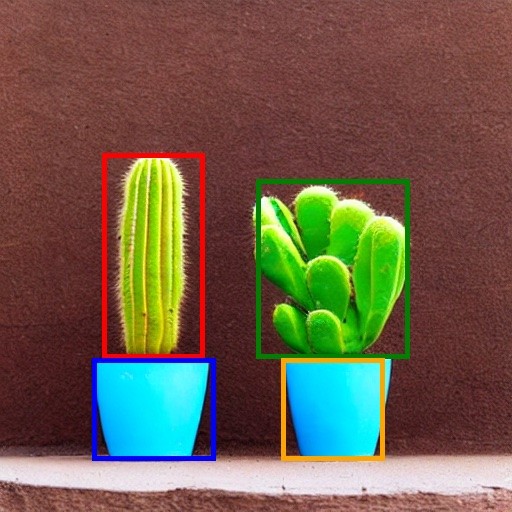} &
        \includegraphics[width=0.135\textwidth]
        {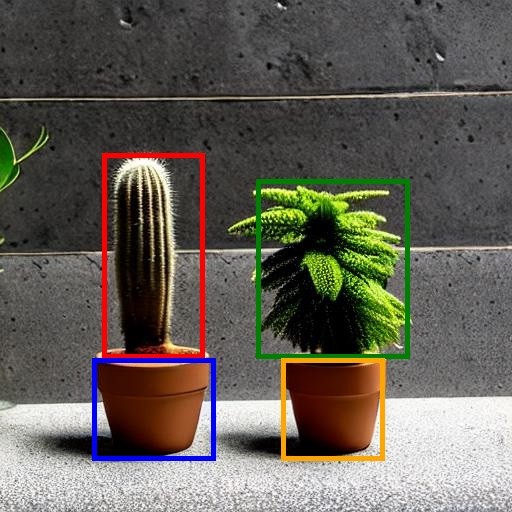} &
        \includegraphics[width=0.135\textwidth]
        {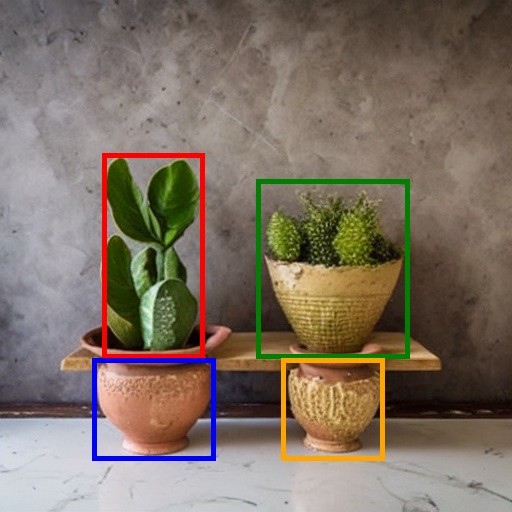} \\
        
        \multicolumn{7}{c}{``A \textcolor{red}{\textit{\underline{cow}}} and a \textcolor{blue}{\textit{\underline{horse}}} eating \textcolor{green}{\textit{\underline{hay}}} in a farm."} \\
        \includegraphics[width=0.135\textwidth]{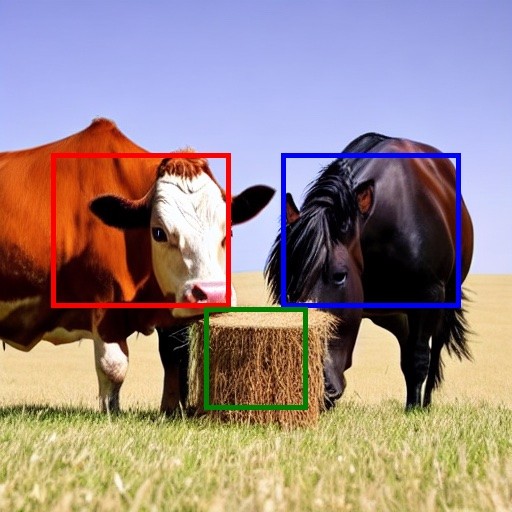} &
        \includegraphics[width=0.135\textwidth]
        {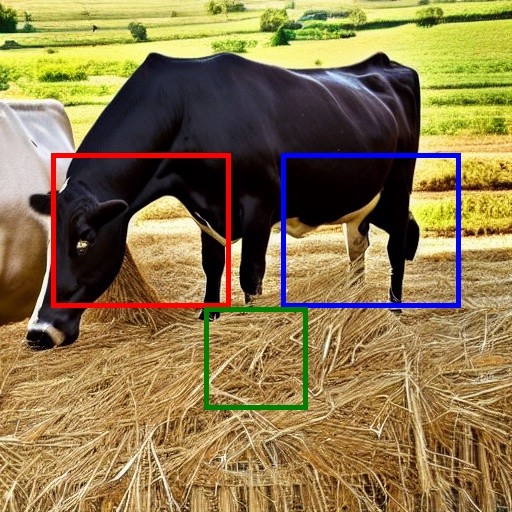} &
        \includegraphics[width=0.135\textwidth]
        {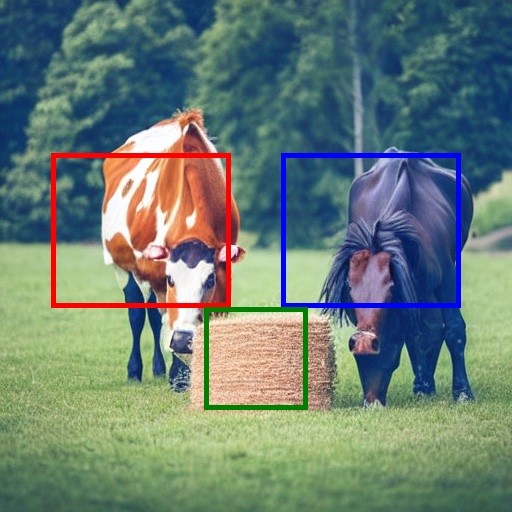} &
        \includegraphics[width=0.135\textwidth]
        {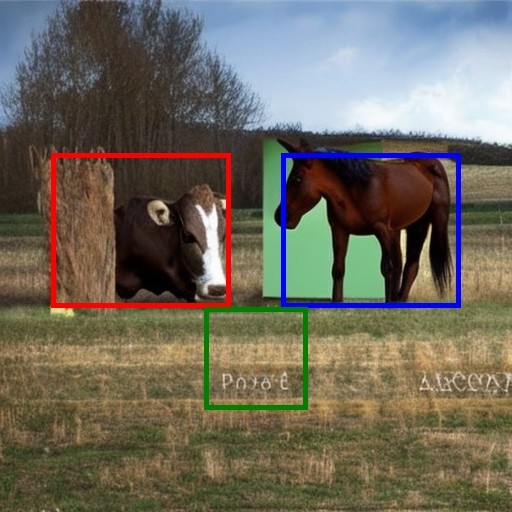} &
        \includegraphics[width=0.135\textwidth]
        {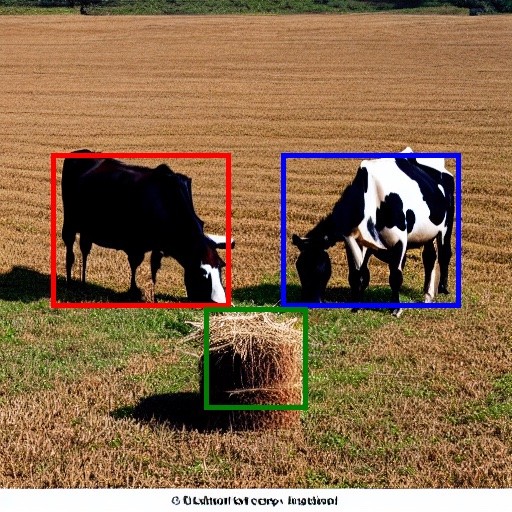} &
        \includegraphics[width=0.135\textwidth]
        {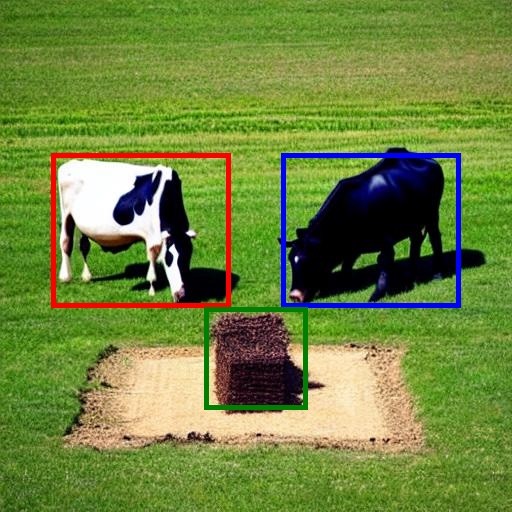} &
        \includegraphics[width=0.135\textwidth]
        {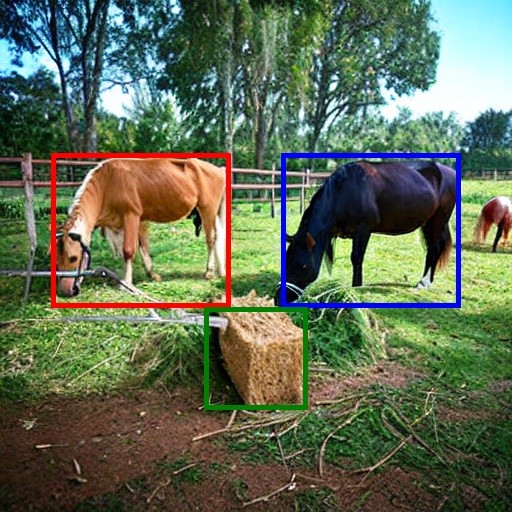} \\
        
        \multicolumn{7}{c}{``A \textcolor{red}{\textit{\underline{white plate}}} with \textcolor{blue}{\textit{\underline{ravioli}}} next to a \textcolor{green}{\textit{\underline{blue bowl}}} of \textcolor{orange}{\textit{\underline{gnocchi}}} on a table."} \\
        \includegraphics[width=0.135\textwidth]{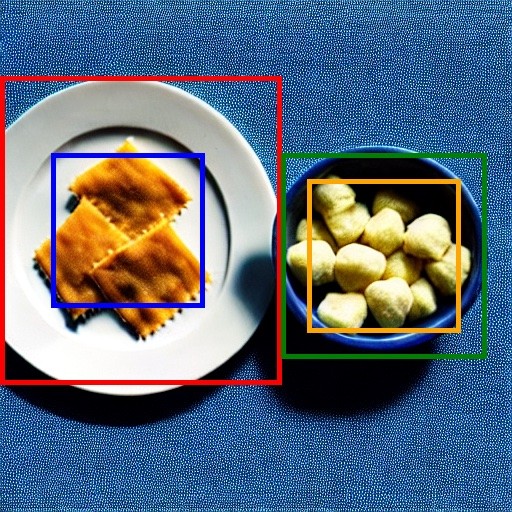} &
        \includegraphics[width=0.135\textwidth]
        {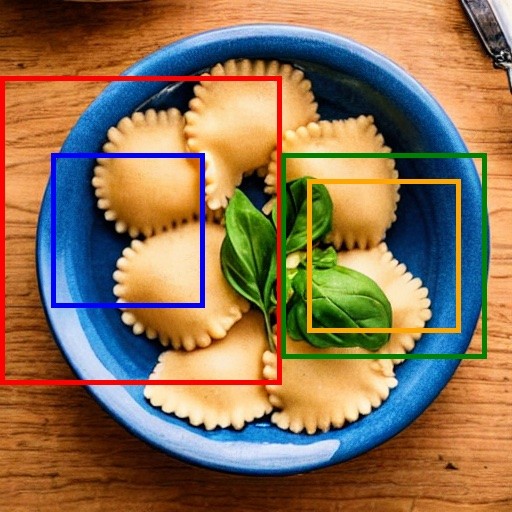} &
        \includegraphics[width=0.135\textwidth]
        {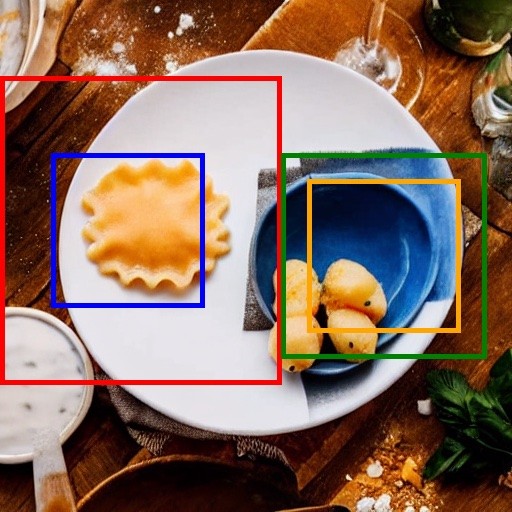} &
        \includegraphics[width=0.135\textwidth]
        {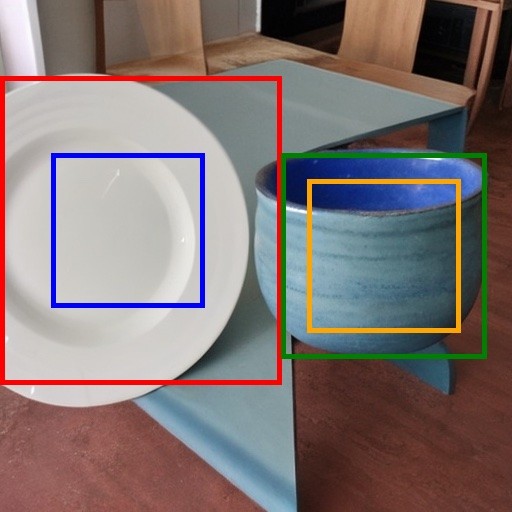} &
        \includegraphics[width=0.135\textwidth]
        {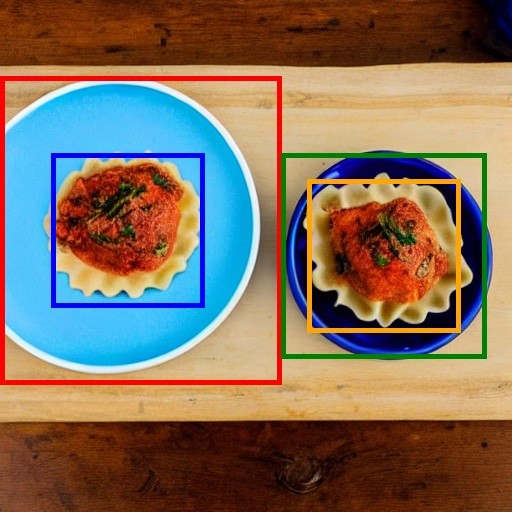} &
        \includegraphics[width=0.135\textwidth]
        {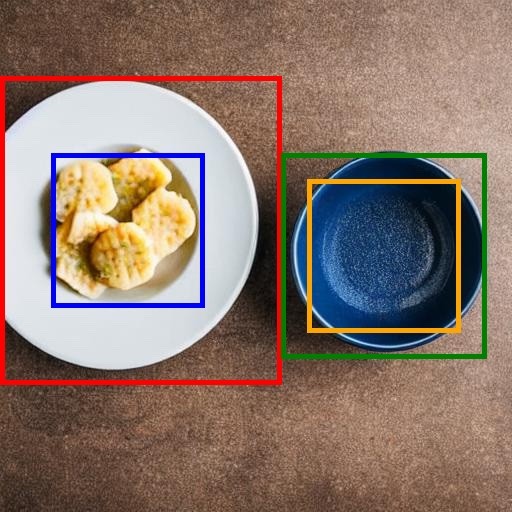} &
        \includegraphics[width=0.135\textwidth]
        {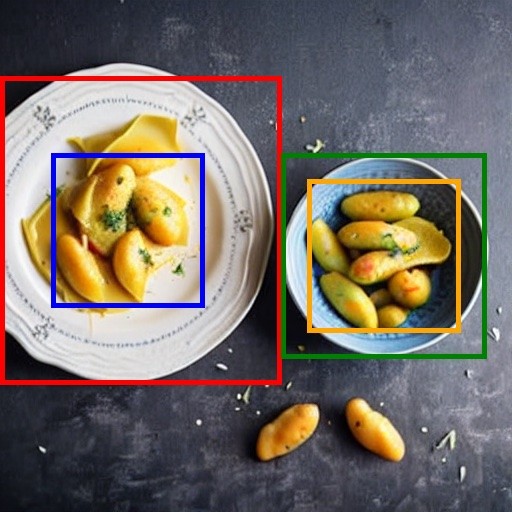} \\

        \multicolumn{7}{c}{``A \textcolor{red}{\textit{\underline{marble statue}}} and two \textcolor{blue}{\textit{\underline{watercolor paintings}}} in a room."} \\
        \includegraphics[width=0.135\textwidth]{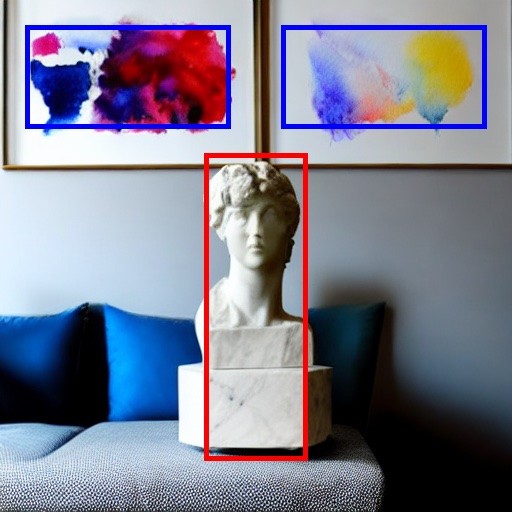} &
        \includegraphics[width=0.135\textwidth]
        {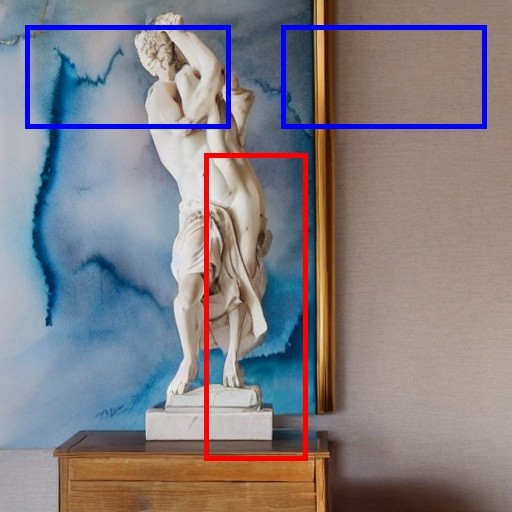} &
        \includegraphics[width=0.135\textwidth]
        {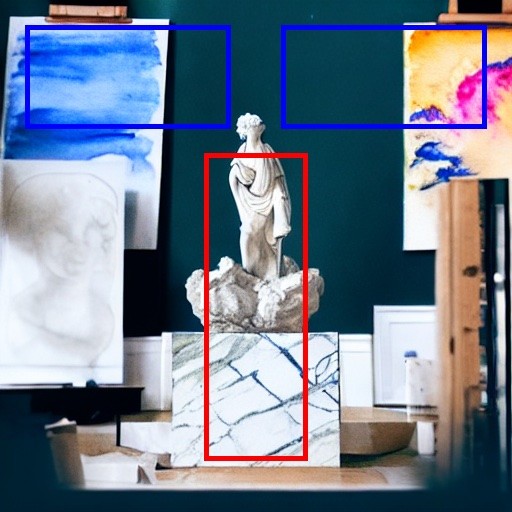} &
        \includegraphics[width=0.135\textwidth]
        {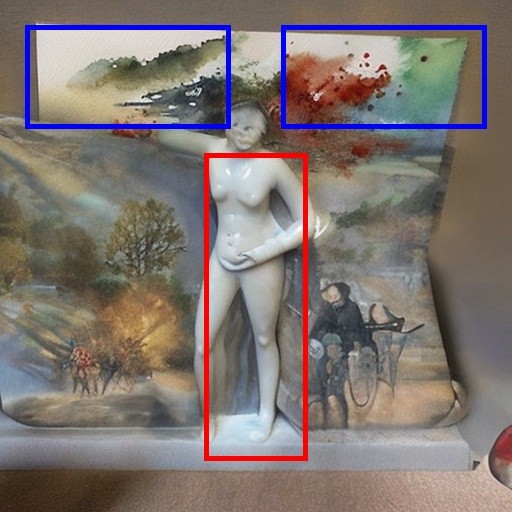} &
        \includegraphics[width=0.135\textwidth]
        {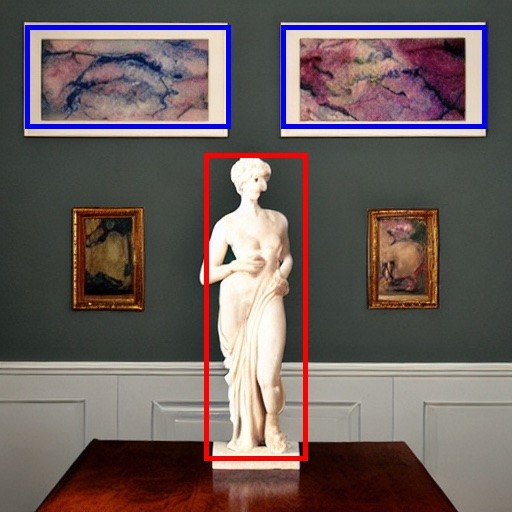} &
        \includegraphics[width=0.135\textwidth]
        {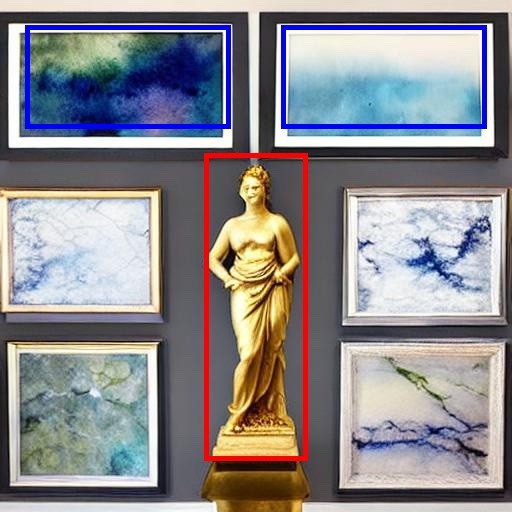} &
        \includegraphics[width=0.135\textwidth]
        {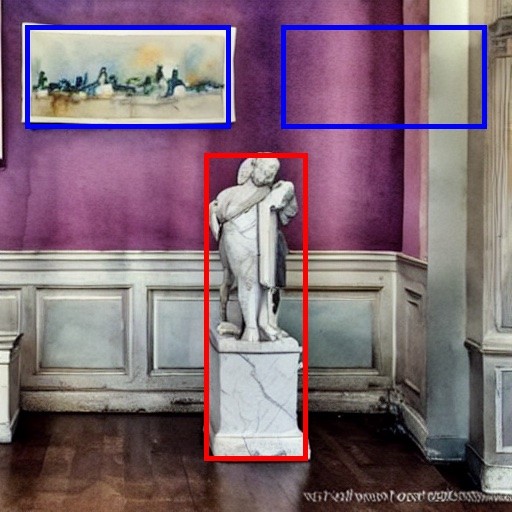} \\

        \multicolumn{7}{c}{``A \textcolor{red}{\textit{\underline{shark chasing}}} a \textcolor{blue}{\textit{\underline{dolphin}}} in the ocean."} \\
        \includegraphics[width=0.135\textwidth]{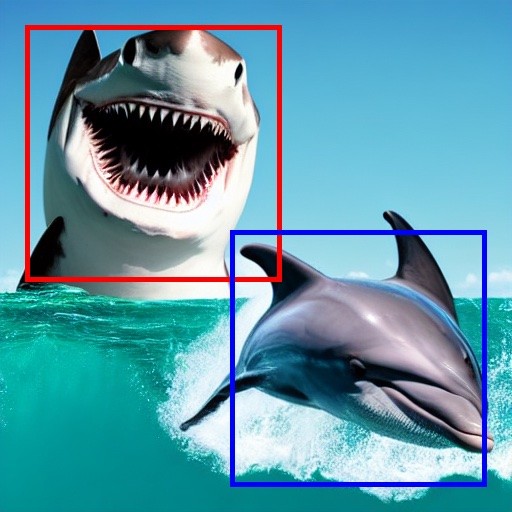} &
        \includegraphics[width=0.135\textwidth]
        {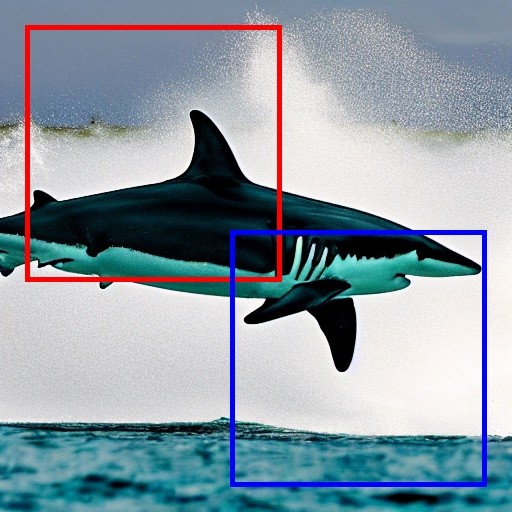} &
        \includegraphics[width=0.135\textwidth]
        {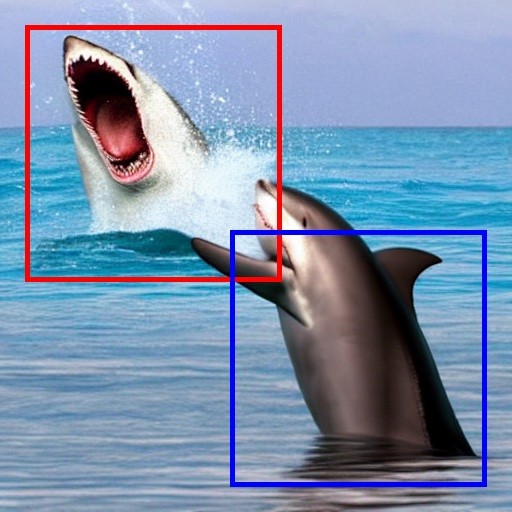} &
        \includegraphics[width=0.135\textwidth]
        {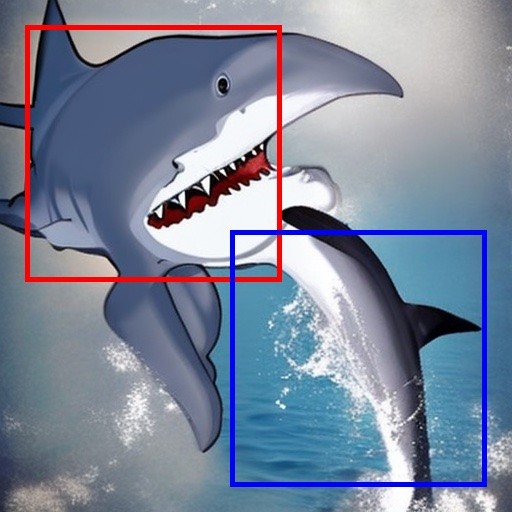} &
        \includegraphics[width=0.135\textwidth]
        {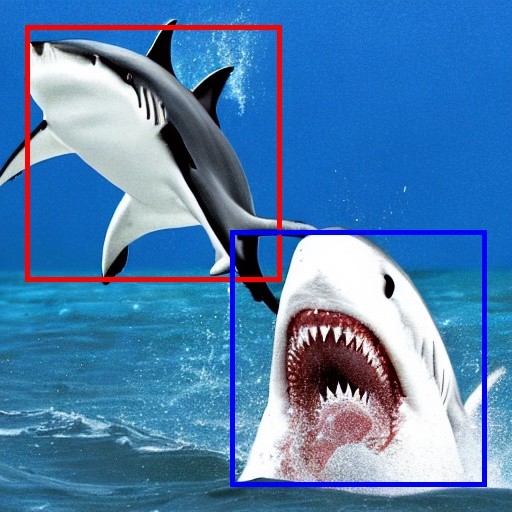} &
        \includegraphics[width=0.135\textwidth]
        {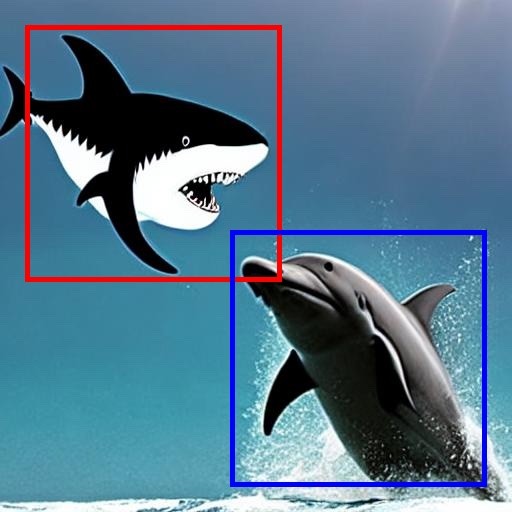} &
        \includegraphics[width=0.135\textwidth]
        {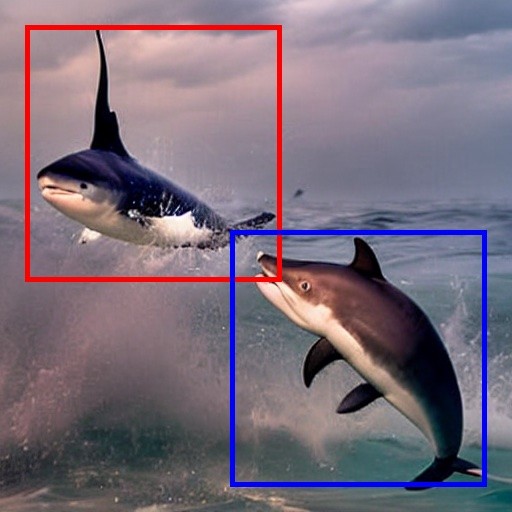} \\
        
        BA (ours) & LG~\cite{chen2023training} & BD~\cite{xie2023boxdiff} & MD~\cite{bar2023multidiffusion} & GLIGEN~\cite{li2023gligen} & AR~\cite{phung2023grounded} & ReCo~\cite{yang2023reco} \\
    \end{tabular}
    \vspace{-4pt}
    }
    \captionof{figure}{More qualitative results of our method in comparison to baseline methods.}
    \vspace{-12pt}
    \label{fig:comparisons2}
\end{figure*}

%% file: figures/sup/sdxl_results3.tex
\begin{figure*}[t]
    \setlength{\tabcolsep}{1pt}
    {\small\centering
    \begin{tabular}{c c c c c c}
        &
        \multicolumn{5}{c}{``A realistic photo of a \textcolor{red}{\textit{\underline{turtle}}} and a \textcolor{blue}{\textit{\underline{jellyfish}}} and an \textcolor{green}{\textit{\underline{octopus}}} and a \textcolor{orange}{\textit{\underline{starfish}}} in the ocean depths.''}
        \\
       \raisebox{15pt}{\rotatebox{90}{Bounded Attention}} &
        \includegraphics[width=0.2\textwidth]{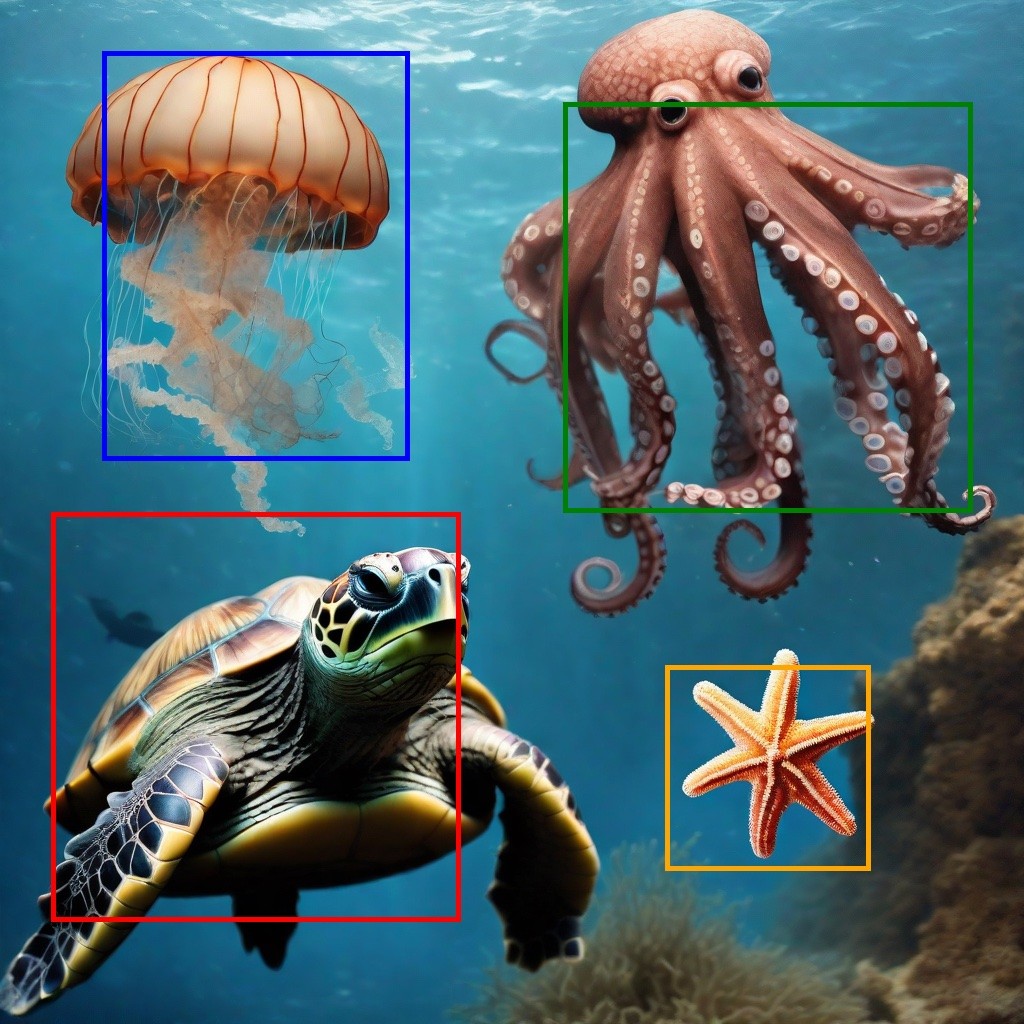} &
        \includegraphics[width=0.2\textwidth]{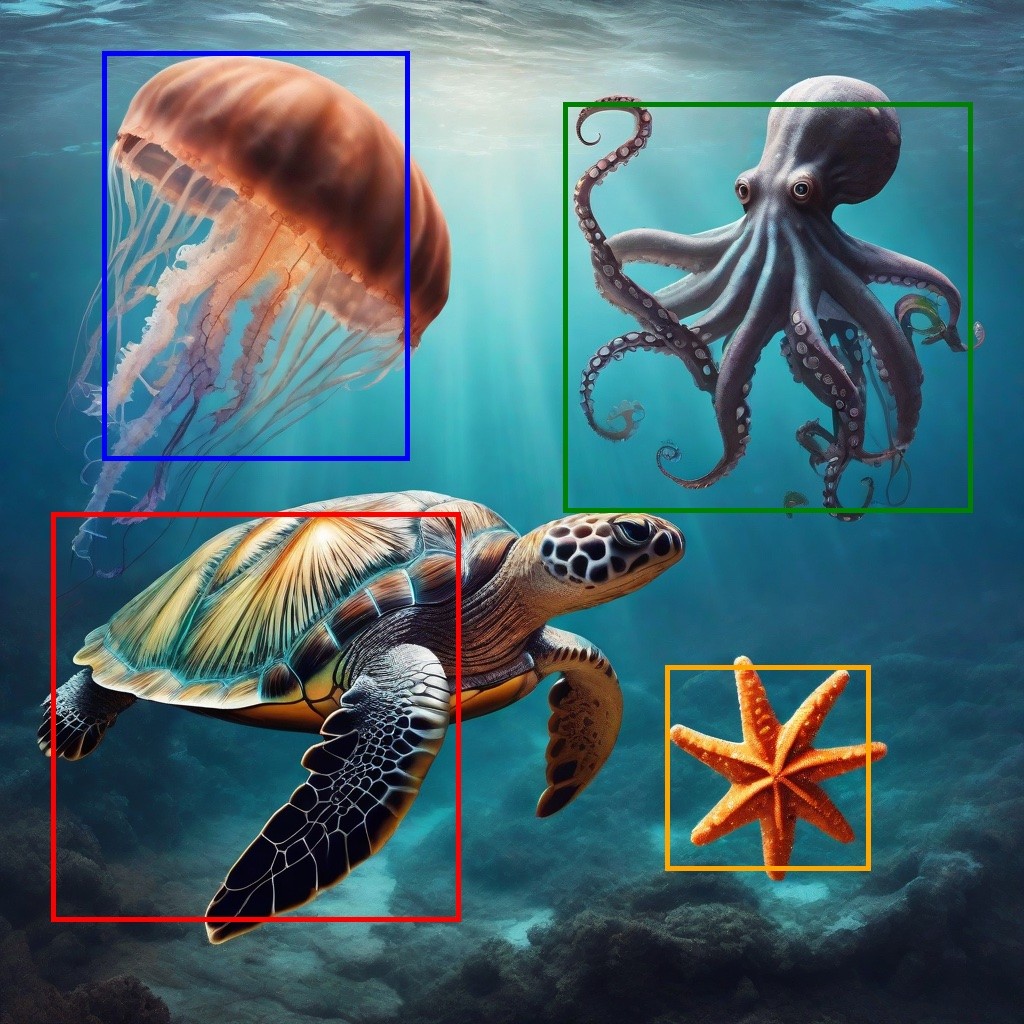} &
        \includegraphics[width=0.2\textwidth]{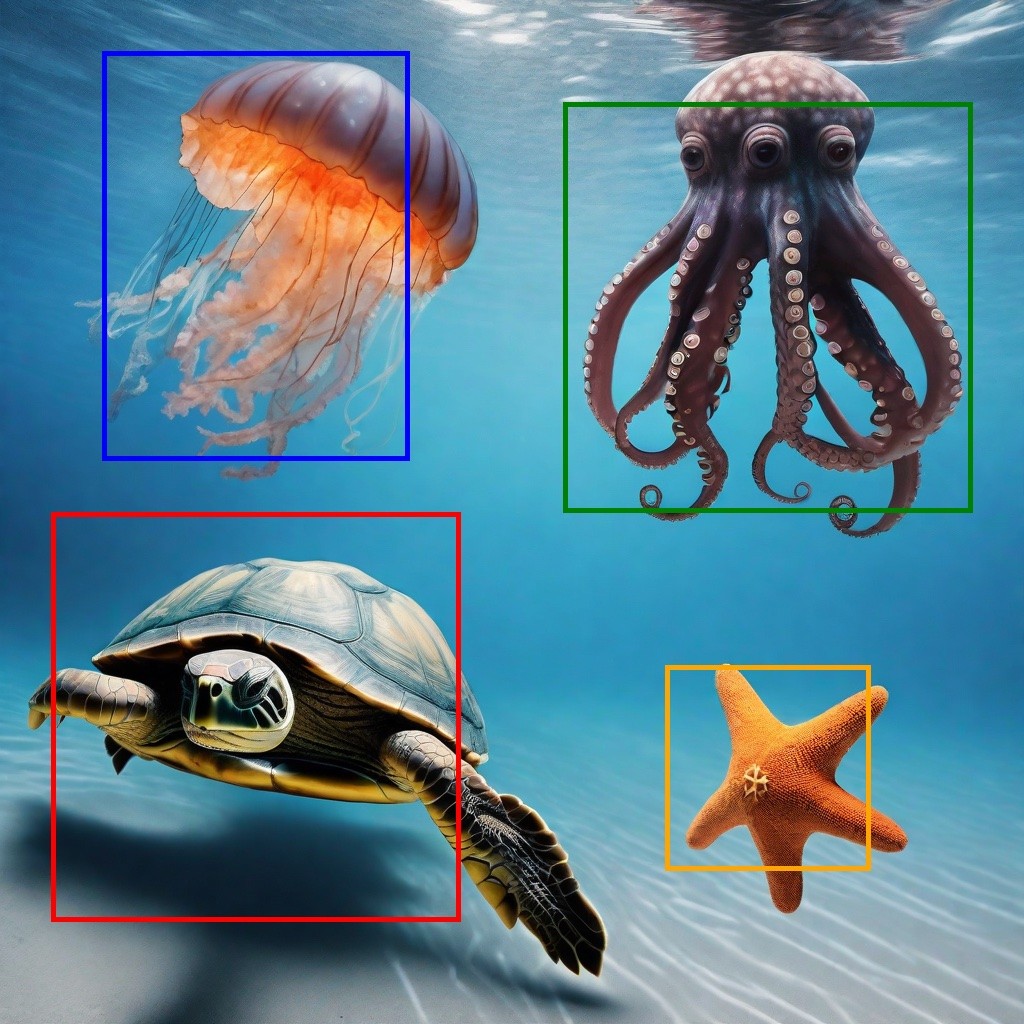} &
        \includegraphics[width=0.2\textwidth]{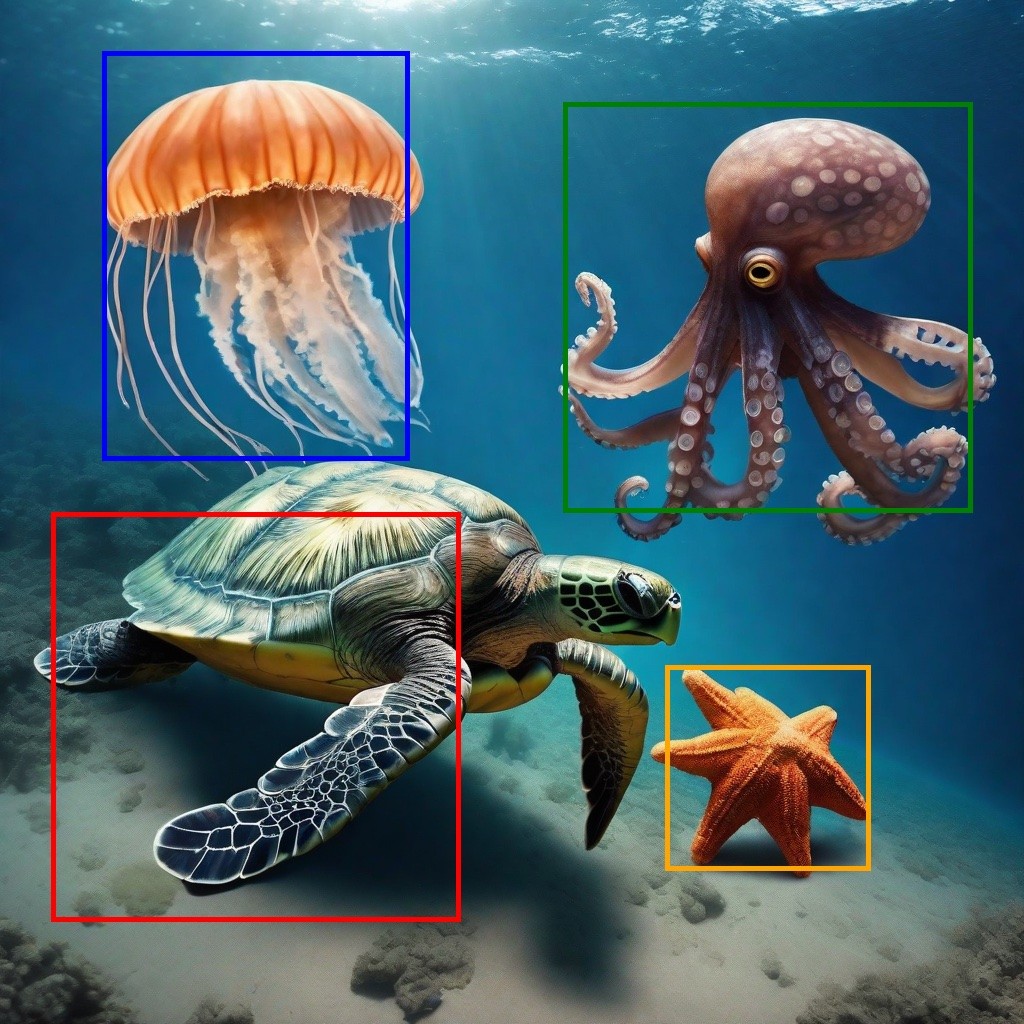} &
        \includegraphics[width=0.2\textwidth]{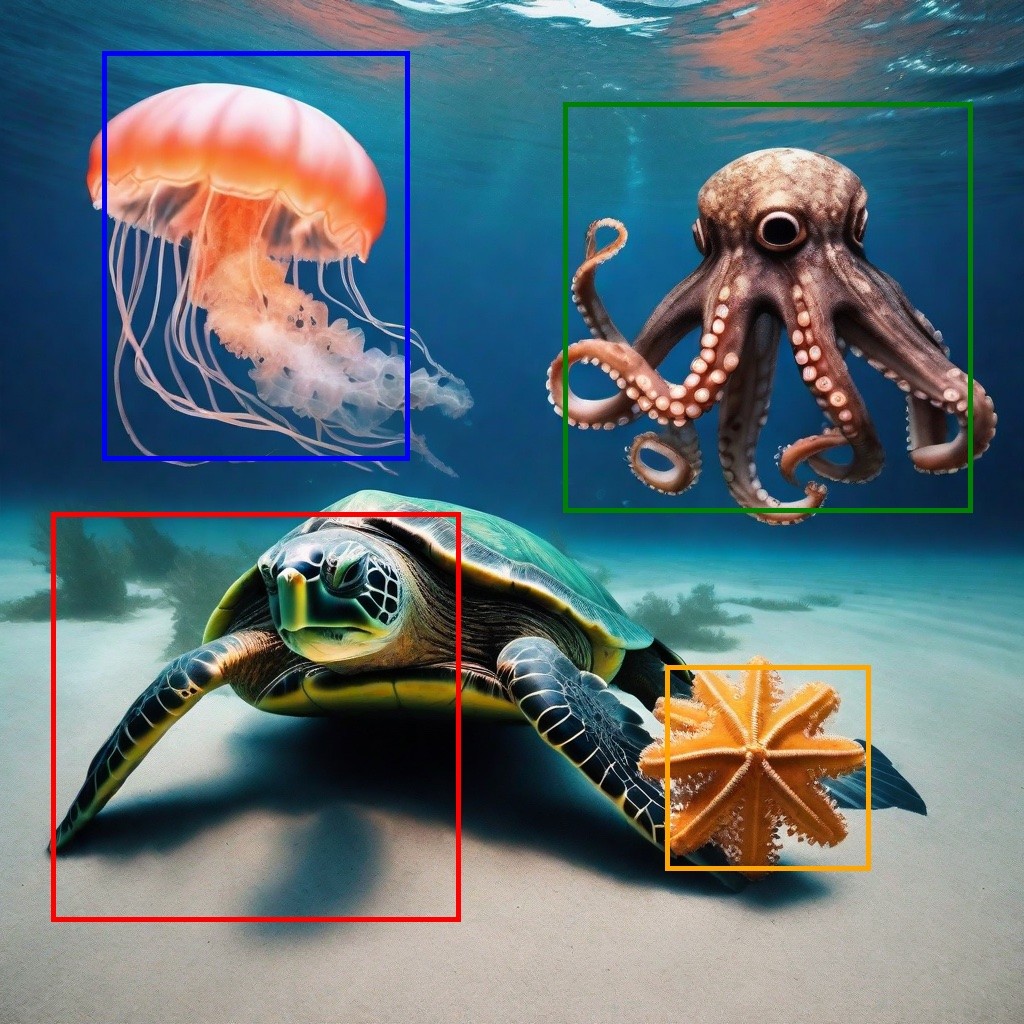} \\
        \raisebox{24pt}{\rotatebox{90}{Vanilla SDXL}} &
        \includegraphics[width=0.2\textwidth]{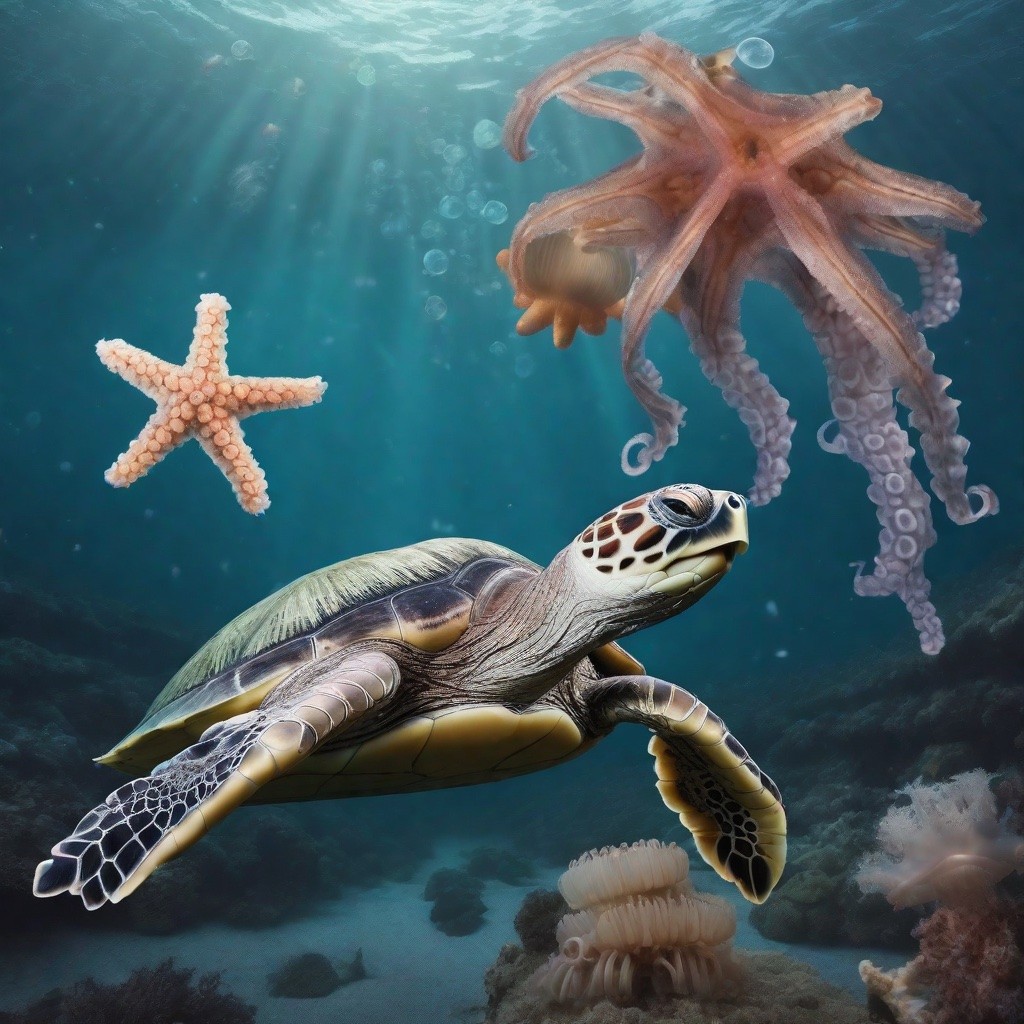} &
        \includegraphics[width=0.2\textwidth]{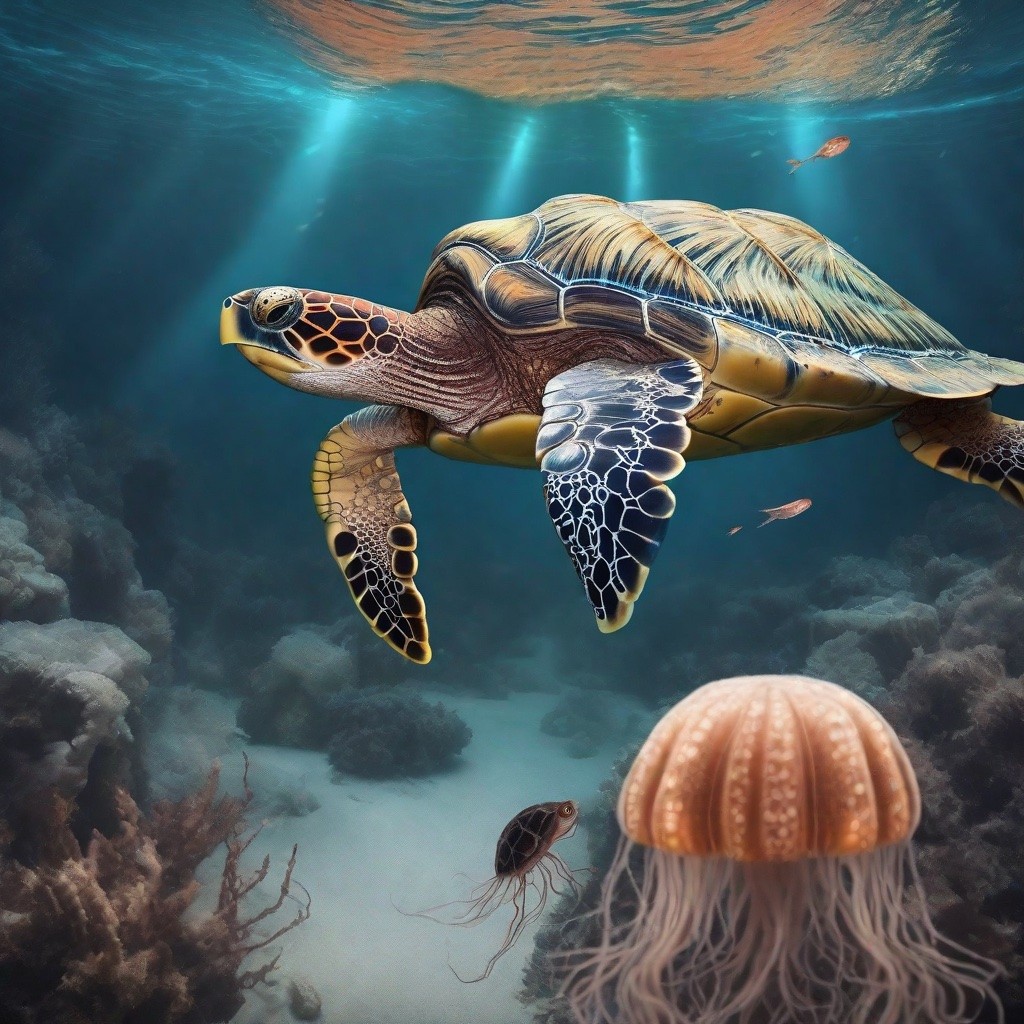} &
        \includegraphics[width=0.2\textwidth]{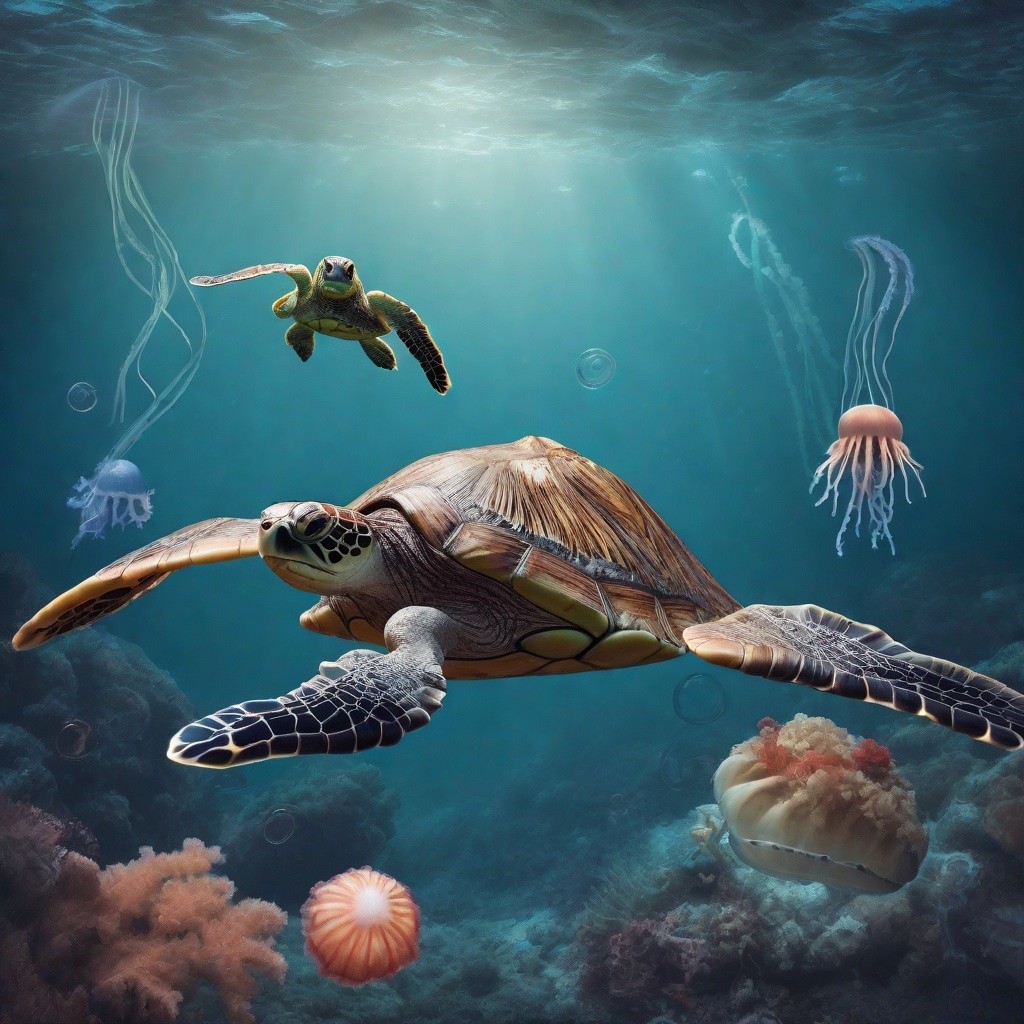} &
        \includegraphics[width=0.2\textwidth]{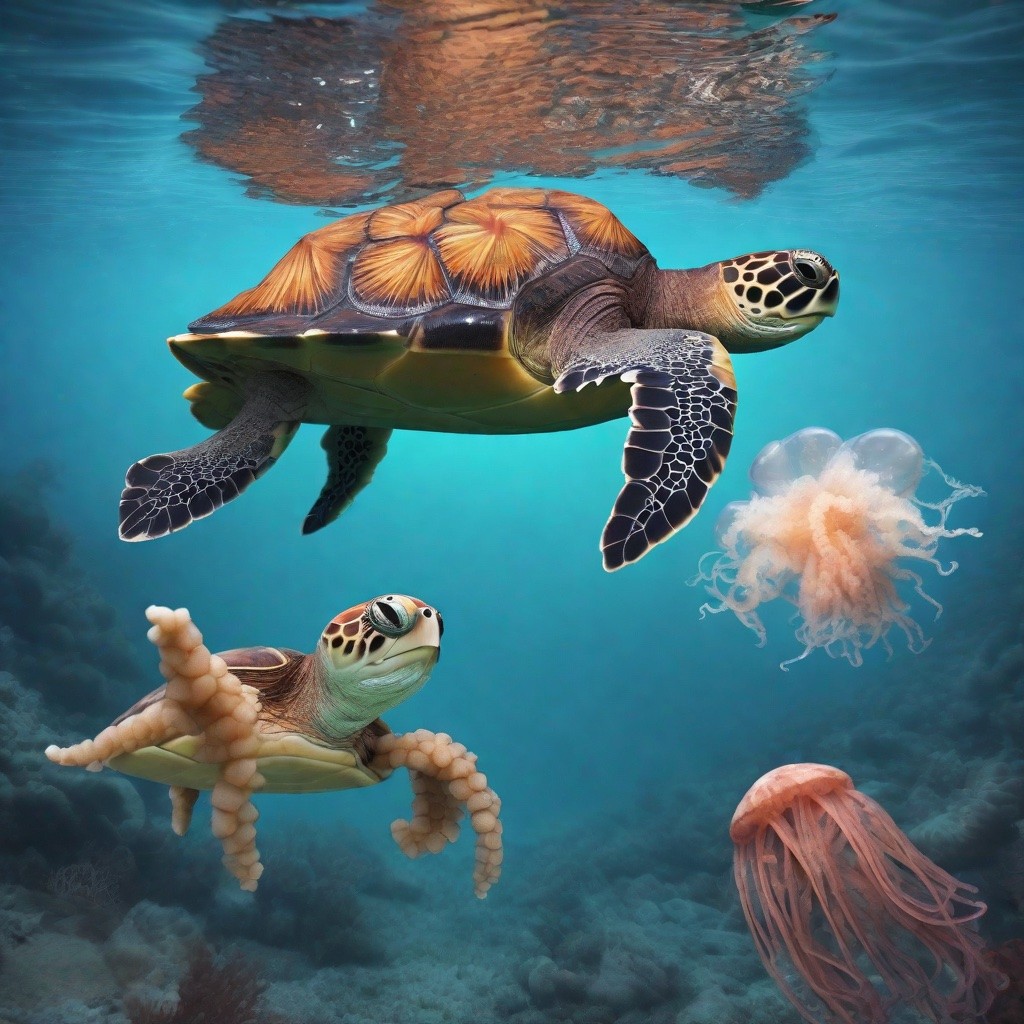} &
        \includegraphics[width=0.2\textwidth]{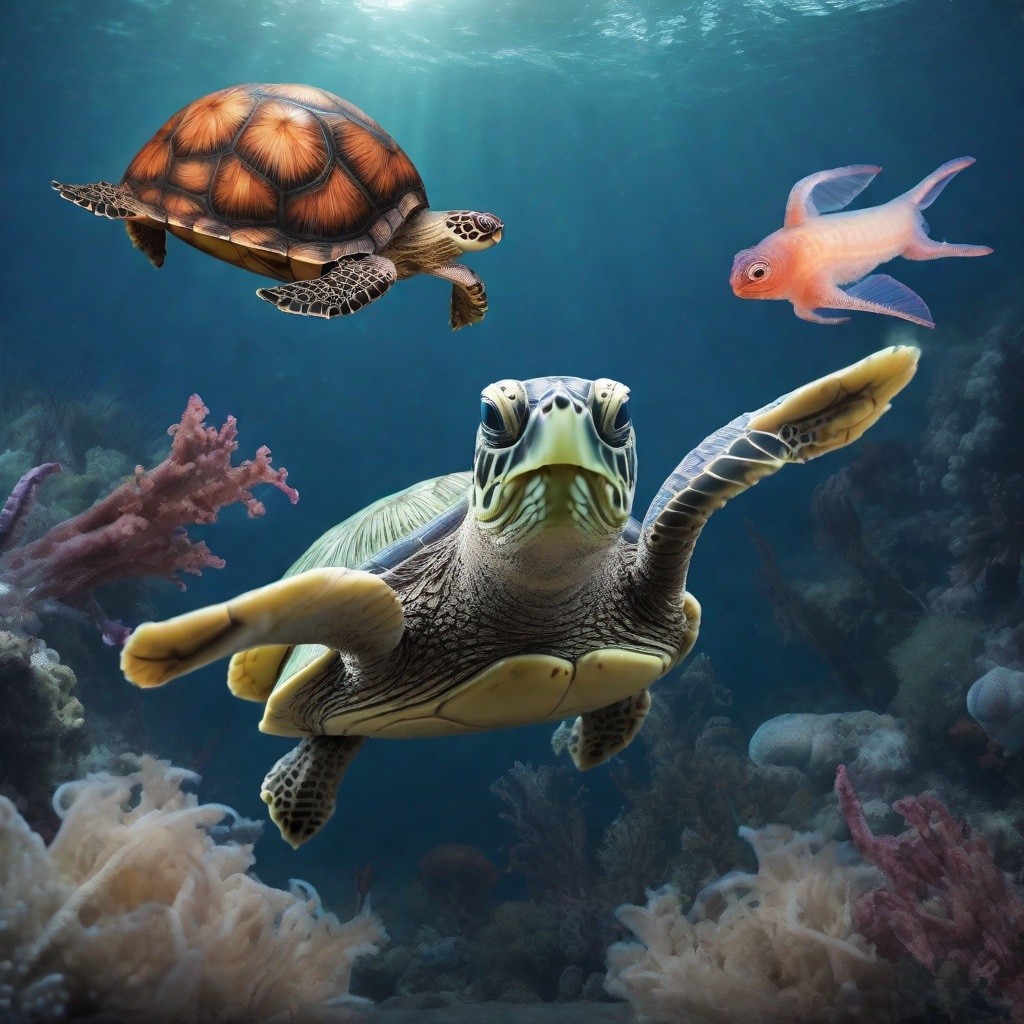} \\
        \\[-10pt]

        &
        \multicolumn{5}{c}{``A science fiction movie poster with an \textcolor{red}{\textit{\underline{astronaut}}} and a \textcolor{blue}{\textit{\underline{robot}}} and a \textcolor{green}{\textit{\underline{green alien}}} and a \textcolor{orange}{\textit{\underline{spaceship}}}.''}
        \\
       \raisebox{15pt}{\rotatebox{90}{Bounded Attention}} &
        \includegraphics[width=0.2\textwidth]{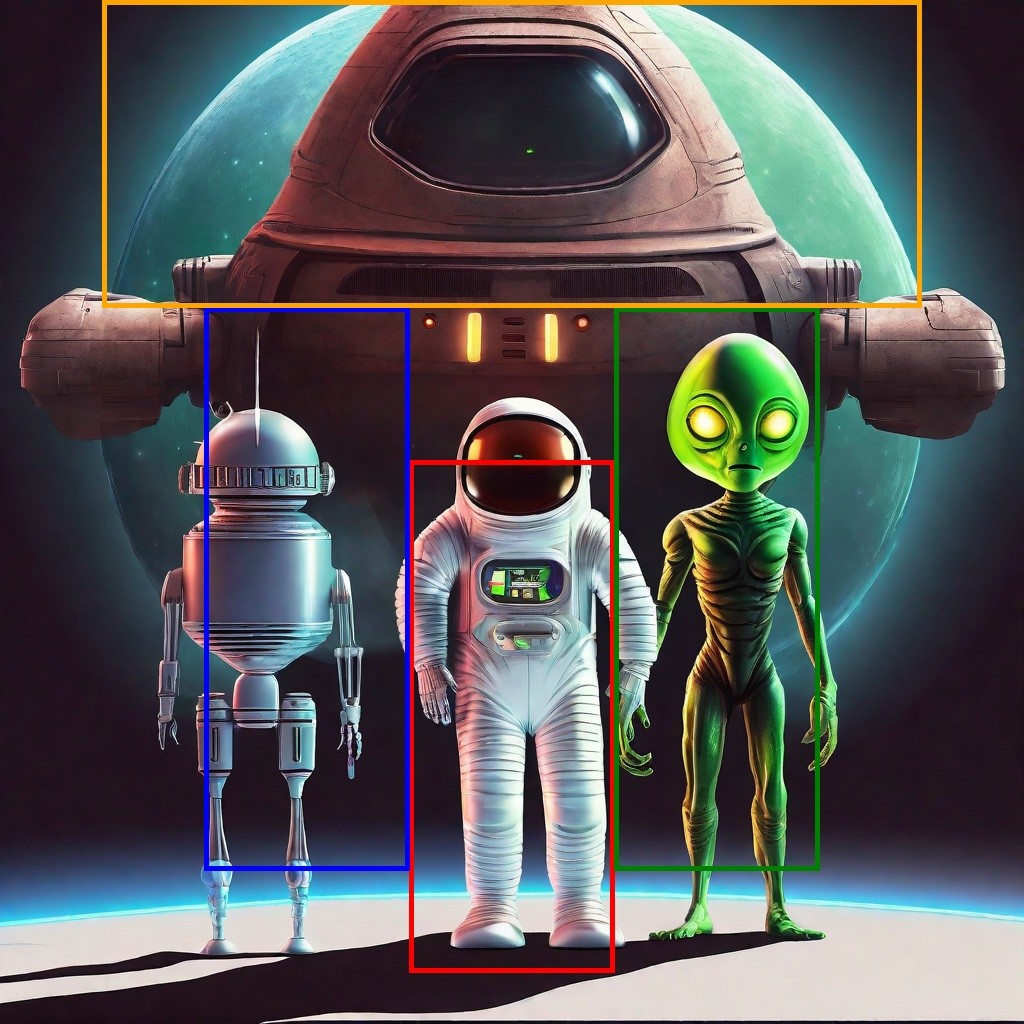} &
        \includegraphics[width=0.2\textwidth]{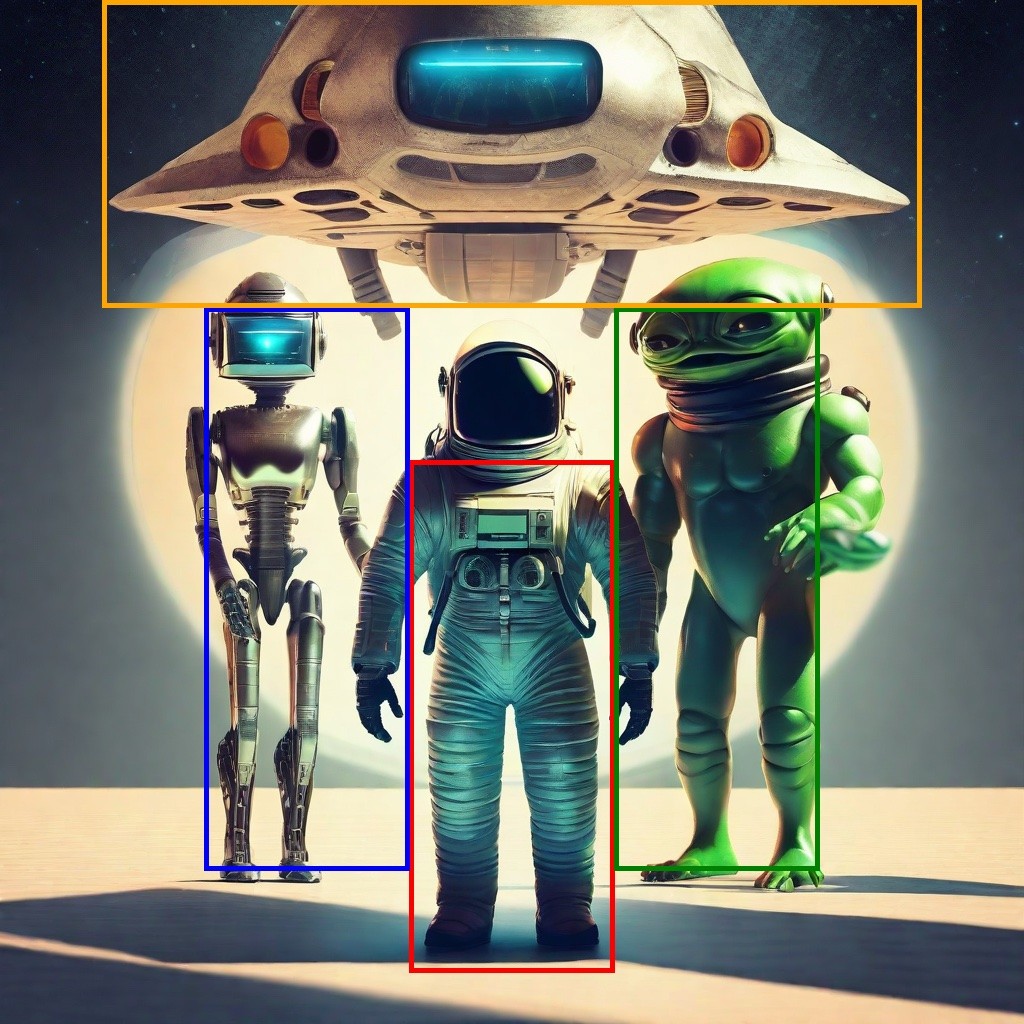} &
        \includegraphics[width=0.2\textwidth]{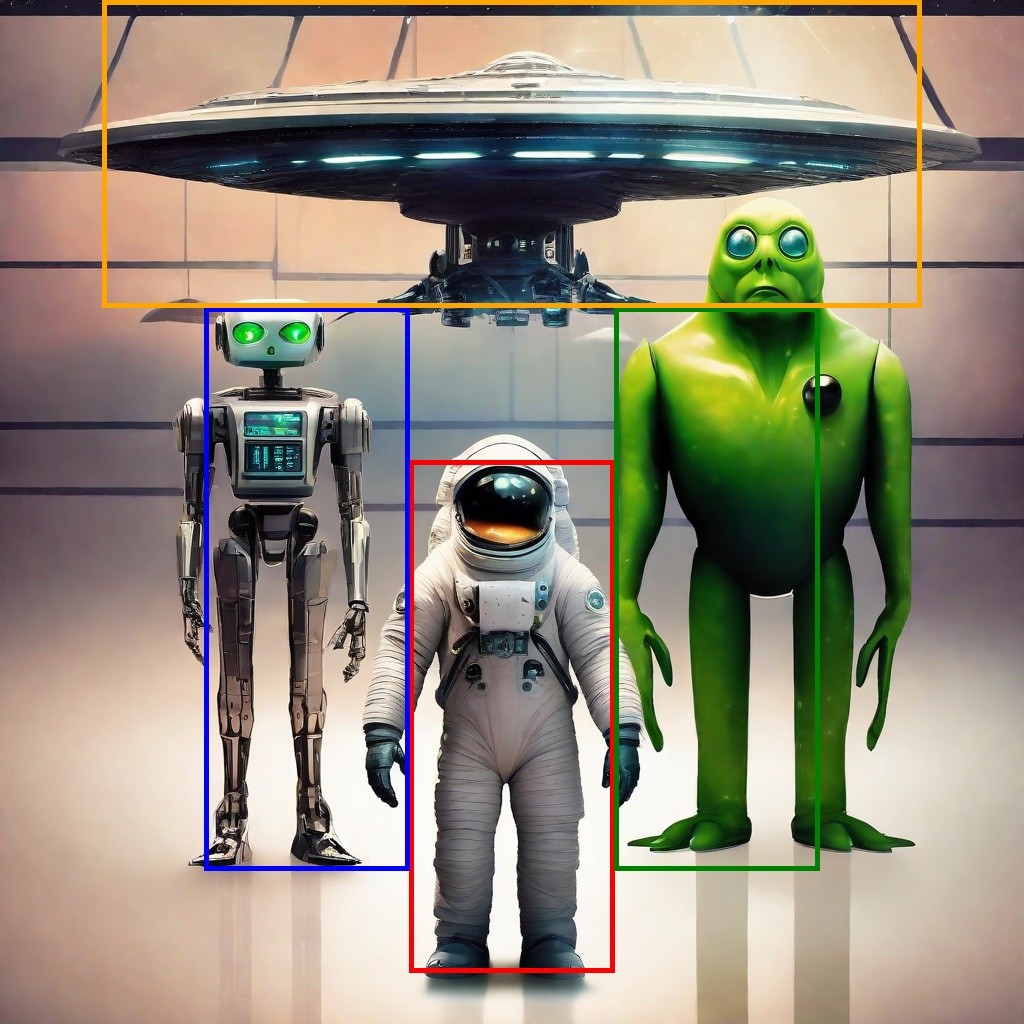} &
        \includegraphics[width=0.2\textwidth]{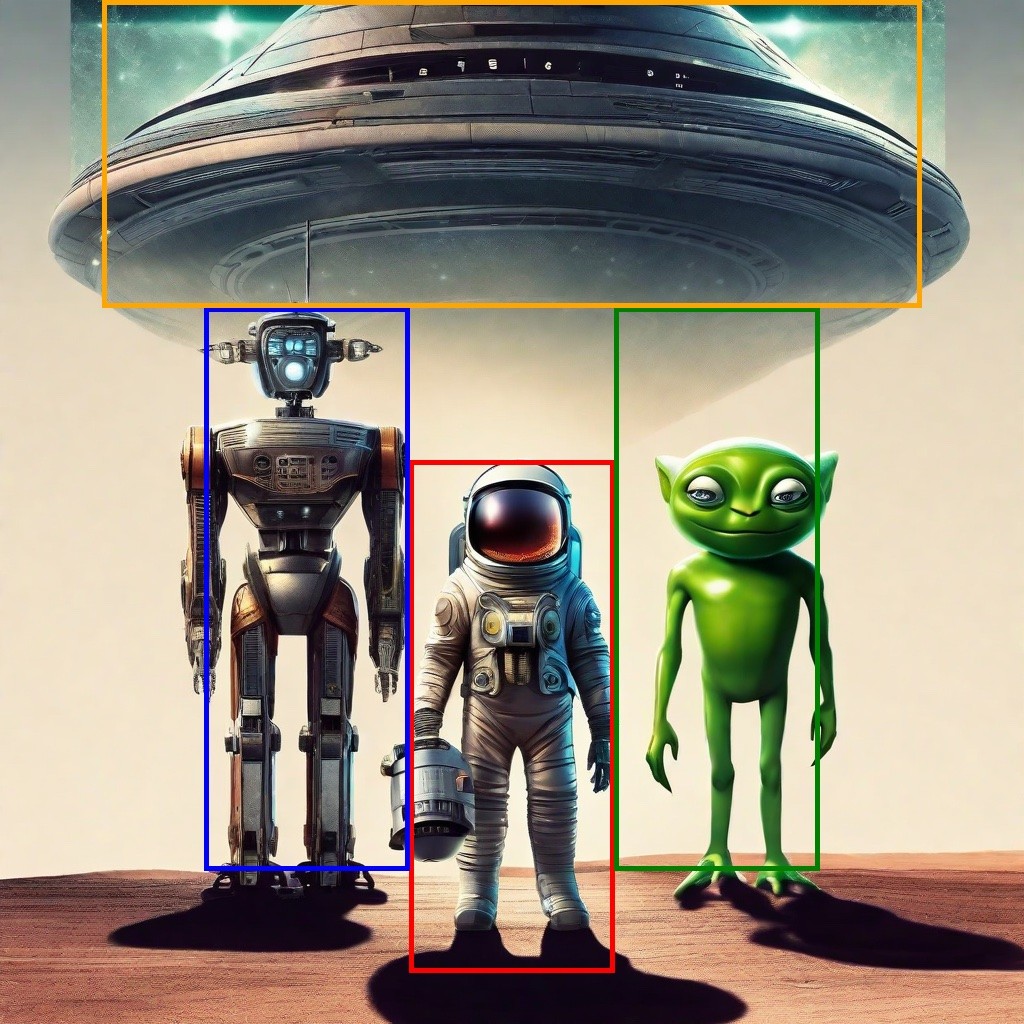} &
        \includegraphics[width=0.2\textwidth]{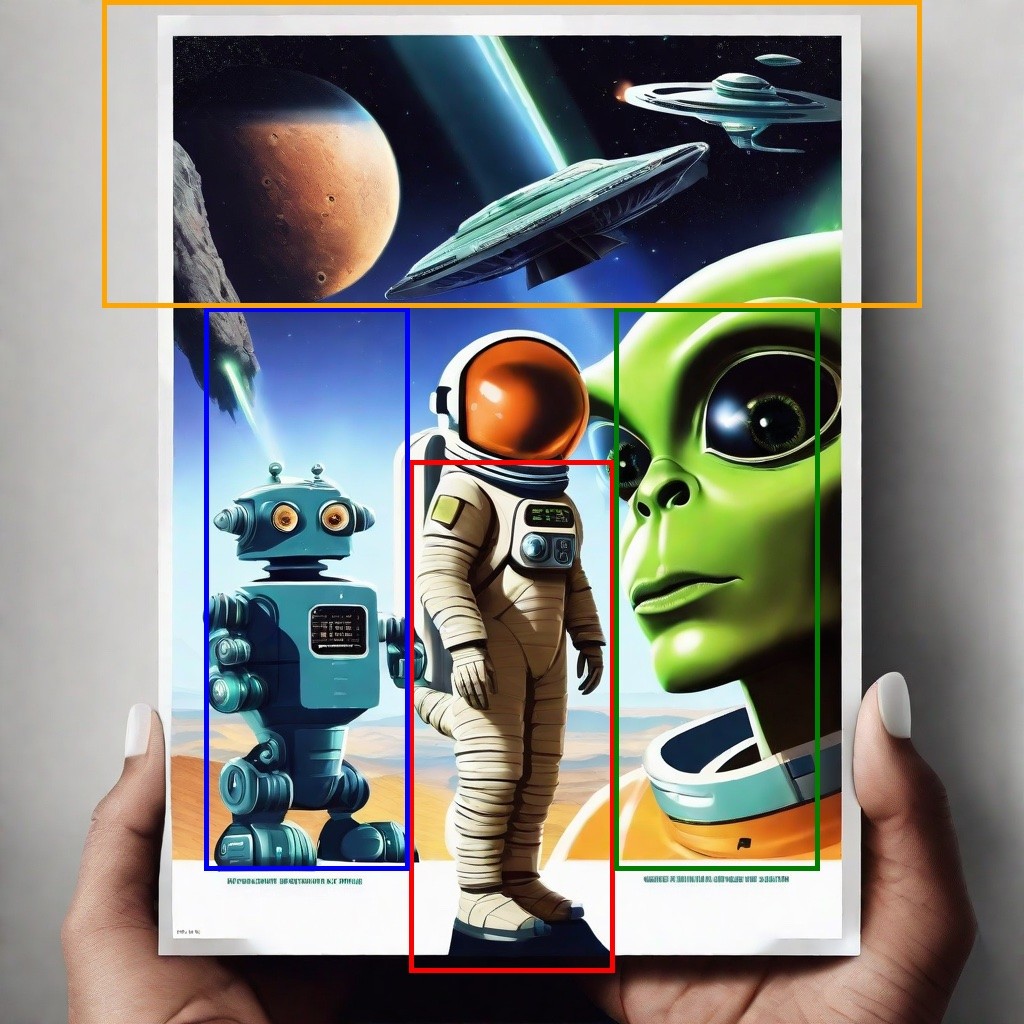} \\
        \raisebox{24pt}{\rotatebox{90}{Vanilla SDXL}} &
        \includegraphics[width=0.2\textwidth]{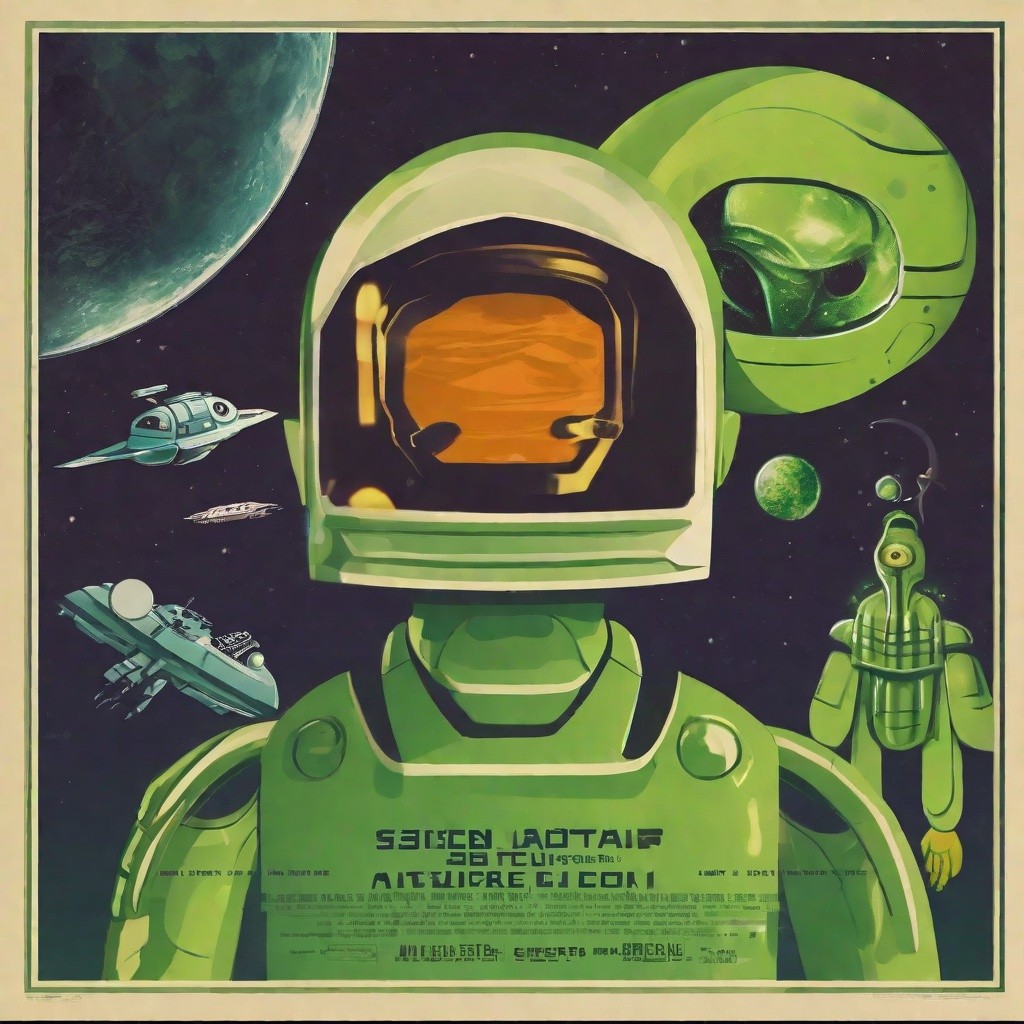} &
        \includegraphics[width=0.2\textwidth]{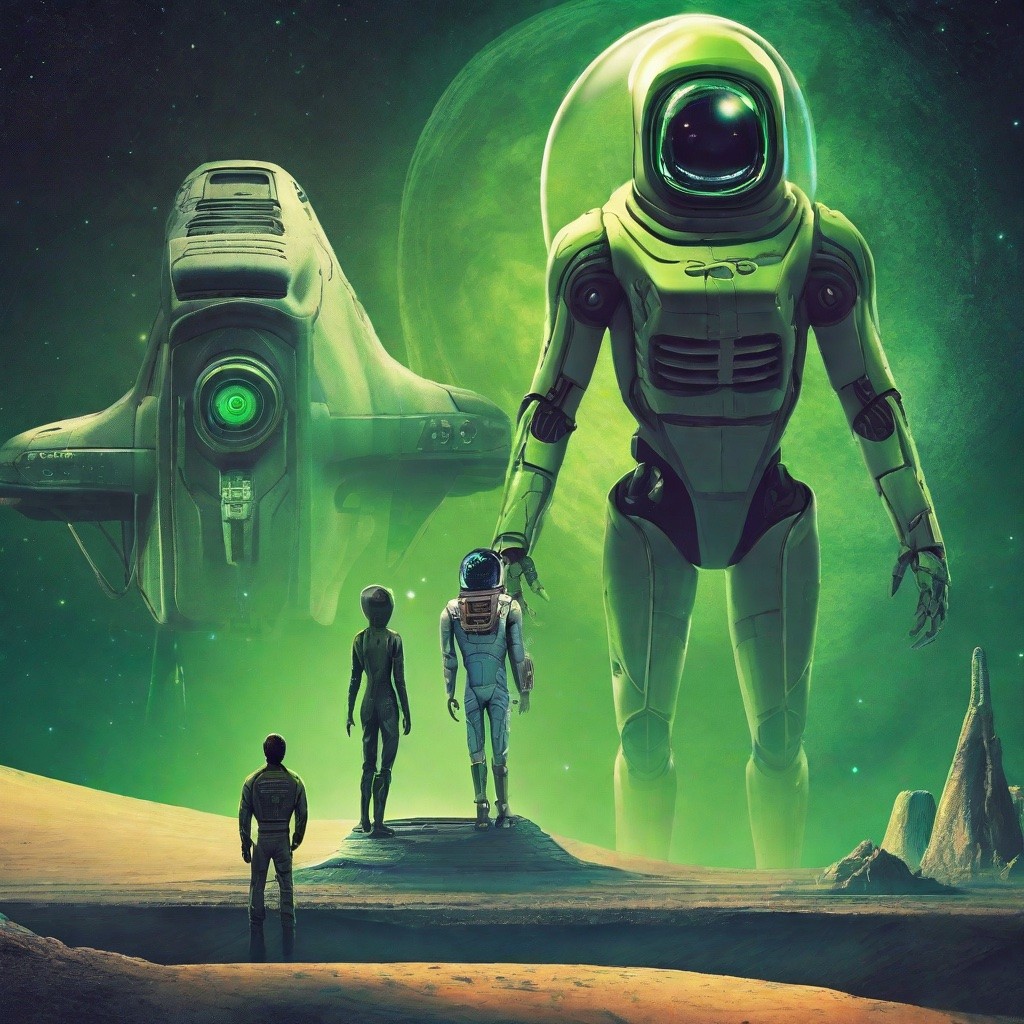} &
        \includegraphics[width=0.2\textwidth]{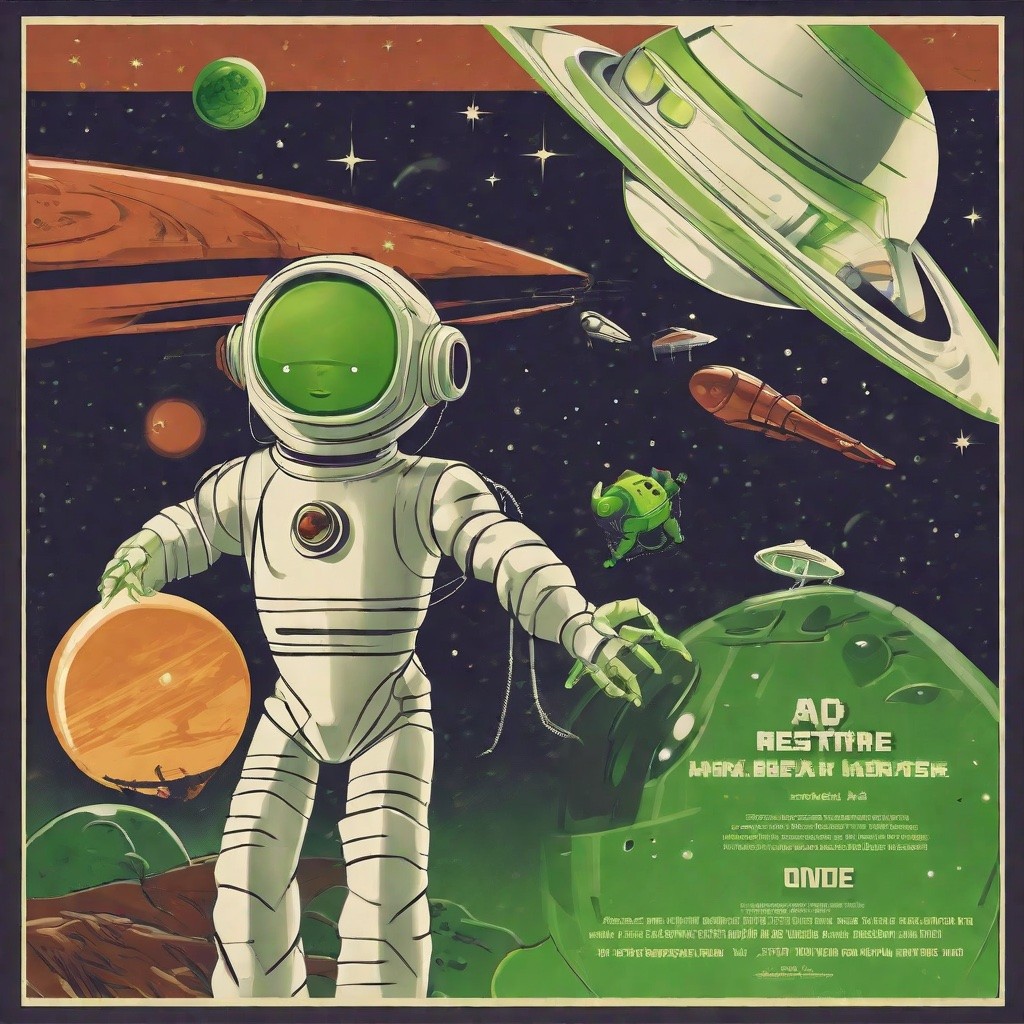} &
        \includegraphics[width=0.2\textwidth]{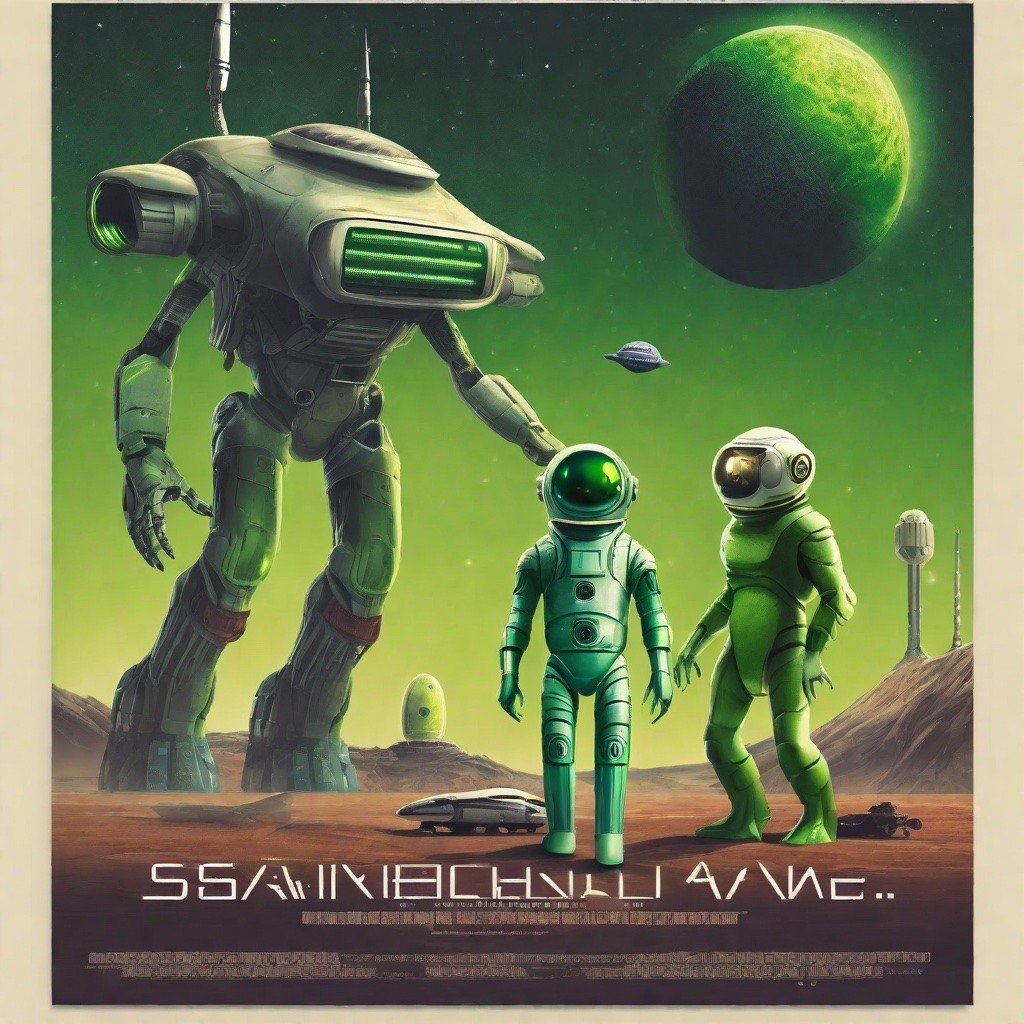} &
        \includegraphics[width=0.2\textwidth]{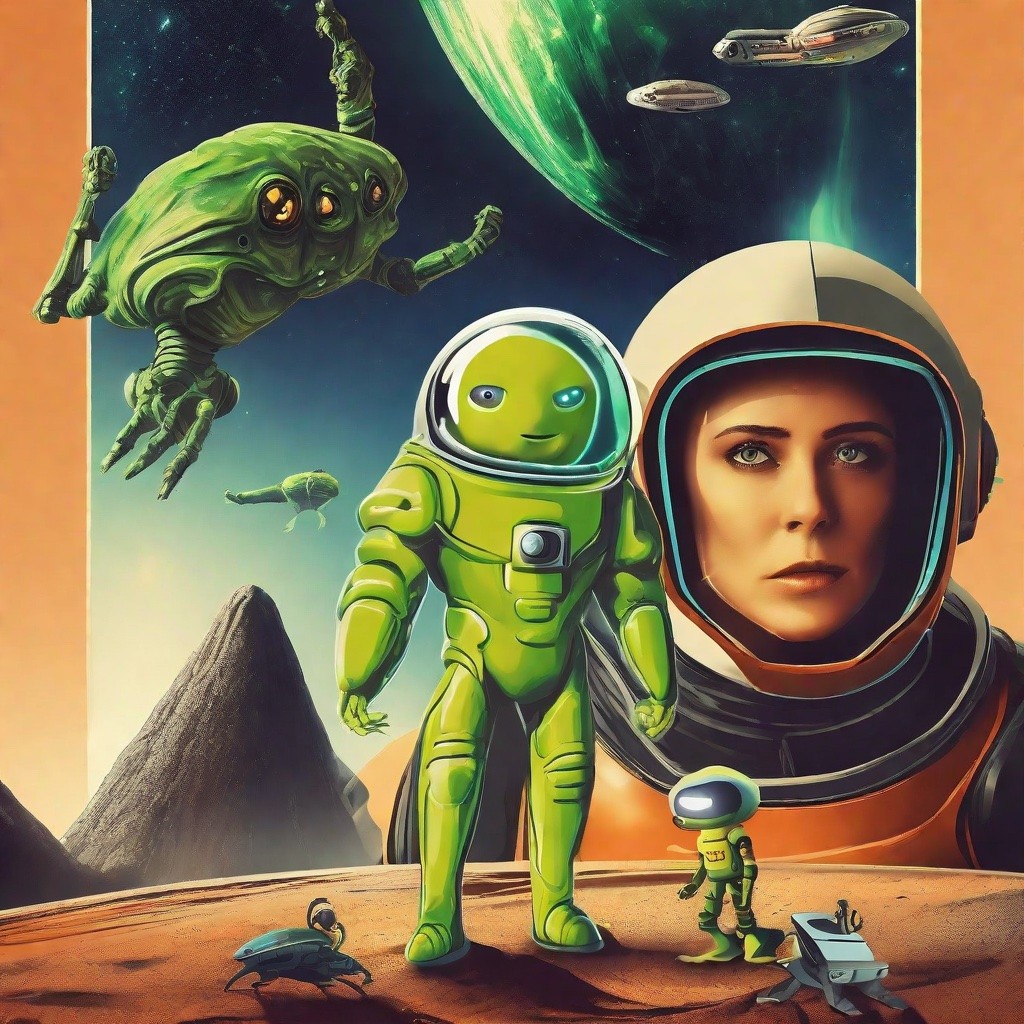} \\
        \\[-10pt]

        &
        \multicolumn{5}{c}{``A \textcolor{red}{\textit{\underline{porcelain pot}}} with \textcolor{blue}{\textit{\underline{tulips}}} and a \textcolor{green}{\textit{\underline{metal can}}} with \textcolor{orange}{\textit{\underline{orchids}}} and a \textcolor{purple}{\textit{\underline{glass jar}}} with \textcolor{Turquoise}{\textit{\underline{sunflowers}}}''} \\
        \raisebox{15pt}{\rotatebox{90}{Bounded Attention}} &
        \includegraphics[width=0.2\textwidth]{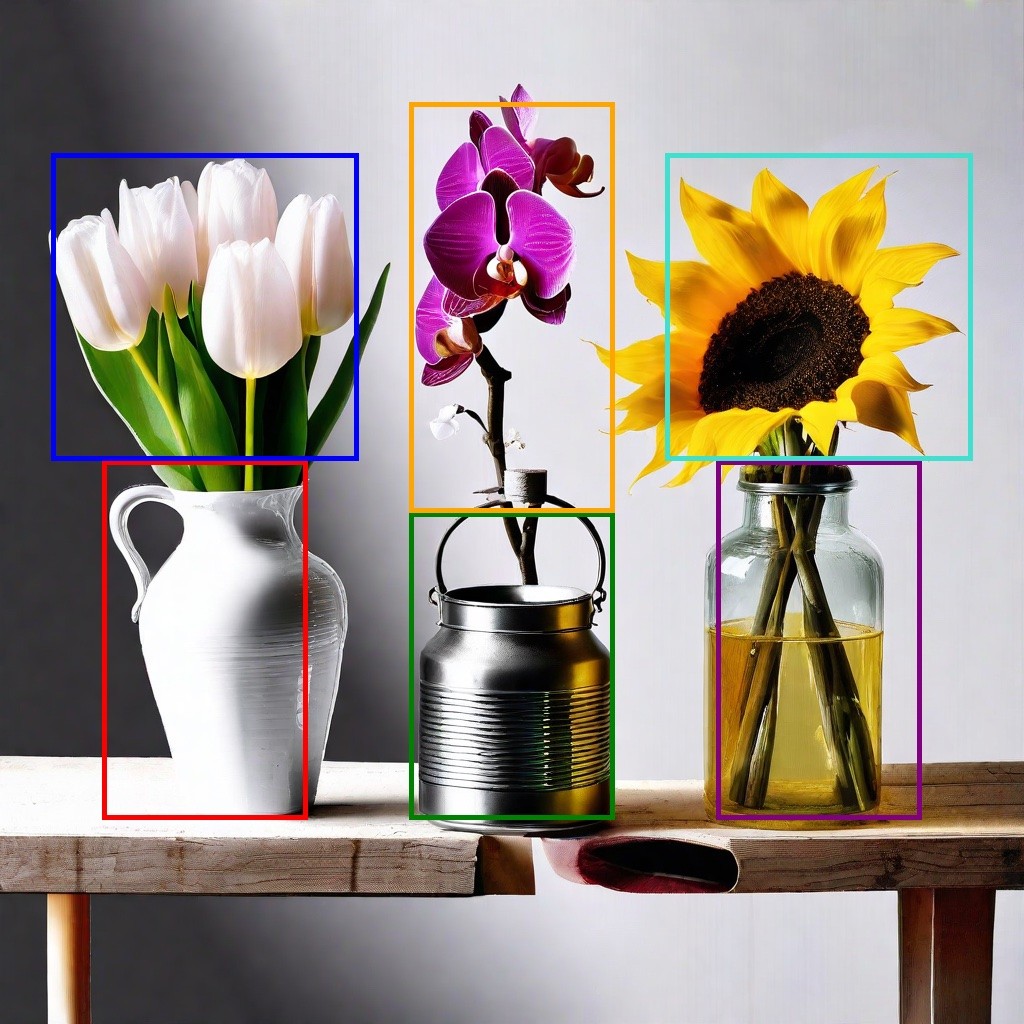} &
        \includegraphics[width=0.2\textwidth]{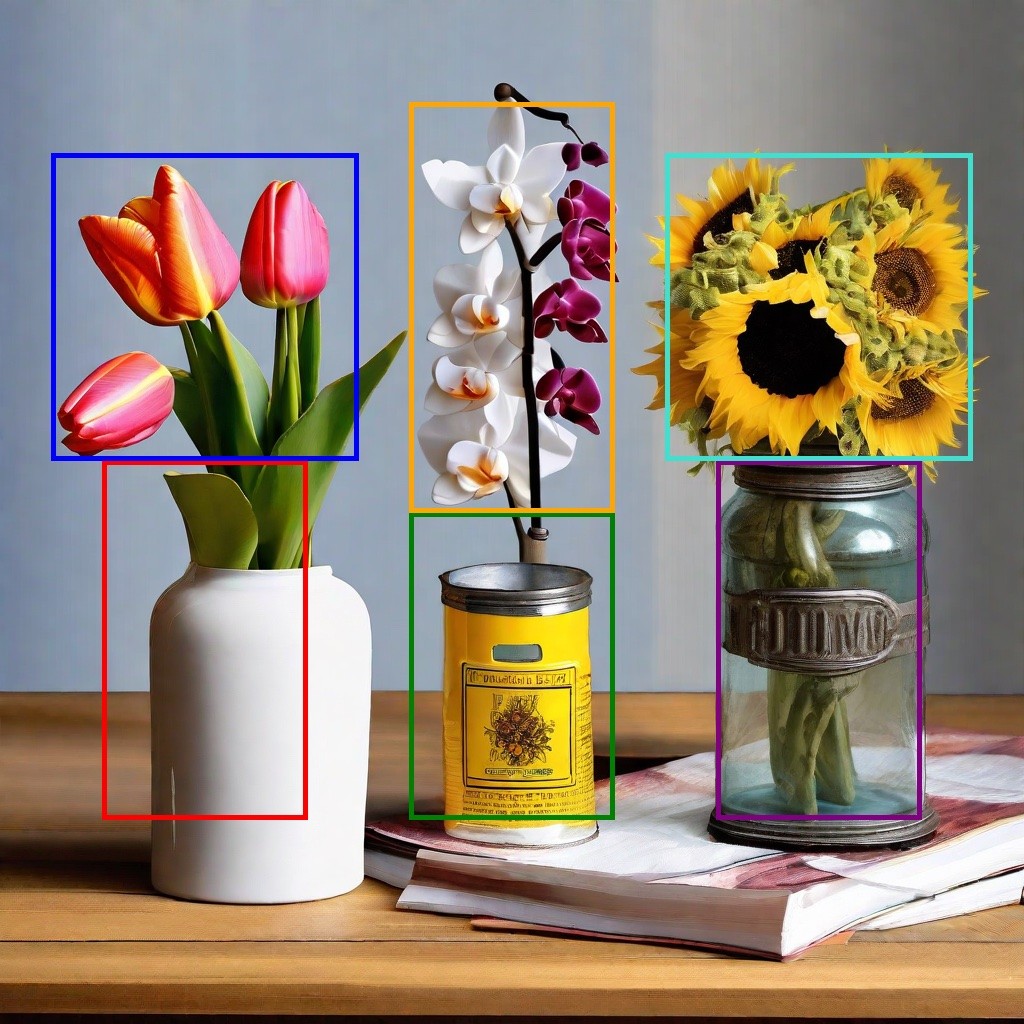} &
        \includegraphics[width=0.2\textwidth]{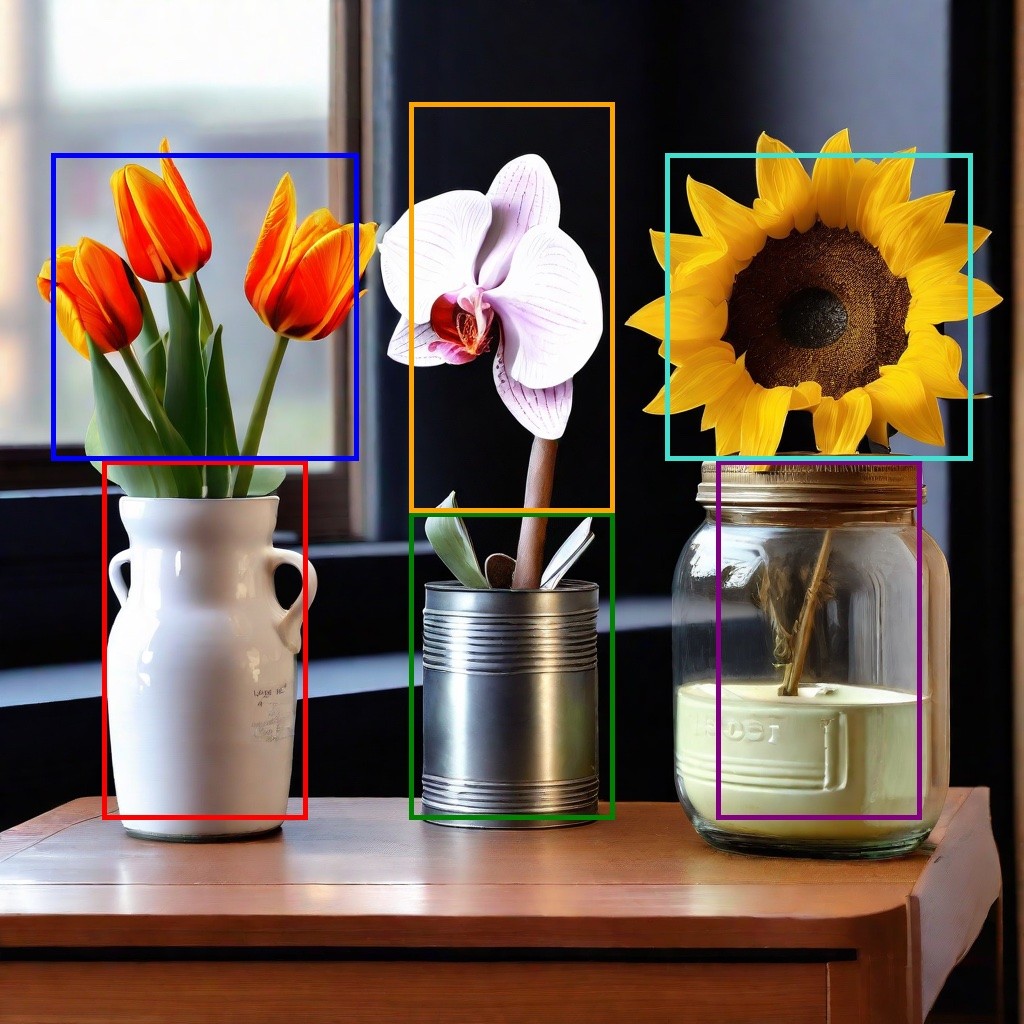} &
        \includegraphics[width=0.2\textwidth]{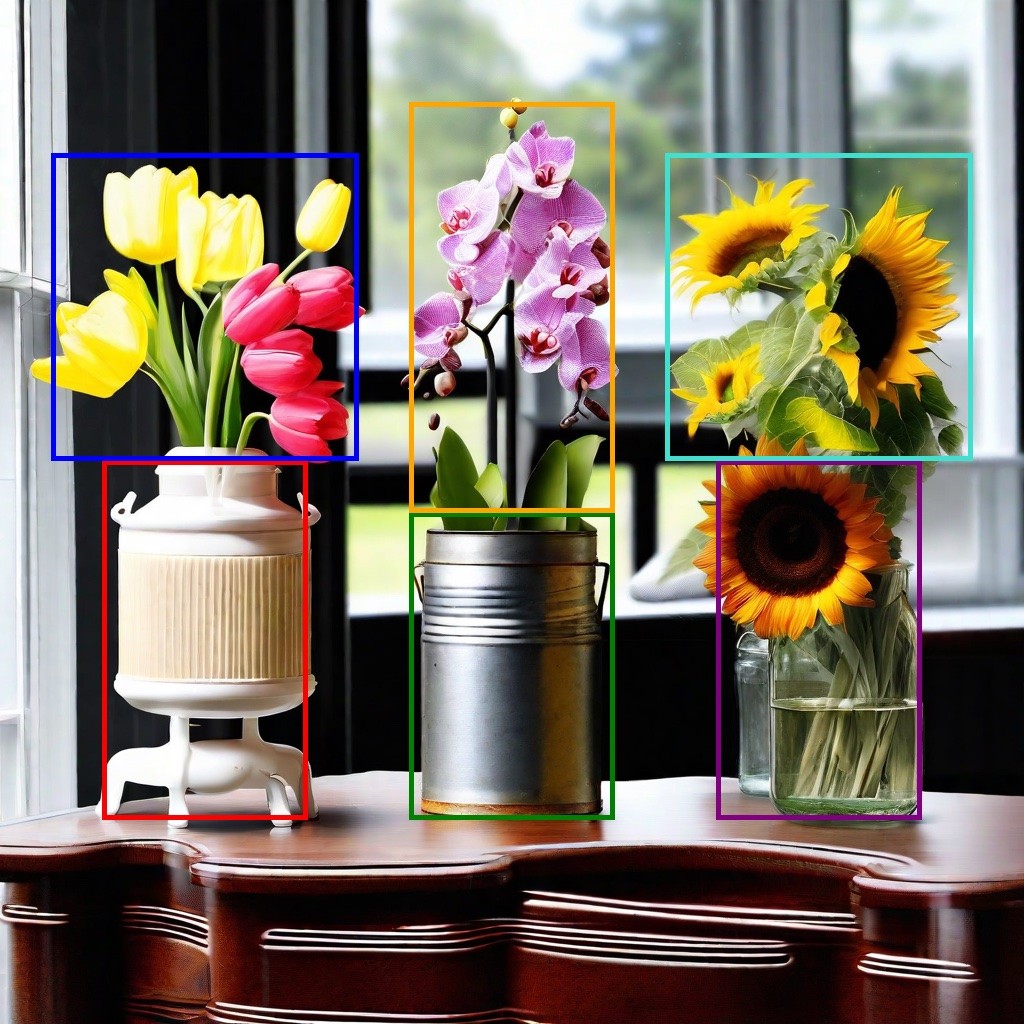} &
        \includegraphics[width=0.2\textwidth]{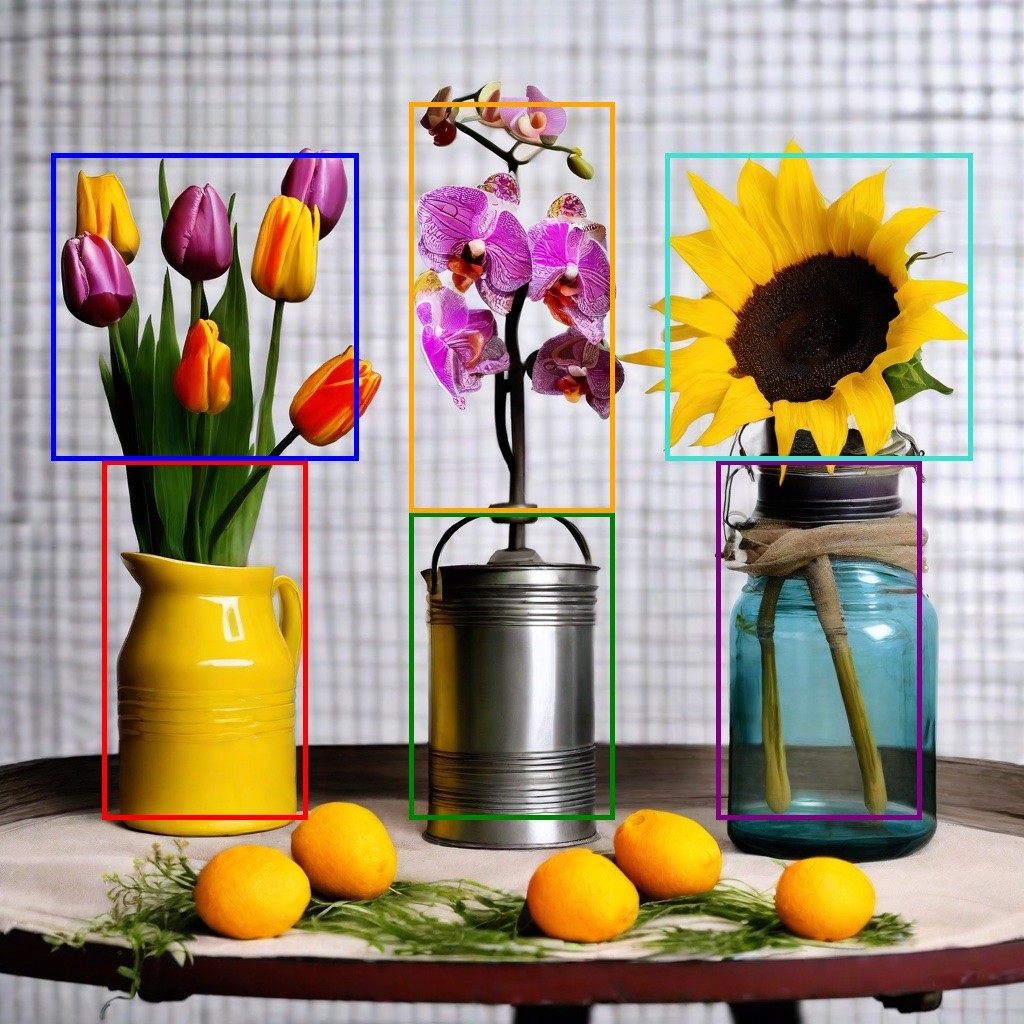} \\
        \raisebox{24pt}{\rotatebox{90}{Vanilla SDXL}} &
        \includegraphics[width=0.2\textwidth]{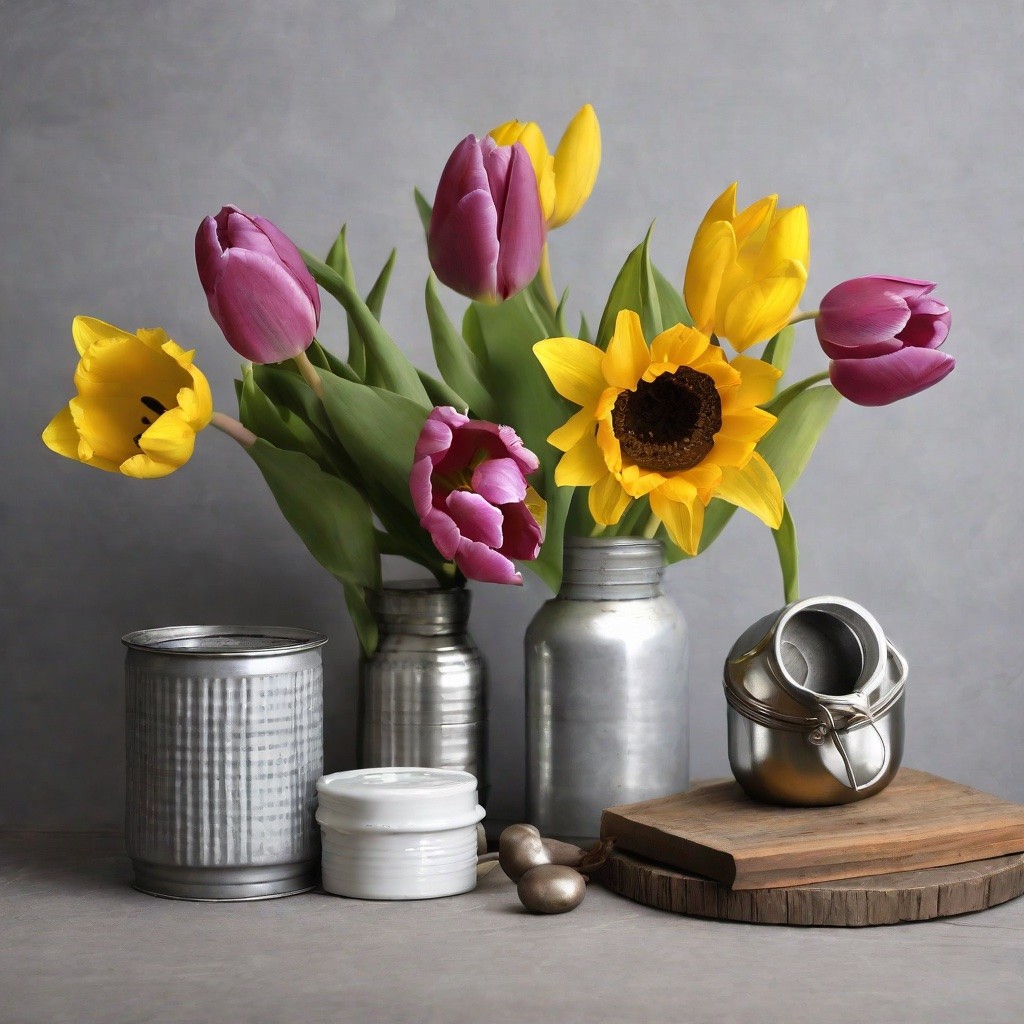} &
        \includegraphics[width=0.2\textwidth]{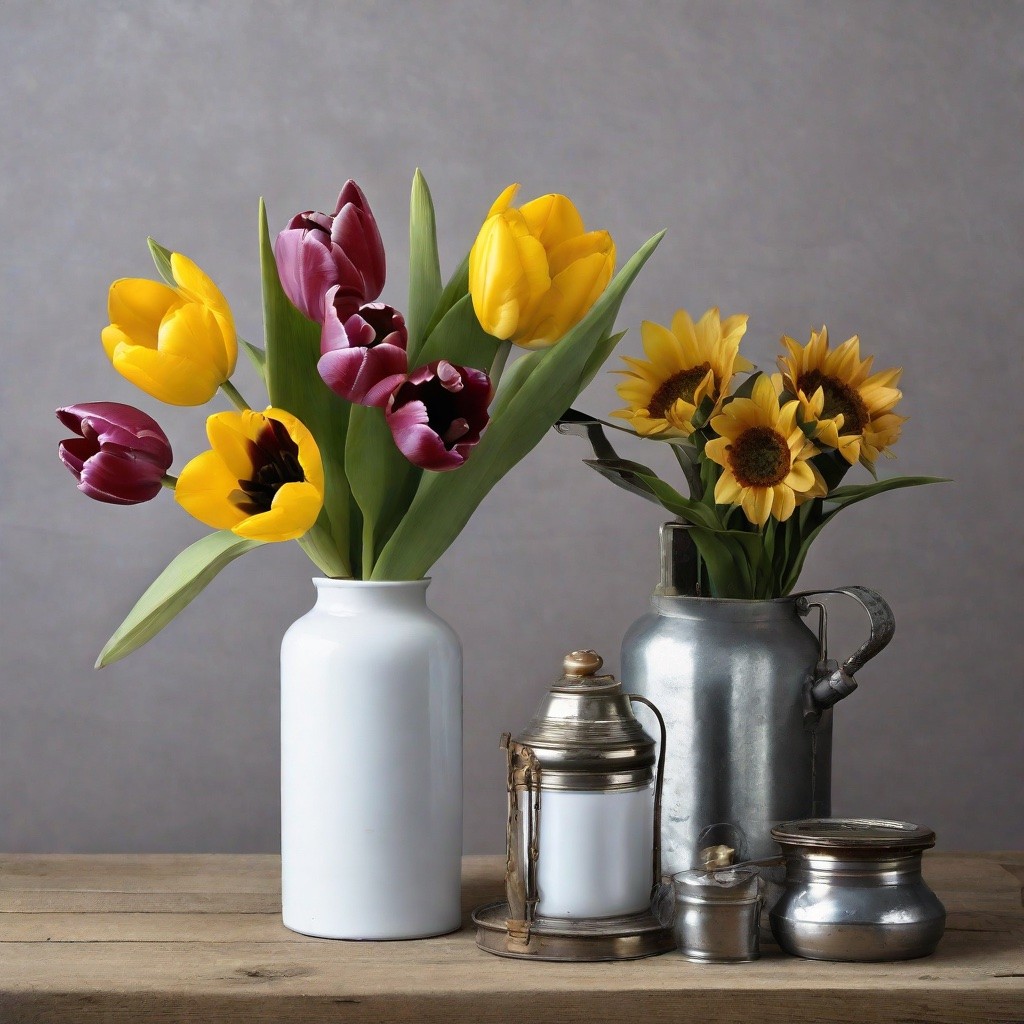} &
        \includegraphics[width=0.2\textwidth]{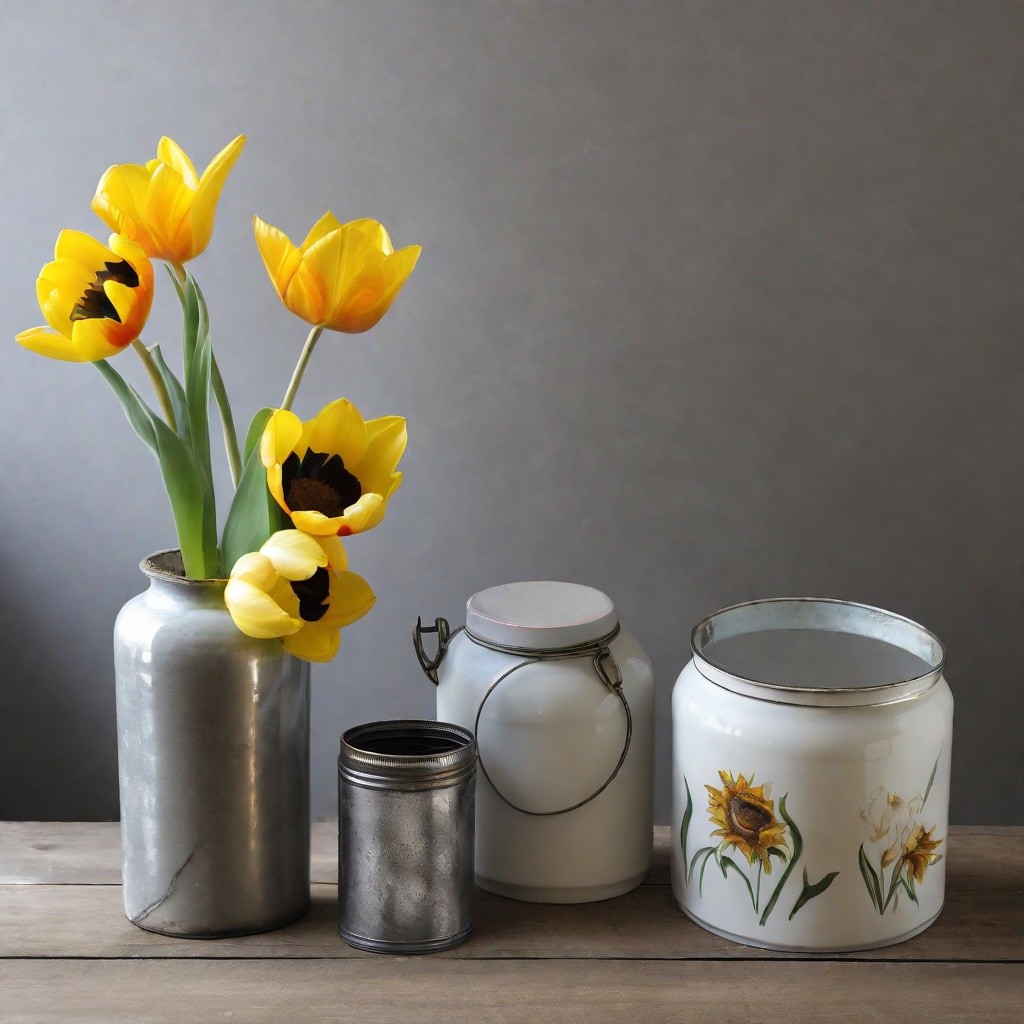} &
        \includegraphics[width=0.2\textwidth]{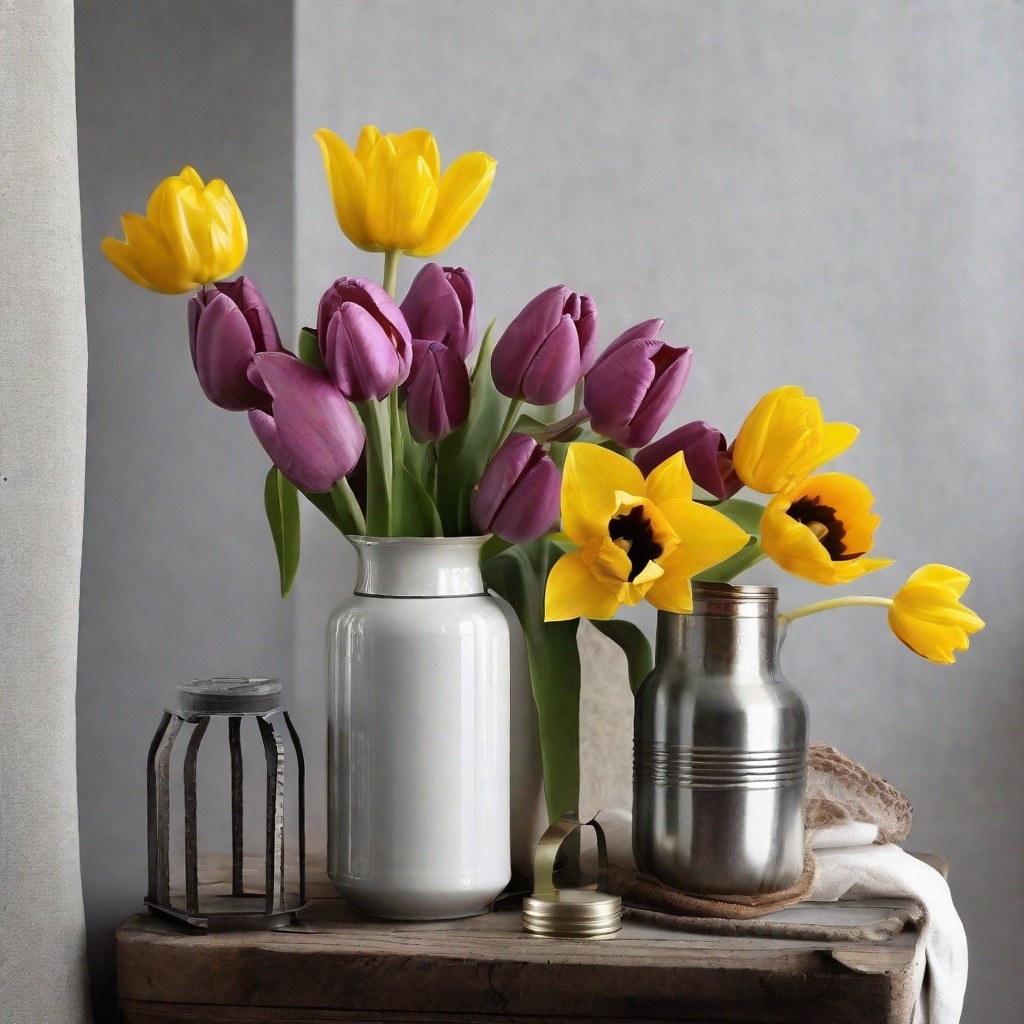} &
        \includegraphics[width=0.2\textwidth]{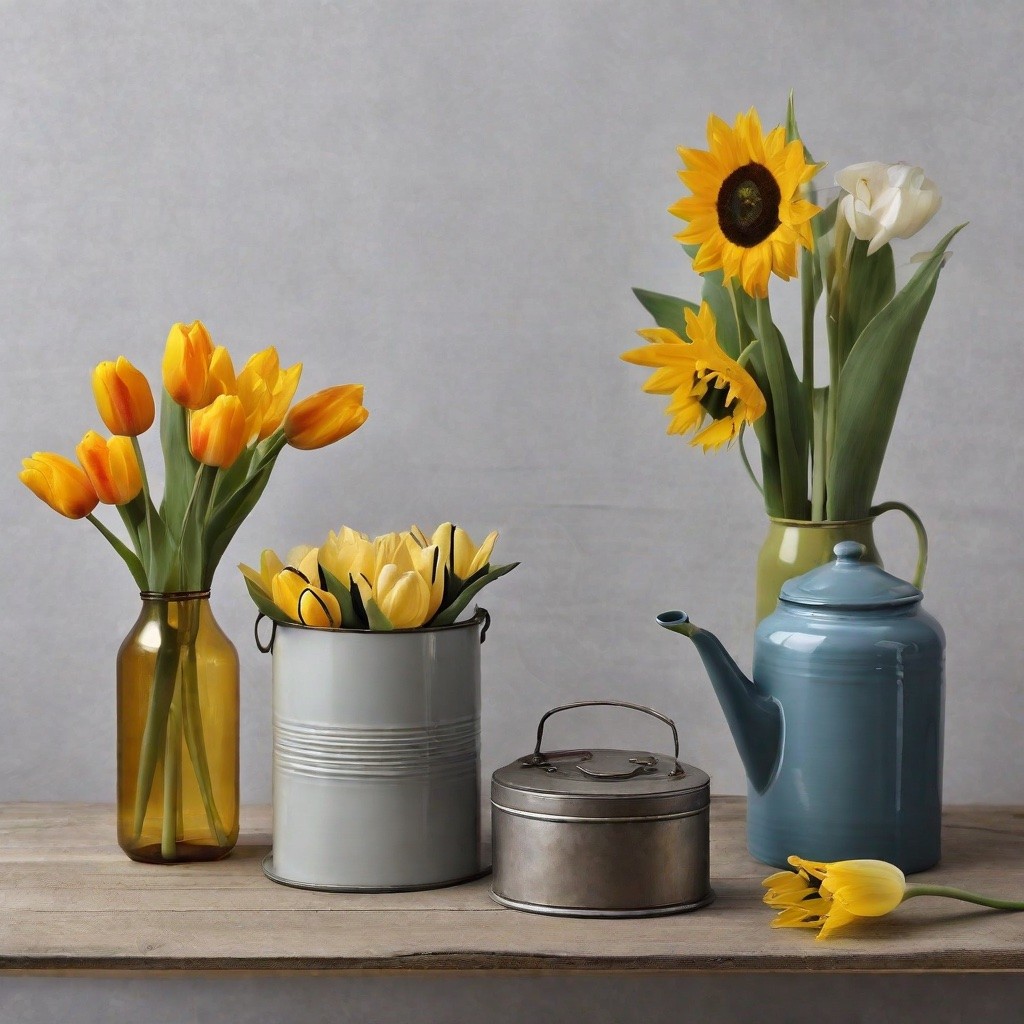} \\
        \\
        
    \end{tabular}
    }
    \vspace{-24pt}
    \captionof{figure}{
        More results of our method using SDXL.
    }
    \label{fig:sdxl3}
\end{figure*}

%% file: figures/sup/sdxl_results4.tex
\begin{figure*}[t]
    \setlength{\tabcolsep}{1pt}
    {\small\centering
    \begin{tabular}{c c c c c c}
        &
        \multicolumn{5}{c}{``A realistic photo of a highway with a \textcolor{red}{\textit{\underline{tourist bus}}} and a \textcolor{blue}{\textit{\underline{school bus}}} and a \textcolor{green}{\textit{\underline{fire engine}}}.''}
        \\
       \raisebox{15pt}{\rotatebox{90}{Bounded Attention}} &
        \includegraphics[width=0.2\textwidth]{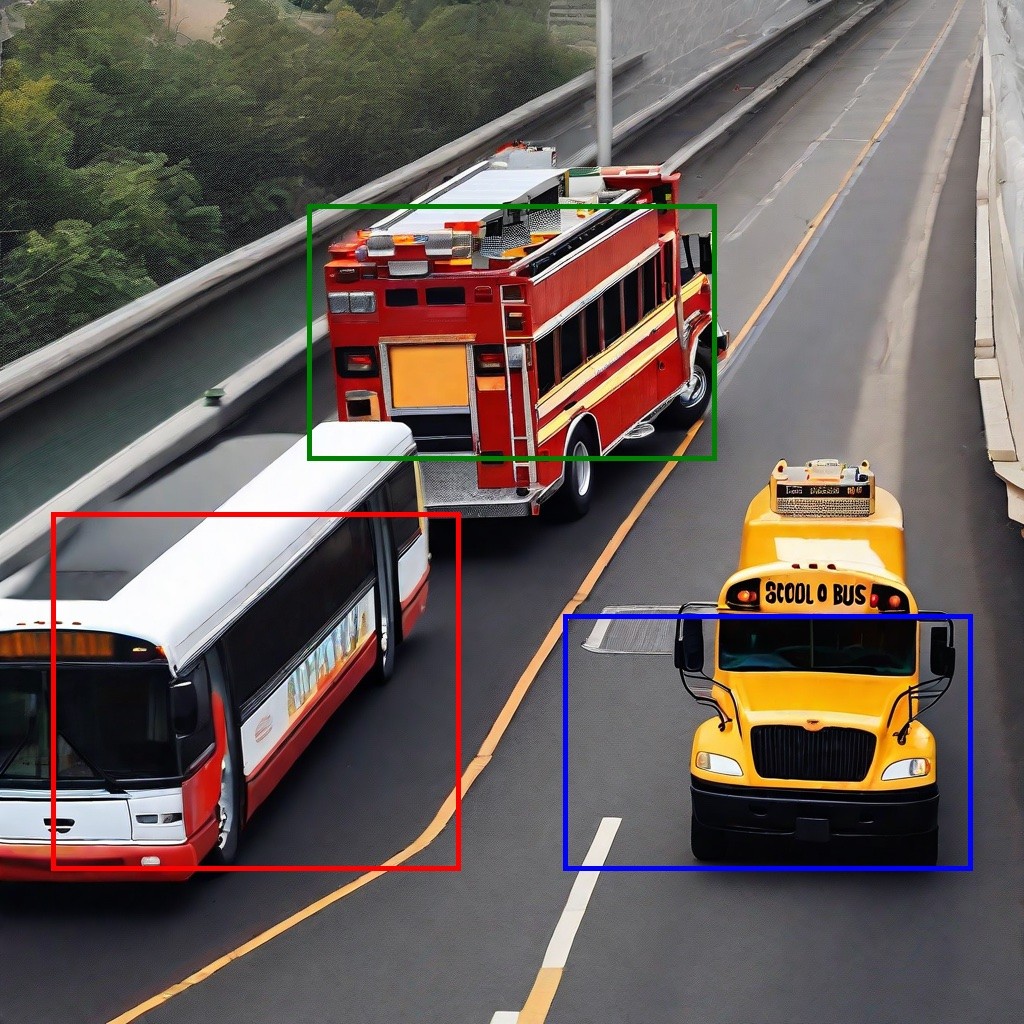} &
        \includegraphics[width=0.2\textwidth]{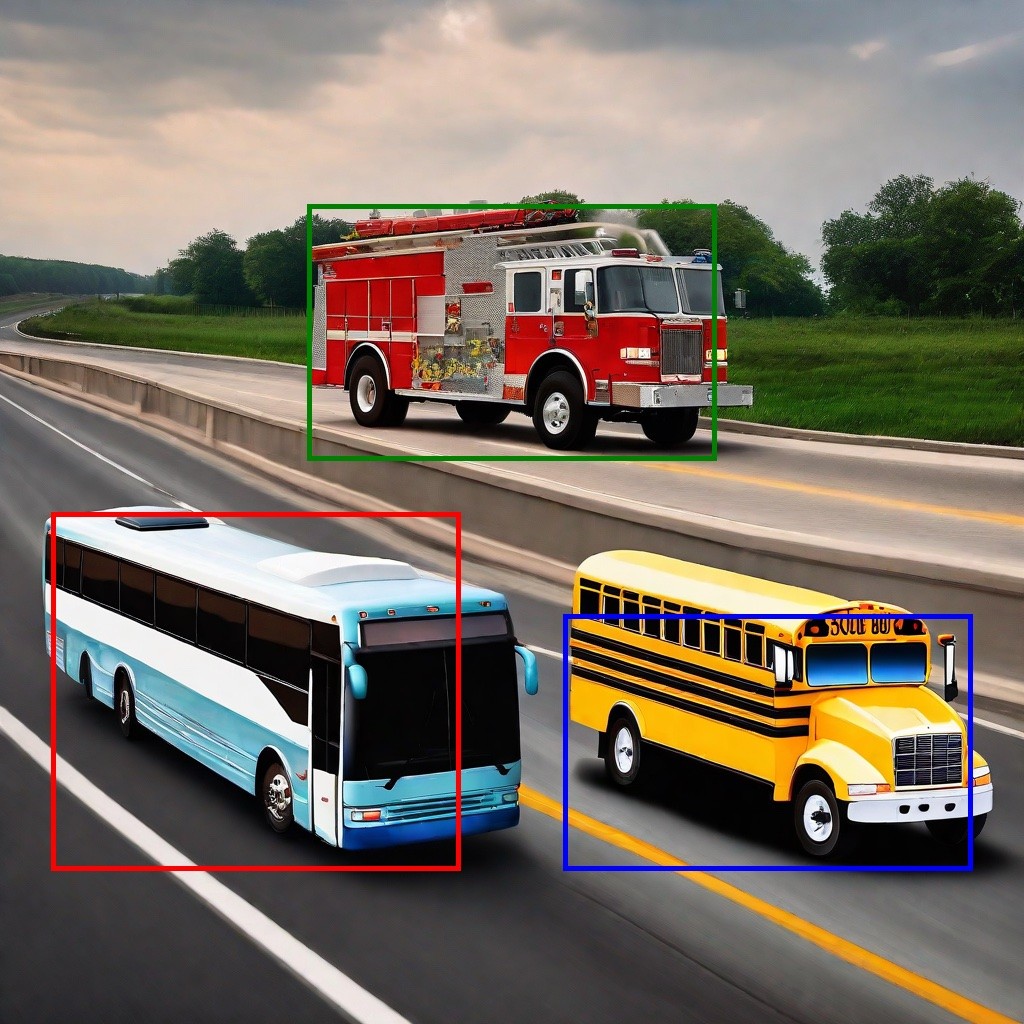} &
        \includegraphics[width=0.2\textwidth]{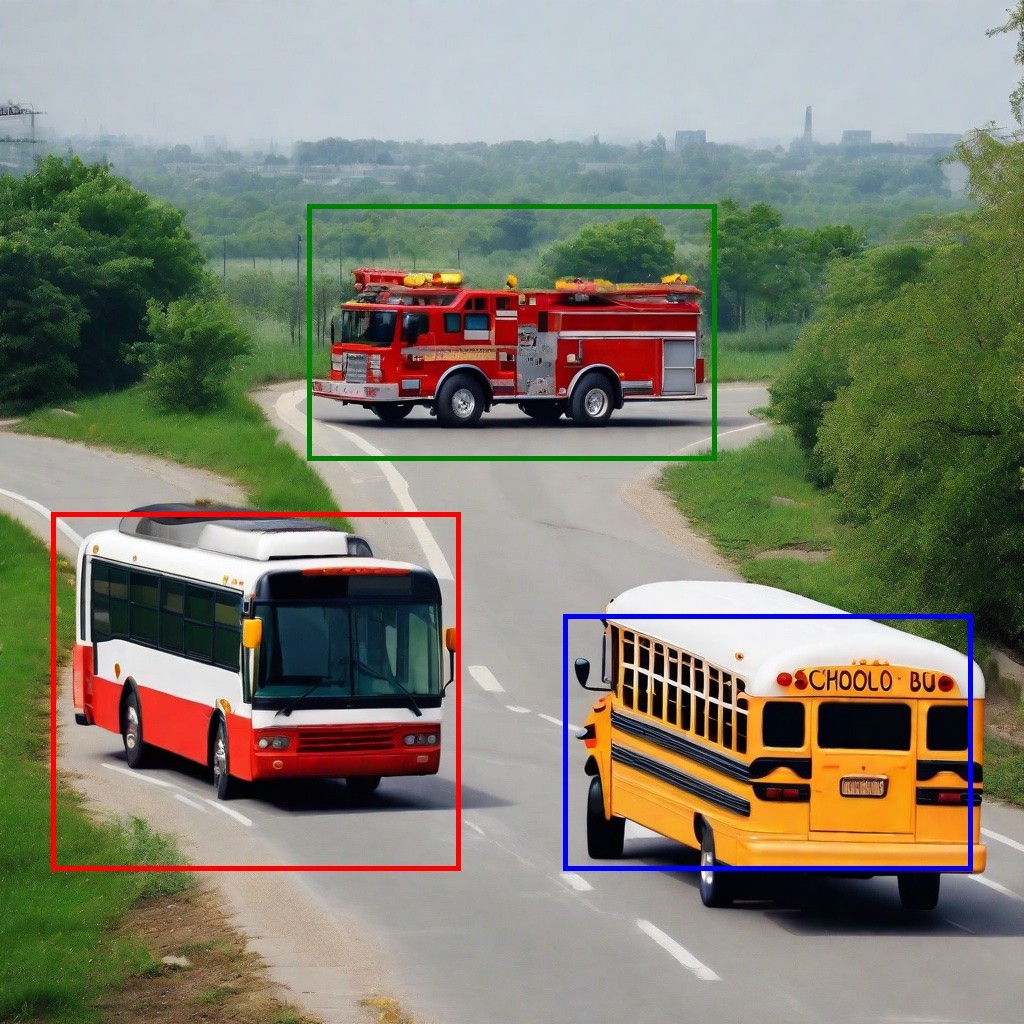} &
        \includegraphics[width=0.2\textwidth]{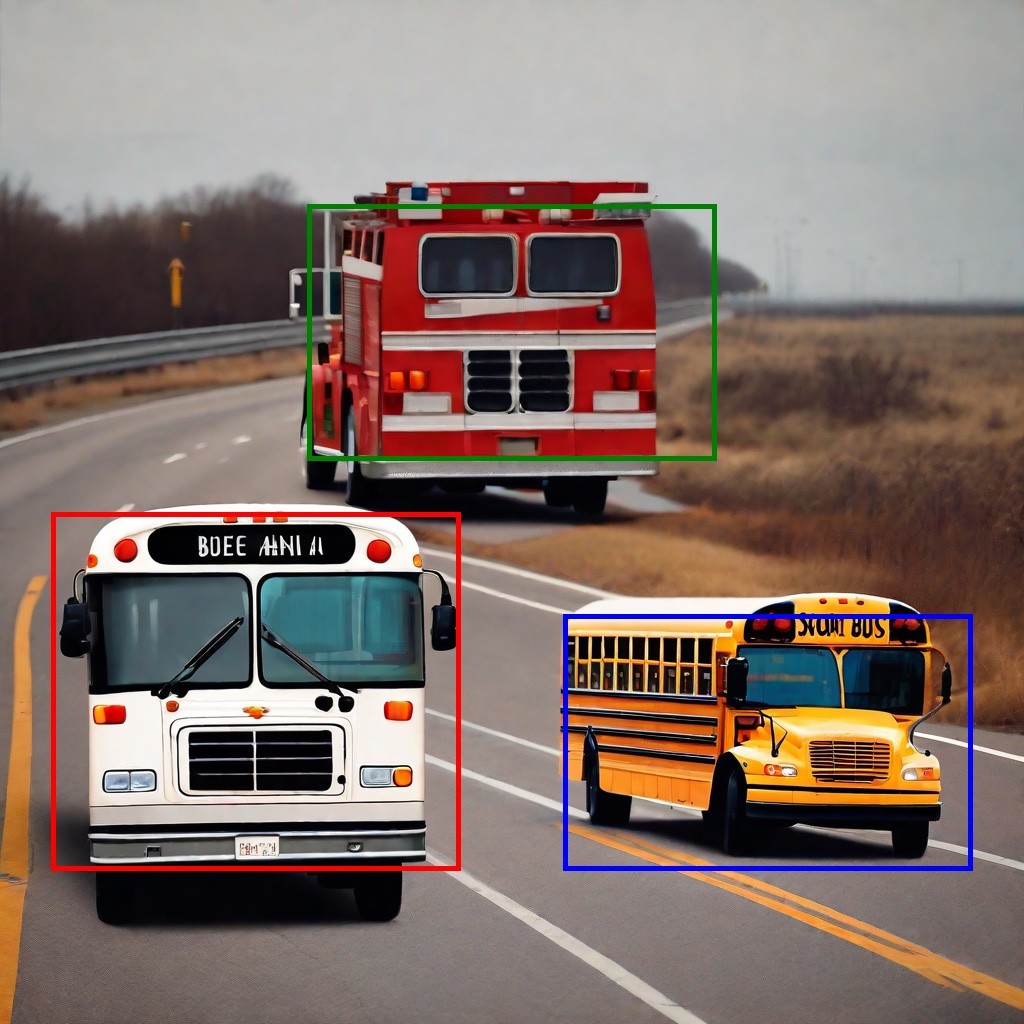} &
        \includegraphics[width=0.2\textwidth]{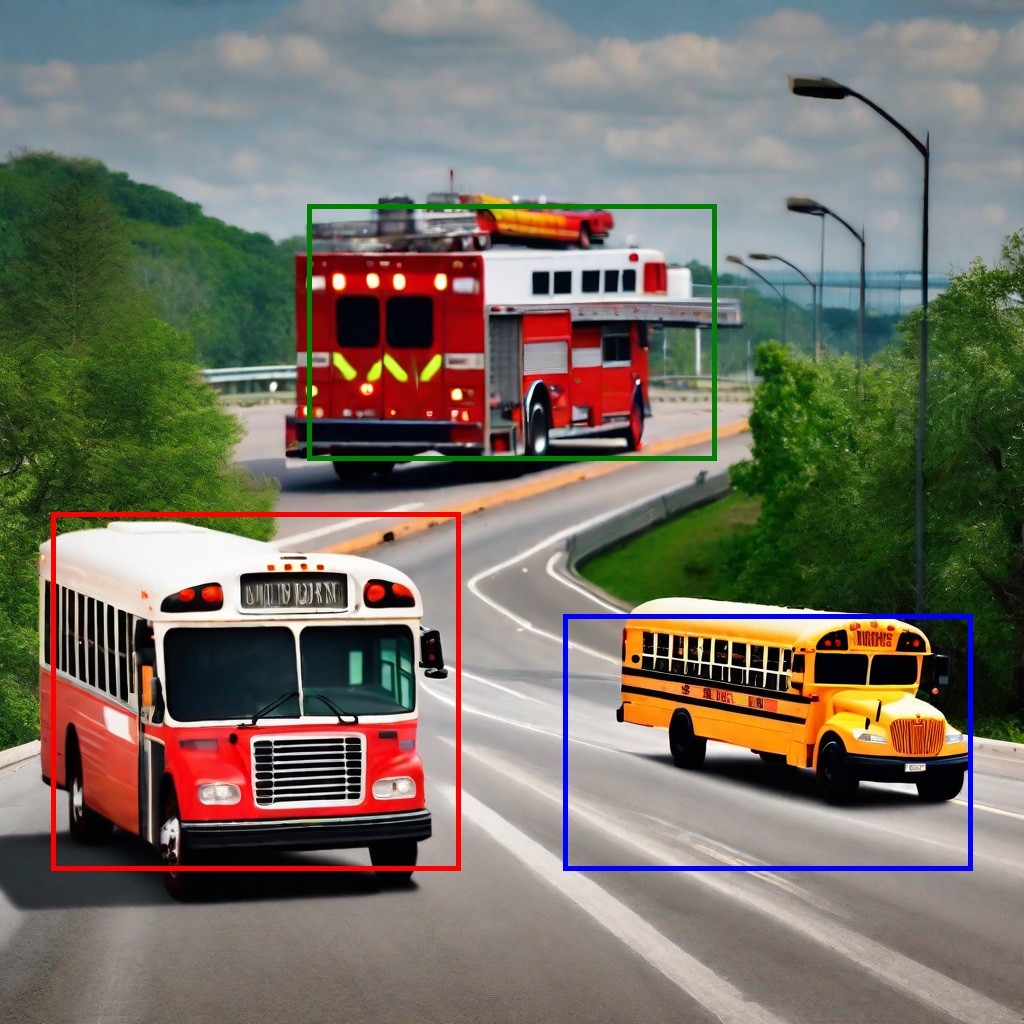} \\
        \raisebox{24pt}{\rotatebox{90}{Vanilla SDXL}} &
        \includegraphics[width=0.2\textwidth]{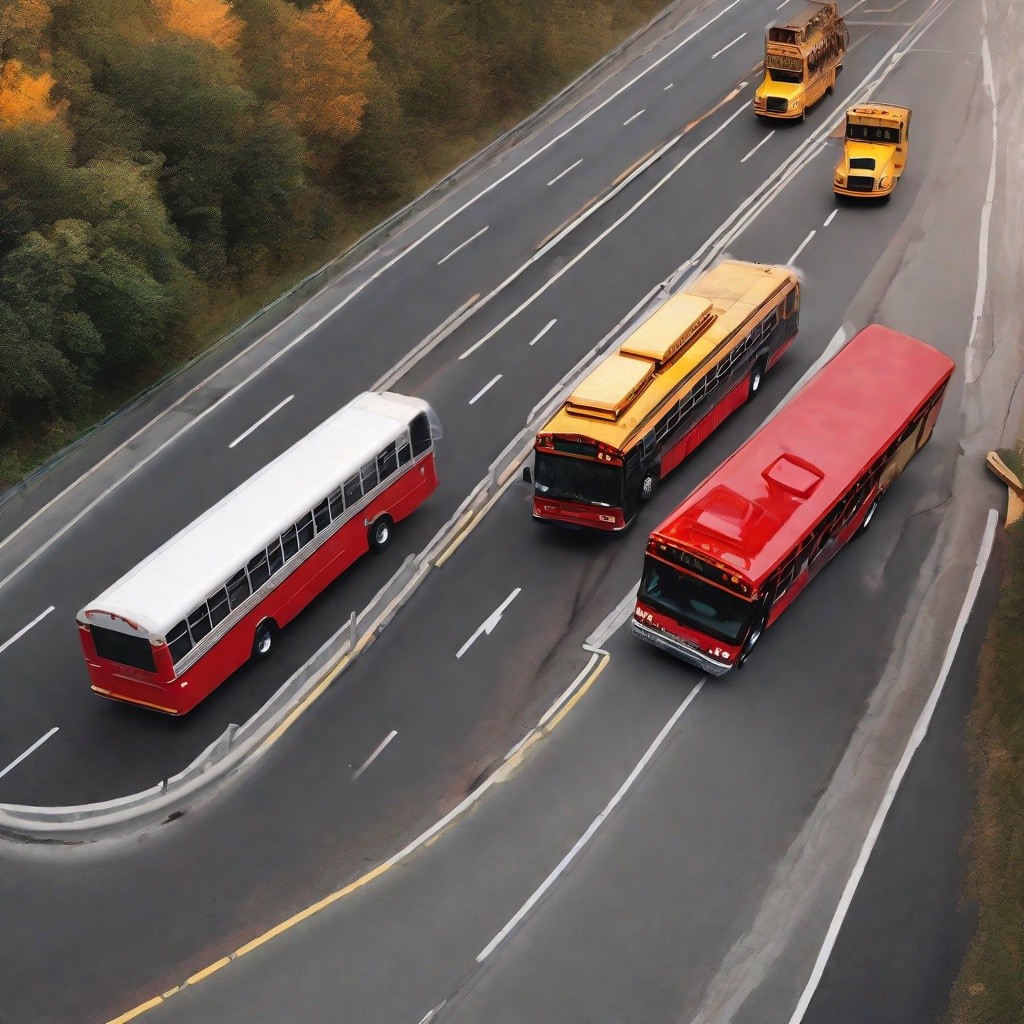} &
        \includegraphics[width=0.2\textwidth]{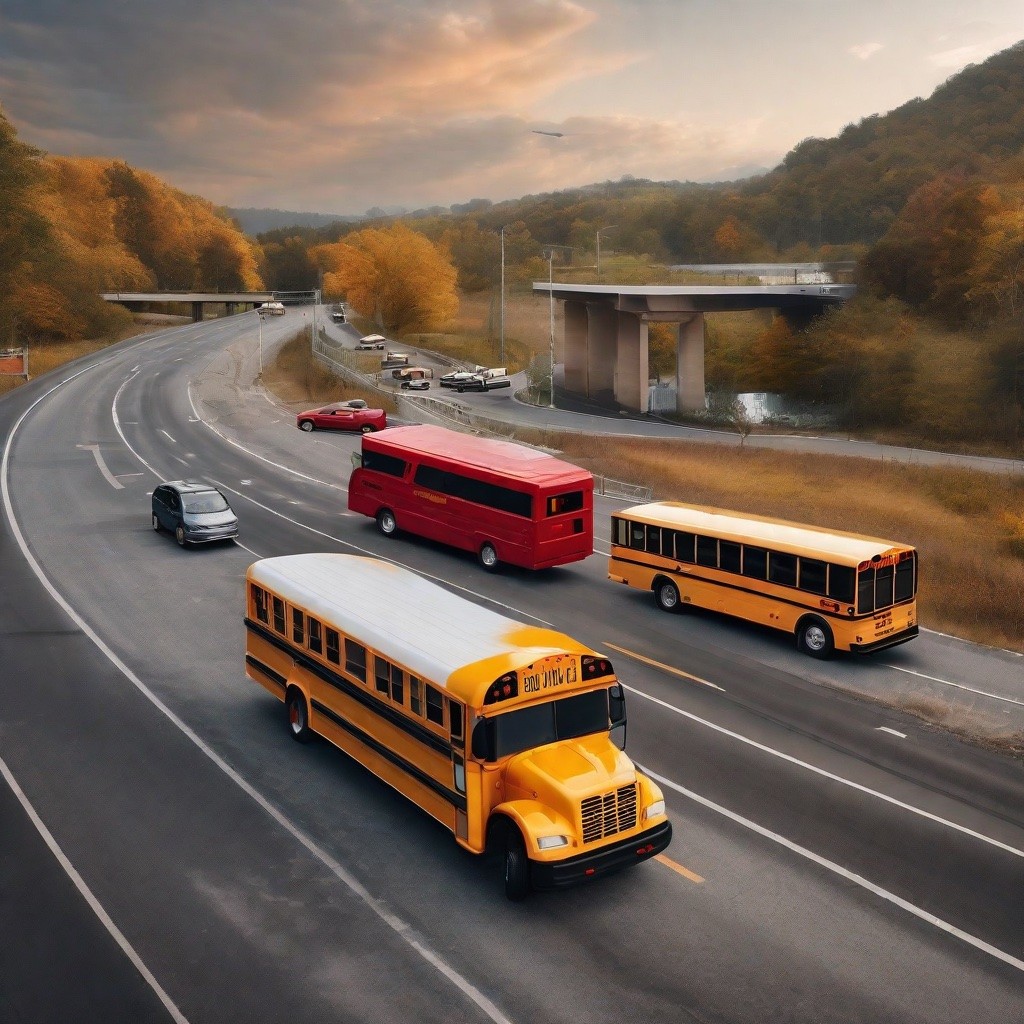} &
        \includegraphics[width=0.2\textwidth]{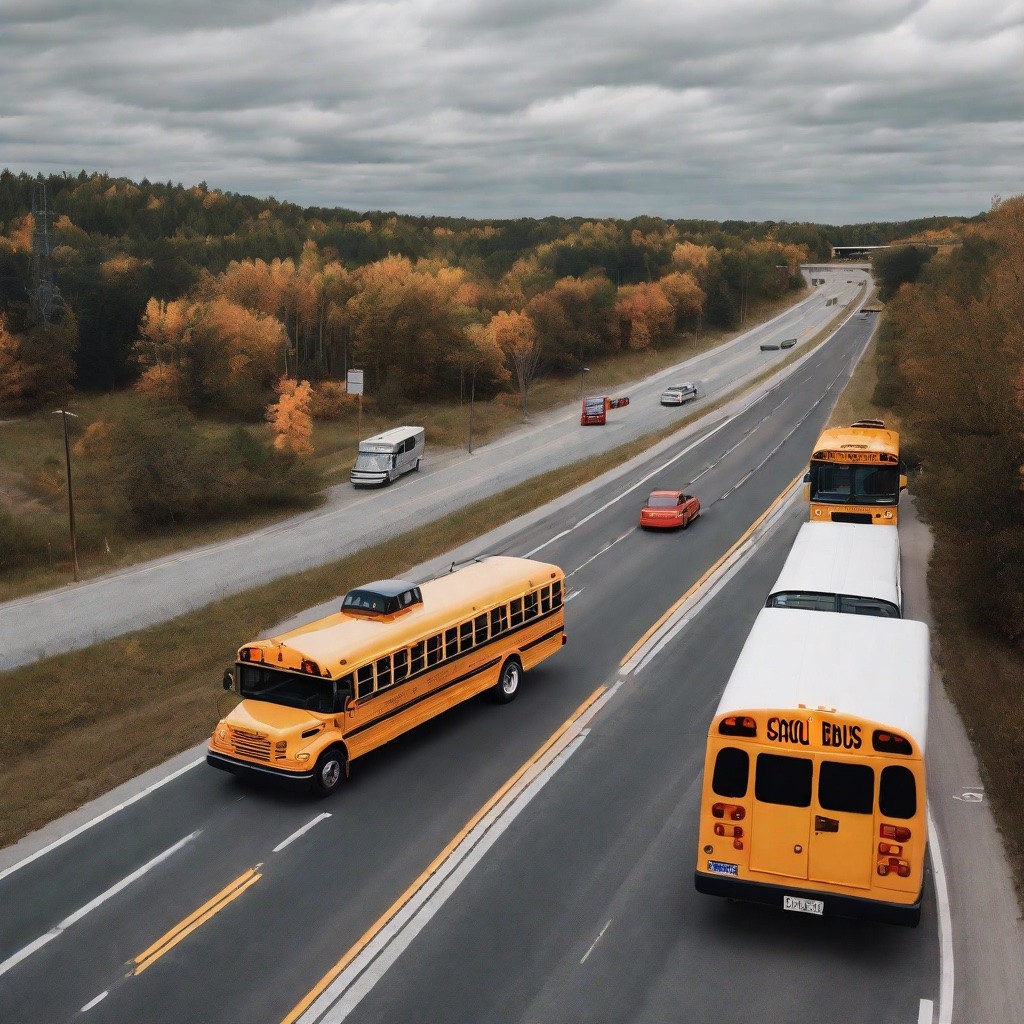} &
        \includegraphics[width=0.2\textwidth]{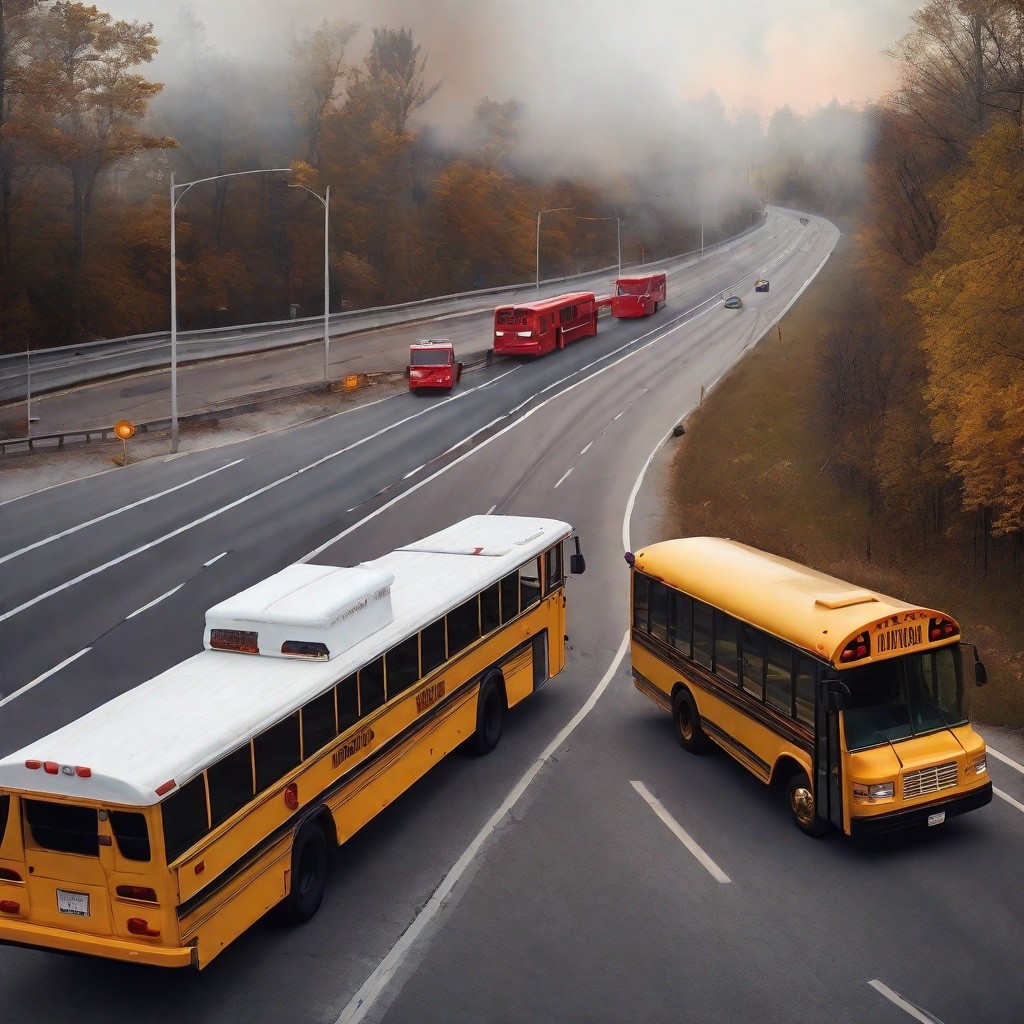} &
        \includegraphics[width=0.2\textwidth]{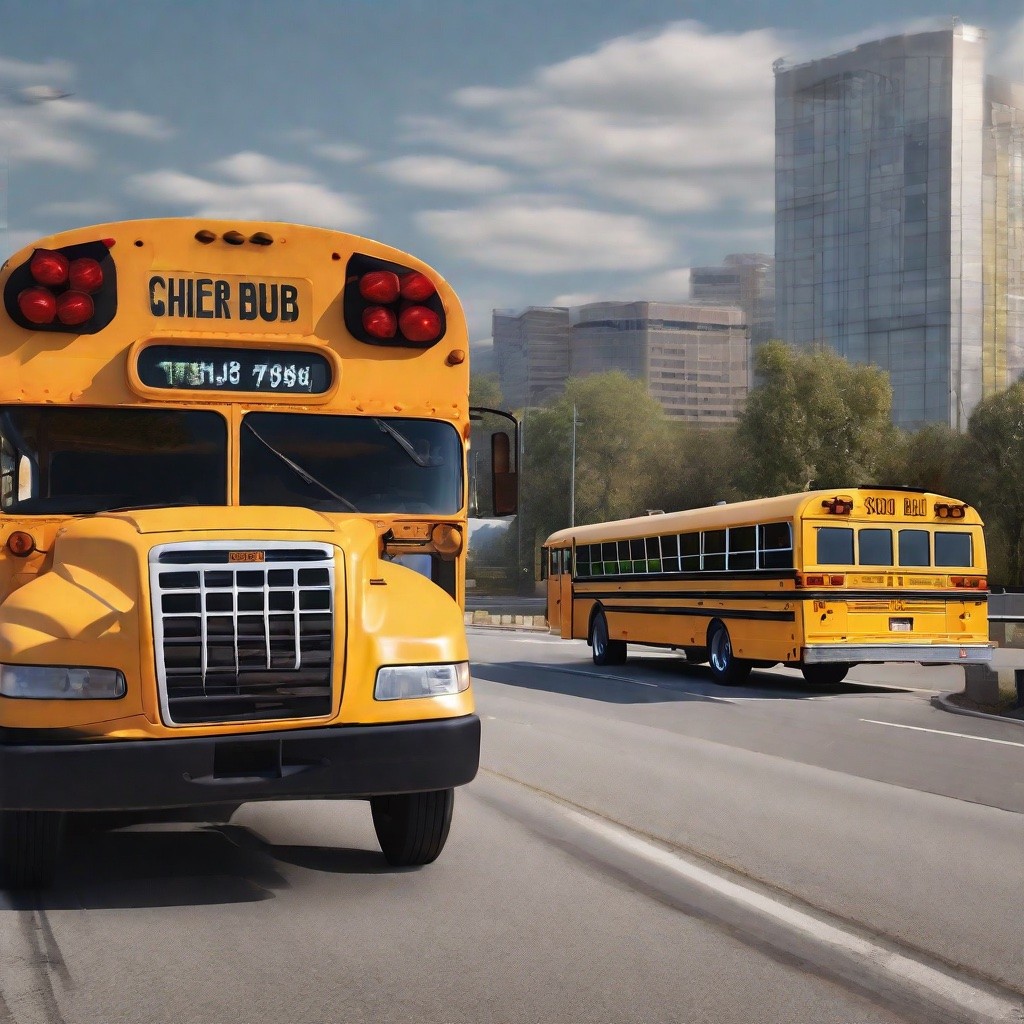} \\
        \\[-10pt]

        &
        \multicolumn{5}{c}{``A realistic photo of a highway with a \textcolor{red}{\textit{\underline{semi trailer}}} and a \textcolor{blue}{\textit{\underline{concrete mixer}}} and a \textcolor{green}{\textit{\underline{helicopter}}}.''}
        \\
       \raisebox{15pt}{\rotatebox{90}{Bounded Attention}} &
        \includegraphics[width=0.2\textwidth]{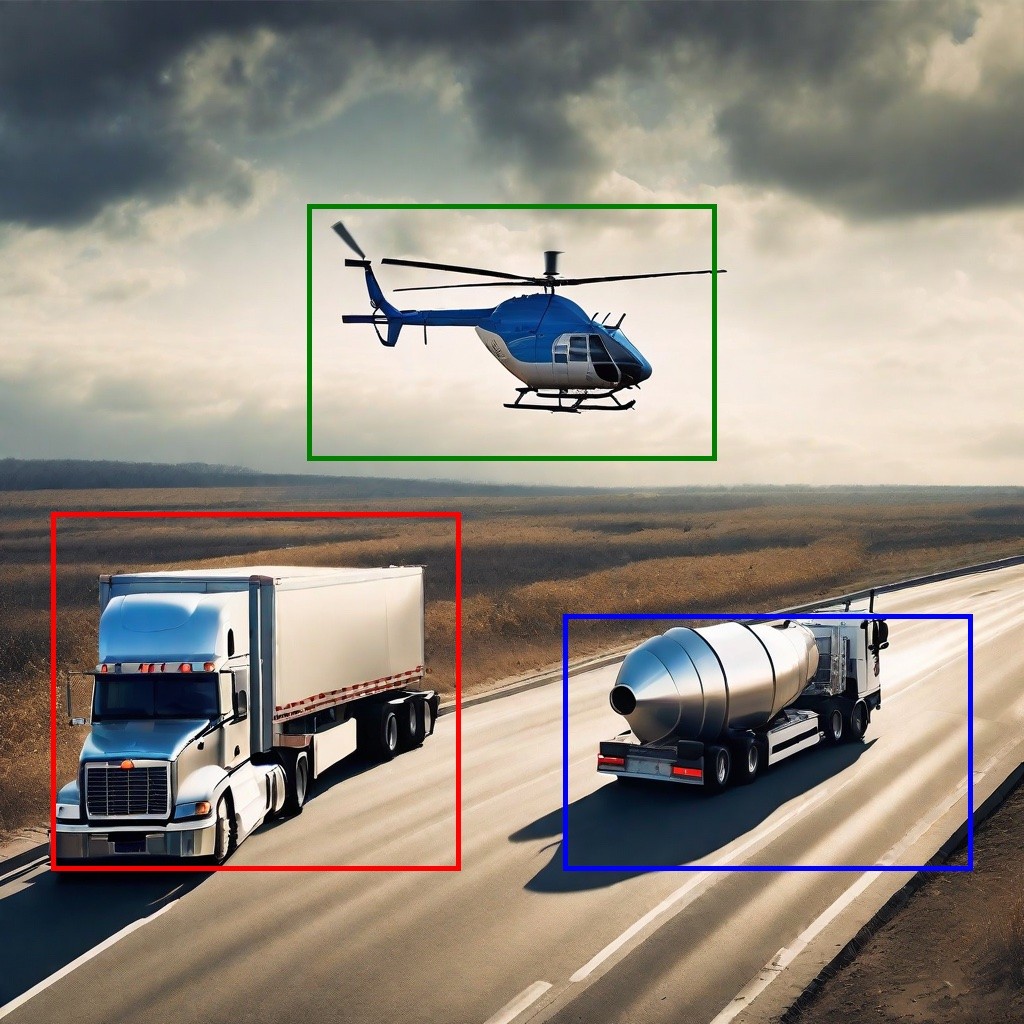} &
        \includegraphics[width=0.2\textwidth]{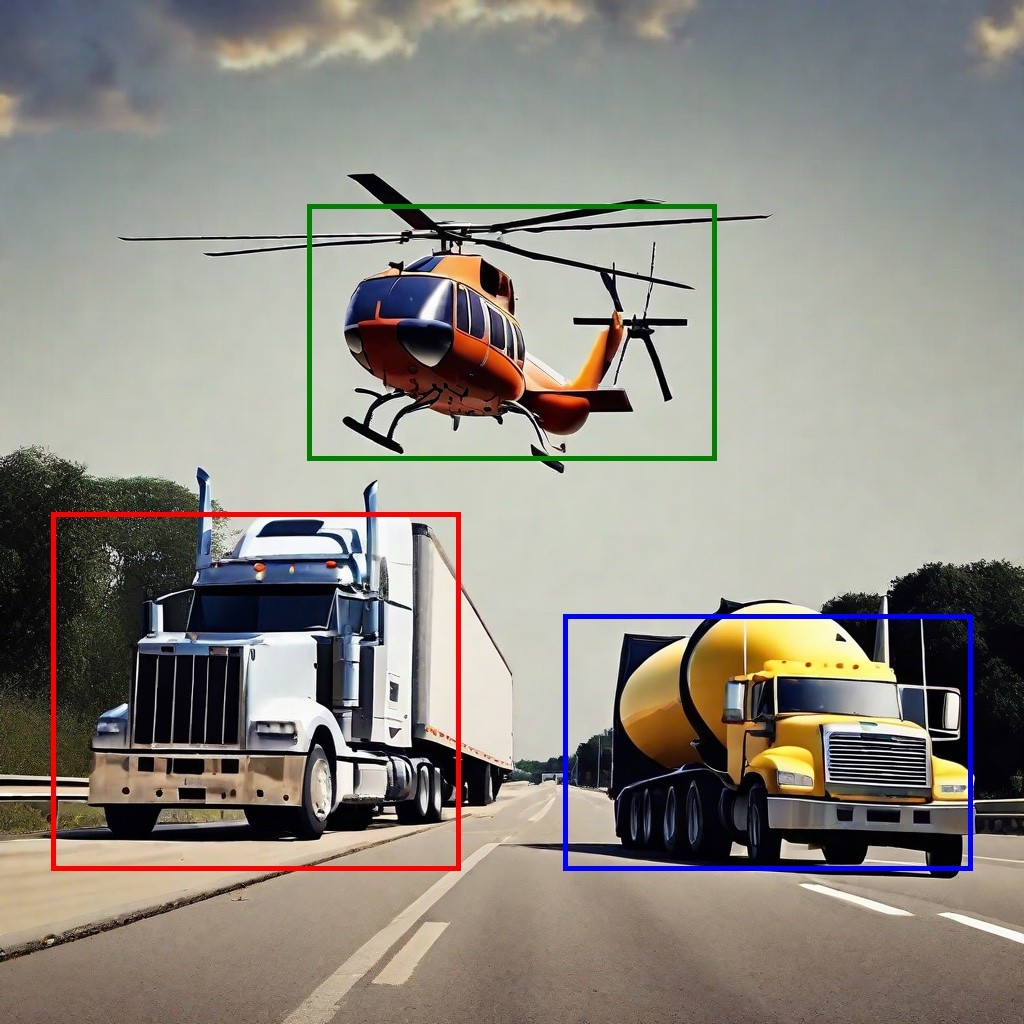} &
        \includegraphics[width=0.2\textwidth]{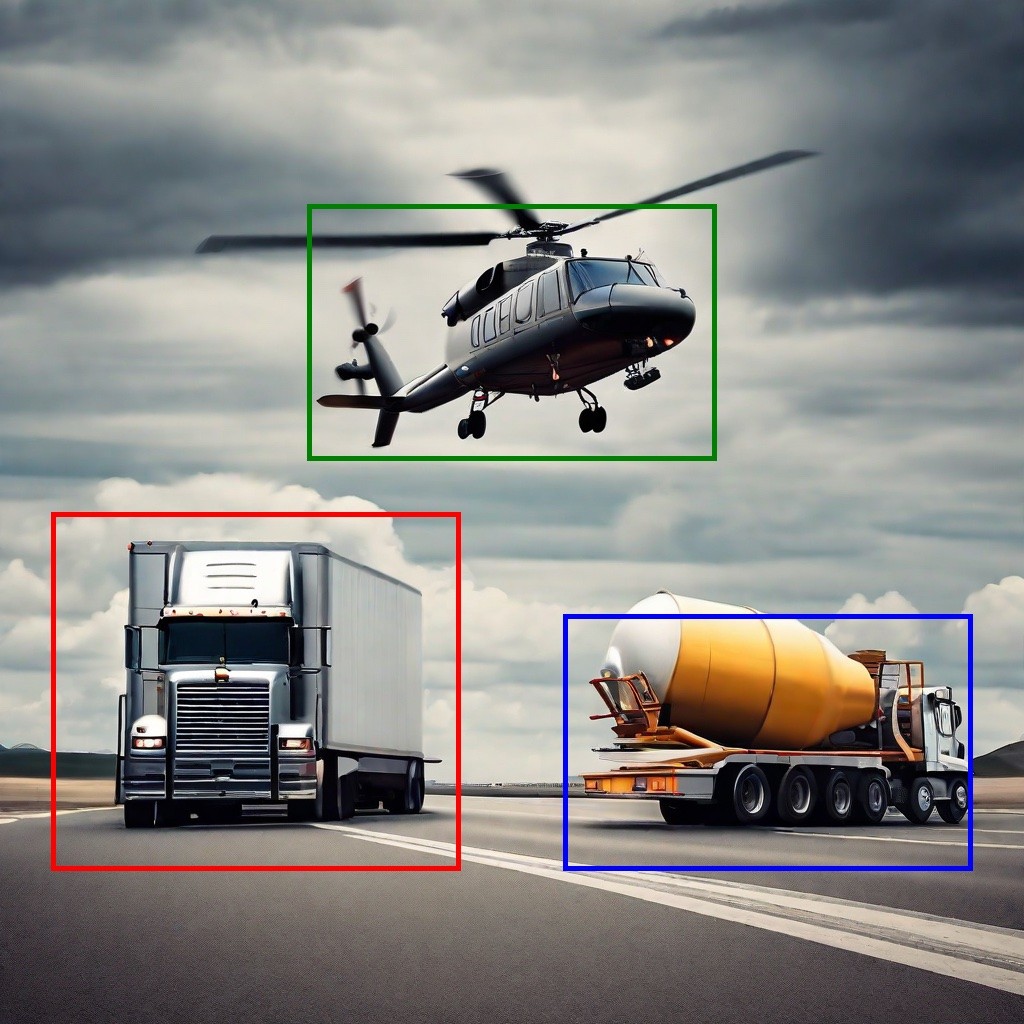} &
        \includegraphics[width=0.2\textwidth]{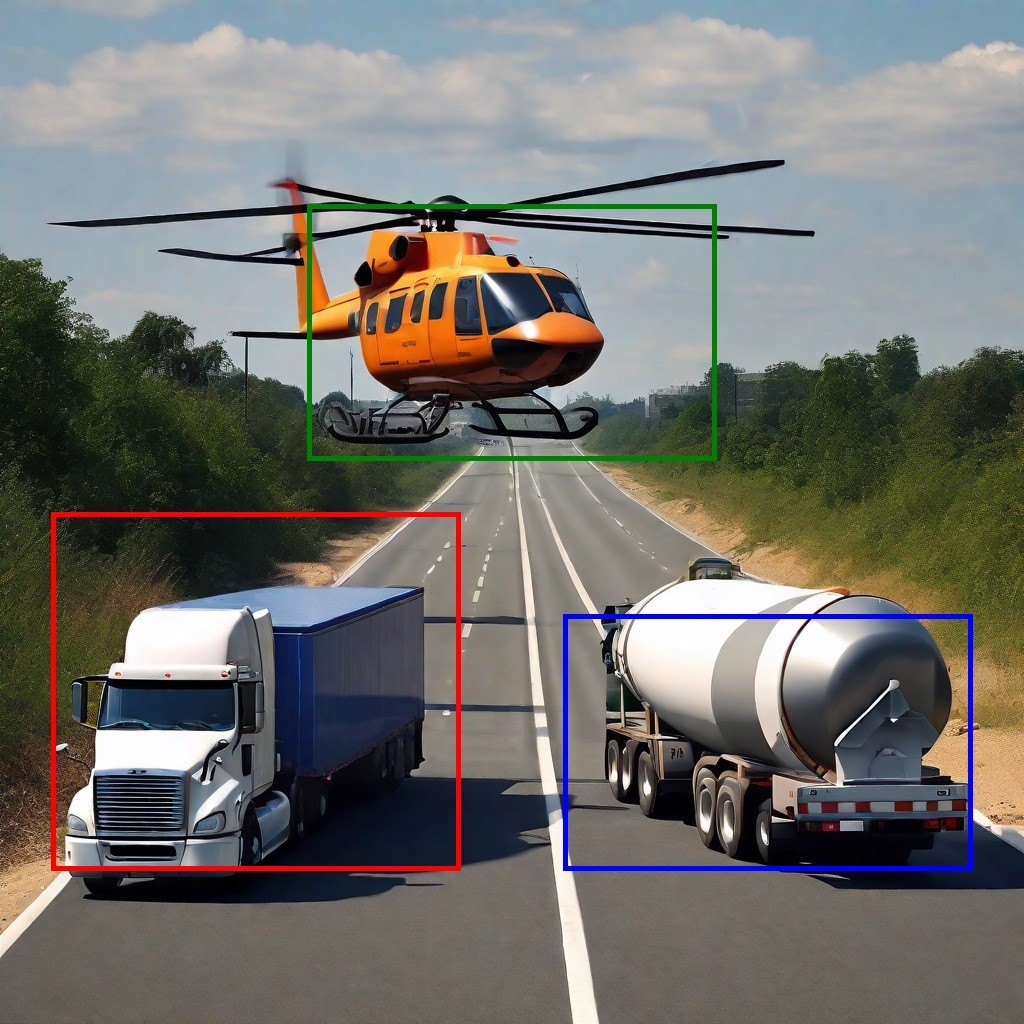} &
        \includegraphics[width=0.2\textwidth]{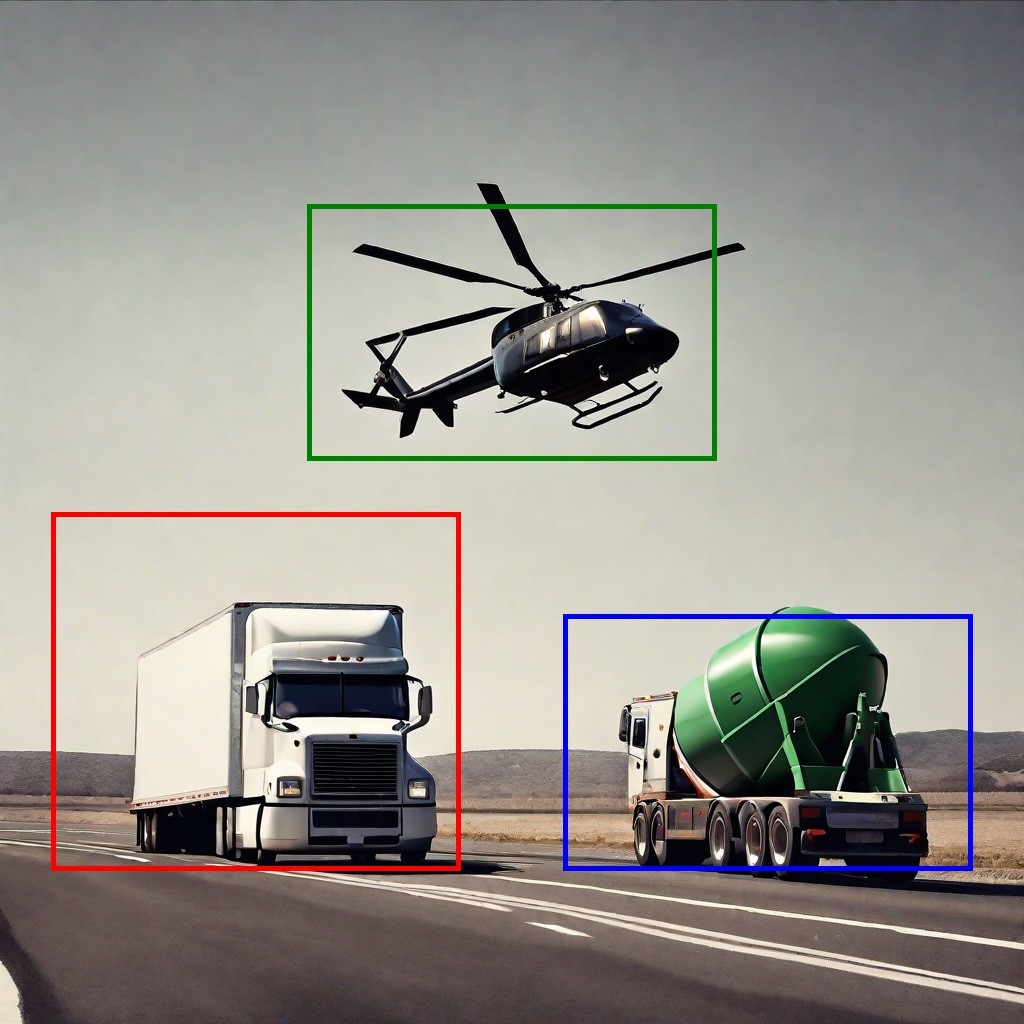} \\
        \raisebox{24pt}{\rotatebox{90}{Vanilla SDXL}} &
        \includegraphics[width=0.2\textwidth]{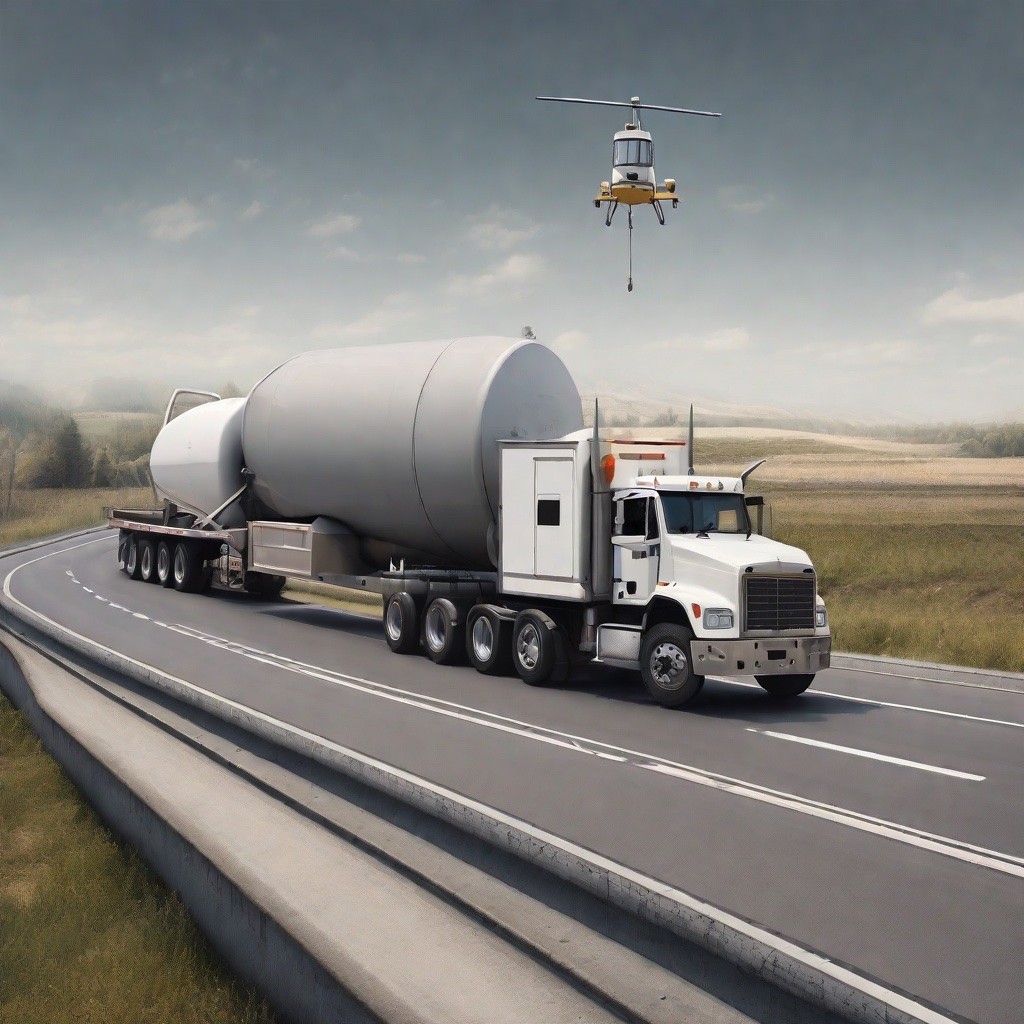} &
        \includegraphics[width=0.2\textwidth]{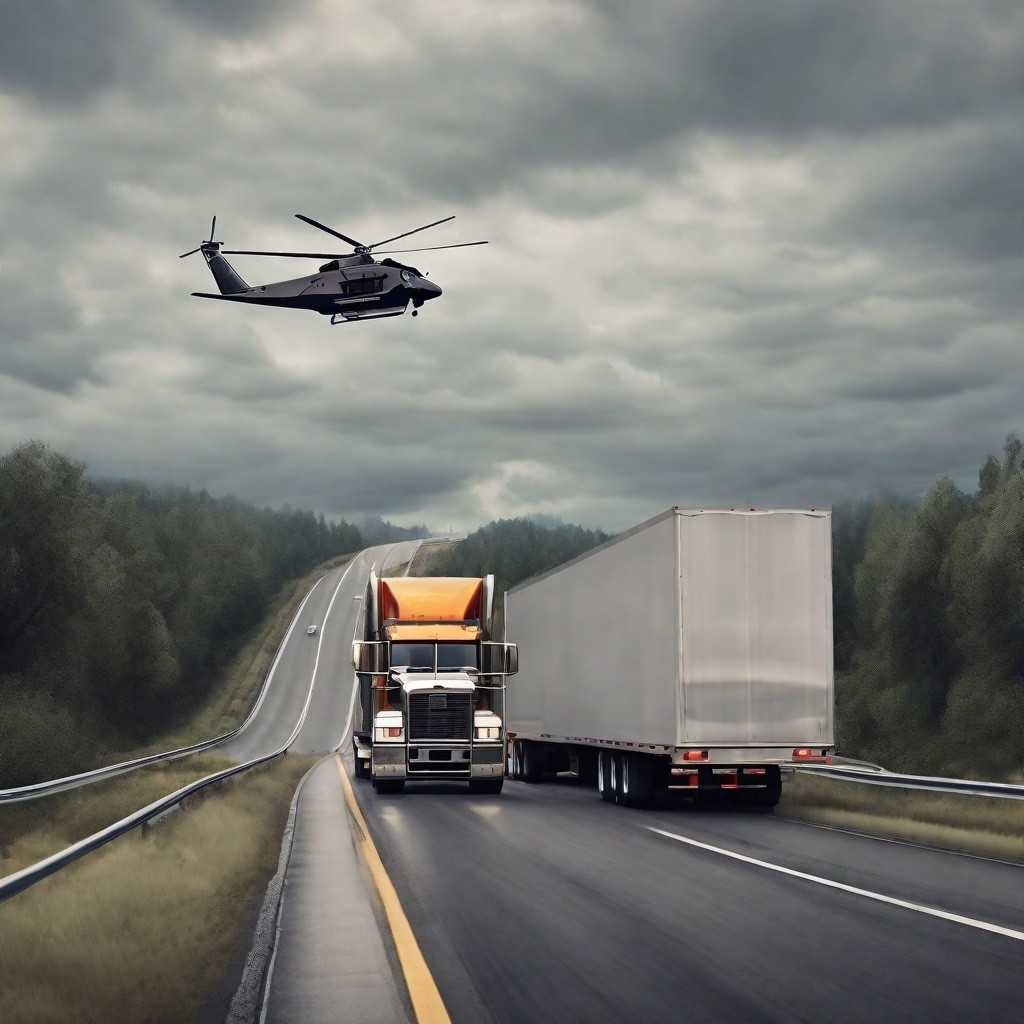} &
        \includegraphics[width=0.2\textwidth]{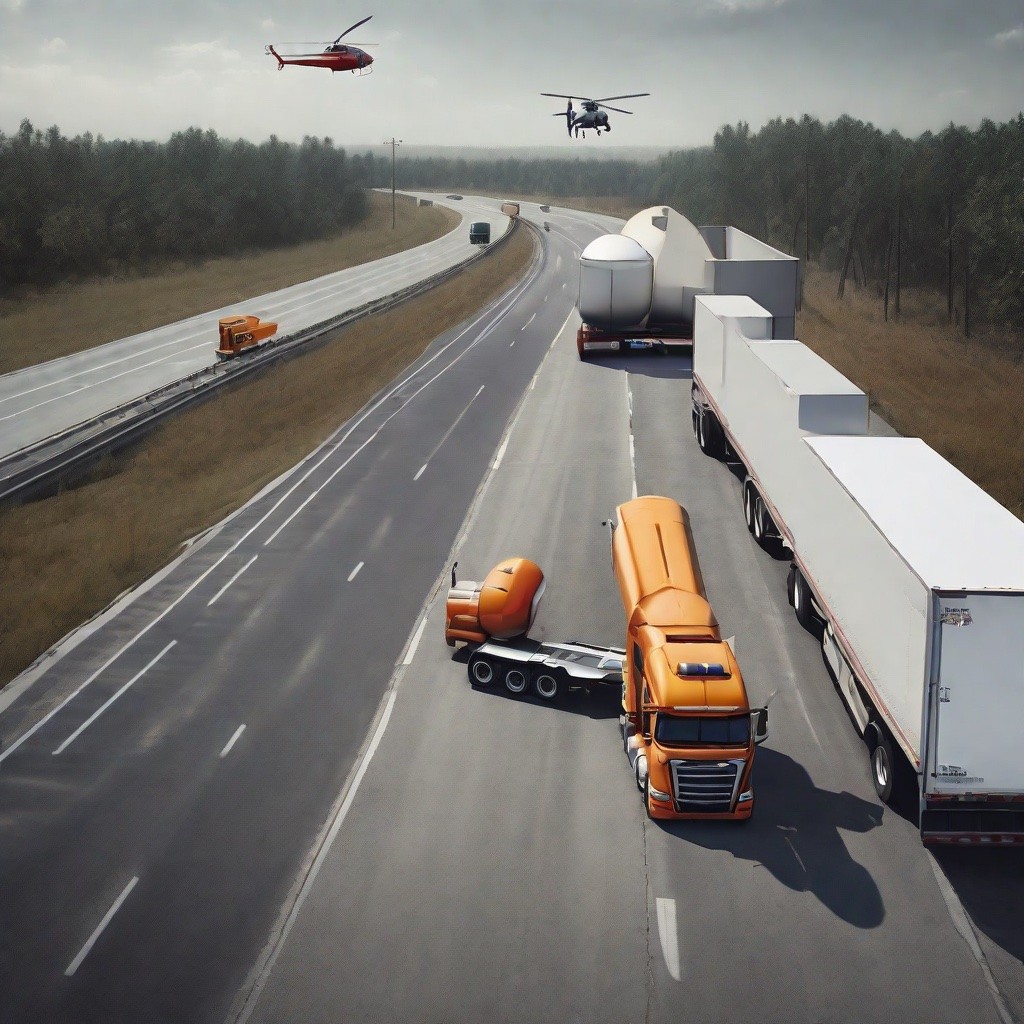} &
        \includegraphics[width=0.2\textwidth]{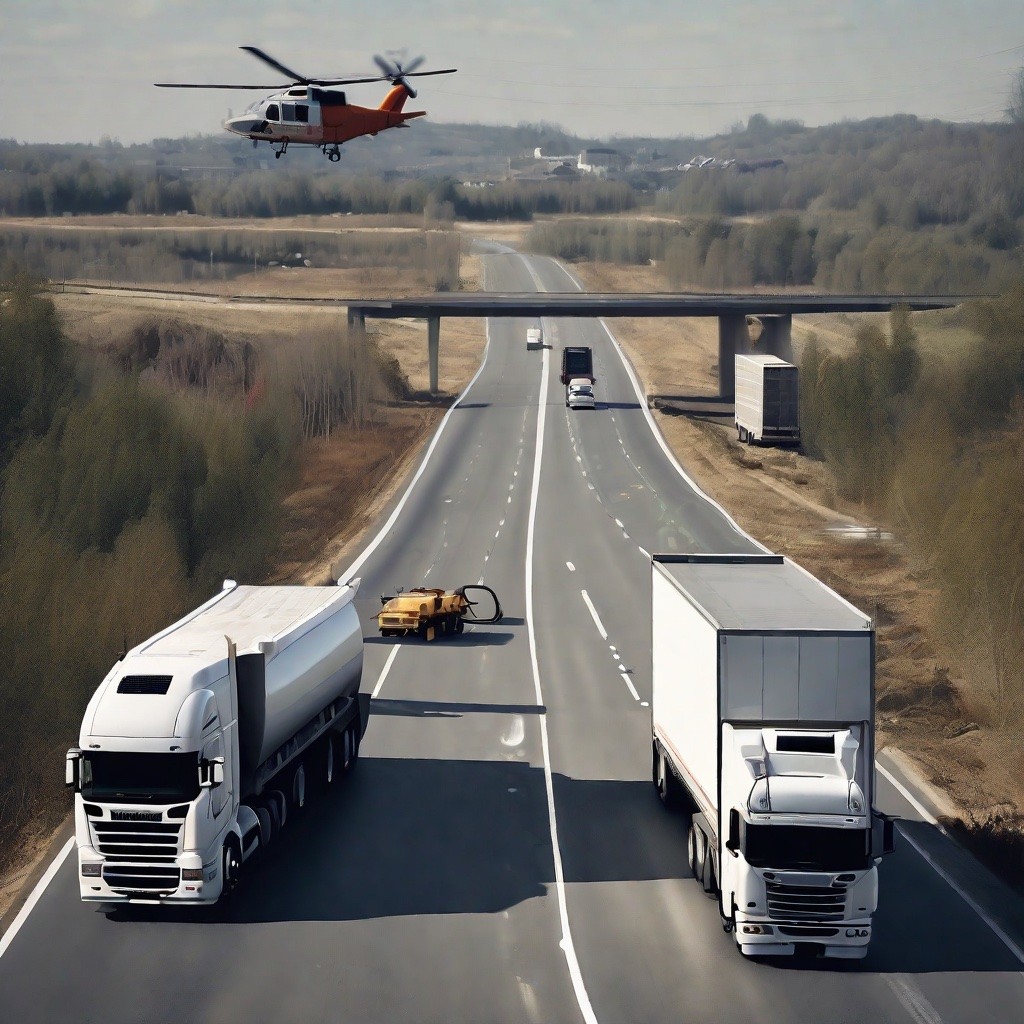} &
        \includegraphics[width=0.2\textwidth]{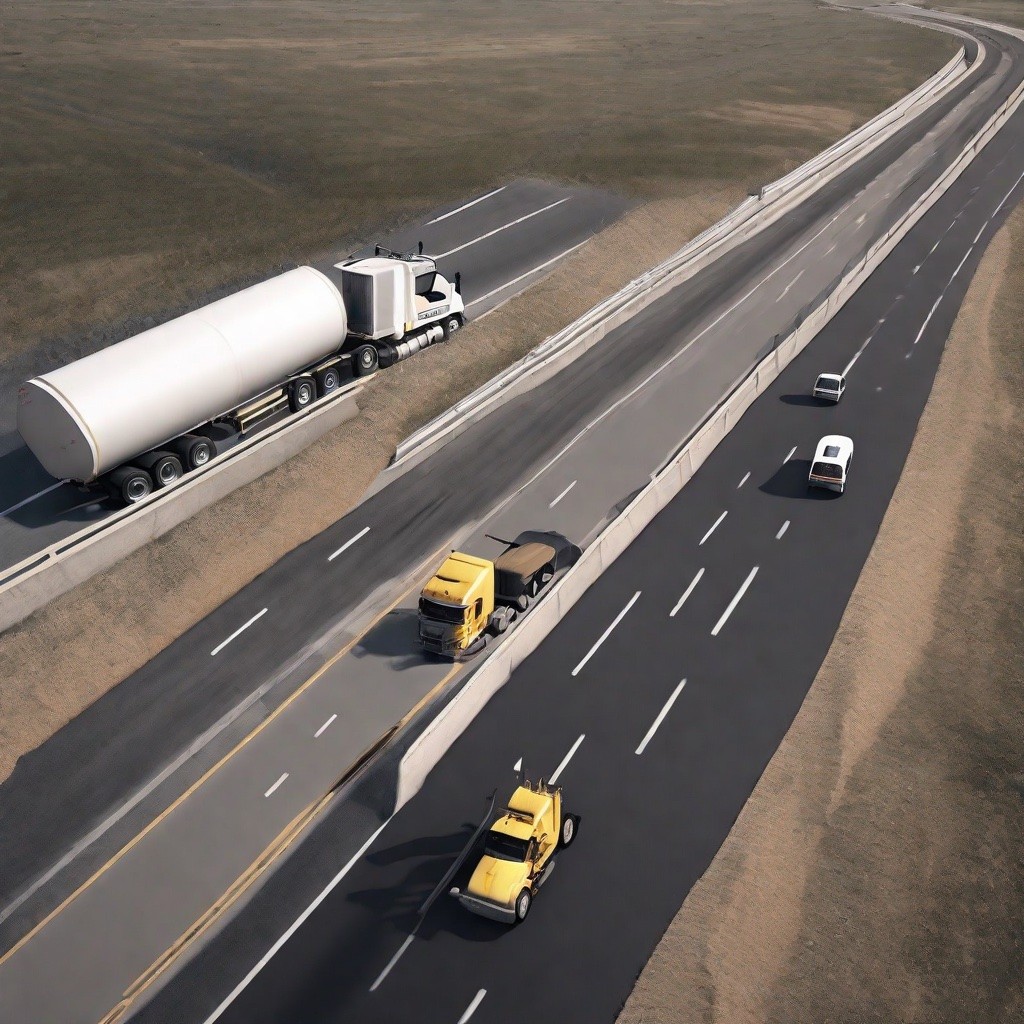} \\
        \\[-10pt]

        &
        \multicolumn{5}{c}{``A realistic photo of a tool shed with a \textcolor{red}{\textit{\underline{lawn mower}}} and a \textcolor{blue}{\textit{\underline{bucket}}} and a \textcolor{green}{\textit{\underline{ladder}}} and \textcolor{orange}{\textit{\underline{tools attached}}} to the wall.''}
        \\
       \raisebox{15pt}{\rotatebox{90}{Bounded Attention}} &
        \includegraphics[width=0.2\textwidth]{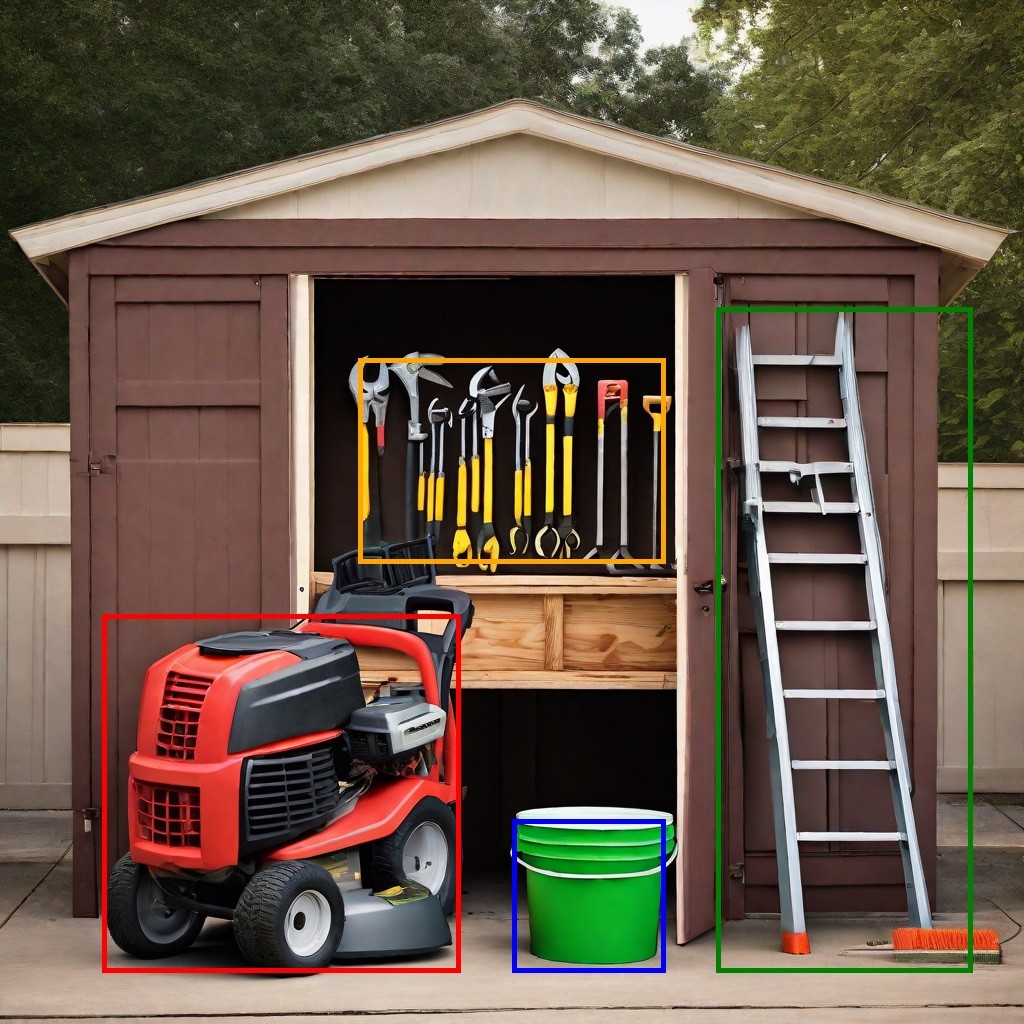} &
        \includegraphics[width=0.2\textwidth]{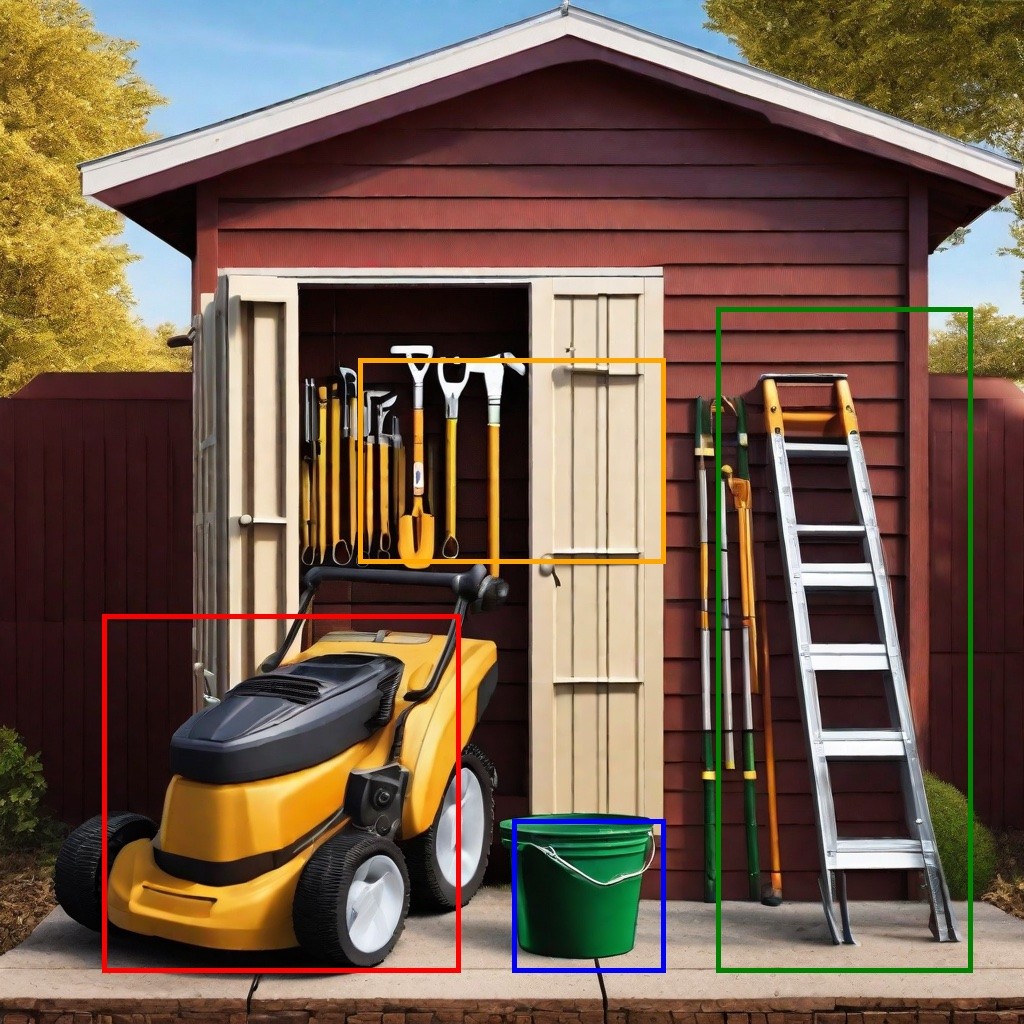} &
        \includegraphics[width=0.2\textwidth]{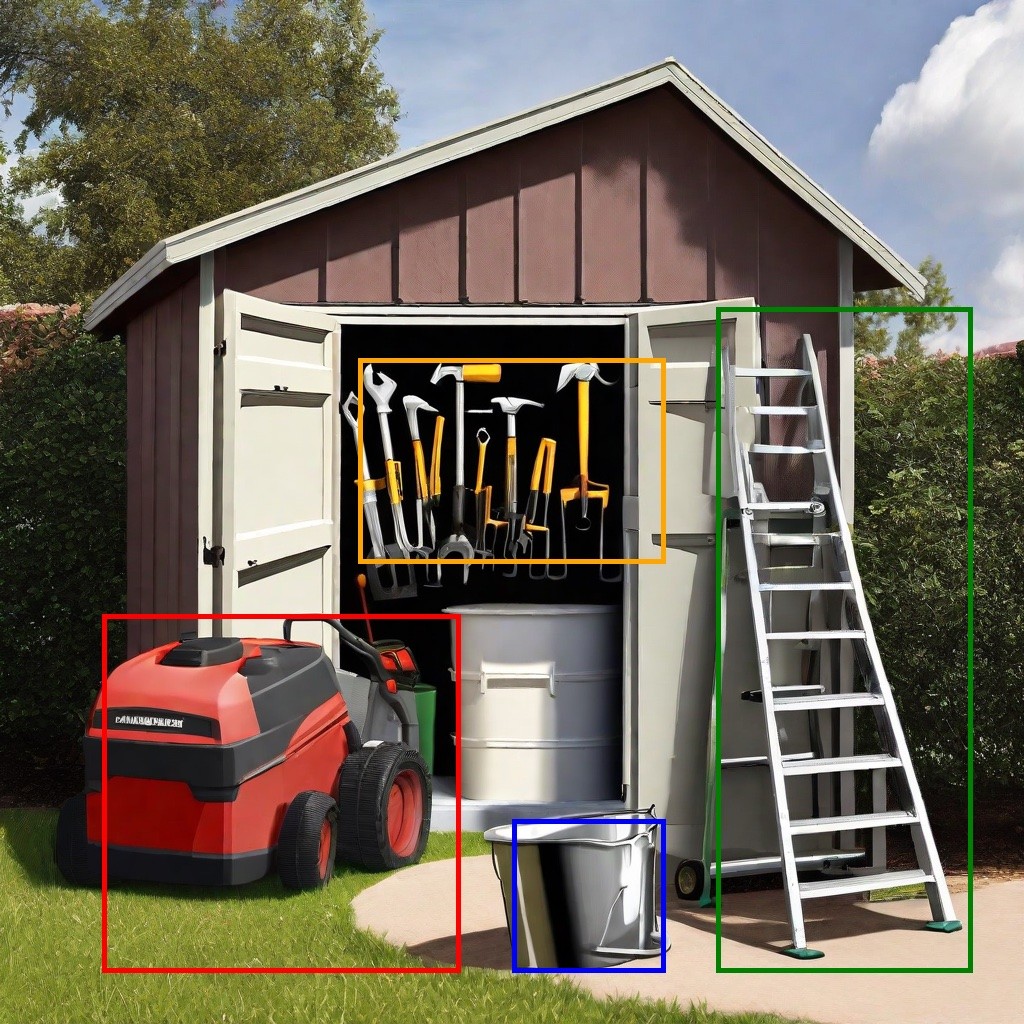} &
        \includegraphics[width=0.2\textwidth]{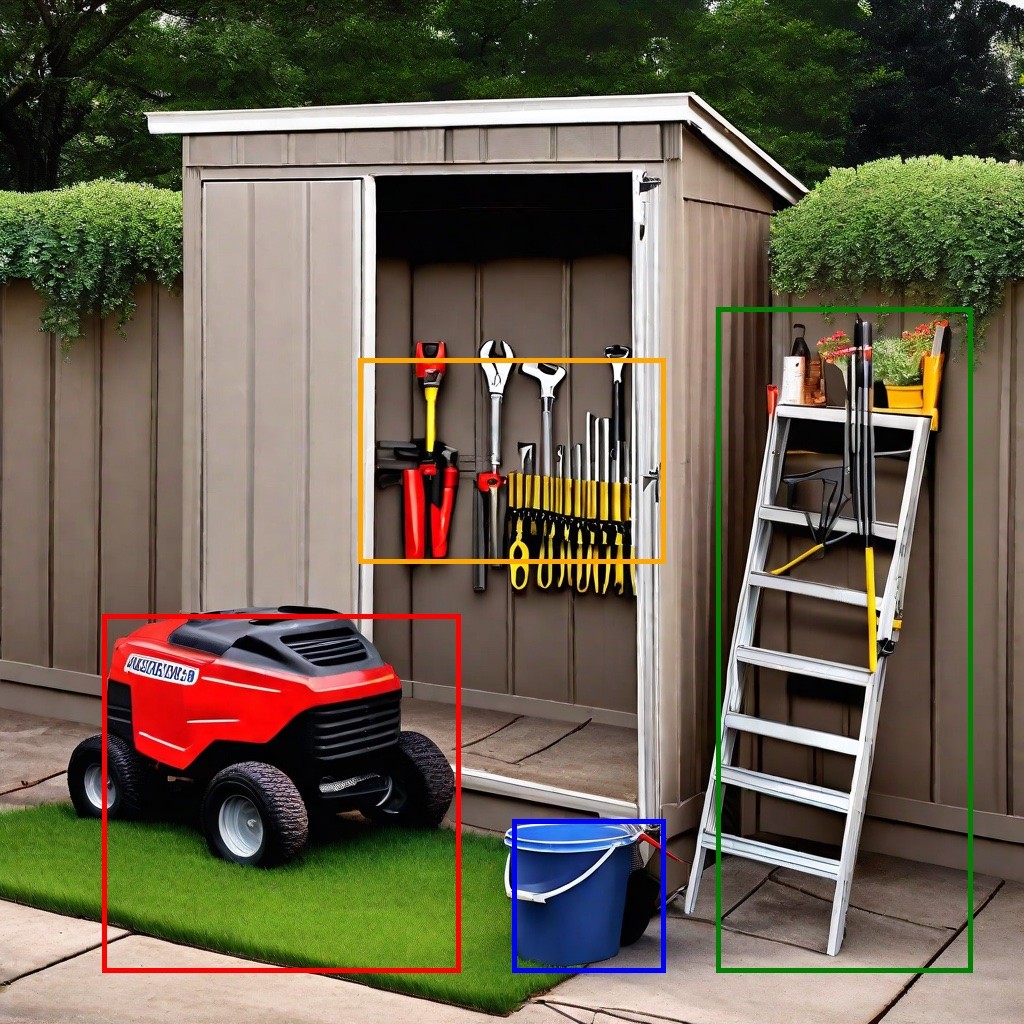} &
        \includegraphics[width=0.2\textwidth]{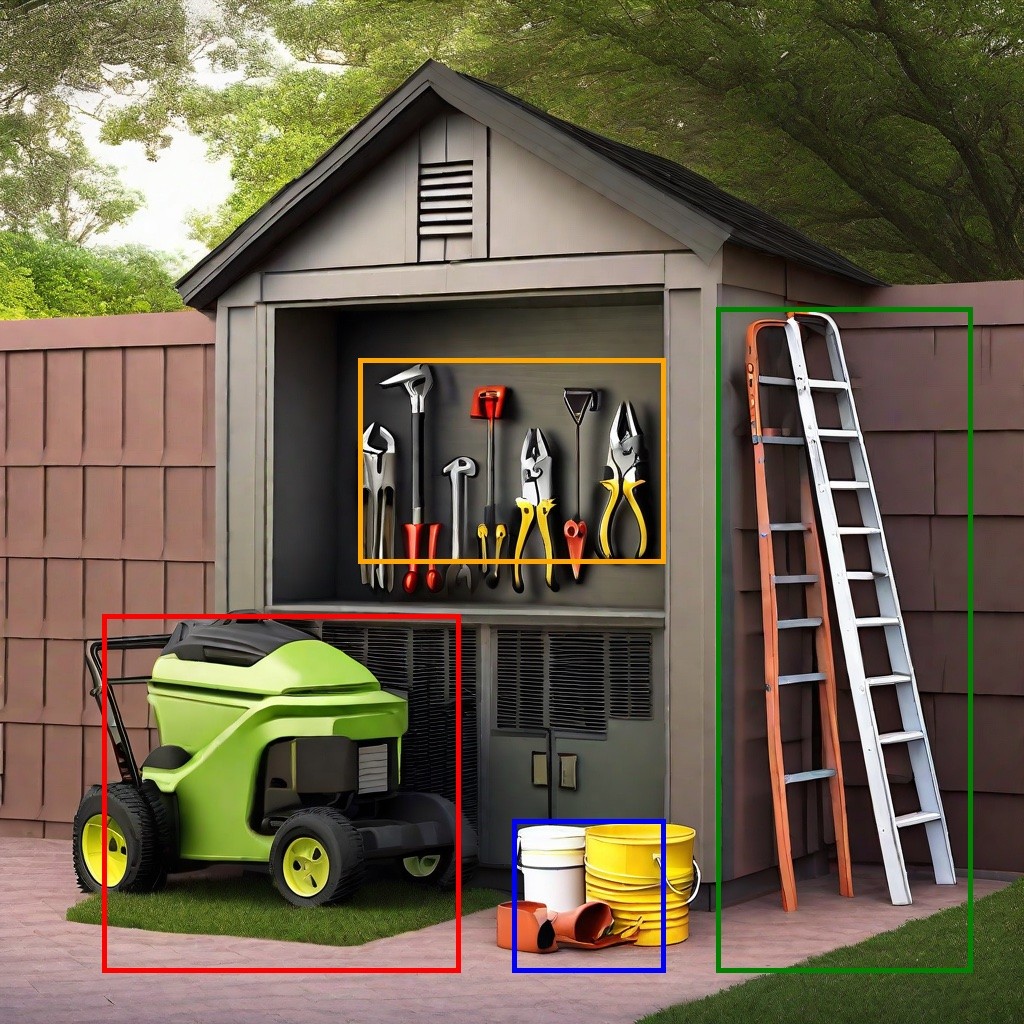} \\
        \raisebox{24pt}{\rotatebox{90}{Vanilla SDXL}} &
        \includegraphics[width=0.2\textwidth]{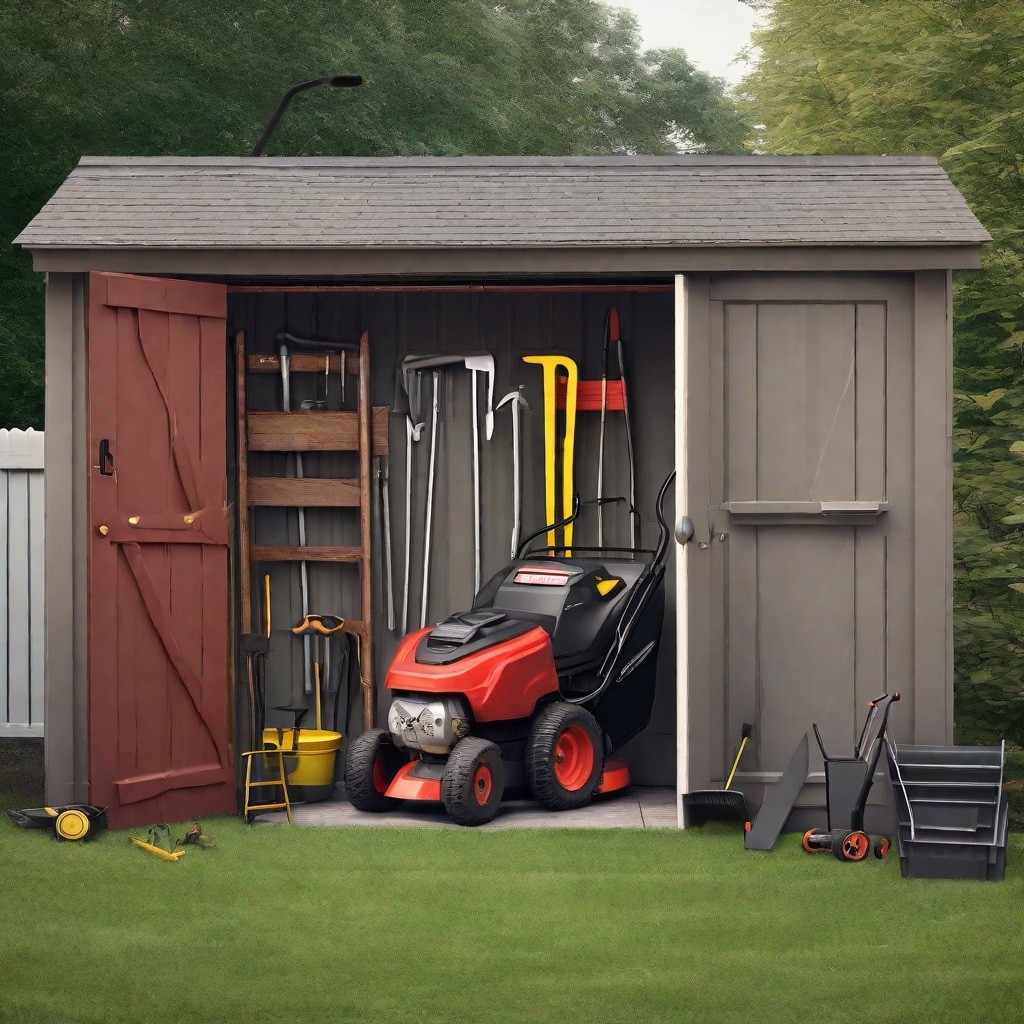} &
        \includegraphics[width=0.2\textwidth]{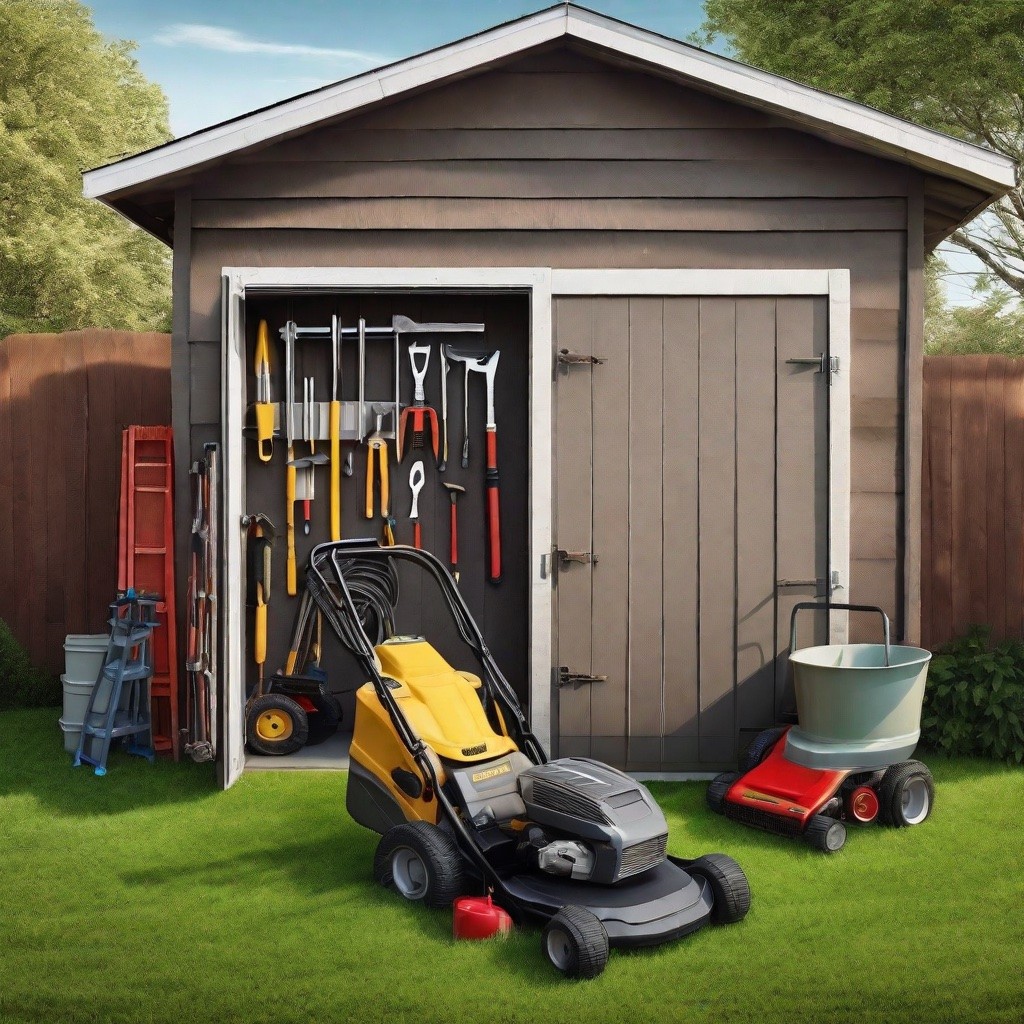} &
        \includegraphics[width=0.2\textwidth]{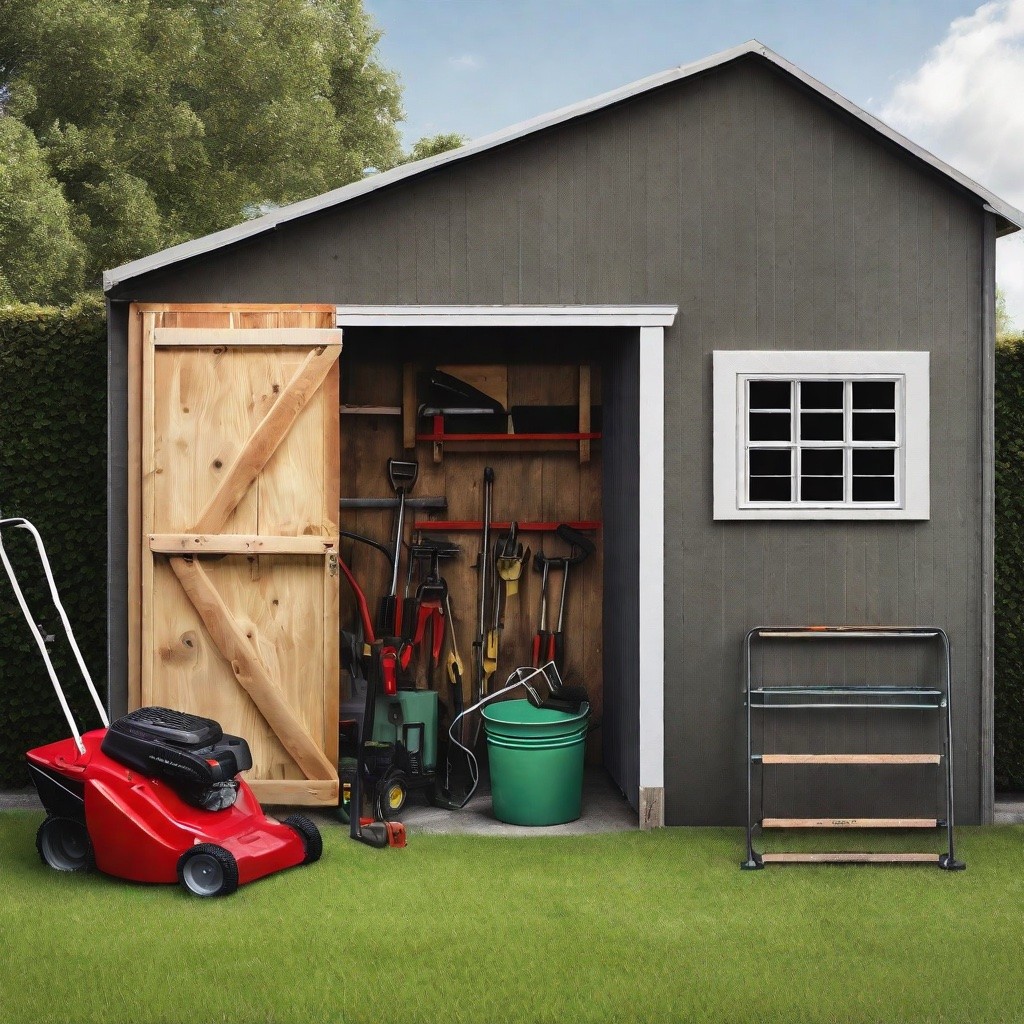} &
        \includegraphics[width=0.2\textwidth]{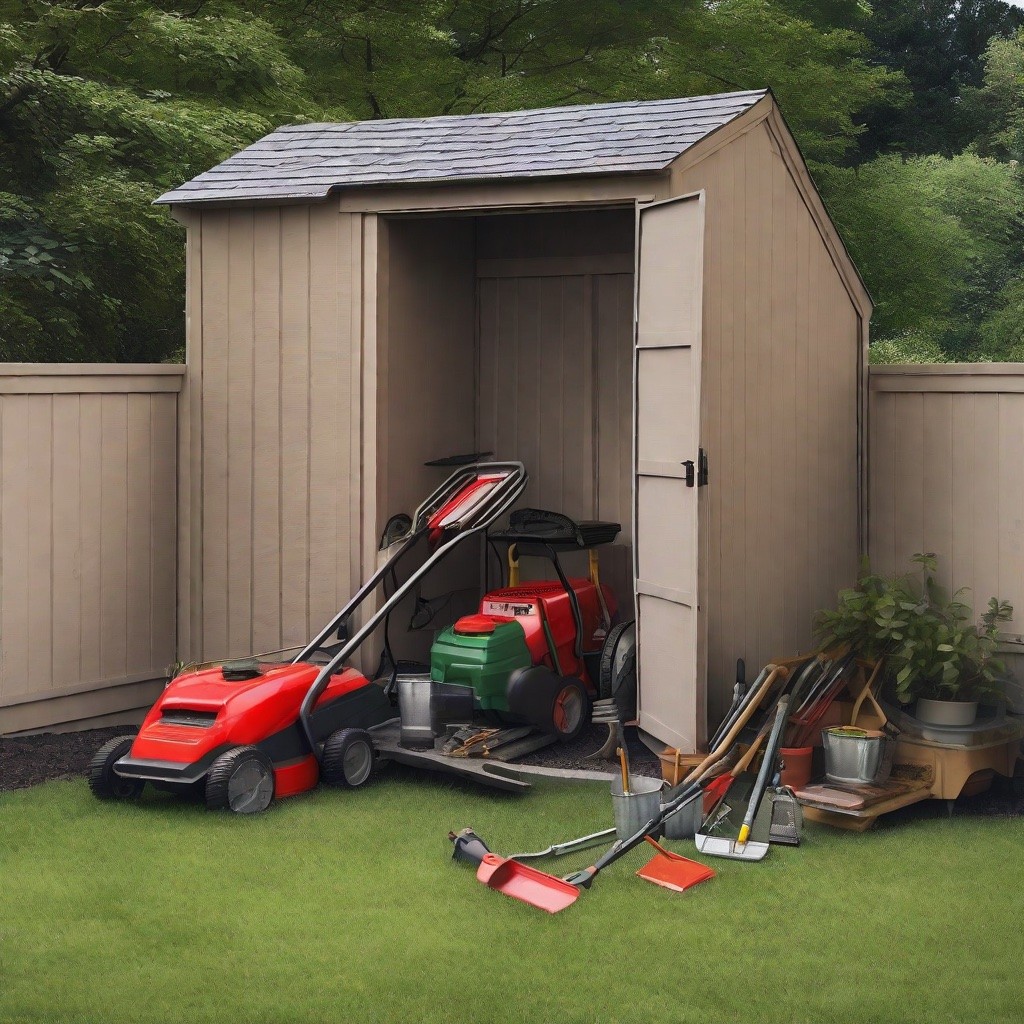} &
        \includegraphics[width=0.2\textwidth]{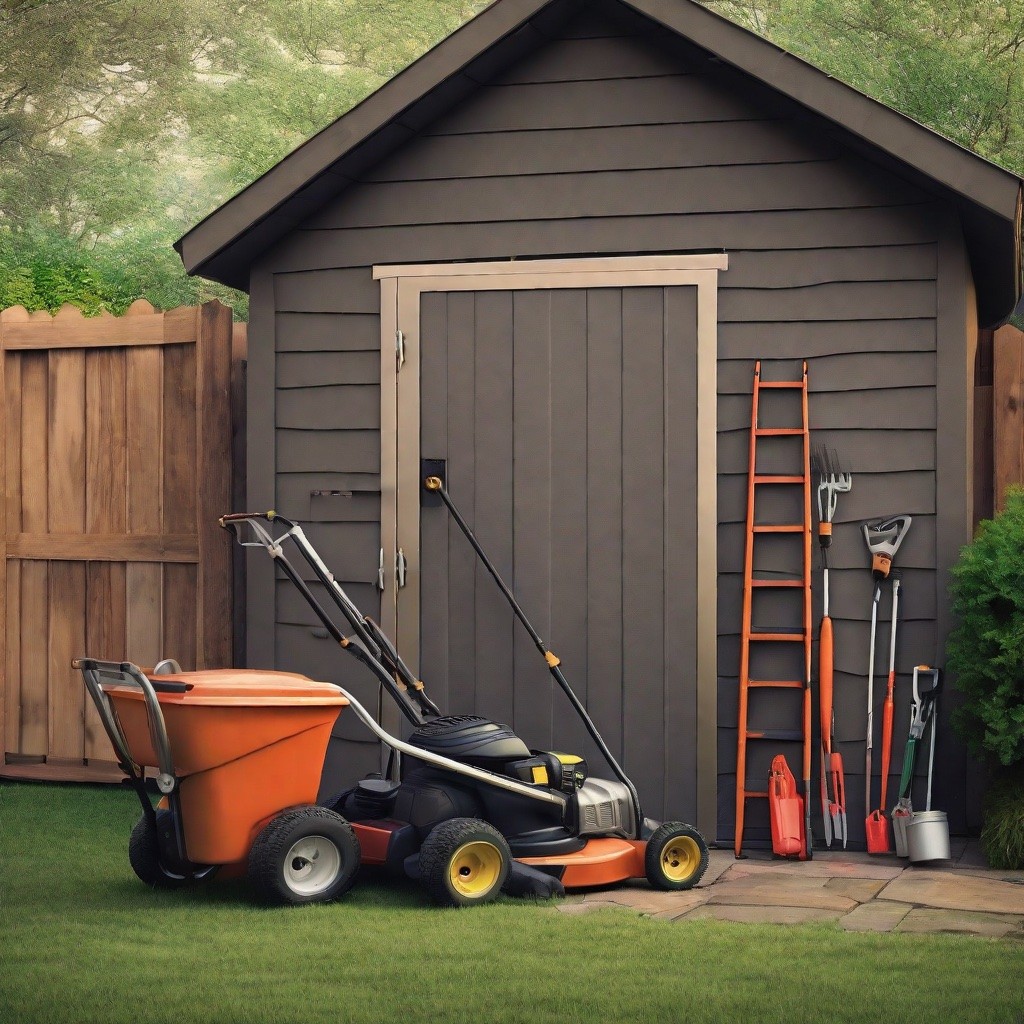} \\
        \\
        
    \end{tabular}
    }
    \vspace{-24pt}
    \captionof{figure}{
        More results of our method using SDXL.
    }
    \label{fig:sdxl4}
\end{figure*}

%% file: main.bbl
\begin{thebibliography}{40}
\providecommand{\natexlab}[1]{#1}
\providecommand{\url}[1]{\texttt{#1}}
\expandafter\ifx\csname urlstyle\endcsname\relax
  \providecommand{\doi}[1]{doi: #1}\else
  \providecommand{\doi}{doi: \begingroup \urlstyle{rm}\Url}\fi

\bibitem[Alaluf et~al.(2023)Alaluf, Garibi, Patashnik, Averbuch-Elor, and Cohen-Or]{alaluf2023cross}
Yuval Alaluf, Daniel Garibi, Or Patashnik, Hadar Averbuch-Elor, and Daniel Cohen-Or.
\newblock Cross-image attention for zero-shot appearance transfer.
\newblock \emph{arXiv preprint arXiv:2311.03335}, 2023.

\bibitem[Avrahami et~al.(2023)Avrahami, Hayes, Gafni, Gupta, Taigman, Parikh, Lischinski, Fried, and Yin]{avrahami2023spatext}
Omri Avrahami, Thomas Hayes, Oran Gafni, Sonal Gupta, Yaniv Taigman, Devi Parikh, Dani Lischinski, Ohad Fried, and Xi Yin.
\newblock Spatext: Spatio-textual representation for controllable image generation.
\newblock In \emph{Proceedings of the IEEE/CVF Conference on Computer Vision and Pattern Recognition}, pages 18370--18380, 2023.

\bibitem[Balaji et~al.(2022)Balaji, Nah, Huang, Vahdat, Song, Kreis, Aittala, Aila, Laine, Catanzaro, et~al.]{balaji2022ediffi}
Yogesh Balaji, Seungjun Nah, Xun Huang, Arash Vahdat, Jiaming Song, Karsten Kreis, Miika Aittala, Timo Aila, Samuli Laine, Bryan Catanzaro, et~al.
\newblock ediffi: Text-to-image diffusion models with an ensemble of expert denoisers.
\newblock \emph{arXiv preprint arXiv:2211.01324}, 2022.

\bibitem[Bar-Tal et~al.(2023)Bar-Tal, Yariv, Lipman, and Dekel]{bar2023multidiffusion}
Omer Bar-Tal, Lior Yariv, Yaron Lipman, and Tali Dekel.
\newblock Multidiffusion: Fusing diffusion paths for controlled image generation.
\newblock 2023.

\bibitem[Cao et~al.(2023)Cao, Wang, Qi, Shan, Qie, and Zheng]{cao2023masactrl}
Mingdeng Cao, Xintao Wang, Zhongang Qi, Ying Shan, Xiaohu Qie, and Yinqiang Zheng.
\newblock Masactrl: Tuning-free mutual self-attention control for consistent image synthesis and editing.
\newblock \emph{arXiv preprint arXiv:2304.08465}, 2023.

\bibitem[Chefer et~al.(2023)Chefer, Alaluf, Vinker, Wolf, and Cohen-Or]{chefer2023attend}
Hila Chefer, Yuval Alaluf, Yael Vinker, Lior Wolf, and Daniel Cohen-Or.
\newblock Attend-and-excite: Attention-based semantic guidance for text-to-image diffusion models.
\newblock \emph{ACM Transactions on Graphics (TOG)}, 42\penalty0 (4):\penalty0 1--10, 2023.

\bibitem[Chen et~al.(2023)Chen, Laina, and Vedaldi]{chen2023training}
Minghao Chen, Iro Laina, and Andrea Vedaldi.
\newblock Training-free layout control with cross-attention guidance.
\newblock \emph{arXiv preprint arXiv:2304.03373}, 2023.

\bibitem[Couairon et~al.(2023)Couairon, Careil, Cord, Lathuiliere, and Verbeek]{couairon2023zero}
Guillaume Couairon, Marlene Careil, Matthieu Cord, St{\'e}phane Lathuiliere, and Jakob Verbeek.
\newblock Zero-shot spatial layout conditioning for text-to-image diffusion models.
\newblock In \emph{Proceedings of the IEEE/CVF International Conference on Computer Vision}, pages 2174--2183, 2023.

\bibitem[Endo(2023)]{endo2023masked}
Yuki Endo.
\newblock Masked-attention diffusion guidance for spatially controlling text-to-image generation.
\newblock \emph{arXiv preprint arXiv:2308.06027}, 2023.

\bibitem[Epstein et~al.(2024)Epstein, Jabri, Poole, Efros, and Holynski]{epstein2024diffusion}
Dave Epstein, Allan Jabri, Ben Poole, Alexei Efros, and Aleksander Holynski.
\newblock Diffusion self-guidance for controllable image generation.
\newblock \emph{Advances in Neural Information Processing Systems}, 36, 2024.

\bibitem[Feng et~al.(2022)Feng, He, Fu, Jampani, Akula, Narayana, Basu, Wang, and Wang]{feng2022training}
Weixi Feng, Xuehai He, Tsu-Jui Fu, Varun Jampani, Arjun Akula, Pradyumna Narayana, Sugato Basu, Xin~Eric Wang, and William~Yang Wang.
\newblock Training-free structured diffusion guidance for compositional text-to-image synthesis.
\newblock \emph{arXiv preprint arXiv:2212.05032}, 2022.

\bibitem[He et~al.(2023)He, Salakhutdinov, and Kolter]{he2023localized}
Yutong He, Ruslan Salakhutdinov, and J~Zico Kolter.
\newblock Localized text-to-image generation for free via cross attention control.
\newblock \emph{arXiv preprint arXiv:2306.14636}, 2023.

\bibitem[Hertz et~al.(2022)Hertz, Mokady, Tenenbaum, Aberman, Pritch, and Cohen-Or]{hertz2022prompt}
Amir Hertz, Ron Mokady, Jay Tenenbaum, Kfir Aberman, Yael Pritch, and Daniel Cohen-Or.
\newblock Prompt-to-prompt image editing with cross attention control.
\newblock \emph{arXiv preprint arXiv:2208.01626}, 2022.

\bibitem[Ho and Salimans(2022)]{ho2022classifier}
Jonathan Ho and Tim Salimans.
\newblock Classifier-free diffusion guidance.
\newblock \emph{arXiv preprint arXiv:2207.12598}, 2022.

\bibitem[Kang et~al.(2023)Kang, Zhu, Zhang, Park, Shechtman, Paris, and Park]{kang2023gigagan}
Minguk Kang, Jun-Yan Zhu, Richard Zhang, Jaesik Park, Eli Shechtman, Sylvain Paris, and Taesung Park.
\newblock Scaling up gans for text-to-image synthesis.
\newblock In \emph{Proceedings of the IEEE Conference on Computer Vision and Pattern Recognition (CVPR)}, 2023.

\bibitem[Kim et~al.(2023)Kim, Lee, Kim, Ha, and Zhu]{kim2023dense}
Yunji Kim, Jiyoung Lee, Jin-Hwa Kim, Jung-Woo Ha, and Jun-Yan Zhu.
\newblock Dense text-to-image generation with attention modulation.
\newblock In \emph{Proceedings of the IEEE/CVF International Conference on Computer Vision}, pages 7701--7711, 2023.

\bibitem[Li et~al.(2023)Li, Liu, Wu, Mu, Yang, Gao, Li, and Lee]{li2023gligen}
Yuheng Li, Haotian Liu, Qingyang Wu, Fangzhou Mu, Jianwei Yang, Jianfeng Gao, Chunyuan Li, and Yong~Jae Lee.
\newblock Gligen: Open-set grounded text-to-image generation.
\newblock In \emph{Proceedings of the IEEE/CVF Conference on Computer Vision and Pattern Recognition}, pages 22511--22521, 2023.

\bibitem[Lian et~al.(2023)Lian, Li, Yala, and Darrell]{lian2023llm}
Long Lian, Boyi Li, Adam Yala, and Trevor Darrell.
\newblock Llm-grounded diffusion: Enhancing prompt understanding of text-to-image diffusion models with large language models.
\newblock \emph{arXiv preprint arXiv:2305.13655}, 2023.

\bibitem[Meng et~al.(2021)Meng, He, Song, Song, Wu, Zhu, and Ermon]{meng2021sdedit}
Chenlin Meng, Yutong He, Yang Song, Jiaming Song, Jiajun Wu, Jun-Yan Zhu, and Stefano Ermon.
\newblock Sdedit: Guided image synthesis and editing with stochastic differential equations.
\newblock \emph{arXiv preprint arXiv:2108.01073}, 2021.

\bibitem[Paiss et~al.(2023)Paiss, Ephrat, Tov, Zada, Mosseri, Irani, and Dekel]{paiss2023teaching}
Roni Paiss, Ariel Ephrat, Omer Tov, Shiran Zada, Inbar Mosseri, Michal Irani, and Tali Dekel.
\newblock Teaching clip to count to ten.
\newblock \emph{arXiv preprint arXiv:2302.12066}, 2023.

\bibitem[Patashnik et~al.(2023)Patashnik, Garibi, Azuri, Averbuch-Elor, and Cohen-Or]{patashnik2023localizing}
Or Patashnik, Daniel Garibi, Idan Azuri, Hadar Averbuch-Elor, and Daniel Cohen-Or.
\newblock Localizing object-level shape variations with text-to-image diffusion models.
\newblock \emph{arXiv preprint arXiv:2303.11306}, 2023.

\bibitem[Phung et~al.(2023)Phung, Ge, and Huang]{phung2023grounded}
Quynh Phung, Songwei Ge, and Jia-Bin Huang.
\newblock Grounded text-to-image synthesis with attention refocusing.
\newblock \emph{arXiv preprint arXiv:2306.05427}, 2023.

\bibitem[Podell et~al.(2023)Podell, English, Lacey, Blattmann, Dockhorn, M{\"u}ller, Penna, and Rombach]{podell2023sdxl}
Dustin Podell, Zion English, Kyle Lacey, Andreas Blattmann, Tim Dockhorn, Jonas M{\"u}ller, Joe Penna, and Robin Rombach.
\newblock Sdxl: Improving latent diffusion models for high-resolution image synthesis.
\newblock \emph{arXiv preprint arXiv:2307.01952}, 2023.

\bibitem[Qu et~al.(2023)Qu, Wu, Fei, Nie, and Chua]{qu2023layoutllm}
Leigang Qu, Shengqiong Wu, Hao Fei, Liqiang Nie, and Tat-Seng Chua.
\newblock Layoutllm-t2i: Eliciting layout guidance from llm for text-to-image generation.
\newblock \emph{arXiv preprint arXiv:2308.05095}, 2023.

\bibitem[Radford et~al.(2021)Radford, Kim, Hallacy, Ramesh, Goh, Agarwal, Sastry, Askell, Mishkin, Clark, et~al.]{radford2021learning}
Alec Radford, Jong~Wook Kim, Chris Hallacy, Aditya Ramesh, Gabriel Goh, Sandhini Agarwal, Girish Sastry, Amanda Askell, Pamela Mishkin, Jack Clark, et~al.
\newblock Learning transferable visual models from natural language supervision.
\newblock In \emph{International conference on machine learning}, pages 8748--8763. PMLR, 2021.

\bibitem[Ramesh et~al.(2022)Ramesh, Dhariwal, Nichol, Chu, and Chen]{ramesh2022hierarchical}
Aditya Ramesh, Prafulla Dhariwal, Alex Nichol, Casey Chu, and Mark Chen.
\newblock Hierarchical text-conditional image generation with clip latents.
\newblock \emph{arXiv preprint arXiv:2204.06125}, 1\penalty0 (2):\penalty0 3, 2022.

\bibitem[Rassin et~al.(2023)Rassin, Hirsch, Glickman, Ravfogel, Goldberg, and Chechik]{rassin2023linguistic}
Royi Rassin, Eran Hirsch, Daniel Glickman, Shauli Ravfogel, Yoav Goldberg, and Gal Chechik.
\newblock Linguistic binding in diffusion models: Enhancing attribute correspondence through attention map alignment.
\newblock \emph{arXiv preprint arXiv:2306.08877}, 2023.

\bibitem[Rombach et~al.(2022)Rombach, Blattmann, Lorenz, Esser, and Ommer]{rombach2022high}
Robin Rombach, Andreas Blattmann, Dominik Lorenz, Patrick Esser, and Bj{\"o}rn Ommer.
\newblock High-resolution image synthesis with latent diffusion models.
\newblock In \emph{Proceedings of the IEEE/CVF conference on computer vision and pattern recognition}, pages 10684--10695, 2022.

\bibitem[Saharia et~al.(2022)Saharia, Chan, Saxena, Li, Whang, Denton, Ghasemipour, Gontijo~Lopes, Karagol~Ayan, Salimans, et~al.]{saharia2022photorealistic}
Chitwan Saharia, William Chan, Saurabh Saxena, Lala Li, Jay Whang, Emily~L Denton, Kamyar Ghasemipour, Raphael Gontijo~Lopes, Burcu Karagol~Ayan, Tim Salimans, et~al.
\newblock Photorealistic text-to-image diffusion models with deep language understanding.
\newblock \emph{Advances in Neural Information Processing Systems}, 35:\penalty0 36479--36494, 2022.

\bibitem[Schuhmann et~al.(2022)Schuhmann, Beaumont, Vencu, Gordon, Wightman, Cherti, Coombes, Katta, Mullis, Wortsman, Schramowski, Kundurthy, Crowson, Schmidt, Kaczmarczyk, and Jitsev]{schuhmann2022laion5b}
Christoph Schuhmann, Romain Beaumont, Richard Vencu, Cade Gordon, Ross Wightman, Mehdi Cherti, Theo Coombes, Aarush Katta, Clayton Mullis, Mitchell Wortsman, Patrick Schramowski, Srivatsa Kundurthy, Katherine Crowson, Ludwig Schmidt, Robert Kaczmarczyk, and Jenia Jitsev.
\newblock Laion-5b: An open large-scale dataset for training next generation image-text models, 2022.

\bibitem[Tumanyan et~al.(2023)Tumanyan, Geyer, Bagon, and Dekel]{tumanyan2023plug}
Narek Tumanyan, Michal Geyer, Shai Bagon, and Tali Dekel.
\newblock Plug-and-play diffusion features for text-driven image-to-image translation.
\newblock In \emph{Proceedings of the IEEE/CVF Conference on Computer Vision and Pattern Recognition}, pages 1921--1930, 2023.

\bibitem[Tunanyan et~al.(2023)Tunanyan, Xu, Navasardyan, Wang, and Shi]{tunanyan2023multi}
Hazarapet Tunanyan, Dejia Xu, Shant Navasardyan, Zhangyang Wang, and Humphrey Shi.
\newblock Multi-concept t2i-zero: Tweaking only the text embeddings and nothing else.
\newblock \emph{arXiv preprint arXiv:2310.07419}, 2023.

\bibitem[Voynov et~al.(2023)Voynov, Chu, Cohen-Or, and Aberman]{voynov2023p+}
Andrey Voynov, Qinghao Chu, Daniel Cohen-Or, and Kfir Aberman.
\newblock $ p+ $: Extended textual conditioning in text-to-image generation.
\newblock \emph{arXiv preprint arXiv:2303.09522}, 2023.

\bibitem[White and Cotterell(2022)]{white2022schr}
Jennifer~C White and Ryan Cotterell.
\newblock Schr$\backslash$"$\{$o$\}$ dinger's bat: Diffusion models sometimes generate polysemous words in superposition.
\newblock \emph{arXiv preprint arXiv:2211.13095}, 2022.

\bibitem[Wu et~al.(2019)Wu, Kirillov, Massa, Lo, and Girshick]{wu2019detectron2}
Yuxin Wu, Alexander Kirillov, Francisco Massa, Wan-Yen Lo, and Ross Girshick.
\newblock Detectron2.
\newblock \url{https://github.com/facebookresearch/detectron2}, 2019.

\bibitem[Xie et~al.(2023)Xie, Li, Huang, Liu, Zhang, Zheng, and Shou]{xie2023boxdiff}
Jinheng Xie, Yuexiang Li, Yawen Huang, Haozhe Liu, Wentian Zhang, Yefeng Zheng, and Mike~Zheng Shou.
\newblock Boxdiff: Text-to-image synthesis with training-free box-constrained diffusion.
\newblock In \emph{Proceedings of the IEEE/CVF International Conference on Computer Vision}, pages 7452--7461, 2023.

\bibitem[Yang et~al.(2023)Yang, Wang, Gan, Li, Lin, Wu, Duan, Liu, Liu, Zeng, et~al.]{yang2023reco}
Zhengyuan Yang, Jianfeng Wang, Zhe Gan, Linjie Li, Kevin Lin, Chenfei Wu, Nan Duan, Zicheng Liu, Ce Liu, Michael Zeng, et~al.
\newblock Reco: Region-controlled text-to-image generation.
\newblock In \emph{Proceedings of the IEEE/CVF Conference on Computer Vision and Pattern Recognition}, pages 14246--14255, 2023.

\bibitem[Zhang et~al.(2023)Zhang, Rao, and Agrawala]{zhang2023adding}
Lvmin Zhang, Anyi Rao, and Maneesh Agrawala.
\newblock Adding conditional control to text-to-image diffusion models.
\newblock In \emph{Proceedings of the IEEE/CVF International Conference on Computer Vision}, pages 3836--3847, 2023.

\bibitem[Zhao et~al.(2023)Zhao, Li, Jin, and Zhou]{zhao2023loco}
Peiang Zhao, Han Li, Ruiyang Jin, and S~Kevin Zhou.
\newblock Loco: Locally constrained training-free layout-to-image synthesis.
\newblock \emph{arXiv preprint arXiv:2311.12342}, 2023.

\bibitem[Zheng et~al.(2023)Zheng, Zhou, Li, Qi, Shan, and Li]{zheng2023layoutdiffusion}
Guangcong Zheng, Xianpan Zhou, Xuewei Li, Zhongang Qi, Ying Shan, and Xi Li.
\newblock Layoutdiffusion: Controllable diffusion model for layout-to-image generation.
\newblock In \emph{Proceedings of the IEEE/CVF Conference on Computer Vision and Pattern Recognition}, pages 22490--22499, 2023.

\end{thebibliography}
